%% file: main.tex
\documentclass[10pt,twocolumn,letterpaper]{article}

\usepackage{cvpr}              

\input{preamble}
\definecolor{cvprblue}{rgb}{0.21,0.49,0.74}
\usepackage[pagebackref,breaklinks,colorlinks,allcolors=cvprblue]{hyperref}
\usepackage{stfloats}
\usepackage{float}
\usepackage{adjustbox}
\usepackage{import}
\usepackage{svg}
\usepackage{subcaption}   
\usepackage{comment}
\usepackage{graphicx}
\usepackage[nolist]{acronym}
\usepackage{amsfonts}
\usepackage[ruled,vlined,linesnumbered]{algorithm2e}
\input{acronyms}
\usepackage{amsmath, amssymb}
\usepackage{tikz}
\usetikzlibrary{arrows.meta, bending, calc, positioning}
\usepackage{pgfplots}
\pgfplotsset{compat=1.18}
\usepackage{multirow}
\usepackage{bm}
\usepackage{xspace}
\newsavebox{\samplefig}

\newsavebox{\imagegrid}

\usepackage{colortbl}
\usepackage{xcolor}
\usepackage{tikz}
\usetikzlibrary{spy,calc}
\usepackage{placeins}
\usepackage{siunitx}

\usepackage{graphicx}
\usepackage{subcaption}   
\newcommand{\rowlabel}[1]{%
  \begin{subfigure}[c]{0.03\textwidth}
    \centering
    \rotatebox[origin=c]{90}{XYZ#1}
  \end{subfigure}\hfill}
 \def\method{SIMILR\xspace}
\def\task{SIMI\xspace}
\def\they{SfD\xspace}
\newcommand{\Point}[1]{{\ensuremath{\textbf{\textit{#1}}}}}
\newcommand{\Vector}[1]{{\ensuremath{\textbf{\textit{#1}}}}}
\newcommand{\Matrix}[1]{{\ensuremath{\textnormal{\textbf{\textit{#1}}}}}}
\newcommand{\TMatrix}[3]{{{}^{\mathrm{#1}}\Matrix{#2}_{\mathrm{#3}}}}

\def\confName{CVPR}
\def\confYear{2027}
\title{SIMILR: Single-Image Multi-Instance Joint 3D Object Reconstruction and Pose Estimation Leveraging Recurrence}
\author{Robert Langendörfer \quad Markus Hillemann \quad Markus Ulrich\\
        Karlsruhe Institute of Technology\\
        {\tt\small \{robert.langendoerfer, markus.hillemann, markus.ulrich\}@kit.edu}
}
\begin{document}
\maketitle
\input{sec/0_abstract} 
\input{sec/1_intro}
\input{sec/02_related_work}
\input{sec/03_method}
\input{sec/04_experiments}

\input{sec/05_conclusion}
\FloatBarrier
\input{sec/llm_usage_statement}
{
    \small
    \bibliographystyle{ieeenat_fullname}
    \bibliography{egbib2}
}
\input{sec/X_suppl}

\end{document}

%% file: preamble.tex
\usepackage{cuted} 

\usepackage{currfile} 

\usepackage{caption} 

%% file: acronyms.tex
\begin{acronym}
\acro{3dgs}[3DGS]{3D Gaussian Splatting}
\acro{2dgs}[2DGS]{2D Gaussian Splatting}
\acro{nvs}[NVS]{Novel View Synthesis}
\acro{psnr}[PSNR]{Peak Signal to Noise Ratio}
\acro{ssim}[SSIM]{Structural Similarity Index Measure}
\acro{nerf}[NeRF]{Neural Radiance Field}
\acroplural{nerf}[NeRFs]{Neural Radiance Fields} 
\acro {sh}[SH]{Spherical Harmonic}  
\acro {gt}[GT]{Ground Truth}
\acro{GAN}[GAN]{Generative Adversarial Network}
\acro{adc}[ADC]{Adaptive Density Control}
\acro{sfm}[SfM]{Structure from Motion}
\end{acronym}

%% file: sec/0_abstract.tex
\begin{abstract}
Joint 3D object reconstruction and 6D object pose estimation from a single image is an inherently ill-posed problem. In industrial settings, however, multiple instances of an object are often randomly arranged in bins, implicitly providing several views of the same object within a single image. We leverage this implicit multi-view geometry to jointly reconstruct the object in 3D and estimate the 6D pose of each visible object instance. 
To this end, we present \method, a new Gaussian-Splatting-based approach that inverts the original Gaussian Splatting formulation: instead of rendering a single scene from multiple cameras, we render multiple object instances from a single camera. Our method relies on an instance segmentation and subsequent Structure from Motion initialization. In contrast to learned monocular depth estimation, we perform true geometry-based reconstruction from image evidence, mitigating hallucinations caused by training data priors, which is crucial in industrial applications that require reliable and verifiable geometric measurements.
We evaluate \method on synthetic and real bin-picking scenarios, and demonstrate accurate reconstruction of previously unseen objects as well as reliable pose estimation of individual instances. The source code is available at \url{https://pimpilimpo.github.io/projects/SIMILR/}.
\end{abstract}

%% file: sec/1_intro.tex
\section{Introduction}
\label{sec:intro}
\begin{figure}

    \centering
    \begin{subfigure}[b]{0.5\linewidth}
        \includegraphics[width=\textwidth,trim={12cm 6cm 12cm 2cm},clip]{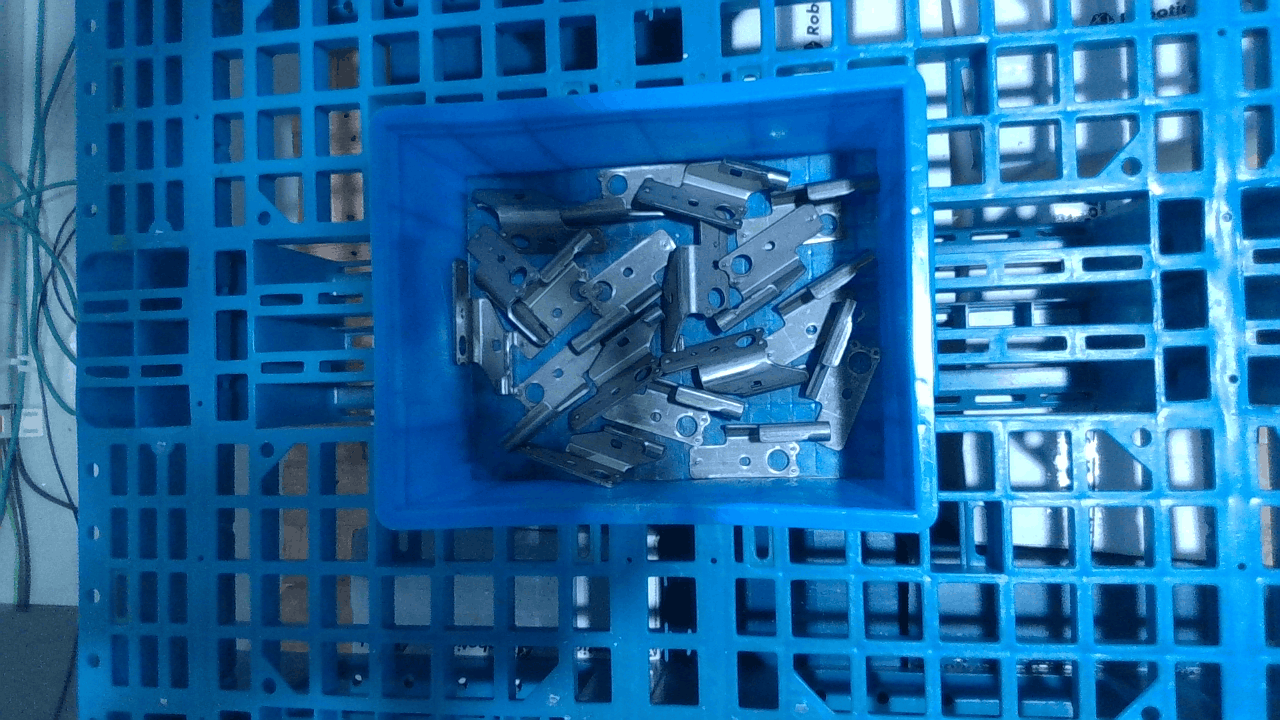}
        \caption{}
        \label{fig:problem_formulation1}
    \end{subfigure}\\
    \begin{subfigure}[b]{0.48\linewidth}
        \includegraphics[width=\textwidth,trim={12cm 6cm 12cm 2cm},clip]{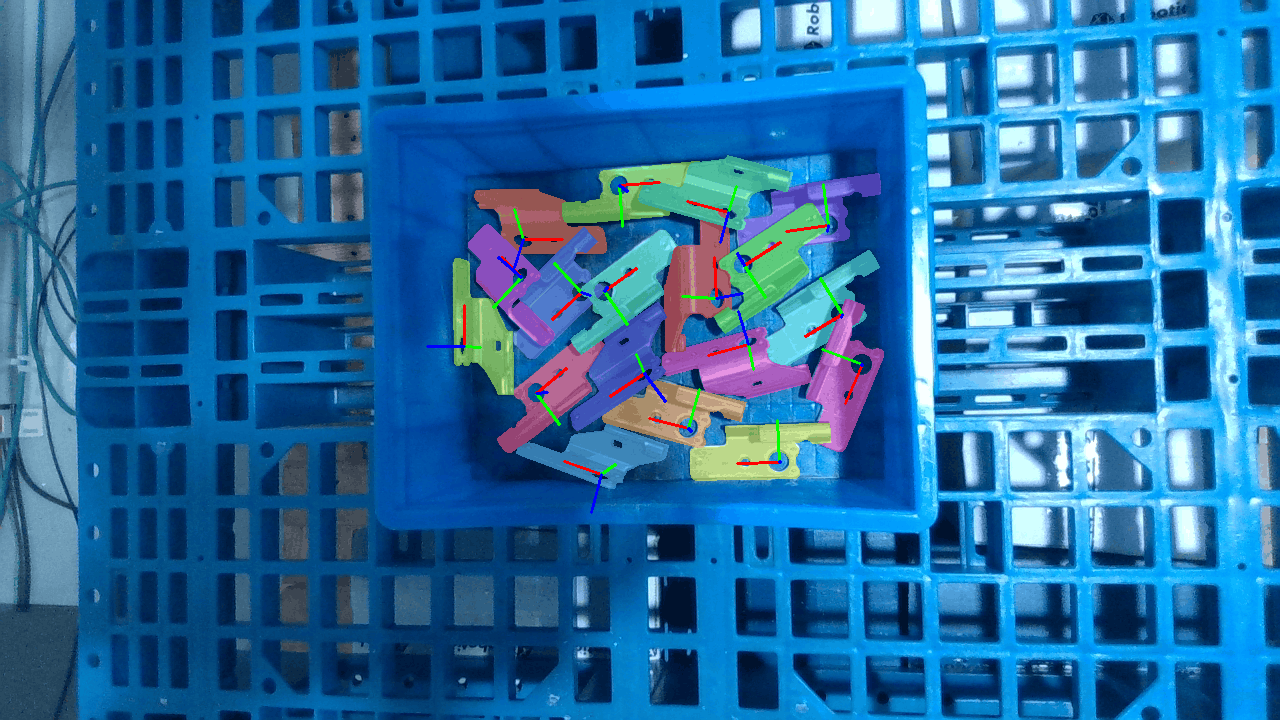}
        \caption{}
        \label{fig:problem_formulation2}
    \end{subfigure}
    \begin{subfigure}[b]{0.48\linewidth}
    
        \raisebox{0.5cm}{\includegraphics[width=\textwidth,trim={1cm 0cm 1cm 0cm},clip]{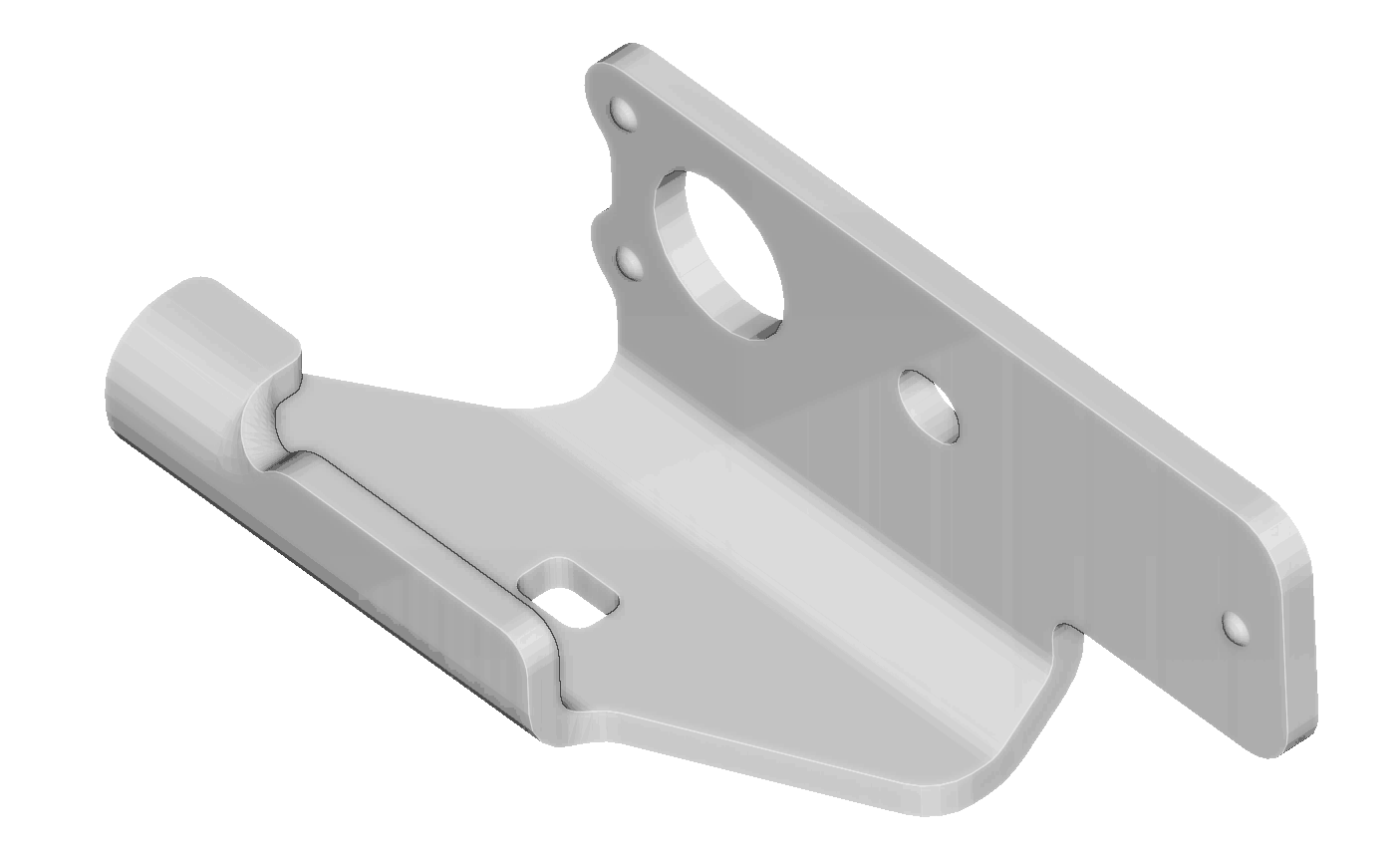}}
        \caption{}
        \label{fig:problem_formulation3}
    \end{subfigure}
    \caption{Visualization of the \task task. Given a multi-instance image, i.e., a single image showing multiple instances of the same object from different perspectives (a), the task is to determine the 6D object poses along with the masks (b) of all instances, and a 3D reconstruction of the canonical object (c). The multi-instance image shown is part of the XYZ-IBD dataset \cite{huangXYZIBDHighprecisionBinpicking2025}.}
    \label{fig:problem_formulation}
\end{figure}
Recovering 3D object geometry and 6D poses from visual data is a fundamental problem in computer vision. 
However, inferring 3D geometry and object pose from a single image is inherently ill-posed, as depth information is not directly observed. Existing single-image reconstruction methods attempt to address this ambiguity either through learned priors or generative models \cite{wang2025vggt,teamSAM3D3Dfy}. Though effective for common scenes, which are well represented in training data, these approaches often generalize poorly and risk hallucinating geometry when image cues are scarce \cite{langendorferIndustryrelevantBenchmarkDataset2025}. This is especially critical in industrial vision, where underrepresented objects demand strictly evidence-based reconstructions.

As previously pointed out \citep{NEURIPS2023_3764a9c8, zhang2023seeing}, many practical scenarios provide an underexplored source of geometric information that can alleviate this ambiguity. For example, in industrial applications, images often contain multiple recurrent instances of the same object, observed under different poses, as illustrated in \cref{fig:problem_formulation}. We call such images \textit{multi-instance images} throughout this paper. Multi-instance images arise naturally in applications like bin-picking, bulk material inspection, or inventory analysis. Because all instances share similar geometry and appearance, they implicitly provide multiple views of the same object within a single image. Inspired by this observation, we exploit such implicit multi-view observations to simultaneously reconstruct object geometry and estimate the 6D pose of each instance. Given sufficient pose diversity across instances, different object surfaces become observable without requiring hallucination, enabling a geometry-driven reconstruction from actual image evidence. 
We refer to the task of zero-shot 3D object reconstruction and pose estimation from a single RGB image with multiple instances as \task in the following.
We propose Single-Image Multi-Instance joint 3D object reconstruction and pose estimation Leveraging Recurrence (\method), a reformulation of conventional Gaussian Splatting (GS) \cite{kerbl20233d} tailored for the \task task. Unlike the conventional GS approach, which relies on multiple images of the same scene, \method requires only a single multi-instance image. 
By assuming that all object instances share the same geometry and radiometry, only a single object needs to be reconstructed, which we refer to as the canonical object. The canonical object is placed and rendered at the poses of the object instances, and both the object's canonical geometry and instance poses are jointly optimized. Our main contributions are:
\begin{itemize}
    \item We reformulate \task as a GS-based joint optimization problem, representing multiple object instances as duplicated, transformed observations of a shared canonical object within a common scene.
    \item Based on this formulation, we introduce \method, a framework to jointly refine the canonical object geometry and all instance poses, given segmentation masks,  coarse pose estimates, and camera intrinsics from a single multi-instance image.
    \item \method outperforms the previous SOTA-method \they \citep{NEURIPS2023_3764a9c8}, while requiring a fraction of the computation time.
    \item We demonstrate that explicitly modeling recurrences can refine monocular foundation-model predictions, improving upon Marigold V2 \cite{pavlovic2026marigold} predictions.
\end{itemize}

%% file: sec/02_related_work.tex
\section{Related Work}
\label{sec:Related work}
\paragraph{GS and 3D Reconstruction.}
In recent years, 3D-GS ~\cite{kerbl20233d} has continuously been improved, and numerous extensions focus especially on the quality of the generated 3D reconstruction \citep{huang2DGaussianSplatting2024,chen2024pgsr,zhuGaussianSplattingDiscretized2025}. Subsequent adaptations incorporate inverse rendering  or "in the wild" approaches \cite{zhangGaussianWild3D2024a, kulhanek2024wildgaussians,kaleta2025lumigauss,sarioSphericalVoronoiDirectional2025} into the original pipeline. They enable the object's appearance to be  modeled separately from the scene illumination. Some GS methods focus on the 3D reconstruction of single objects viewed from multiple camera poses \cite{yangGaussianObjectHighQuality3D2024,roggeObjectCentric2DGaussian2025}, but only a few GS methods explore the prior of repetitive objects. Notably, Splat and Replace \cite{violanteSplatReplace3D2025} and InstGS \cite{lu2026instgs} construct each individual instance from a template geometry with instance specific offset parameters from multiple camera poses. Instead, \method focuses on 3D reconstruction and object pose estimation of multiple similar instances from a \textit{single} multi-instance image. Furthermore, \method is able to model differences in appearance among the instances by inverse rendering, instead of using instance specific offset parameters, which struggle with strong differences caused, e.g., by specular reflections \cite{violanteSplatReplace3D2025}. All mentioned GS approaches are based on multi-view images. In contrast,  3D reconstruction and pose estimation from a single multi-instance image is much more challenging as the image background cannot be used for pose estimation, and hence substantially less information is available to solve the task.
\paragraph{6D Object Pose Estimation.} 6D object pose estimation approaches can be broadly categorized into model-based \cite{caraffa2024freeze,ausserlechner2024zs6d} and model-free approaches \cite{liu2025one2any}. Model-based approaches require an accurate 3D model, e.g., a CAD model, of the object as prior knowledge, which makes their practical application inflexible and costly. Model-free approaches either utilize multiple images \cite{liu2022gen6d,he2022onepose++} or create a 3D model for subsequent model-based approaches in a generative manner \cite{lee2025any6d}. In contrast to \method, these methods are not able to leverage the information of a multi-instance image. Recent work uses RGB-D input to recover relative poses between object instances in multi-instance scenes \cite{khodabandehloo2026prior}. 
\paragraph{Single-Image Inverse Rendering.} Inverse Rendering from a single image is inherently ill-posed \cite{yu2019inverserendernet,Sengupta_2019_ICCV}. Nevertheless, recent methods leveraging foundation models, such as Marigold V2 \cite{pavlovic2026marigold} achieve impressive results. However, these approaches neither utilize an explicit multi-instance prior nor estimate object poses. Furthermore, hallucinations due to training data priors cannot be excluded.
\paragraph{Multi-Shot Joint 3D Reconstruction and Object Pose Estimation from a Single Multi-Instance Image.} The methods presented in \cite{wuUnsupervisedJoint3D2019,cherianInSeGANGenerativeApproach2021} output a 3D reconstruction and object poses of a single multi-instance depth image at inference, however, they use multiple multi-instance images showing the same object to train the model. Similarly, \cite{liSimtoRealObjectRecognition2022} relies on a known 3D model to generate synthetic multi-instance images, which are used for training.
\method differs from the discussed methods in two key aspects: First, it operates on RGB images alone, without relying on additional depth information. This poses a greater challenge for 3D reconstruction, as radiometry and geometry must be disentangled from the same observation. At the same time, this reduces hardware requirements and broadens the range of potential applications. Second, it is a zero-shot method, while previous methods require some sort of additional prior such as the 3D model or training images. 
\paragraph{Zero-Shot Joint 3D Reconstruction and Object Pose Estimation from a Single Multi-Instance Image.} \citet{abbeloos20173d} detects instances in multi-instance images by matching keypoints within recurrent regions in a single RGB-D image. Therefore, objects need to be well textured, and only a sparse representation is created. In contrast, \method can handle textureless objects after initialization, does not require depth as input, and produces a dense 3D reconstruction.
The method presented in \cite{zhang2023seeing} learns probabilistic object intrinsics such as the geometry and albedo of instances of an object class, and environmental extrinsics such as 6D pose and scene illumination. For initialization, the method requires a coarse specification of the expected pose distribution (e.g., a plane for objects that stand on a table) and instance segmentation masks as priors. \they \cite{NEURIPS2023_3764a9c8} consists of a modified SfM initialization stage and subsequently trains geometry, visibility, and material composition in three consecutive stages. Like \cite{zhang2023seeing}, \they models object intrinsics and environmental extrinsics, albeit in a non-probabilistic manner. \they parametrizes the object surface with a Signed Distance Field (SDF) represented as an MLP \cite{wang2021neus}. As \they uses a stage-wise procedure, errors in the first stage propagate to later stages. Both methods \cite{NEURIPS2023_3764a9c8,zhang2023seeing} are computationally intensive, with optimization of the neural representations for a single image requiring several hours. Furthermore, both do not perform any refinement of the initial, coarse object poses, thereby limiting the accuracy.

\method uses the SfM-based initialization of \they, but differs in several aspects. Next to the geometry, \method also refines the initial object poses. It estimates these quantities jointly with illumination, avoiding error propagation across different stages. Furthermore, \method uses a reformulated GS representation rather than an SDF-based neural surface model, reducing computation times from several hours to several minutes. 
Finally, a fundamental difference lies in the scene representation. \they reformulates the problem as a multi-view reconstruction task by interpreting duplicate objects as virtual observations of a single canonical object. Samples along each ray are mapped and evaluated in the canonical object frame using the instance's object pose of each visible mask, while in background regions, a random instance is used. Since each ray evaluates only one instance, the instances are never combined into a joint 3D scene. Therefore, errors in geometry and object poses are not rendered consistently.
In contrast, \method learns a canonical object representation that is instantiated multiple times within one global scene frame. Consequently, all instances are rendered jointly in one image, so that errors in geometry and object poses are consistently reflected in the rendered image and, hence, are directly punished by our optimization.

%% file: sec/03_method.tex
\section{Methodology of \method}
\label{sec:methodology}
\begin{figure*}[tbp]
    \centering
    \setcounter{subfigure}{0}
    \begin{tikzpicture}
        \node[anchor=south west, inner sep=0] (image) at (0,0) {
       \includegraphics[width=\textwidth]{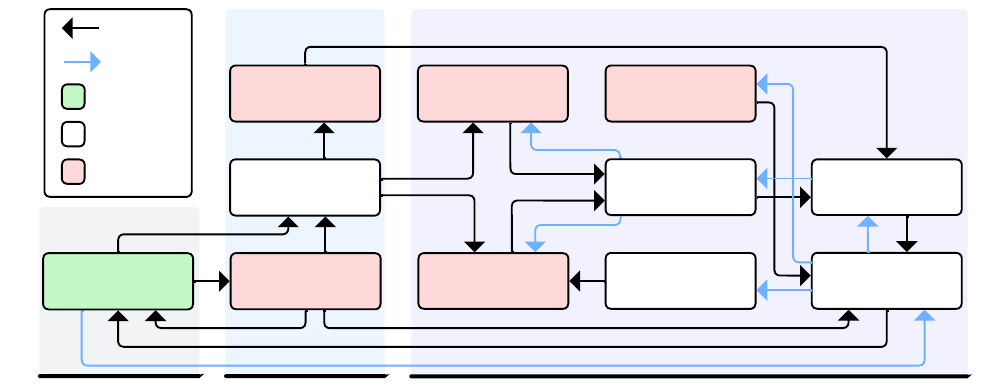}
        };
        \begin{scope}[x={(image.south east)}, y={(image.north west)}]
            \node at (0.305,0.27){\footnotesize Masks $\Matrix{M}^{}_{}$};
            \node at (0.45, 0.67) {\footnotesize };
            \node[anchor=west, align=left, text width=3cm] at (0.1, 0.925) {\footnotesize Operation\\[-5pt] Flow};
            \node[anchor=west, align=left, text width=3cm] at (0.1,0.8355){\footnotesize Gradient\\[-5pt] Flow};
            \node[anchor=west, align=left, text width=3cm] at (0.1, 0.745) {\footnotesize Input};
            \node[anchor=west, align=left, text width=3cm] at (0.1,0.65){\footnotesize Processes};
            \node[anchor=west, align=left, text width=3cm] at (0.1,0.55){\footnotesize Output};
            \node[align=center,  text width=3cm] at (0.305, 0.52) {\footnotesize SfM \\[-5pt] Reconstruction};
            \node at (0.49,0.76){\footnotesize Object Poses $\mathcal{P}$};
            \node[ align=center, text width=3cm] at (0.49,0.27){\footnotesize Canonical Model\\[-5pt] $\mathcal{G}_\mathrm{c}$};
            \node at (0.678, 0.76) {\footnotesize Illumination $\mathcal{U}$};
            \node[ align=center, text width=3cm] at (0.678, 0.27) {\footnotesize Adaptive Density\\[-5pt] Control};
            \node[ align=center, text width=3cm] at (0.884, 0.27) {\footnotesize Tile Based\\[-5pt] Rasterizer};
            \node at (0.884, 0.52) {\footnotesize Projection};
            \node[align=center,  text width=3cm] at (0.678, 0.52) {\footnotesize Scene \\[-5pt]Composition};
            \node at (0.305, 0.76) {\footnotesize Interior Orientation $\mathcal{C}_1$};

            \node at (0.12, 0.27) {\footnotesize Input Image $\Matrix{I}^{}$};
        \end{scope}
        \vspace{35pt}
 
        \node[above= -10pt] at ($(image.south west)!0.12!(image.south east)$)
            {\footnotesize (a\alph{subfigure})};

 
        \node[above=-10pt] at ($(image.south west)!0.305!(image.south east)$)
            {\footnotesize (b\alph{subfigure})};


        \node[above=-10pt] at ($(image.south west)!0.678!(image.south east)$)
            {\footnotesize (c\alph{subfigure})};
    \end{tikzpicture}
\caption{Overview of \method. (a) is the input, (b) the initialization (adopted from \they \cite{NEURIPS2023_3764a9c8}), and (c) the reformulated GS pipeline.}
    \label{fig:method_overview}
\end{figure*}

    \cref{fig:method_overview} gives an overview of \method. The multi-instance image is passed without any other priors to the initialization phase, which generates initial segmentation masks of the object instances in the image, an initial point cloud of the canonical object, and initial poses of the object instances. These initial values are subsequently refined within our GS pipeline (\cref{sec:multi_instance_rendering}), which is our core contribution. Conceptually, \method places each object instance at its pose in 3D space and renders all instances at once in a global coordinate frame, while optimizing the geometry of a single canonical object. This allows to model occlusions and scene lighting correctly, while exploiting the maximum amount of data per iteration, which -- as a byproduct -- leads to a smoother convergence than conventional GS.

\subsection{Preliminary}
\label{sec:preliminary}
\citet{zhuGaussianSplattingDiscretized2025} show that the lighting effects need to be untangled from the geometry, in order to generate an accurate 3D reconstruction with GS. Therefore, our method builds upon a reflection-based Spherical Voronoi (SV) parametrization \cite{sarioSphericalVoronoiDirectional2025}, which is based on 2D-GS \cite{huang2DGaussianSplatting2024}. Reflection-based SV is a deferred rendering novel view synthesis approach that represents a scene using 2D Gaussian primitives. 2D Gaussians are beneficial for accurate 3D reconstruction and deferred rendering, as they generate more reliable normals and object surfaces than 3D Gaussian primitives \cite{huang2DGaussianSplatting2024}: The set of 2D Gaussians in the scene is denoted by $
\mathcal{G}=\{\Point{p}_i,\Matrix{R}_i,\Matrix{S}_i,\alpha_i,\Vector{d}_i,r_i\}_{i=1}^N,
    \label{eq:canonical_object}$
with position $\Point{p}_i$, rotation $\Matrix{R}_i$, scaling along the rotation axis $\Matrix{S}_i$, opacity $\alpha_i$, diffuse albedo $\Vector{d}_i$, roughness $r_i$, and the number of Gaussians $N$. Reflection-based SV decomposes the geometry and material properties represented in $\mathcal{G}$ from the scene illumination, $
\mathcal{U}=\{\Matrix{U}_{\mathrm{far}},\Matrix{U}_{\mathrm{near}}\}
    \label{eq:notation_lighting}
$, with the far-field illumination $\Matrix{U}_{\mathrm{far}}$ 
parametrized as a cubemap and the near-field illumination $\Matrix{U}_{\mathrm{near}}$ parametrized as a set of light probes placed in the scene. We chose this representation because, particularly in close-range applications, the assumption of an infinitely distant light source is often violated due to light ray divergence and complex inter- and intra-object reflections. In GS, the set of $J$ cameras is represented as $
    {}^{\mathrm{O}}\mathcal{C}_{\mathrm{C}}=\{\Matrix{K}_j,\TMatrix{O}{R}{C,\mathit{j}},\TMatrix{O}{t}{C,\mathit{j}} \}_{j=1}^J,
    \label{eq:notation_cameras}
$ where $\Matrix{K}_j$ denotes the calibration matrix, $\TMatrix{O}{R}{C,\mathit{j}}$ the rotation, and $\TMatrix{O}{t}{C,\mathit{j}}$ the  translation from the Camera Coordinate System (CCS) to the Object Coordinate System (OCS). 
By perspective projection, $\mathcal{G}$ can be splatted onto the image space of ${}^{\mathrm{O}}\mathcal{C}_{\mathrm{C},j}$ \cite{huang2DGaussianSplatting2024,zwicker2002ewa}. Subsequently, the 2D Gaussians are rasterized in a geometry pass, to obtain per pixel diffuse color \cite{kerbl20233d,huang2DGaussianSplatting2024,sarioSphericalVoronoiDirectional2025}. The geometry pass also outputs 3D position, roughness $r$ and surface normal per pixel. In combination with the viewing direction and scene illumination $\mathcal{U}$ these values are used to calculate the specular component, which is added to the diffuse color to generate the final rendered image $\hat{\Matrix{I}^{}}$.
\subsection{Scene Representation}
\label{sec:theory}
\begin{figure}[tbp]
    \centering
    \begin{subfigure}[b]{0.1\textwidth}
        \centering
        \includegraphics[height=0.09\textheight]{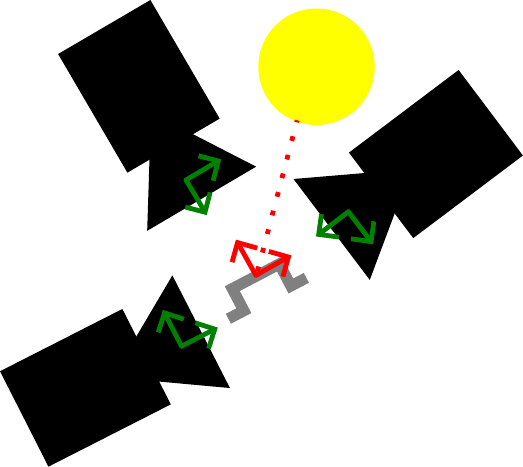}
        \caption{}
        \label{fig:gaussian_cases_1}
    \end{subfigure}
    \hfill
    \begin{subfigure}[b]{0.1\textwidth}
        \centering
        \includegraphics[height=0.09\textheight]{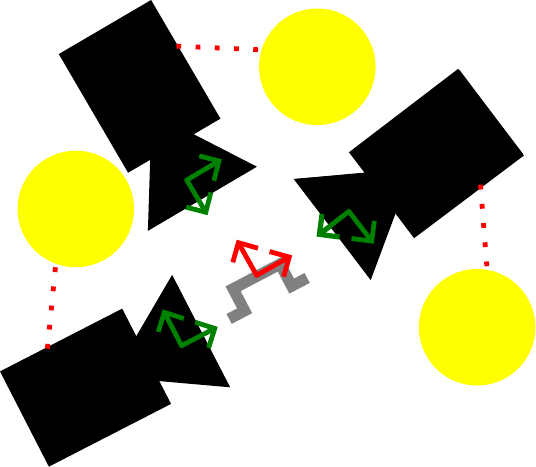}
        \caption{}
        \label{fig:gaussian_cases_2}
    \end{subfigure}
    \hfill
    \begin{subfigure}[b]{0.1\textwidth}
        \centering
        \includegraphics[height=0.09\textheight]{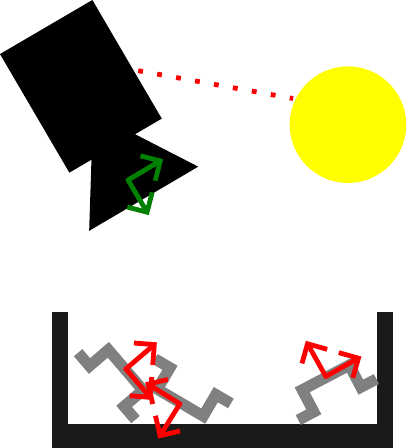}
        \caption{}
        \label{fig:gaussian_cases_3}
    \end{subfigure}
            \caption{Comparison of scene representations. The CCS is shown in green, the OCS in red, and the illumination is represented by a single light source for the sake of clarity, while the actual illumination can be more complex. (a) The conventional GS case: A single scene with scene-specific illumination (visualized by the dotted red line) and multiple observing views. (b) The naive solution with virtual camera poses: The spatial relation between object instances is lost and illumination is no longer fixed to the scene, but rather to the views. (c) \method : illumination is consistently modeled by rendering a single view with multiple object instances, preserving spatial relations.}
    \label{fig:gaussian_cases}
\end{figure}
The naive solution for 3D reconstruction and 6D object pose estimation from a multi-instance image with a GS backbone would be to adopt the conventional GS paradigm. As GS renders a single scene, the scene must consist of the Gaussians of a \textit{single} canonical object $\mathcal{G}_\mathrm{c}$. $\mathcal{G}_\mathrm{c}$ could be optimized by masking out all parts of the image except a single object instance, repeating this for all object instances, and rendering the image from $J$ virtual camera poses ${}^{\mathrm{O}}\mathcal{C}_{\mathrm{C},j}$, one for each object instance. This can be achieved via a masked photometric loss and a background loss \cite{yangGaussianObjectHighQuality3D2024}. In the following,  we refer to this approach as the \textit{naive} solution. \cref{fig:gaussian_cases_1} and \cref{fig:gaussian_cases_2} illustrate the scene composition and rendering of conventional GS and the naive solution with virtual camera poses, respectively. By introducing virtual camera poses, the illumination is assumed to move together with the virtual camera instead of remaining fixed in the scene, resulting in a camera-collocated light source analogous to flash photography \cite{namPracticalSVBRDFAcquisition2018,luanUnifiedShapeSVBRDF2021}. 
Furthermore, the naive solution loses the spatial relation between different instances, and occlusions are not handled by the background loss.\\
Therefore, instead of rendering \textit{one} canonical object from $J$ virtual cameras ${}^{\mathrm{O}}\mathcal{C}_{\mathrm{C},j}$, we place each of the $J$ object instances at its object pose ${}^{\mathrm{C}}\mathcal{P}_{\mathrm{O}}=\{\TMatrix{C}{R}{O,\mathit{j}},\TMatrix{C}{t}{O,\mathit{j}}\}_{j=1}^J$ in the global coordinate frame, and render \textit{all} instances from the single real camera pose ${}^{\mathrm{O}}\mathcal{C}_{\mathrm{C}}$. Conceptually, we invert the conventional GS approach with a single scene viewed from multiple cameras and instead model multiple scenes viewed from a single camera. This approach has several advantages: First, the scene illumination is modeled consistently, as depicted in \cref{fig:gaussian_cases_3}. This also allows to correctly model inter- and intra-object reflections, as the spatial relation between instances is preserved. 
Second, all instances are rendered simultaneously instead of sequentially, implicitly introducing multi-view constraints for a consistent geometry estimation. Third, in each iteration of our reformulated GS pipeline, the entire image with all object instances is utilized, ensuring smooth convergence. Conventional GS, in contrast, processes a different image per iteration, covering only a subset of the scene information, which often results in unstable convergence with pronounced fluctuations \cite{choi2025efficient}.

\subsection{Reformulated GS Pipeline}
\label{sec:multi_instance_rendering}

This section describes the core methodological contribution, i.e., the reformulated GS pipeline with all modifications necessary to adapt the SV \cite{sarioSphericalVoronoiDirectional2025} backbone to the \task task. 
\paragraph{Initialization. }Analogously to conventional GS pipelines \cite{kerbl20233d}, \method requires a sufficiently good initialization. An ablation study on the convergence radius of \method is conducted in the supplement. For a fair comparison, we choose the same SfM-based initialization method as in \they \cite{NEURIPS2023_3764a9c8}. It requires known visible instance segmentation masks $\Matrix{M}_{j}^{}$ beforehand, which are estimated by applying panoptic segmentation \cite{cheng2021mask2former}.

\paragraph{Scene Generation.} \method optimizes the Gaussians $\mathcal{G}_\mathrm{c}$ of a single canonical object, which is never rendered directly. Instead, it is transformed $J$ times to the object poses ${}^{\mathrm{C}}\mathcal{P}_{\mathrm{O,}j}$.
 After transforming $\mathcal{G}_\mathrm{c}$ to the set of object poses $\mathcal{P}$, the composed scene $\mathcal{G}_{J}$ can be described as
\begin{equation}
\mathcal{G}_{J}=\{\TMatrix{C}{R}{O,\mathit{j}}\Point{p}_i+\TMatrix{C}{t}{O,\mathit{j}},\TMatrix{C}{R}{O,\mathit{j}}\Matrix{R}_i,\Matrix{S}_i,\alpha_i,\Vector{d}_i,r_i,j \}_{i=1}^{N} \, {}_{j=1}^{J}.
    \label{eq:notation_all_gaussians}
\end{equation}
Note that each Gaussian in $\mathcal{G}_{J}$ is assigned an instance identifier $j$, enabling association of Gaussians with their respective instance. \cref{eq:notation_all_gaussians} motivates the use of a deferred rendering pipeline, such as SV \cite{sarioSphericalVoronoiDirectional2025}, as the parameters $\Vector{d}_i$ and $r_i$ do not depend on the view direction and can therefore directly be copied from $\mathcal{G}_\mathrm{c}$, just like $\Matrix{S}_i$ and $\alpha_i$. A color representation that depends on the view direction, such as \acp{sh} used in \cite{huang2DGaussianSplatting2024,kerbl20233d} fails to learn high-frequency color changes, as in \method, the \acp{sh} would encode an object-pose dependent color instead of a view-direction dependent color. The single static camera ${}^{\mathrm{O}}\mathcal{C}_{\mathrm{C}}$ is defined by the interior orientation obtained during initialization, while the extrinsics are set to $\TMatrix{O}{R}{C}=\Matrix{I}_{3\times3}$ and $\TMatrix{O}{t}{C}=\Vector{0}_{1\times3}$. For \method, we use only the near-field illumination $\Matrix{U}_{\mathrm{near}}$, as modeling both $\Matrix{U}_{\mathrm{near}}$ and $\Matrix{U}_{\mathrm{far}}$ from a single multi-instance image yields worse results (\cref{sec:evaluation}). In each training iteration in the forward pass, the newly generated scene $\mathcal{G}_{J}$ is rendered from ${}^{\mathrm{O}}\mathcal{C}_{\mathrm{C}}$, along with the learnable lighting $\Matrix{U}_{\mathrm{near}}$, resulting in the estimated image $\hat{\Matrix{I}^{}}$.  In the backward pass, only the parameters of ${}^{\mathrm{C}}\mathcal{P}_{\mathrm{O,j}}$, $\mathcal{G}_\mathrm{c}$, and $\Matrix{U}_{\mathrm{near}}$ receive gradient signals, ensuring consistency by design between object instances.
\paragraph{Object Pose Optimization.}  By optimizing object poses instead of camera poses,  ${}^{\mathrm{C}}\mathcal{P}_{\mathrm{O,j}}$ is never passed to the CUDA rasterizer and all object pose gradients are therefore calculated automatically via autograd \cite{paszke2017automatic}. We use the 6D representation \cite{hempel6dRotationRepresentation2022} for the parametrization of the object rotation, resulting in 9 parameters per object pose in total to be optimized. Pose optimization starts after a warm-up phase of 500 iterations, allowing for a coarse geometry to form beforehand.  Similar to \cite{schmidtLookGaussNo2024}, we employ a $L1$ loss on the mean opacity of all rendered Gaussians $\mathcal{L}_\mathrm{o}$, which not only speeds up the reconstruction, but also leads to more accurate object poses due to less ambiguities between object poses and object shape.

\paragraph{Photometric Loss.} We calculate the photometric loss $\mathcal{L}_\mathrm{p}$ as 
\begin{equation}
    \mathcal{L}_\mathrm{p} = (1-\lambda) \mathcal{L}_1(\Matrix{I}_{\mathrm{m}}^{},\hat{\Matrix{I}_{}^{}}) + \lambda \left(1 - \mathrm{SSIM}(\Matrix{I}_{\mathrm{m}}^{}, \hat{\Matrix{I}_{}^{}})\right),
    \label{eq:photometric_loss}
\end{equation}
with $\lambda=0.2$ \cite{kerbl20233d,yangGaussianObjectHighQuality3D2024}. 
\paragraph{Background Loss.} We use a Binary Cross Entropy loss (BCE) $\mathcal{L}_\mathrm{m}$ \cite{jadonSurveyLossFunctions2020} on the rendered instance opacity $\hat{\Matrix{M}}_{j}^{}$. We observed that calculating an instance-independent BCE over the union of $\Matrix{M}_{j}^{}$, as \they does, does not work for heavily occluded or intersecting object instances. Therefore, we instead calculate instance-wise opacity and BCE, encouraging instance geometry only in the corresponding mask:
\begin{equation}
\mathcal{L}_{\mathrm{m}}=\sum_{j=1}^J\big(-(\Matrix{M}_{j}^{}\odot\mathrm{log}(\hat{\Matrix{M}}_{j}^{})+(1-\Matrix{M}_{j}^{})\odot\mathrm{log}(1-\hat{\Matrix{M}}_{j}^{}) )\big).
    \label{eq:masking_alpha}
\end{equation}
This is achieved by rendering instance-wise opacity maps using the instance identifier $j$ of each Gaussian in $\mathcal{G}_{J}$. This assumes that the canonical object is fully opaque, leveraging the fact that opacity is accumulated only until saturation, and occlusions are caused by other instances.
\paragraph{Normal Consistency Loss.} Like \they, we use a normal consistency loss $\mathcal{L}_\mathrm{nc}$ \cite{NEURIPS2023_3764a9c8}, which calculates the dissimilarity between the rendered normals and a predicted normal image. We use Marigold V2 \cite{pavlovic2026marigold} as our monocular normal predictor.
\paragraph{Final Loss.} Finally, the model is optimized by minimizing the loss
    \begin{equation}
\mathcal{L}=\mathcal{L}_\mathrm{p}+\beta\mathcal{L}_\mathrm{m}+\gamma\mathcal{L}_\mathrm{n}+\delta\mathcal{L}_\mathrm{d}+\epsilon\mathcal{L}_\mathrm{nc}+\zeta\mathcal{L}_\mathrm{o},
    \label{eq:final_loss}
\end{equation}
where $\mathcal{L}_\mathrm{n}$ and $\mathcal{L}_\mathrm{d}$ are 2D-GS regularization terms \cite{huang2DGaussianSplatting2024}. For loss weights, adaptive density control, and further implementation details, see the supplement.

%% file: sec/04_experiments.tex
\section{Experiments}
\label{sec:experiments}
This section presents the experimental results of \method on diverse synthetic and real-world datasets. All experiments were conducted on a system with an NVIDIA RTX 4090 GPU with 24 GB VRAM. We train \method for 7000 iterations.  \cref{sec:metrics} introduces the evaluation metrics, while \cref{sec:datasets} describes the datasets. \cref{sec:baselines} presents the baseline methods used for comparison. \cref{sec:evaluation} reports the results for 3D reconstruction and object pose estimation. Finally, \cref{sec:limitations} discusses current limitations of \method. 
\subsection{Evaluation Metrics}
\label{sec:metrics}
\textbf{2D Evaluation.} For evaluation in 2D, we calculate the Mean angular error (MAE) for estimated normals and the Peak Signal to Noise Ratio (PSNR) \cite{wangImageQualityAssessment2004} for assessing the quality of the rendered image. We report inverse rendering results in the supplement as our focus is on 3D reconstruction and object pose estimation.\\\\
\textbf{Object Pose Evaluation.} We use the evo tool \cite{grupp2017evo} for comparing absolute object pose differences. We report Mean Translation Error (MTE) and Mean Rotation Error (MRE) \cite{langendorferIndustryrelevantBenchmarkDataset2025}. We align the estimated object poses to the \ac{gt} coordinate frame using a similarity transformation, as object poses are not metric, which is estimated via the Umeyama algorithm \cite{umeyama1991least}.\\\\
\textbf{Mesh Extraction and 3D Evaluation.} To evaluate the 3D reconstruction result of the canonical object, we extract its mesh following standard practice \cite{huang2DGaussianSplatting2024,sarioSphericalVoronoiDirectional2025}. Similar to \they \cite{NEURIPS2023_3764a9c8}, we calculate the Chamfer distance $\mathrm{d_c}$ \cite{barrow1977parametric} between the reconstructed mesh and the \ac{gt} mesh after a coarse similarity transformation and a subsequent rigid iterative closest point \cite{besl1992method} transformation for refinement.
\subsection{Datasets}
\label{sec:datasets}
\paragraph{Dup.} The Dup dataset \cite{NEURIPS2023_3764a9c8} consists of 6 real-world (Dup-real) and 9 synthetic (Dup-synthetic) multi-instance images. Objects are well textured and have both diffuse and specular surfaces. The number of instances per image ranges from five to ten with varying degrees of overlap. For a fair comparison, we choose the same SfM-based initialization method as in \they \cite{NEURIPS2023_3764a9c8}. Dup-real is only evaluated qualitatively due to lack of GT data.
\paragraph{XYZ-IBD.} Additionally, we applied all evaluated methods to a subset of four real-world multi-instance images of the industrial bin-picking XYZ-IBD dataset \cite{huangXYZIBDHighprecisionBinpicking2025}. Scenes in the dataset are characterized by severe occlusions, complex shapes, texturelessness, and mostly specular surfaces. For all images, accurate GT is available. However, the SfM-based initialization adopted from \they fails on XYZ-IBD images because of the insufficient texture. Therefore, we instead evaluate the performance in an oracle-style experiment. We initialize $\Matrix{M}^{}_{j}$ with the \ac{gt} visible masks. To the \ac{gt} object poses, we add rotation noise with a magnitude of 5\textdegree around a random axis and translation noise with a magnitude of 30\,mm in a random direction to the \ac{gt} object poses to simulate inaccuracies in the initialization. The geometry is initialized with 10k random points within a unit cube.
\subsection{Baselines}
\label{sec:baselines}
We compare \method against the following baselines: COLMAP MVS \cite{schonbergerPixelwiseViewSelection2016}, \they \cite{NEURIPS2023_3764a9c8}, and a 2DGS implementation of the naive solution described in \cref{sec:theory}. We do not compare against \cite{zhang2023seeing} because the approach cannot handle occlusions and \they \cite{NEURIPS2023_3764a9c8} already showed that \cite{zhang2023seeing} fails to handle the Dup data well. The original \they implementation used Omnidata \cite{eftekhar2021omnidata} monocular normals for the normal consistency loss. We additionally evaluate a modified \they version by replacing Omnidata with the stronger Marigold V2 \cite{pavlovic2026marigold} for fairness reasons, which we refer to as \they\unskip$^*$.

\subsection{Results}
\label{sec:evaluation}

\paragraph{Dup. }\method 3D reconstruction outperforms all other baselines (\cref{tab:chamfer_distance_dup}). With the same input data, our 3D reconstruction beats current SOTA \they\unskip$^*$ by more than 5\,mm on average. The naive solution exhibits the largest $\mathrm{d_c}$ values, primarily because the used background loss \cite{yangGaussianObjectHighQuality3D2024} does not model occlusions correctly. In occluded areas, the correct geometry is penalized as no amodal masks are available leading to wrong geometry in occluded regions. The masked photometric loss fails in a similar manner in occluded regions. The MVS 3D reconstruction suffers from known limitations \cite{topalouglu2026oraclegs}: In Dup, only very sparse and occluded observations are available, resulting in incomplete 3D reconstructions, especially in textureless areas.

\cref{tab:ape} reports MRE and MTE for \they and \method on Dup. As \they does not optimize object poses beyond initialization, the values of \they correspond to the initial object poses. In Dup, \method reduces MRE by a third and almost halves MTE on average compared to the initial values, which also represent the results of \they. Along with the photometric error, the mask geometry is a strong prior for the object pose estimation, explaining why the MRE is worst for objects with rotational symmetry, such as \emph{cleaner} and \emph{tin}, as their masks do not change significantly when rotated around the symmetry axis of the object.

\begin{figure}
    \centering
    \begin{subfigure}[b]{0.32\linewidth}
\includegraphics[width=\linewidth,trim={2cm 12cm 16cm 8cm},clip]{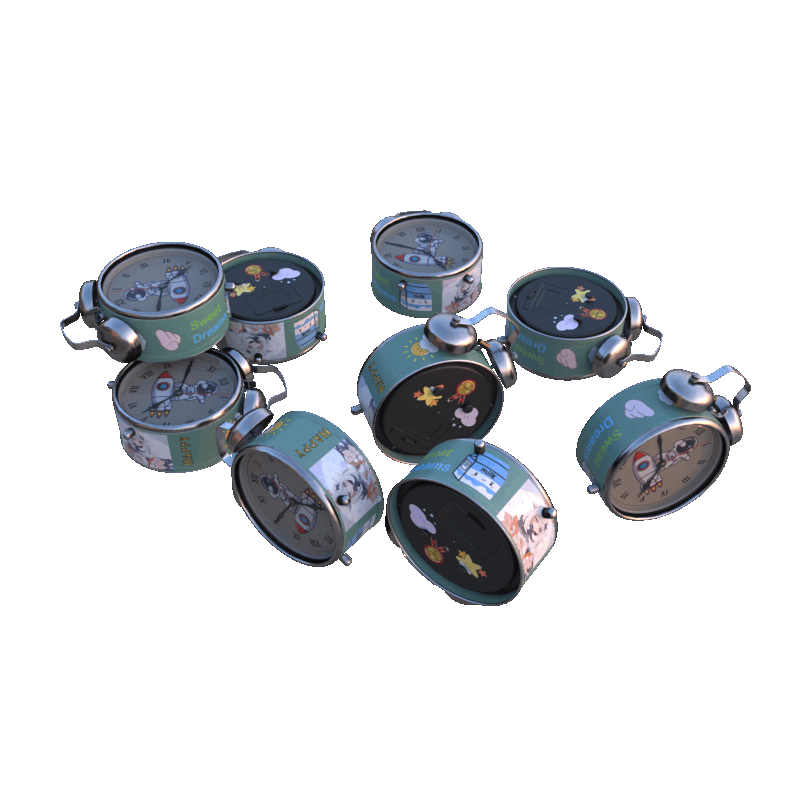}
        \begin{tikzpicture}[overlay,  shift={(0,0)}]
            \draw[color=red!60, very thick]
                (0.6\linewidth, 0.75\linewidth) circle (0.2); 
                    \draw[color=red!60, very thick]
                (0.25\linewidth, 0.25\linewidth) circle (0.2);
                    \draw[color=red!60, very thick]
                (0.59\linewidth, 0.225\linewidth) circle (0.2); 
        \end{tikzpicture}
          \vspace{-12pt}
        \caption{}
        \label{fig:clock_1}
    \end{subfigure}
        \begin{subfigure}[b]{0.32\linewidth}
\includegraphics[width=\linewidth,,trim={2cm 12cm  16cm 8cm},clip]{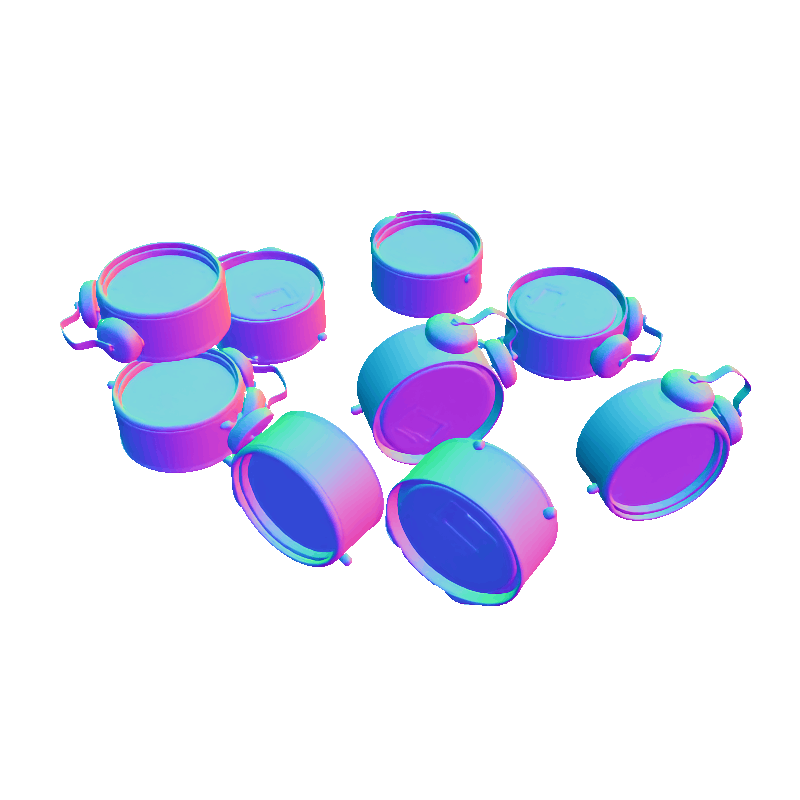}
                    \begin{tikzpicture}[overlay,  shift={(0,0)}]
            \draw[color=red!60, very thick]
                (0.6\linewidth, 0.75\linewidth) circle (0.2); 
                    \draw[color=red!60, very thick]
                (0.25\linewidth, 0.25\linewidth) circle (0.2);
                    \draw[color=red!60, very thick]
                (0.59\linewidth, 0.225\linewidth) circle (0.2); 
        \end{tikzpicture}
          \vspace{-12pt}
                       \caption{}
        \label{fig:clock_2}
    \end{subfigure}
        \begin{subfigure}[b]{0.32\linewidth}
\includegraphics[width=\linewidth,trim={2cm 12cm  16cm 8cm},clip]{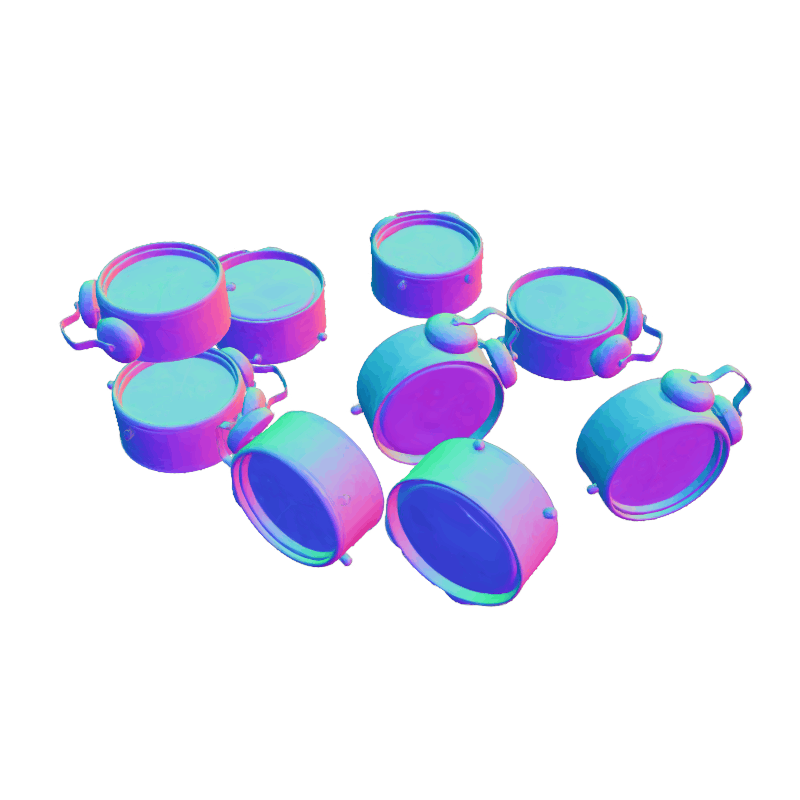}
                    \begin{tikzpicture}[overlay,  shift={(0,0)}]
            \draw[color=red!60, very thick]
                (0.6\linewidth, 0.75\linewidth) circle (0.2); 
                    \draw[color=red!60, very thick]
                (0.25\linewidth, 0.25\linewidth) circle (0.2);
                    \draw[color=red!60, very thick]
                (0.59\linewidth, 0.225\linewidth) circle (0.2); 
        \end{tikzpicture}
          \vspace{-12pt}
                            \caption{}
        \label{fig:clock_3}
    \end{subfigure}
    \caption{Close-up of scene \emph{clock} from Dup-synthetic (a). Hallucination effects lead to the omission of parts of the clock (see red circles) for Marigold V2 (b). In contrast, \method (c) successfully recovers the complete object geometry, despite utilizing the normal image generated by Marigold V2, by leveraging consistency across multiple instances.}
    \vspace{-15pt}
    \label{fig:todo}
\end{figure}

\cref{tab:normal_angular_error} compares the rendered normal images. For all scenes in Dup, we are able to refine the Marigold V2 result, while \they does not improve upon the Marigold V2 normals in all cases. \cref{fig:todo} provides an example of  hallucination effects of Marigold V2 that lead to the omission of object parts. Despite utilizing the normal image generated by Marigold V2, \method successfully recovers the complete object geometry, highlighting the benefit of enforcing consistency across multiple instances.

\begin{table*}[ht]
\begin{minipage}{0.62\textwidth}
\centering
\caption{Chamfer distance $\mathrm{d_c}$ in [mm]$\downarrow$ on Dup-synthetic. $^*$ marks that the monocular normal predictor was changed to Marigold V2 \cite{pavlovic2026marigold}. Best results are bold. }
\label{tab:chamfer_distance_dup}
\resizebox{\linewidth}{!}{
\begin{tabular}{lcccccccccc}
\hline
  Object & box & cash & cleaner & clock & coffee & fire & guitar & sign & tin & Mean \\
\hline
  Naive & 17.0 & 23.5 & 67.8 & 17.9 & 24.3 & 44.1& 81.4 & 28.3 & 173.1 & 53.1 \\
  MVS & 43.0 & 19.4 & 12.9 & 75.4 & 37.8 & 14.1 & 10.0 & 7.6 & 5.9 & 25.1\\
  \they \cite{NEURIPS2023_3764a9c8}  & 15.0 & 30.9 & 6.3 & 12.4 & 22.4 & 18.6 & 7.4 & 19.6 & 3.8 & 15.1 \\ 
  \they\unskip$^*$ & 8.2 & 25.6 & \textbf{4.0} & \textbf{8.1} & 20.0 & 15.3 & 5.6 & 8.9 & 2.0 & 10.8 \\
  \method & \textbf{3.0} & \textbf{9.5} & 5.2 & \textbf{8.1} & \textbf{11.4} & \textbf{5.6} & \textbf{2.0 } & \textbf{3.2} & \textbf{1.9} & \textbf{5.5} \\

\hline
\end{tabular}
}
\end{minipage}
\hfill
\raisebox{0.17cm}{\begin{minipage}{0.36\textwidth}
    \centering
\caption{Chamfer distance $\mathrm{d_c}$ in [mm]$\downarrow$ on XYZ-IBD. $^*$ marks that the monocular normal predictor was changed to Marigold V2 \cite{pavlovic2026marigold}. Best results are bold.}
\label{tab:chamfer_distance_xyz}
\resizebox{\linewidth}{!}{
\begin{tabular}{lccccc}
\hline
 Object &XYZ1&XYZ2&XYZ3&XYZ4&Mean\\
\hline

  \they\unskip$^*$ & 16.9&15.2&13.7&22.0&17.0\\
  \method & \textbf{0.78}&\textbf{0.83}&\textbf{0.66}&\textbf{1.12}&\textbf{0.88} \\
\hline
\end{tabular}
}
\end{minipage}}
\end{table*}
 
\begin{figure*}[tbp]

    \centering
\rowlabel{1}%
\begin{subfigure}[c]{0.15\textwidth}
    \centering
    \includegraphics[width=\textwidth]{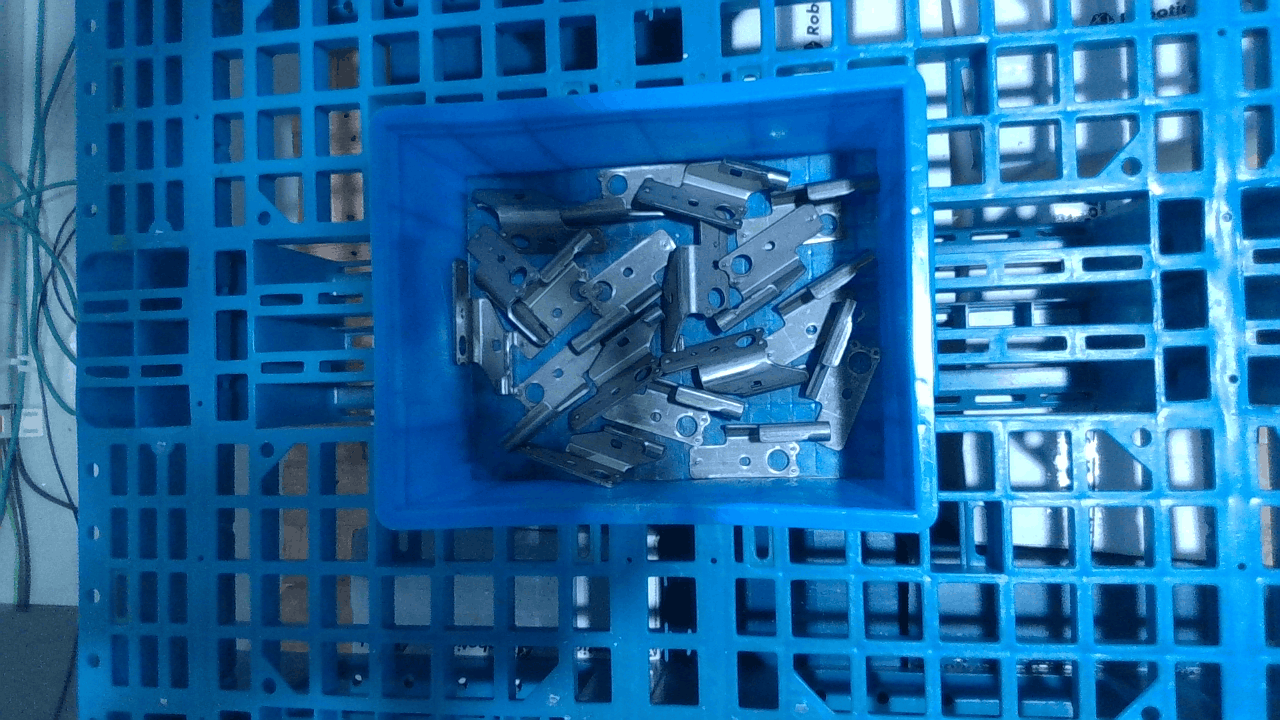}
\end{subfigure}\hfill
\begin{subfigure}[c]{0.15\textwidth}
    \centering
    \includegraphics[height=2cm,trim={16cm 8cm 13cm 4cm},clip]{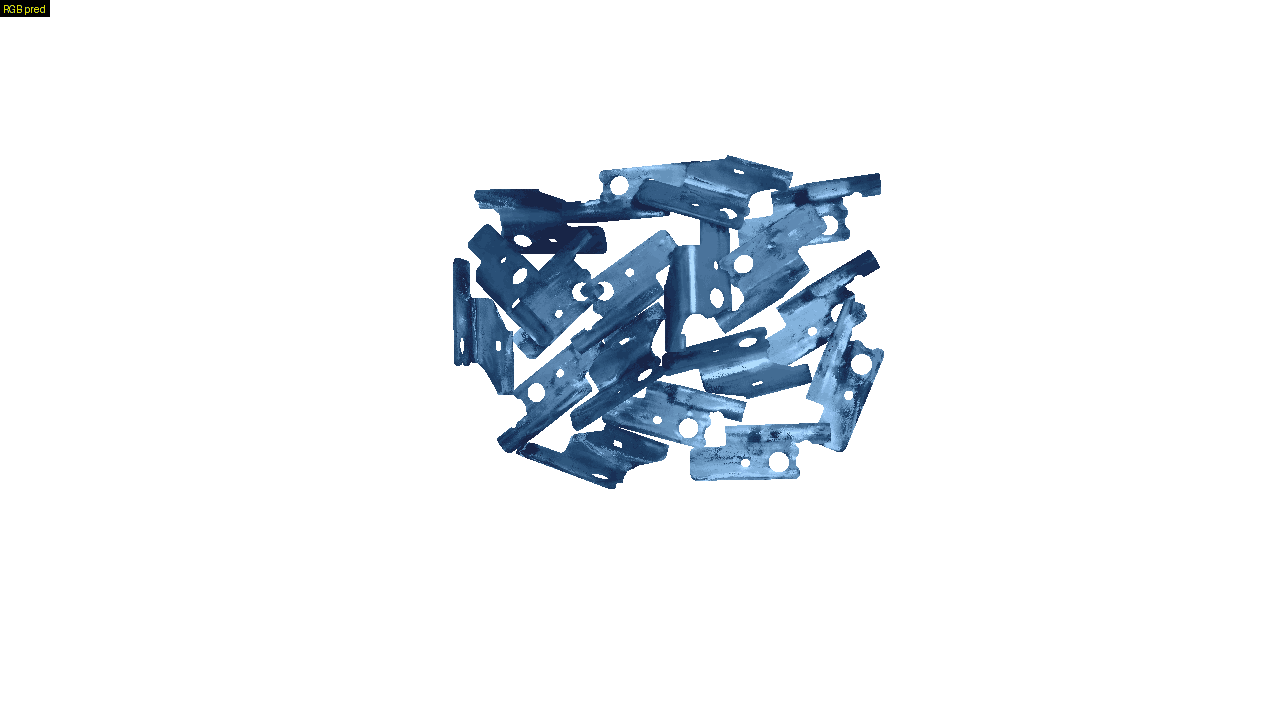}
\end{subfigure}\hfill
\begin{subfigure}[c]{0.15\textwidth}
    \centering
    \includegraphics[height=2cm,trim={16cm 8cm 13cm 4cm},clip]{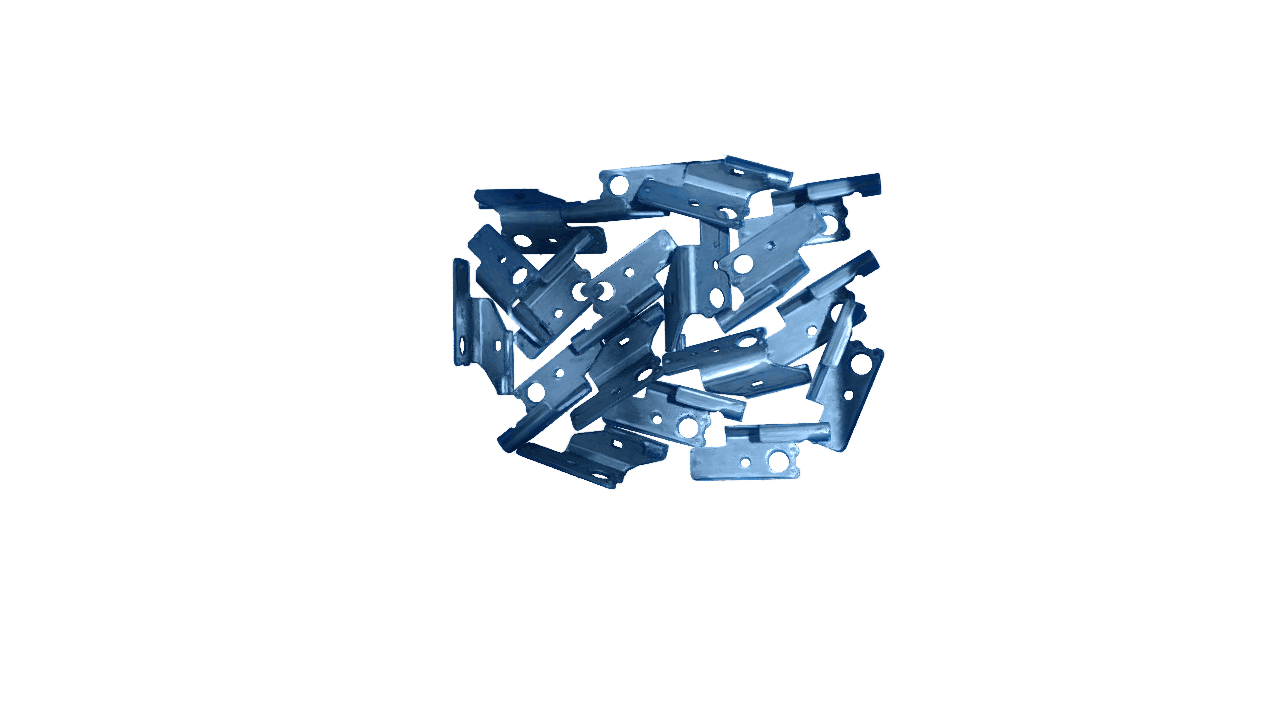}
\end{subfigure}\hfill
\begin{subfigure}[c]{0.15\textwidth}
    \centering
    \includegraphics[height=2cm]{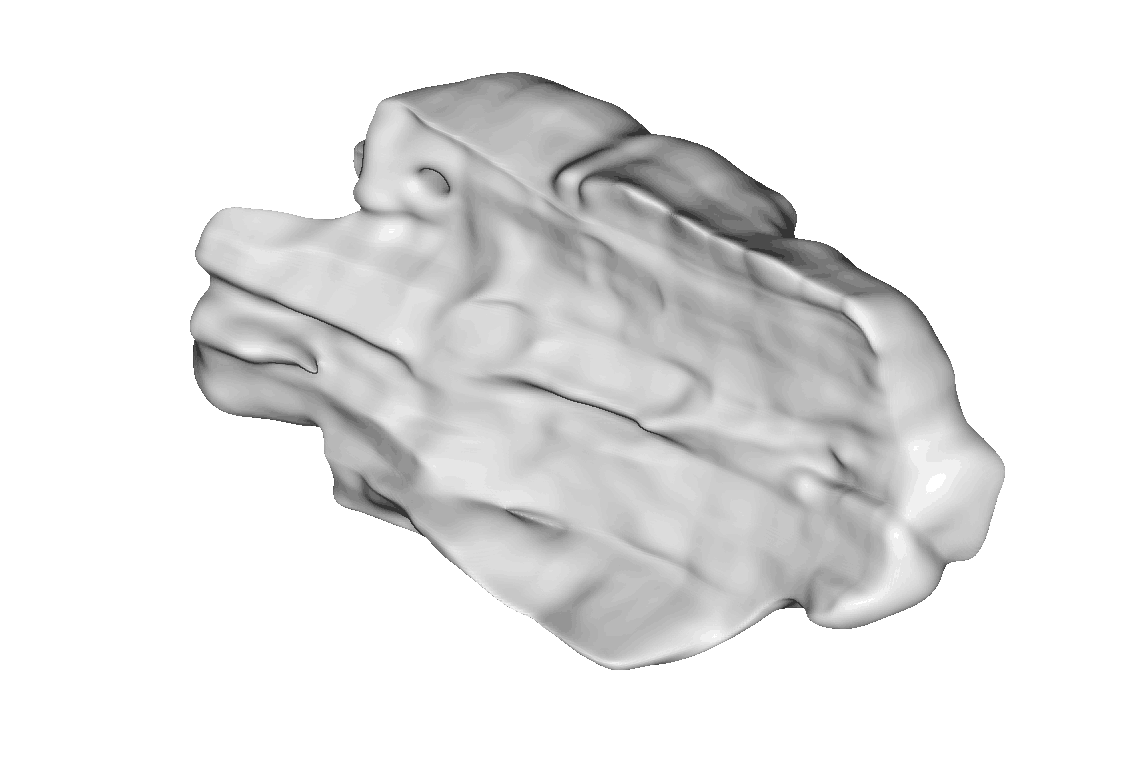}
\end{subfigure}\hfill
\begin{subfigure}[c]{0.15\textwidth}
    \centering
    \includegraphics[height=2cm]{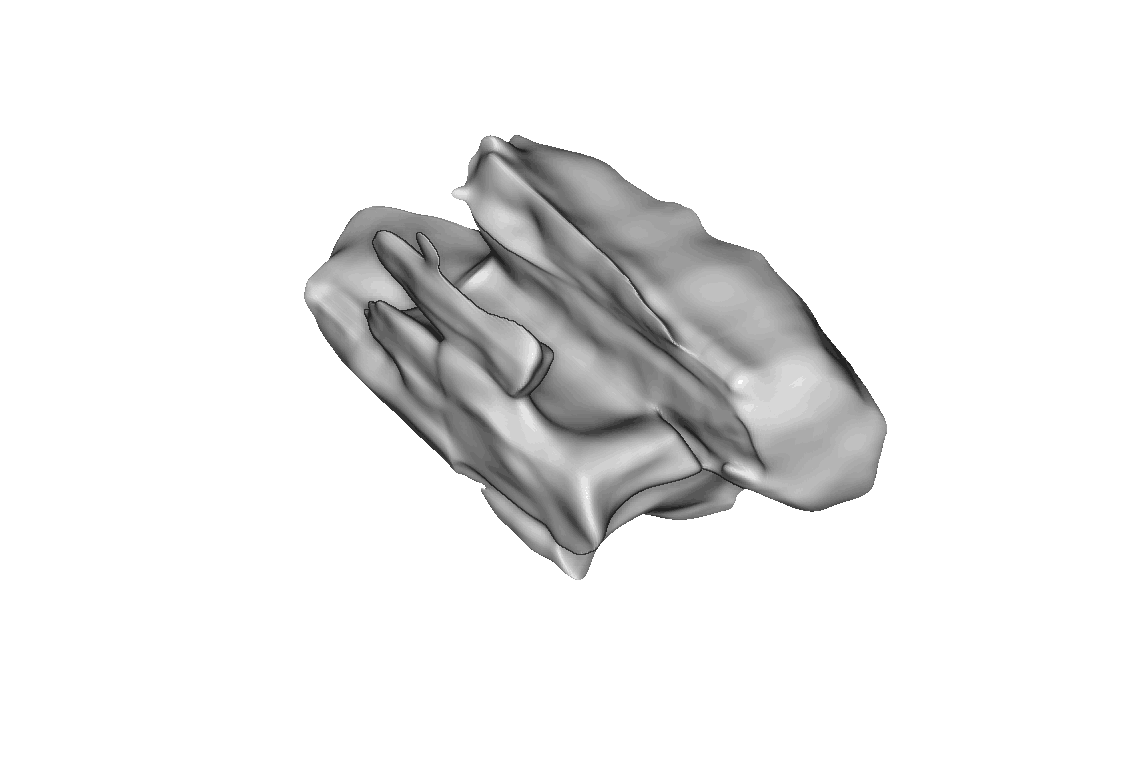}
\end{subfigure}\hfill
\begin{subfigure}[c]{0.15\textwidth}
    \centering
    \includegraphics[height=2cm]{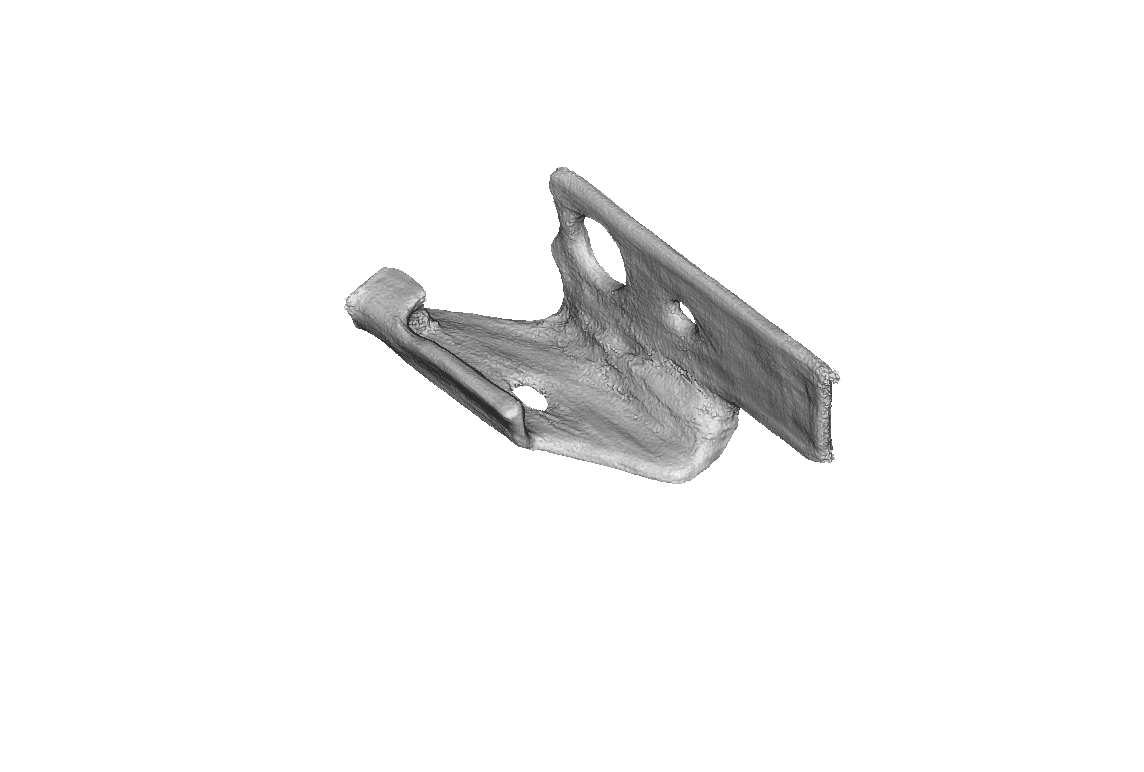}
\end{subfigure}\\
\rowlabel{2}%
\begin{subfigure}[c]{0.15\textwidth}
    \centering
    \includegraphics[width=\textwidth]{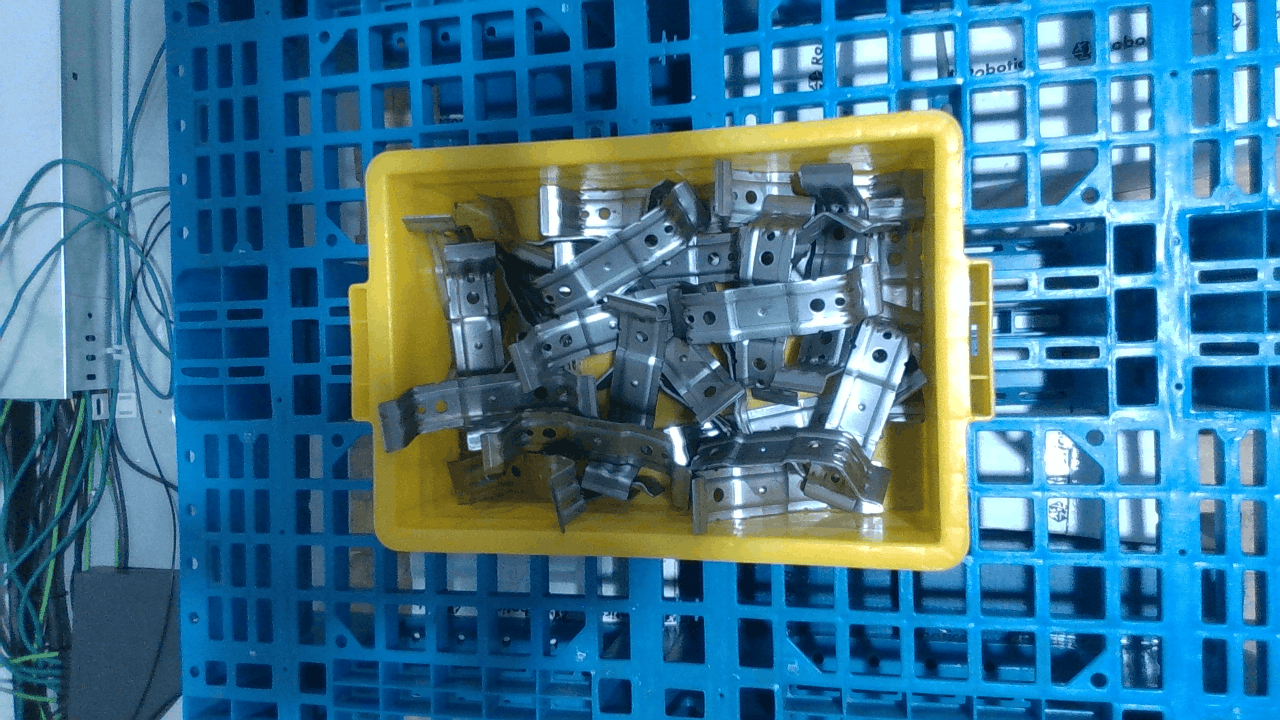}
\end{subfigure}\hfill
\begin{subfigure}[c]{0.15\textwidth}
    \centering
    \includegraphics[height=2cm,trim={14cm 6cm 12cm 4cm},clip]{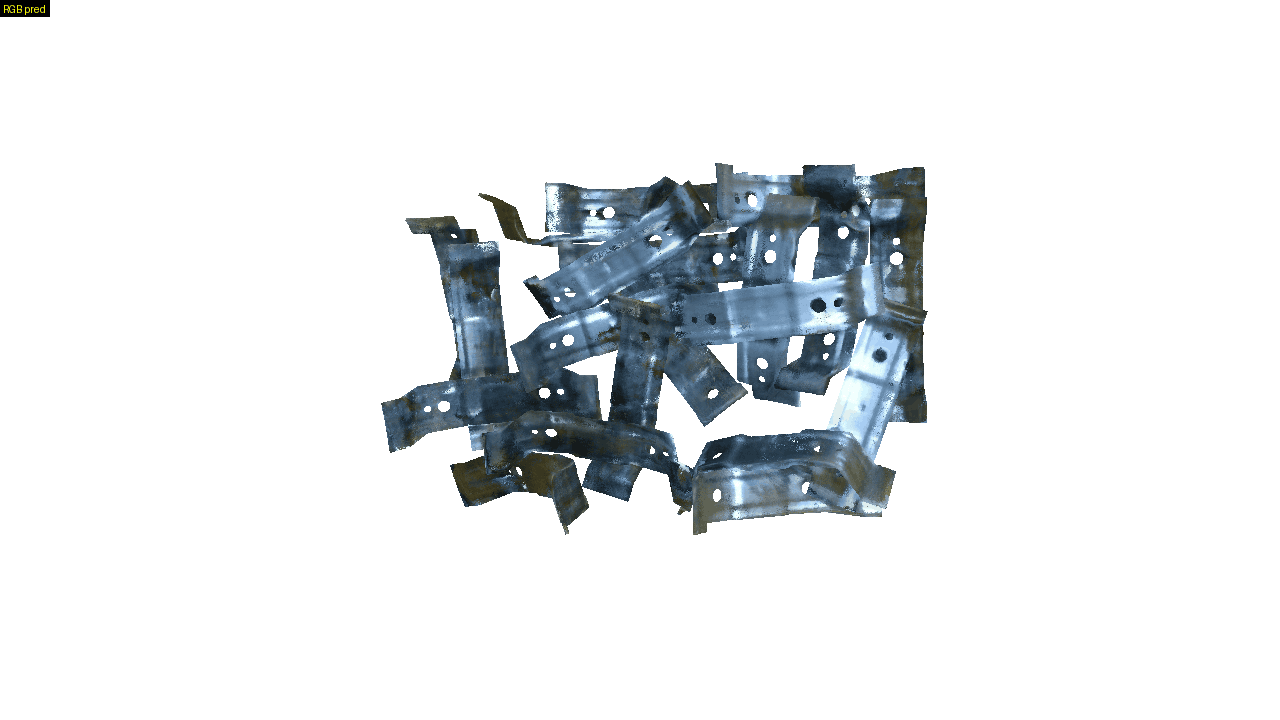}
\end{subfigure}\hfill
\begin{subfigure}[c]{0.15\textwidth}
    \centering
    \includegraphics[height=2cm,trim={14cm 6cm 12cm 4cm},clip]{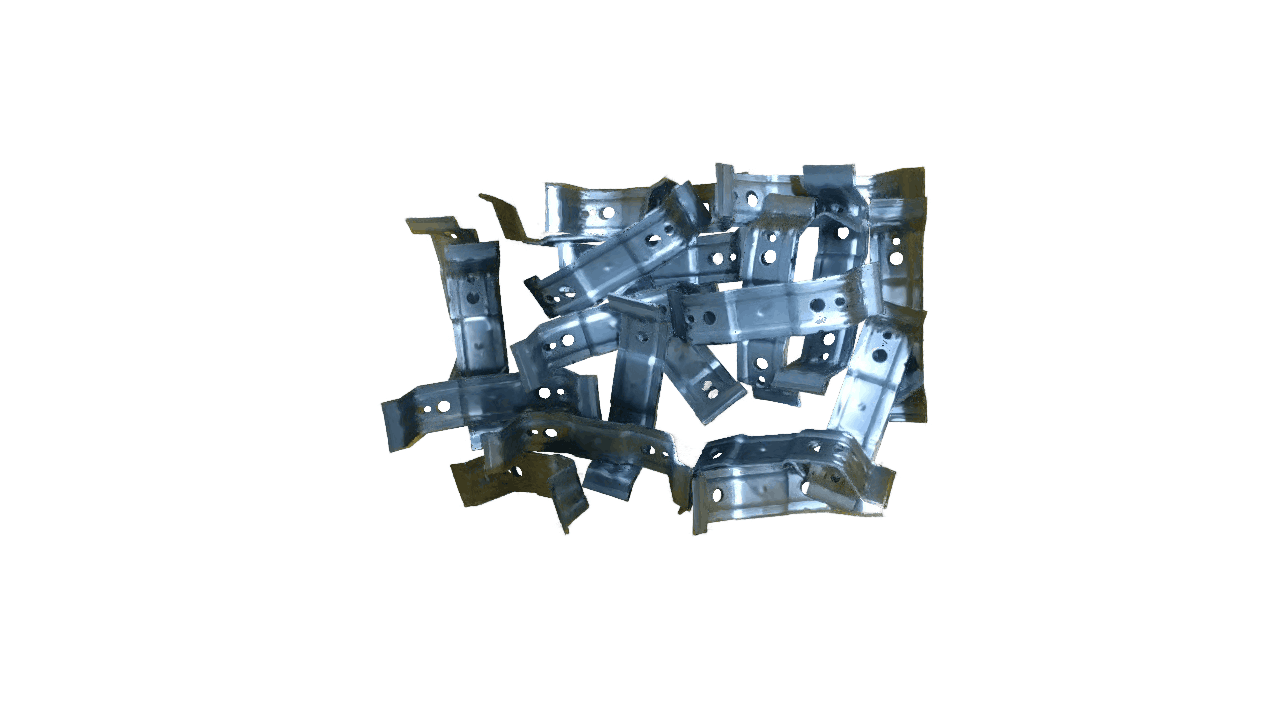}
\end{subfigure}
\hfill
\begin{subfigure}[c]{0.15\textwidth}
    \centering
    \includegraphics[height=2cm]{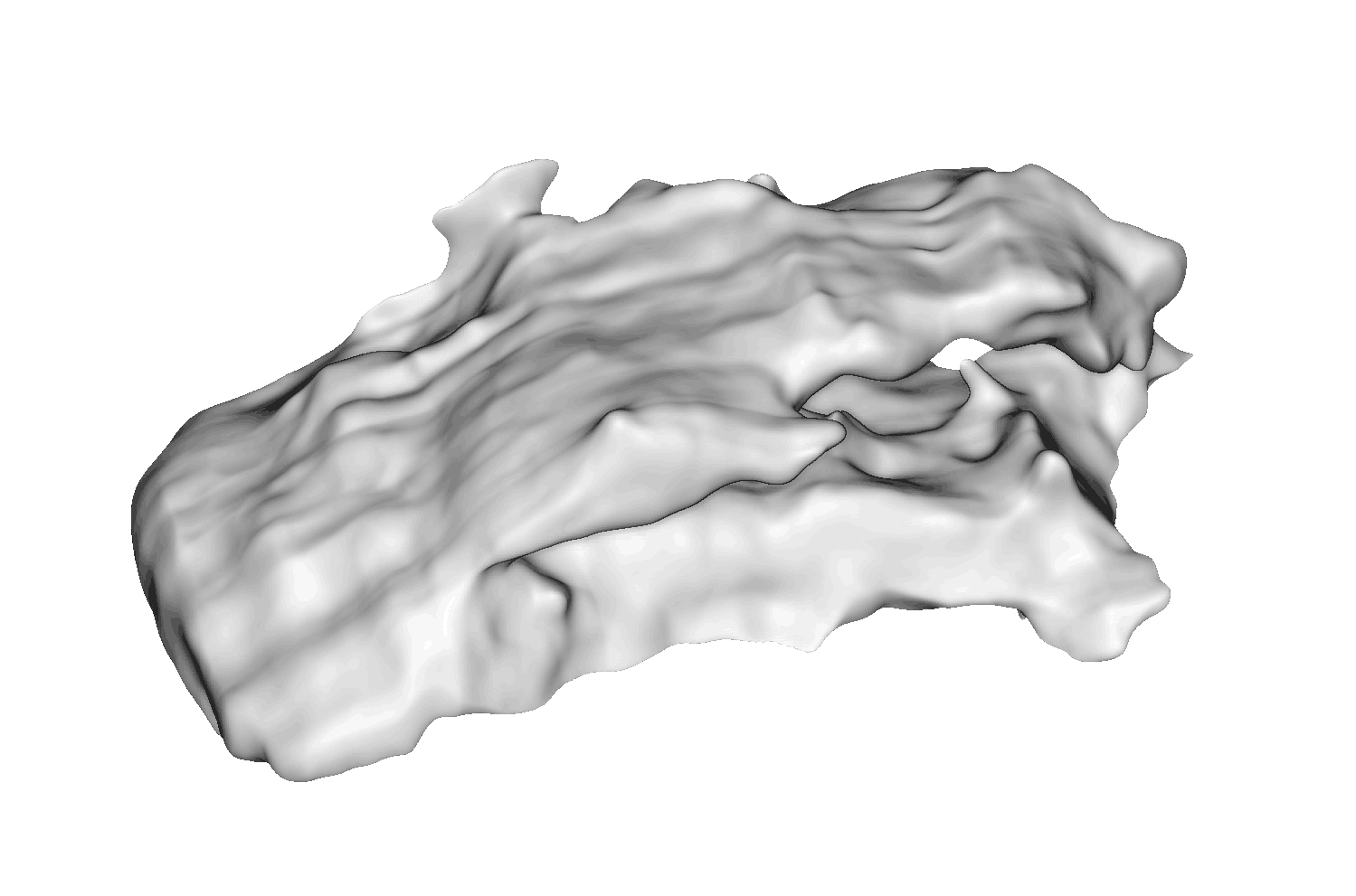}
\end{subfigure}\hfill
\begin{subfigure}[c]{0.15\textwidth}
    \centering
    \includegraphics[height=2cm]{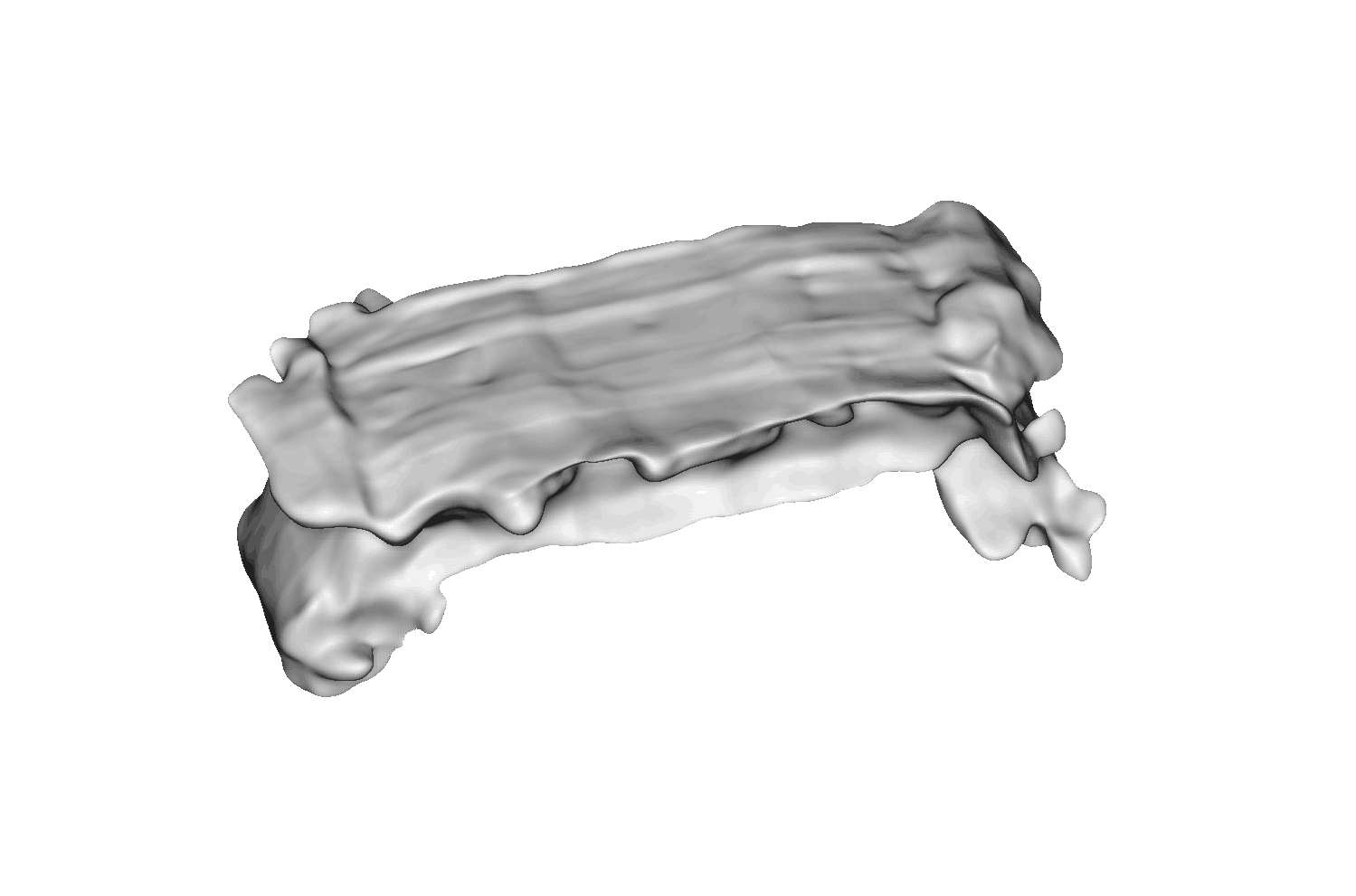}
\end{subfigure}\hfill
\begin{subfigure}[c]{0.15\textwidth}
    \centering
    \includegraphics[height=2cm]{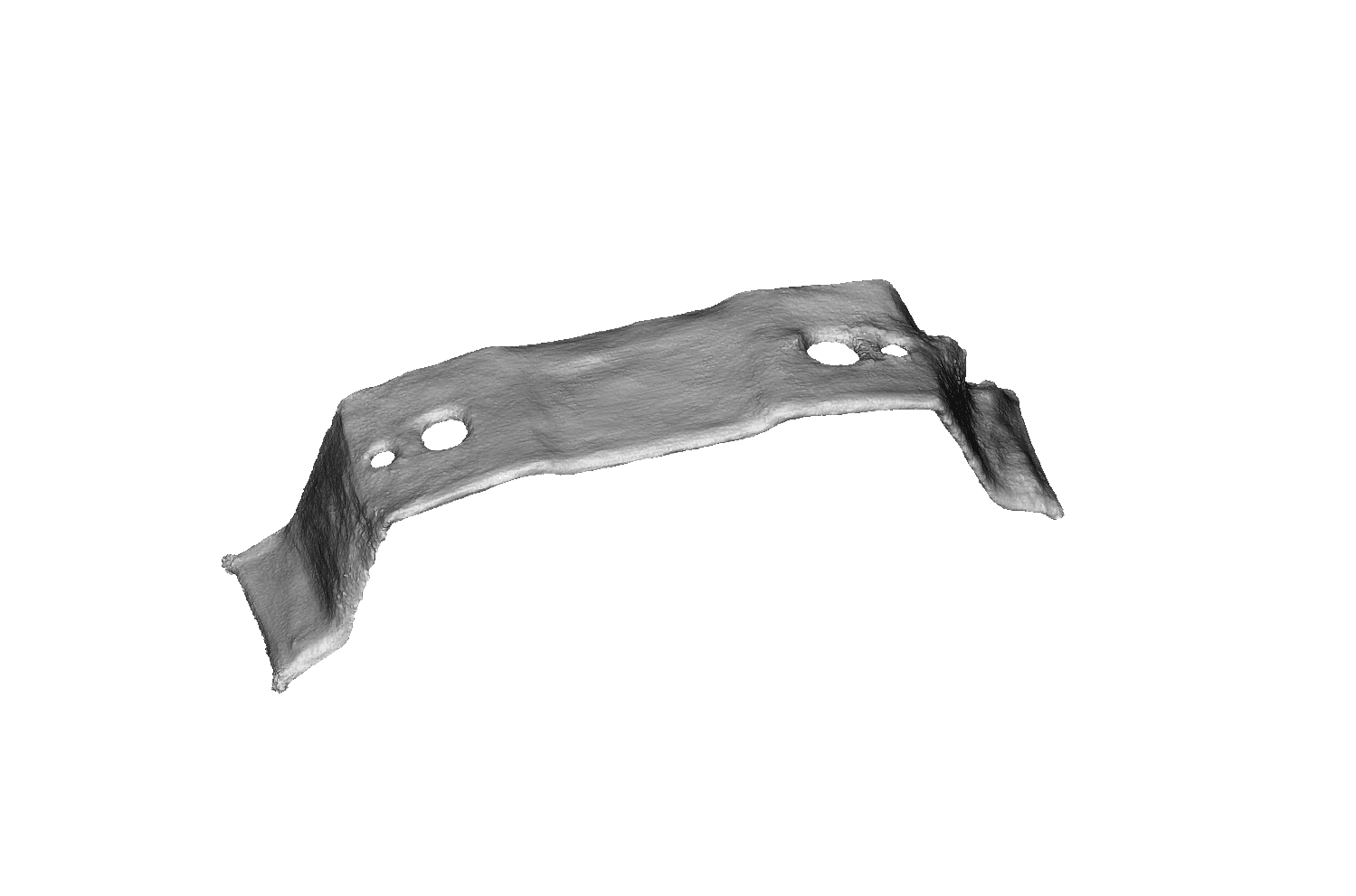}
\end{subfigure}\\
\rowlabel{3}%
\begin{subfigure}[c]{0.15\textwidth}
    \centering
    \includegraphics[width=\textwidth]{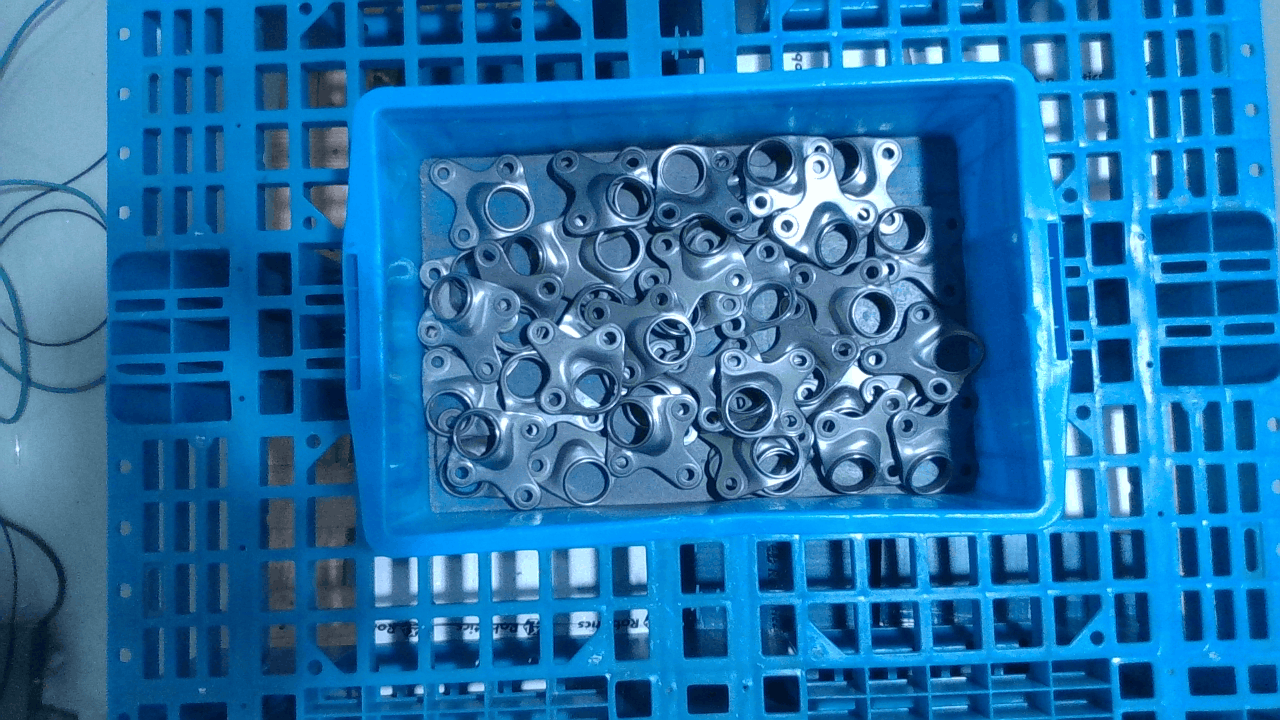}
\end{subfigure}\hfill
\begin{subfigure}[c]{0.15\textwidth}
    \centering
    \includegraphics[height=2cm,trim={14cm 6cm 7cm 4cm},clip]{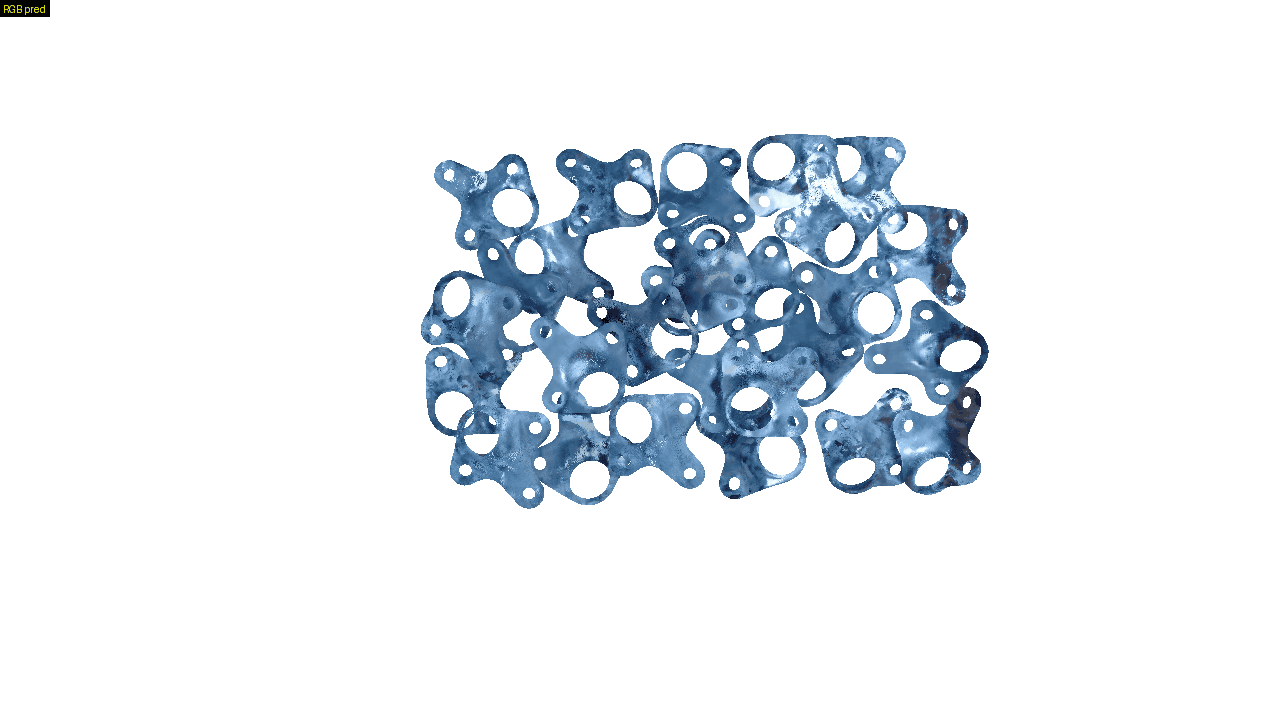}
\end{subfigure}\hfill
\begin{subfigure}[c]{0.15\textwidth}
    \centering
    \includegraphics[height=2cm,trim={14cm 6cm 7cm 4cm},clip]{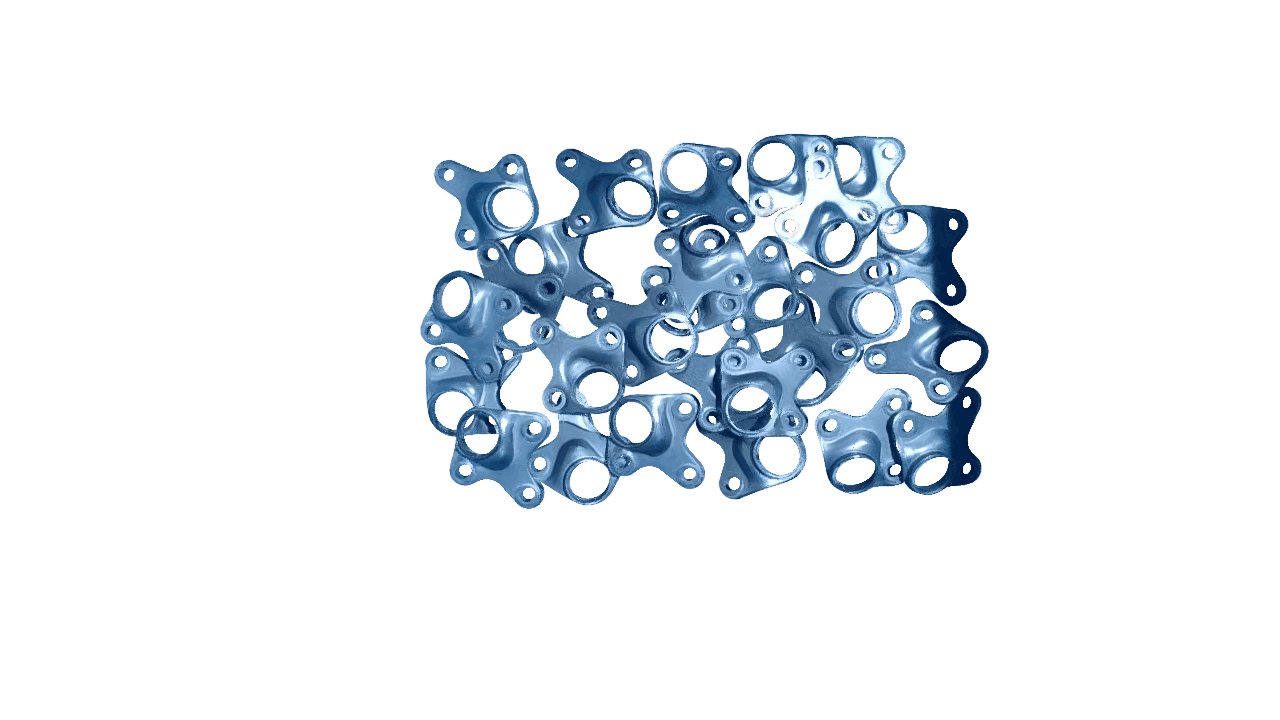}
\end{subfigure}\hfill
\begin{subfigure}[c]{0.15\textwidth}
    \centering
    \includegraphics[height=2cm]{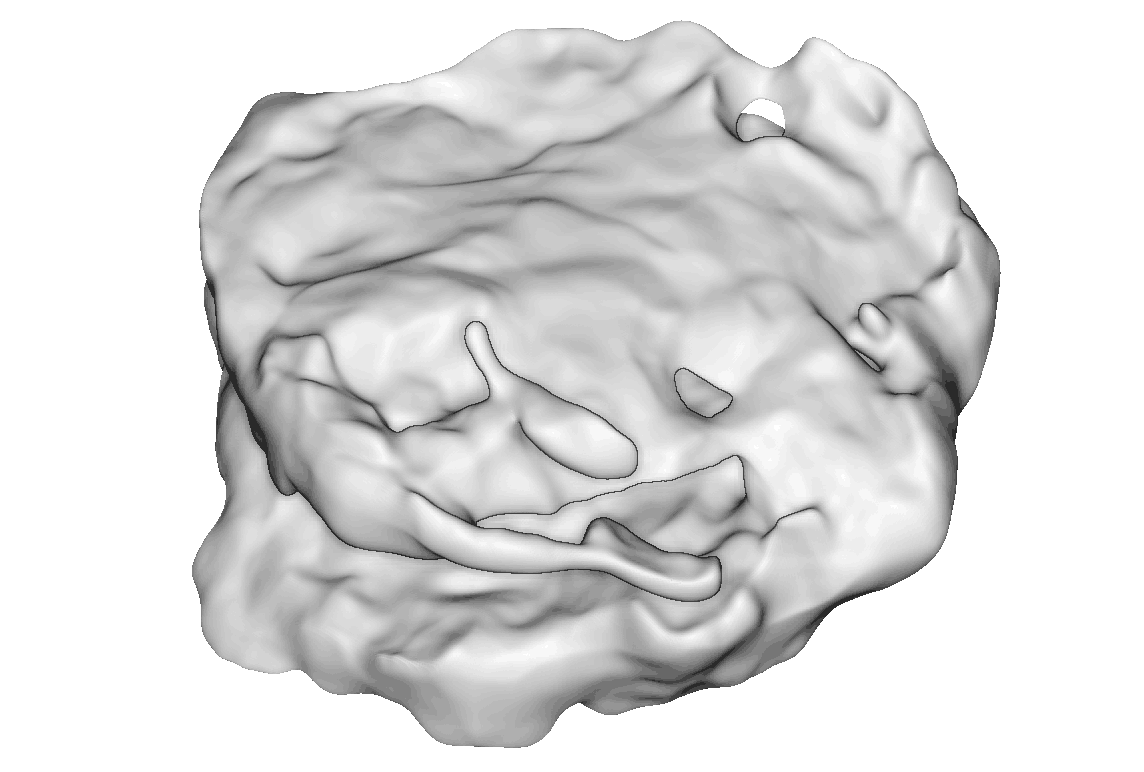}
\end{subfigure}\hfill
\begin{subfigure}[c]{0.15\textwidth}
    \centering
    \includegraphics[height=2cm]{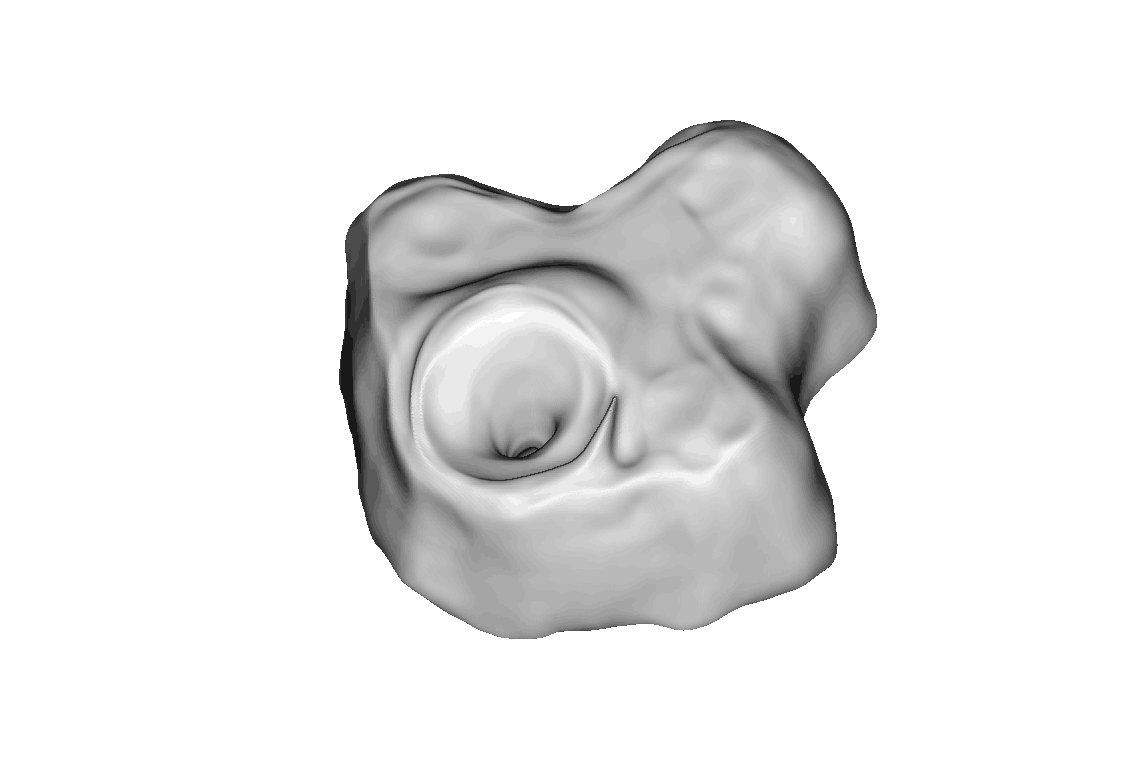}
\end{subfigure}\hfill
\begin{subfigure}[c]{0.15\textwidth}
    \centering
    \includegraphics[height=2cm]{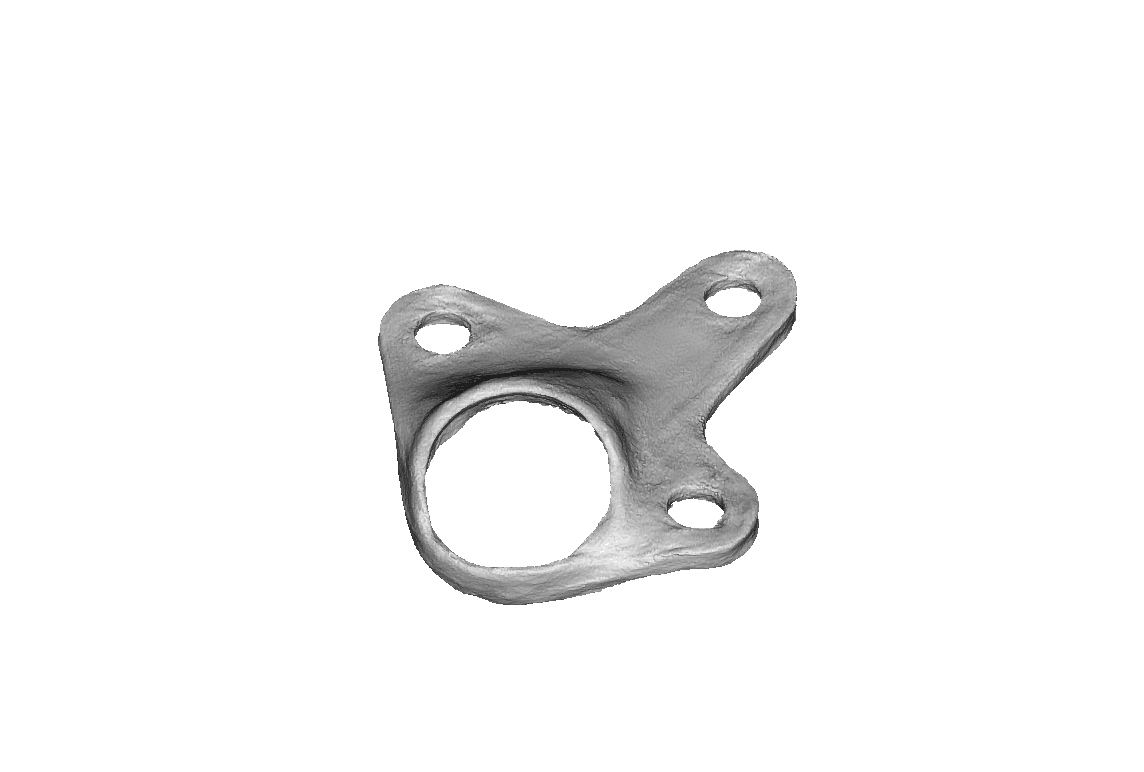}
\end{subfigure}
\\
\rowlabel{4}%
\begin{subfigure}[c]{0.15\textwidth}
    \centering
    \includegraphics[width=\textwidth]{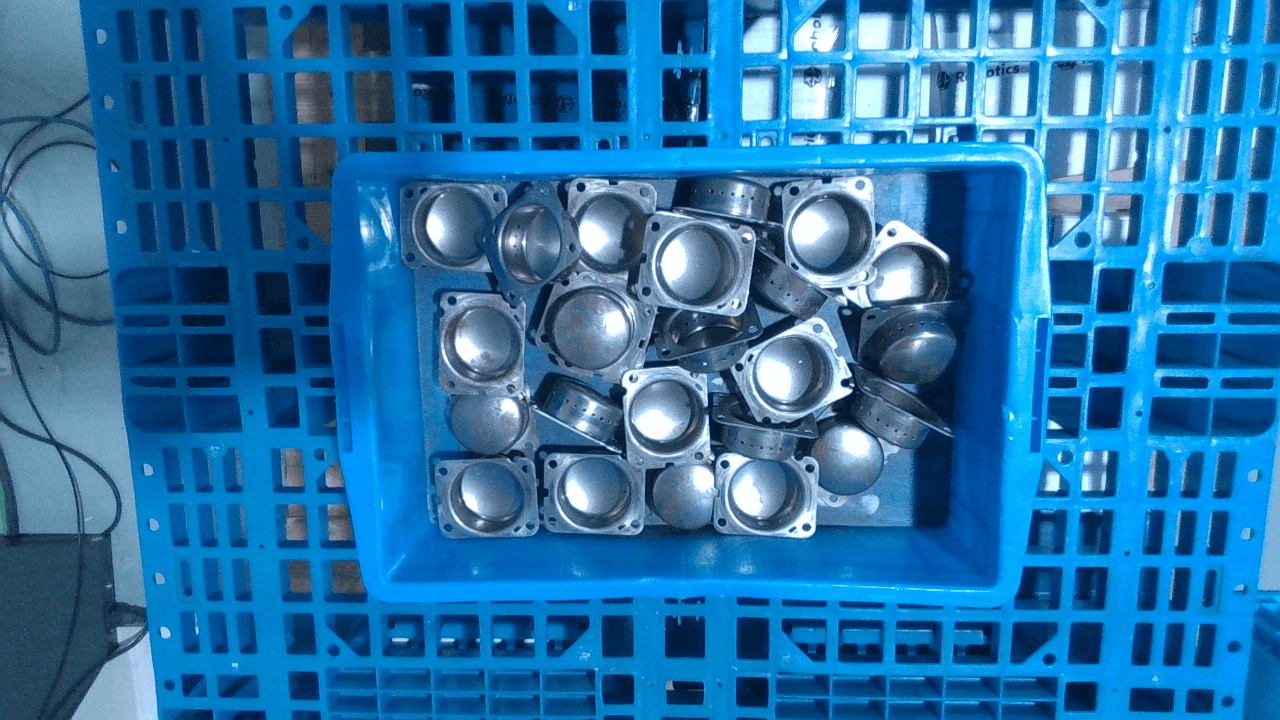}
    \caption*{\centering Input image}
\end{subfigure}\hfill
\begin{subfigure}[c]{0.15\textwidth}
    \centering
    \includegraphics[height=2cm,trim={13cm 6cm 10cm 4cm},clip]{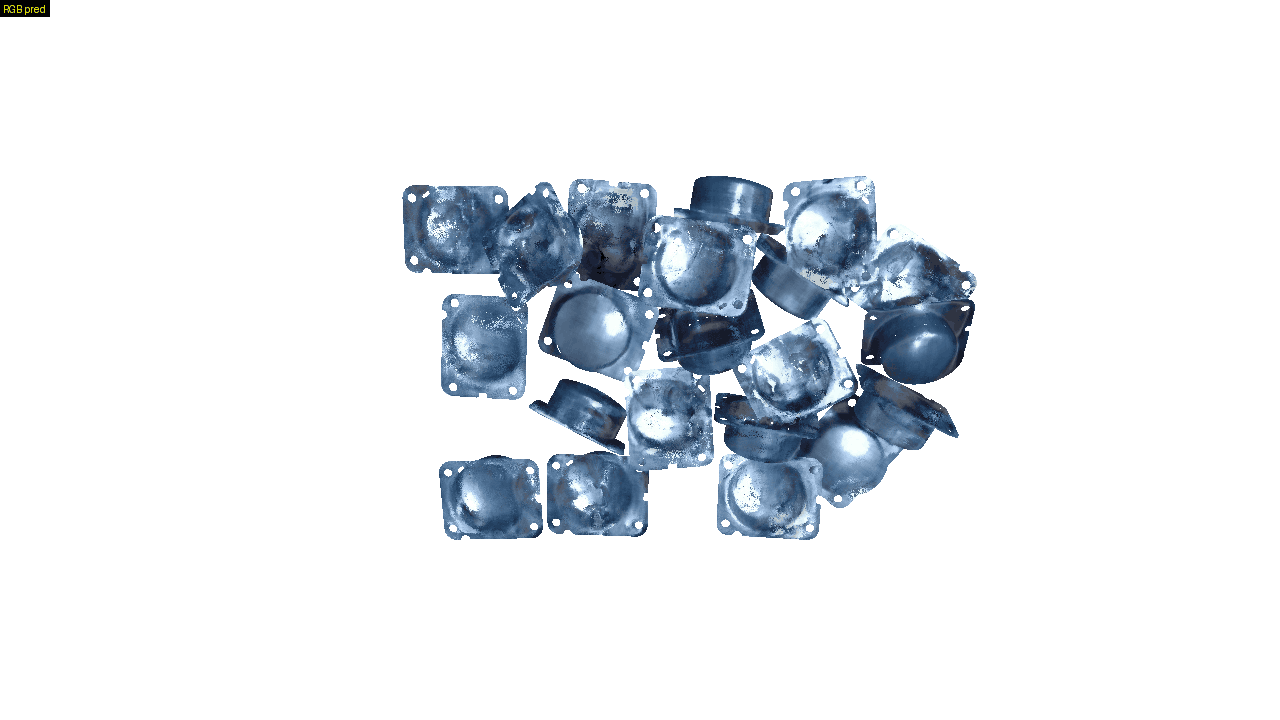}
    \caption*{\centering Rendered RGB \\\they\unskip$^*$}
\end{subfigure}\hfill
\begin{subfigure}[c]{0.15\textwidth}
    \centering
    \includegraphics[height=2cm,trim={13cm 6cm 10cm 4cm},clip]{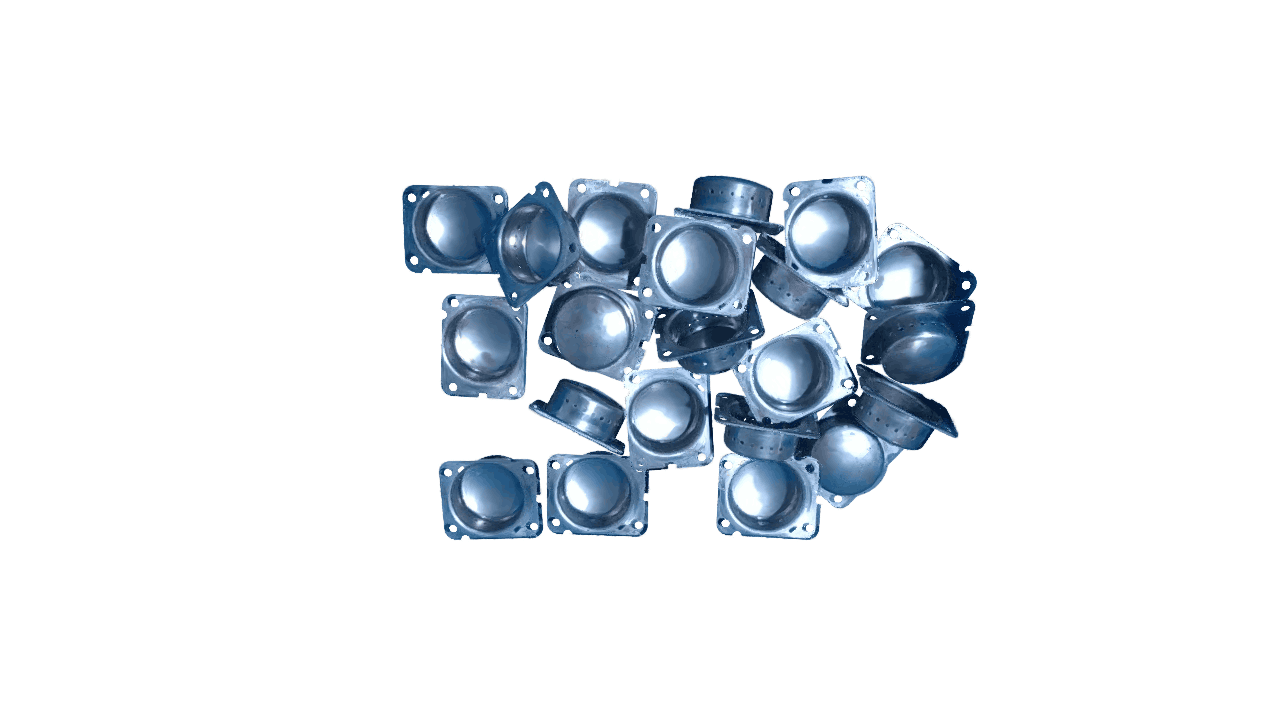}
    \caption*{\centering Rendered RGB \\\method}
\end{subfigure}\hfill
\begin{subfigure}[c]{0.15\textwidth}
    \centering
    \includegraphics[height=2cm]{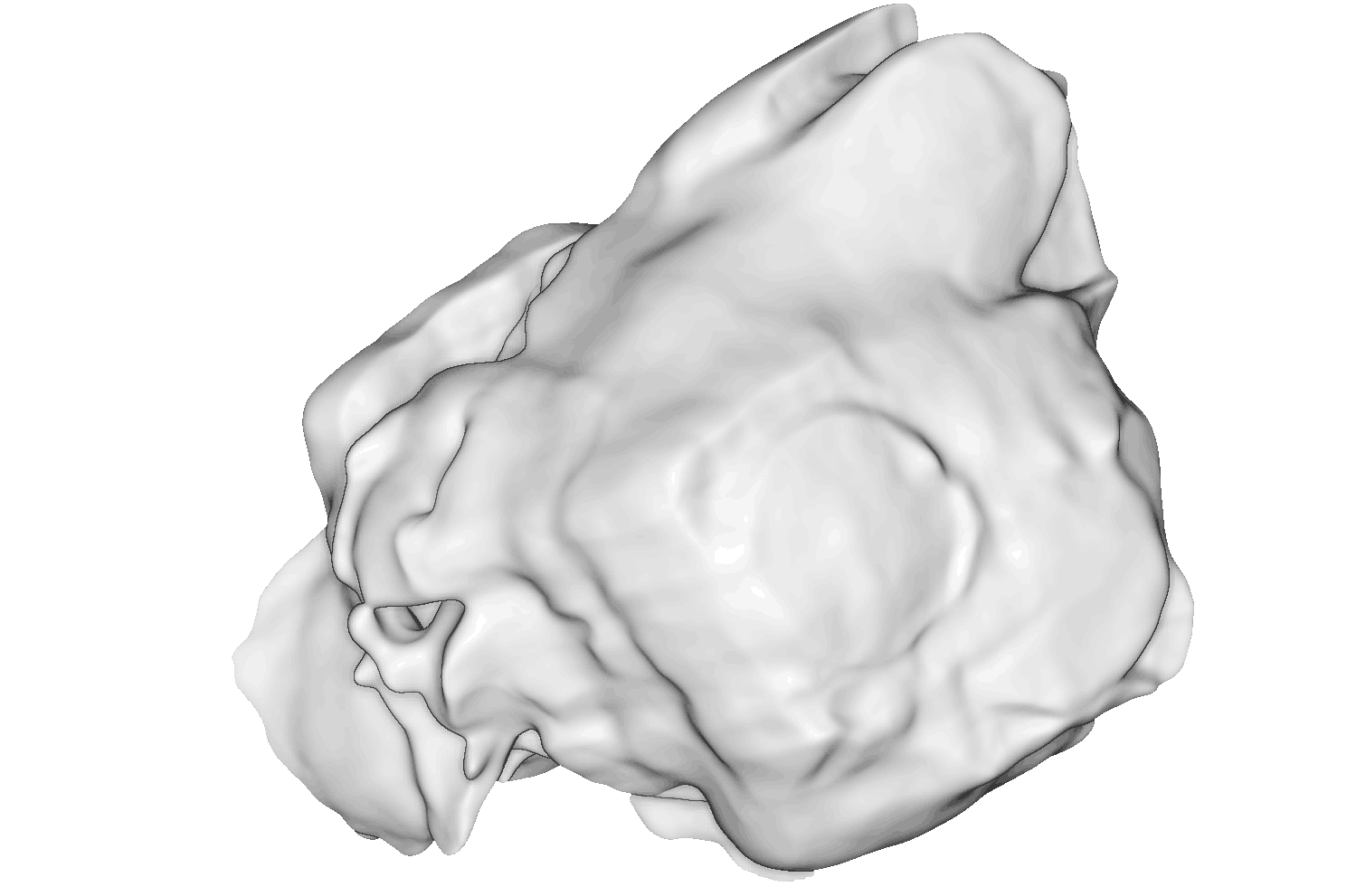}
    \caption*{\centering 3D reconstruction \they\unskip$^*$}
\end{subfigure}\hfill
\begin{subfigure}[c]{0.15\textwidth}
    \centering
    \includegraphics[height=2cm]{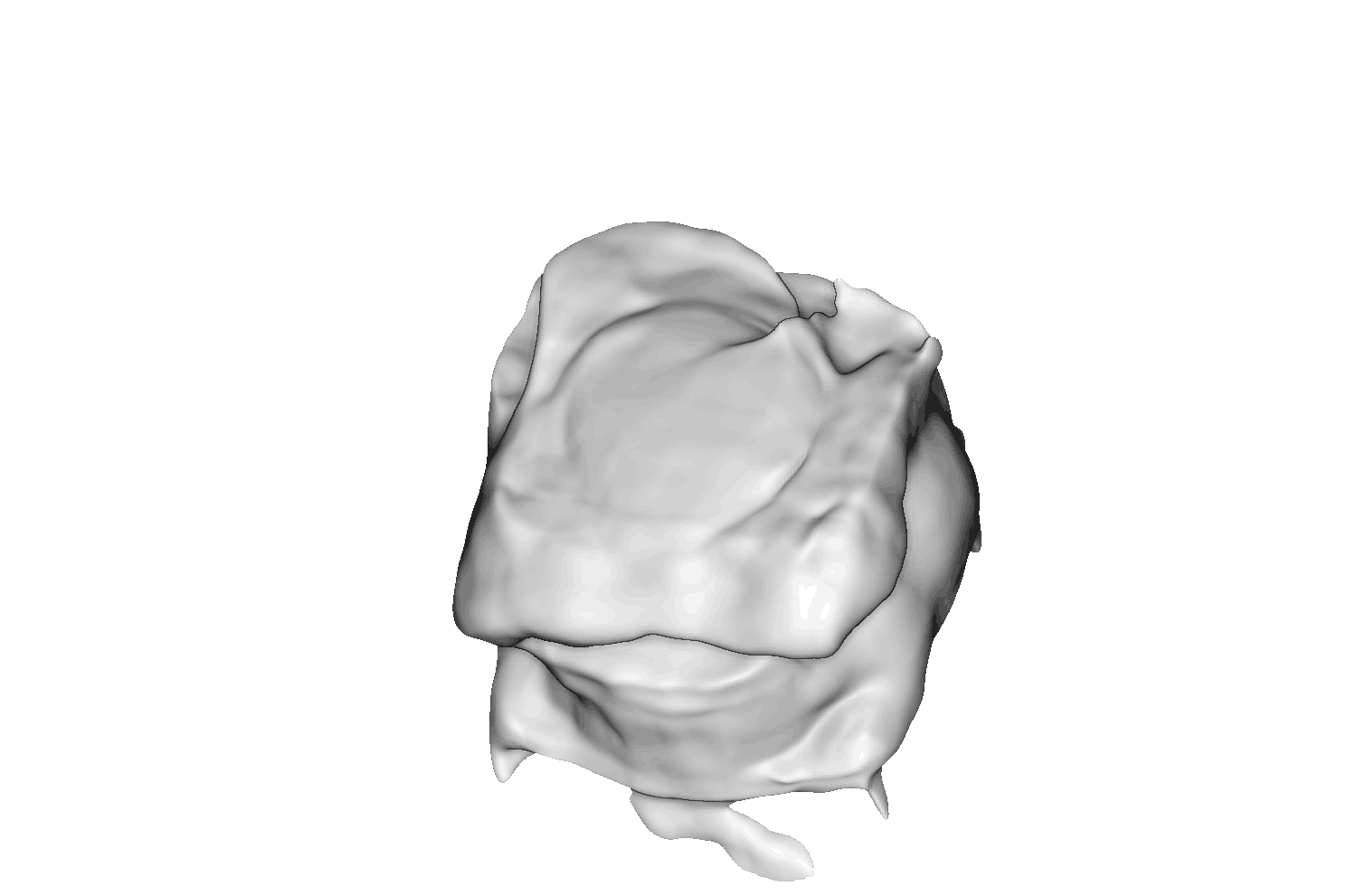}
    \caption*{\centering 3D reconstruction \they\unskip$^*$ with GT poses}
\end{subfigure}\hfill
\begin{subfigure}[c]{0.15\textwidth}
    \centering
    \includegraphics[height=2cm]{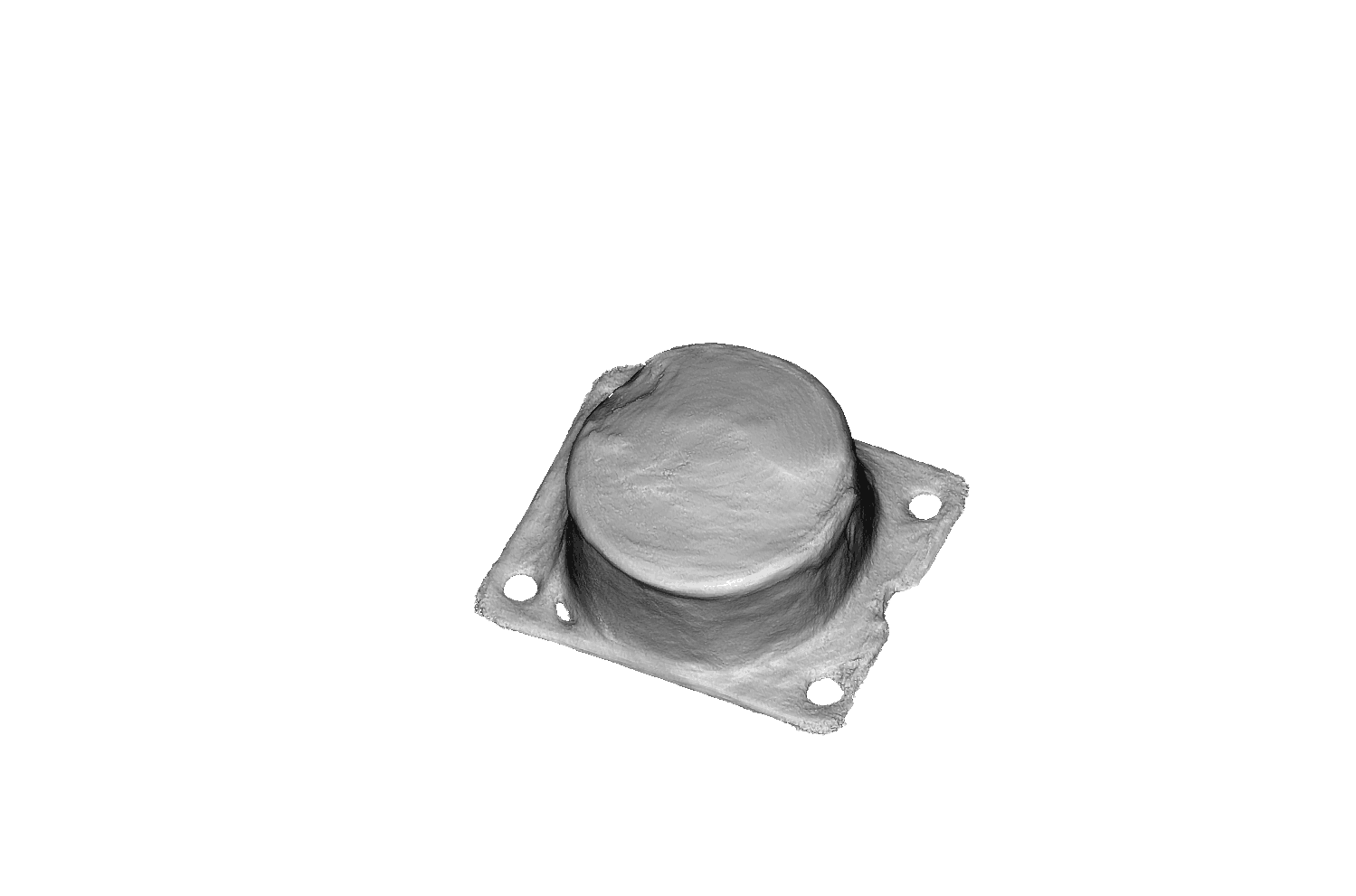}
    \caption*{\centering 3D reconstruction \method}
\end{subfigure}
\\
    \caption{Qualitative results of \they\unskip$^*$, and \method on real-world multi-instance images of XYZ-IBD \cite{huangXYZIBDHighprecisionBinpicking2025}. Both methods use an oracle-style initialization with noisy object poses (see \cref{sec:datasets}). For reference, \they\unskip$^*$ initialized with GT object poses is shown as well.}
    \label{fig:qualitative_comparison_real_xyz}
\end{figure*}

\begin{table*}[tbp]
    \centering
\caption{Object pose quality in terms of MRE [\textdegree]$\downarrow$ and MTE [mm]$\downarrow$ on Dup-synthetic (left) and XYZ-IBD (right). As \they  does not optimize object poses after the initialization, the initialization values correspond to the final results of \they. Best results are bold. }
\label{tab:ape}
\scalebox{0.8}{
\begin{tabular}{c l cccccccccc|ccccc}
\toprule
 & Object & box & cash & cleaner & clock & coffee & fire & guitar & sign & tin & Mean & XYZ1 & XYZ2 & XYZ3 & XYZ4 & Mean \\
\midrule
\multirow{2}{*}{\rotatebox[origin=c]{90}{MRE}}
&Init./ \they&0.83&0.62&\textbf{2.02}&0.84&0.62&1.21&0.88&0.91&1.05&1.00&5.00&5.00&5.00&5.00&5.00\\
&\method&\textbf{0.30}&\textbf{0.24}&2.10&\textbf{0.63}&\textbf{0.44}&\textbf{0.58}&\textbf{0.41}&\textbf{0.30}&\textbf{1.02}&\textbf{0.67}&\textbf{2.02}&\textbf{1.09}&\textbf{0.53}&\textbf{1.34}&\textbf{1.24}\\ 
\midrule
\multirow{2}{*}{\rotatebox[origin=c]{90}{MTE}}
&Init./ \they&9.85&15.04&34.81&14.30&11.05&18.94&19.81&19.24&\textbf{24.85}&18.65&30.00&30.00&30.00&30.00&30.00\\
&\method  &\textbf{4.57}&\textbf{1.72}&\textbf{20.97}&\textbf{9.00}&\textbf{6.19}&\textbf{6.72}&\textbf{11.05}&\textbf{3.70}&25.68&\textbf{9.96}&\textbf{5.24}&\textbf{5.54}&\textbf{1.54}&\textbf{5.50}&\textbf{4.46}\\ 

\bottomrule
\end{tabular}
}
\end{table*}
\begin{table*}[tbp]
\centering
\begin{minipage}{0.68\textwidth}
    \caption{MAE [\textdegree]$\downarrow$ of normal images on Dup-synthetic. $^*$ marks that the monocular normal predictor was changed to Marigold V2 \cite{pavlovic2026marigold}. Best results are bold.}
\label{tab:normal_angular_error}
\resizebox{\linewidth}{!}{
\begin{tabular}{lcccccccccc}
\toprule
Object & box & cash & cleaner & clock & coffee & fire & guitar & sign & tin & Mean \\
\midrule
Marigold V2 \cite{pavlovic2026marigold}& 5.0 & 12.3 & 7.6 & 13.8 & 16.6 & 10.3 & 9.0 & 9.4 & 12.6 & 10.7 \\
Omnidata \cite{eftekhar2021omnidata} & 12.3 & 21.3 & 18.1 & 17.4 & 20.7 & 15.8 & 18.0 & 25.7 & 17.6 & 18.5 \\
\they \cite{NEURIPS2023_3764a9c8}& 5.9 & 19.4 & 11.1& 15.6 & 21.6 & 15.2 & 12.2& 21.1 & 15.7 & 15.3 \\
\they\unskip$^*$& 4.4 & 13.8  & 6.5&  14.5& 17.8 &10.1  & 8.3 & 8.9 & 12.5 &  10.8\\ 

\method&\textbf{3.5}  & \textbf{9.7}  & \textbf{5.9}&\textbf{12.2}  & \textbf{15.6} & \textbf{8.1}& \textbf{6.8} & \textbf{6.3} &\textbf{11.0}  &\textbf{8.8}  \\    

\bottomrule
\end{tabular}
}
\end{minipage}
\hfill
\begin{minipage}{0.3\textwidth}
        \centering
     \caption{Evaluation metrics at different training iterations on Dup-synthetic. $^*$ marks that the monocular normal predictor was changed to Marigold V2 \cite{pavlovic2026marigold}. Best results are bold.}
     \resizebox{\linewidth}{!}{
    \begin{tabular}{lcc|c}
    \toprule
    Method&\multicolumn{2}{c|}{\method}&\they\unskip$^*$ \\
         &@7k&@30k& \\
         \midrule
         MAE [\textdegree]$\downarrow$&\textbf{8.80}&8.85&10.8\\
         MRE [\textdegree]$\downarrow$&\textbf{0.67}&0.67&1.00\\
         MTE [mm]$\downarrow$&\textbf{9.96}&10.44&18.65\\
         PSNR [dB]$\uparrow$&39.55&\textbf{42.45}&22.93\\
         Time [min]$\downarrow$&\textbf{8}&37&360\\
         \bottomrule
    \end{tabular}
   }
    \label{tab:time}
\end{minipage}

\end{table*}\cref{tab:time} examines possible tradeoffs when training \method for 30k instead of 7k iterations. If 3D reconstruction and object poses are sought for, training for 7k iterations is preferable, while further iterations mainly improve PSNR. Further training behavior and qualitative results of \method are discussed in the supplement. 
\paragraph{XYZ-IBD. } \cref{tab:chamfer_distance_xyz} shows the results of 3D reconstruction and \cref{tab:ape} the results of object pose estimation  on the XYZ-IBD dataset in an oracle-style experiment (see \cref{sec:datasets}). 
On all tested scenes, \method is able to significantly improve the object poses. 
Regarding the 3D reconstruction results, \method is able to reconstruct the canonical object shape with a mean $\mathrm{d_c}$ of less than 1\,mm, whereas \they is worse by a factor of roughly twenty. \cref{fig:qualitative_comparison_real_xyz} backs up the quantitative results with qualitative comparisons. 
As depicted, \they\unskip 's 3D reconstruction is degenerate. Even when GT poses are used for \they, the 3D reconstruction is inaccurate with a mean $\mathrm{d_c}$ of 12.9\,mm. This is not reflected by the rendered RGB image, since rendered instances are cropped with the visible masks. In contrast, \method renders all existing geometry without cropping. This demonstrates a fundamental difference between \they and \method: \they casts rays per pixel, and assigns each ray to an instance based on the mask, evaluating the canonical object in the canonical object frame. Because every ray only evaluates its own instance, overlapping or intersecting geometry between instances due to object pose and geometry errors is never rendered at all, leaving them undetected by photometric and background loss. Background pixels are evaluated through a random instance at a time, using only a fraction of available information. In contrast, \method places all object instances at their poses in the global scene frame, making overlaps visible and optimizable with the instance-wise background loss (\cref{sec:multi_instance_rendering}), which also supervises all instances jointly in background regions in each iteration. Essentially, this architectural difference leads to \method revealing such inconsistencies, whereas \they is concealing them.
\paragraph{Ablation Study of Methodological Components. }We assess the contribution of each individual methodological component of \method in \cref{tab:ablation_3d_reconstruction}. Especially on XYZ-IBD, $\mathcal{L}_m$ is indispensable. Using $\mathcal{L}_{nc}$ improves the 3D reconstruction significantly, but comes at the cost of slightly worse object poses. Adding $\Matrix{U}_{far}$ does not improve the result, and removing the scene illumination entirely (w/o $\Matrix{U}_{near}$) leads to worse results. $\mathcal{L}_o$ has a small positive effect on all metrics. Disabling pose optimization leads to a slightly worse 3D reconstruction on Dup-synthetic, while for XYZ-IBD pose optimization is crucial, since the initialization is too inaccurate for a geometry-only optimization. 

\begin{table}[!tbp]
    \centering
    \caption{Contribution of the individual proposed methodological components of \method. We report pose errors MRE [\textdegree], MTE [mm], and reconstruction error $\mathrm{d}_c$ [mm] for the datasets Dup-synthetic and XYZ-IBD. $^*$ marks that the monocular normal predictor was changed to Marigold V2 \cite{pavlovic2026marigold}. Best results are bold. }
\scalebox{0.8}{
\begin{tabular}{l|ccc|ccc}
    \hline
\multirow{2}{*}{Method}
  & \multicolumn{3}{c|}{Dup-synthetic}
  & \multicolumn{3}{c}{XYZ-IBD} \\
   &MRE$\downarrow$&MTE$\downarrow$&$\mathrm{d}_c\downarrow$&MRE$\downarrow$&MTE$\downarrow$&$\mathrm{d}_c\downarrow$\\
         \hline
                 \method &0.67&9.96&\textbf{5.54}&\textbf{1.24}&4.46&\textbf{0.85}  \\
        \hline
        w/o $\mathcal{L}_m$&0.68&10.03&11.56&21.90&56.52&45.58  \\
        w/o $\mathcal{L}_{nc}$&\textbf{0.64}&\textbf{9.59}&9.27&1.61&\textbf{4.40}&1.49  \\
        w/ $\Matrix{U}_{far}$&0.70&10.85&\textbf{5.54}&1.96&5.84&0.86  \\
        w/o $\Matrix{U}_{near}$&0.68&10.67&6.09&1.89&7.40&1.14  \\
        w/o $\mathcal{L}_o$&0.70&10.76&5.76&1.32&5.01&0.86  \\
        w/o pose opt&1.00&18.65&5.72&5.00&30.00&11.78\\
         \hline
         \they\unskip$^*$&1.00&18.65&10.8&5.00&30.00& 17.0 \\
         \bottomrule
\end{tabular}
    }
    \label{tab:ablation_3d_reconstruction}
\end{table}
\subsection{Current Limitations and Future Work}
\label{sec:limitations}

While our method achieves promising results, the current implementation faces two limitations we aim to address in the future. The accuracy of the approach depends on sufficient pose diversity and segmentation quality and, more importantly, the SfM-based initialization, which we adopted from \they \cite{NEURIPS2023_3764a9c8}. This initialization fails for complex multi-instance images such as in XYZ-IBD, identifying it as our main focus for future work. Additionally, our background loss formulation assumes the canonical object to be opaque, which currently prevents the support of transparent objects.

%% file: sec/05_conclusion.tex
\section{Conclusion}
We present \method, a reformulation of conventional GS, adapted to solve the \task task, which consists of zero-shot joint 3D reconstruction and object pose estimation from a single multi-instance image, i.e., an image where multiple instances of the same canonical object are visible in different perspectives.  With \method, we obtain an accurate 3D reconstruction and object poses in the case of well-textured objects from a single RGB image, while outperforming the SOTA on the evaluated benchmarks. In addition, we show that \method, in contrast to the SOTA, recovers accurate 3D reconstruction and object poses for the textureless objects and occluded scenes of XYZ-IBD, given appropriate initial values. We attribute this to our architecture decision, to train the geometry of object instances in the global coordinate frame, paired with the instance-wise background loss. By enforcing consistency between object instances, \method is able to refine rendered normals of SOTA monocular normal estimators. 

%% file: sec/llm_usage_statement.tex
\section*{LLM Usage Statement}
LLMs and generative AI tools were used in this work. For writing and editing text, researching related work, and reviewing, locally hosted open-weight models (GLM-5.3-Flash and Deepseek-V4-Flash), Copilot, and Claude were used non-agentically; content was reviewed and verified by the authors. For coding and experiments, work was partially conducted agentically using GitHub Copilot models and locally hosted open-weight models (GLM-5.3-Flash and Deepseek-V4-Flash); all code was reviewed and validated by the authors. 

%% file: sec/X_suppl.tex
\clearpage
\setcounter{page}{1}
\setcounter{section}{0}
\renewcommand{\thesection}{\Alph{section}}
\maketitlesupplementary
\section{Modifications to the \ac{adc}}
\paragraph{\ac{adc} in conventional GS.} After initialization with from sparse points, the set of Gaussians $\mathcal{G}$ needs to be adapted by adding and removing Gaussians in over- and under-reconstructed areas for an accurate representation. Both cases are characterized by Gaussians with large view-space positional gradients $\nabla\Vector{p}_{i,t}^{2\mathrm{D}}$ \cite{kerbl20233d}. At each densification interval $T$, Gaussians get split or cloned in case
\begin{equation}
\frac{\sum_{t=1}^{T}\mathrm{vis}(\mathcal{G}_{i,t})\vert\vert\nabla\Vector{p}_{i,t}^{2\mathrm{D}}\vert\vert_2}{\sum_{t=1}^{T}\mathrm{vis}(\mathcal{G}_{i,t})}>\tau,
    \label{eq:tresh}
\end{equation}
where 
$\mathrm{vis(\mathcal{G}_{i,t})}=1$, if $\mathcal{G}_{i,t}$ is visible from the current view at iteration $t$ and $\mathrm{vis(\mathcal{G}_{i,t})}=0$ if it is not visible. \citet{kerbl20233d} set $\tau=0.0002$.
\paragraph{\ac{adc} in \method. }The original \ac{adc} and its heuristics in 3D-GS \cite{kerbl20233d}, which are unchanged in 2D-GS \cite{huang2DGaussianSplatting2024} and reflection-based SV \cite{sarioSphericalVoronoiDirectional2025}, do not fit to the proposed reformulated GS approach (\cref{fig:method_overview,sec:methodology}). It relies on the principle, that each individual Gaussian is visible only once in a single image, or iteration, respectively. In our case, this is violated, as each Gaussian can be seen a maximum of $J$ times in a single iteration, since \method uses the same, complete image in every iteration of the reformulated GS pipeline. Therefore, the ADC thresholding in \cref{eq:tresh} is modified to account for how often $\mathcal{G}_{i,j,t}$ is visible within a render: \method splits or clones Gaussians in case
 \begin{equation}
 {\frac{\sum_{t=1}^{T}\sum_{j=1}^{J}\mathrm{vis}(\mathcal{G}_{i,j,t})\vert\vert\nabla\Vector{p}_{i,j,t}^{2\mathrm{D}}\vert\vert_2}{\sum_{t=1}^{T}\sum_{j=1}^{J}\mathrm{vis}(\mathcal{G}_{i,j,t})}>\tau}.
     \label{eq:tau_new}
 \end{equation}
We set $\tau=0.00005$, as \method reconstructs small scale object centric scenes. Additionally, we add a single opacity reset at iteration 1000, to prune incorrect geometry early, before it negatively impacts training. All other hyperparameters of the ADC are left unchanged for \method.
\section{Implementation Details}
\label{sec:implementation_details}
\paragraph{Modifications to the Deferred Rendering.}
A few changes are made to the deferred rendering of SV \cite{sarioSphericalVoronoiDirectional2025}. SV initializes the light probes of $\Matrix{U}_{\mathrm{near}}$ at random positions within the bounding box of the scene $\mathcal{G}_\mathrm{c}$. \method, instead, initializes the light probes randomly within the bounding box of $\mathcal{G}_{J}$. Otherwise, the SV light probes struggle to converge to the positions where specular reflections are needed.

\paragraph{Loss Weights. }We set $\epsilon=1$, and $\zeta=0.001$, while we use the same $\gamma=0.05$, and $\delta=1000$ as in 2D-GS \cite{huang2DGaussianSplatting2024}. For all experiments on Dup, we set $\beta=1$, while on XYZ-IBD, due to significantly higher degree of occlusion and complete lack of texture, we set $\beta=10$, as $\mathcal{L}_\mathrm{m}$ is crucial for object pose optimization. Especially for textureless objects, intersecting geometry does not lead to a significant change in photometric error, while the signal of the $\mathcal{L}_\mathrm{m}$ is independent of texture.

The pose parameters are added to an Adam \cite{kingma2014adam} optimizer and we set $\mathrm{lr}_\mathrm{t}=\mathrm{lr}_\mathrm{R}=0.001$, exponentially decaying to $0.00001$ over the training iterations in all our experiments.
\paragraph{XYZ-IBD Data} For reproducibility we specify exactly which images we used. The four scenes from XYZ-IBD have been randomly selected from the validation set. XYZ1 is image 14 from scene 000015, XYZ2 corresponds to scene 000040 image 17, XYZ3 is image 6 from scene 000060, and XYZ4 is image 34 from scene 000030. In all cases, we used the RGB image captured by the Realsense sensor. Although XYZ-IBD is captured with RGB-D sensors, we only ever used the RGB image for training.
\section{Further Ablation Studies}
\label{sec:ablations}
\begin{figure*}

    \raisebox{12ex}{\rotatebox{90}{XYZ-IBD} }\hfill
    \begin{subfigure}{0.45\textwidth}
        \includegraphics[width=\textwidth]{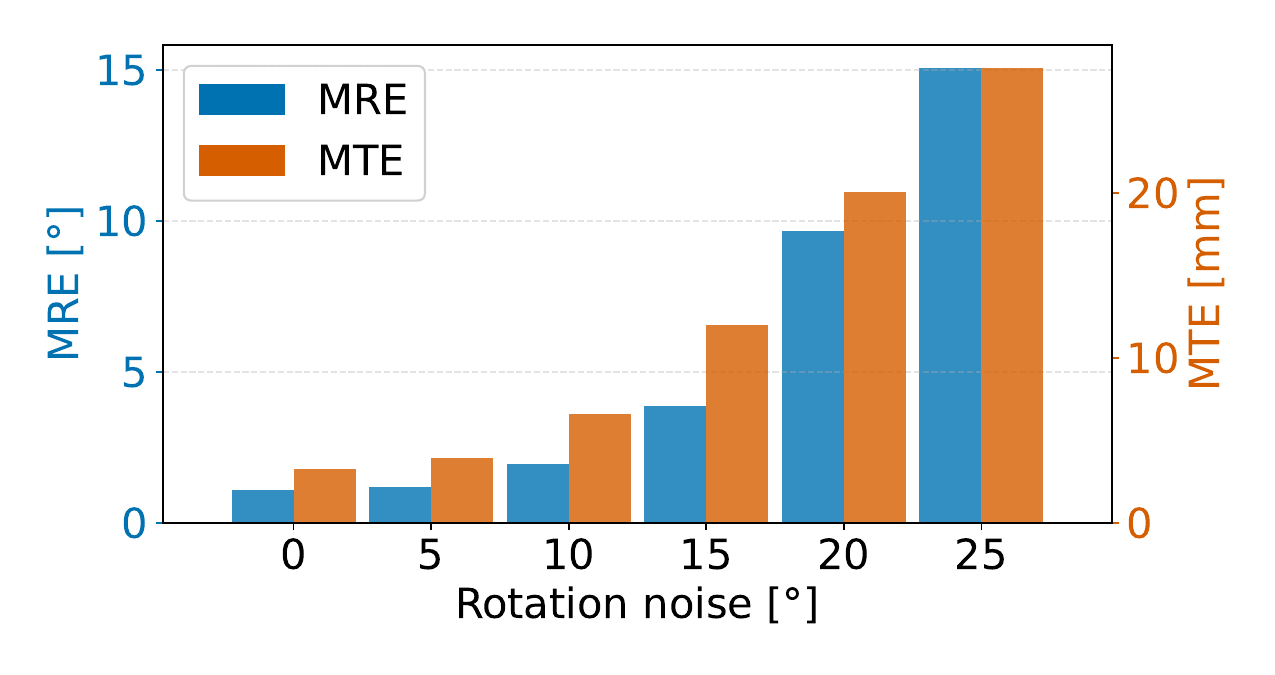}
        
    \end{subfigure} \hfill
        \begin{subfigure}{0.45\textwidth}
        \includegraphics[width=\textwidth]{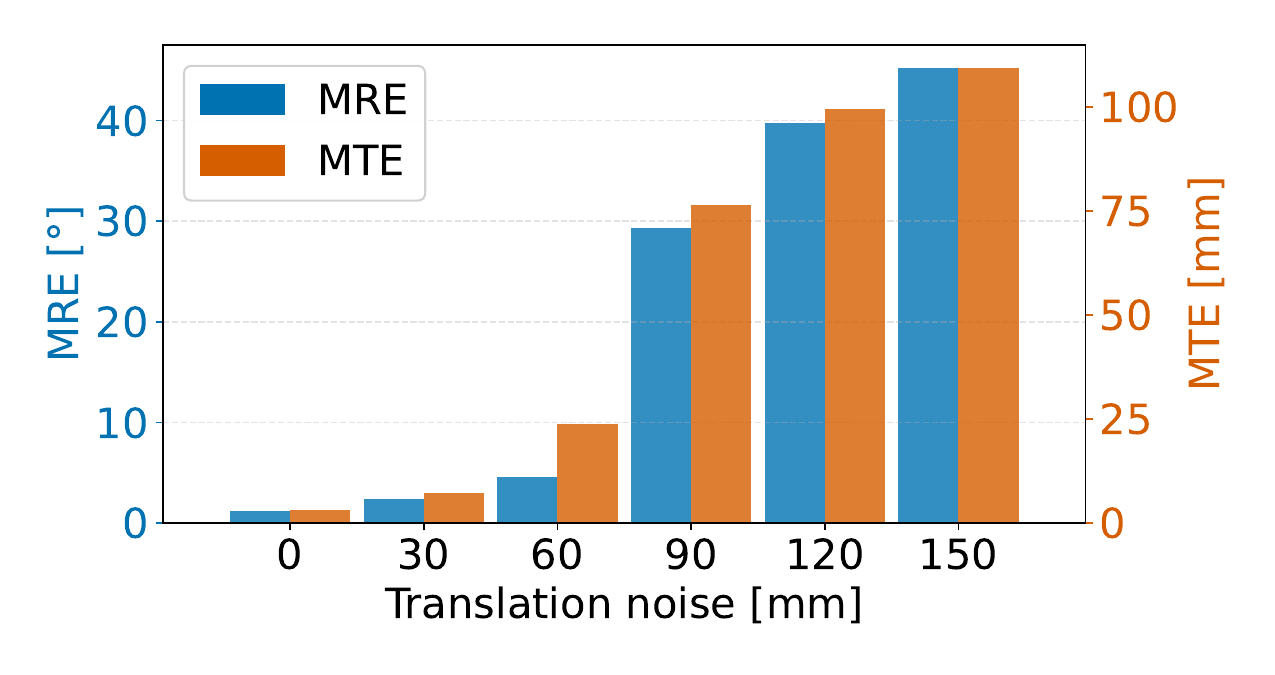}
    \end{subfigure}\\

\raisebox{12ex}{\rotatebox{90}{Dup-synthetic} }\hfill
    \begin{subfigure}{0.45\textwidth}
    \includegraphics[width=\textwidth]{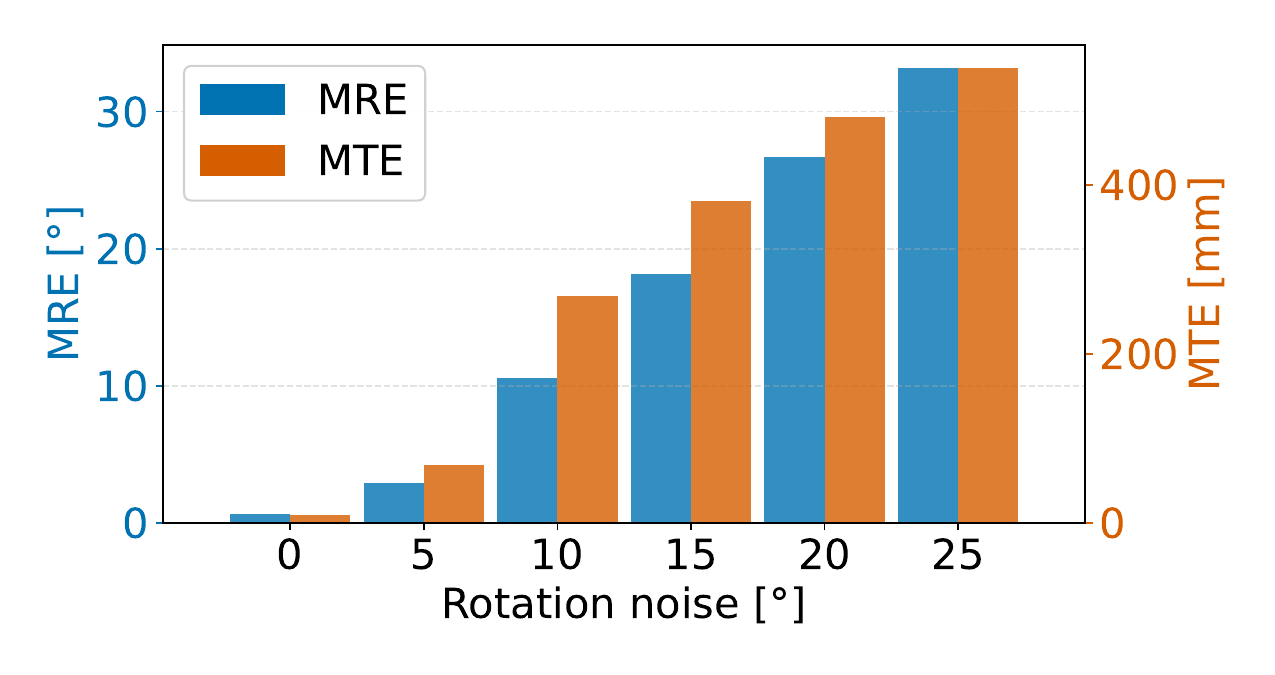}
    \end{subfigure} \hfill
        \begin{subfigure}{0.45\textwidth}
        \includegraphics[width=\textwidth]{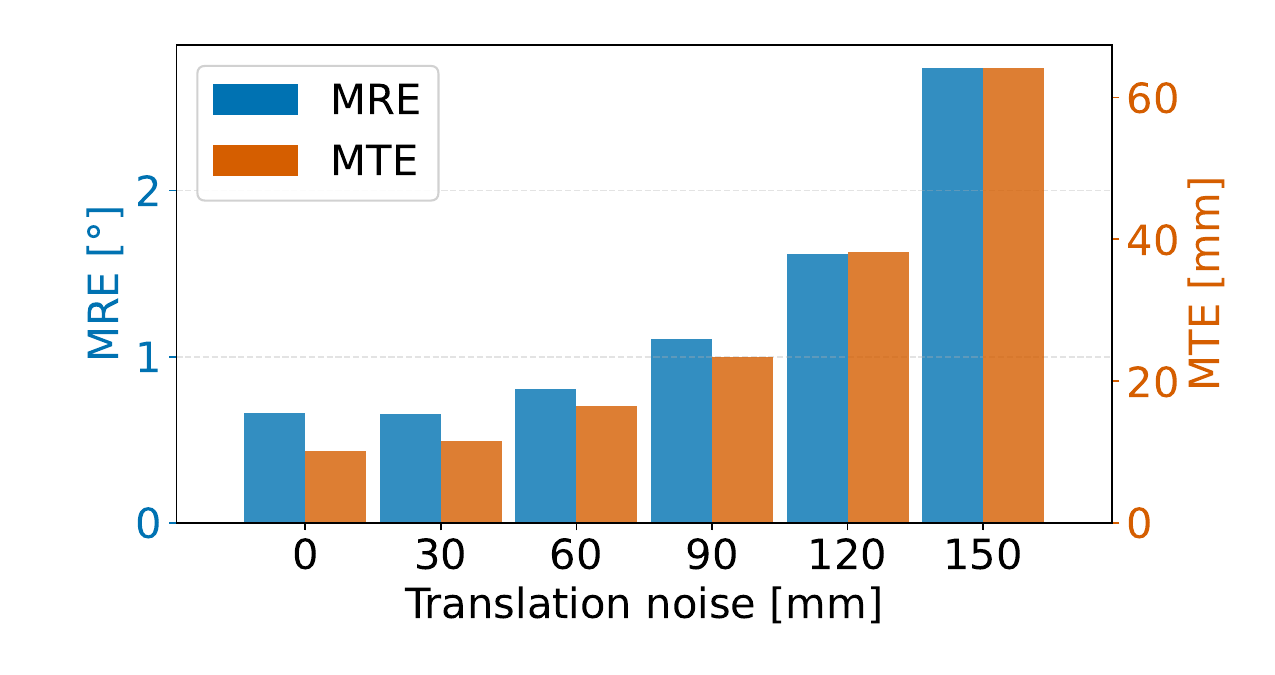}
    \end{subfigure}
    
    \caption{Effect of noisy initial poses on \method. We evaluate MRE [\textdegree] and MTE [mm] on XYZ-IBD (top row) and Dup-synthetic (bottom row). }
    \label{fig:Convergence_radius}
\end{figure*}

\paragraph{Robustness of Our Reformulated GS Pipeline Against Noisy Initial Poses.} To assess the influence of noisy initial object poses on \method, we perturb the rotational and translational component of object poses, as described in \cref{sec:datasets}. The rotational noise is increased from 0 to 25\textdegree in increments of 5\textdegree, while translation noise is increased from 0 to 150\,mm in increments of 30\,mm. The results shown in \cref{fig:Convergence_radius} are averaged over all scenes in XYZ-IBD 1-4 and Dup-synthetic. For XYZ-IBD, \method achieves an MRE of less than 5\textdegree  with initial rotation noise of 15\textdegree. Translational noise smaller than 90\,mm is handled well, while \method fails for larger noise. On Dup-synthetic, \method exhibits a larger convergence basin with respect to translational noise, likely due to larger objects, while rotational noise is more challenging compared to XYZ-IBD. Overall, the experiment demonstrates that \method can operate successfully with only moderately accurate pose initialization. Completely random object poses are not a sufficiently accurate starting point, due to the inherently ill-posed nature of the task. However, the required initialization accuracy highlights practical robustness and applicability.

For very noisy initial poses, we found that a higher learning rate for the object pose translation and rotation can lead to significantly improved results. Similarly, we found that activating the pose optimization only after 500 iterations improves results on XYZ-IBD while doing the contrary on Dup-synthetic, opposed to activating pose optimization immediately. This is most likely due to initialization of geometry, which is initialized with a random point cloud for XYZ-IBD compared to a more accurate sparse point cloud for Dup-synthetic (see \cref{sec:datasets}). These findings suggest that, given a good initialization of the geometry, it is advantageous to immediately optimize both geometry and object poses jointly.

\begin{figure}
    \centering
        \includegraphics[width=\linewidth]{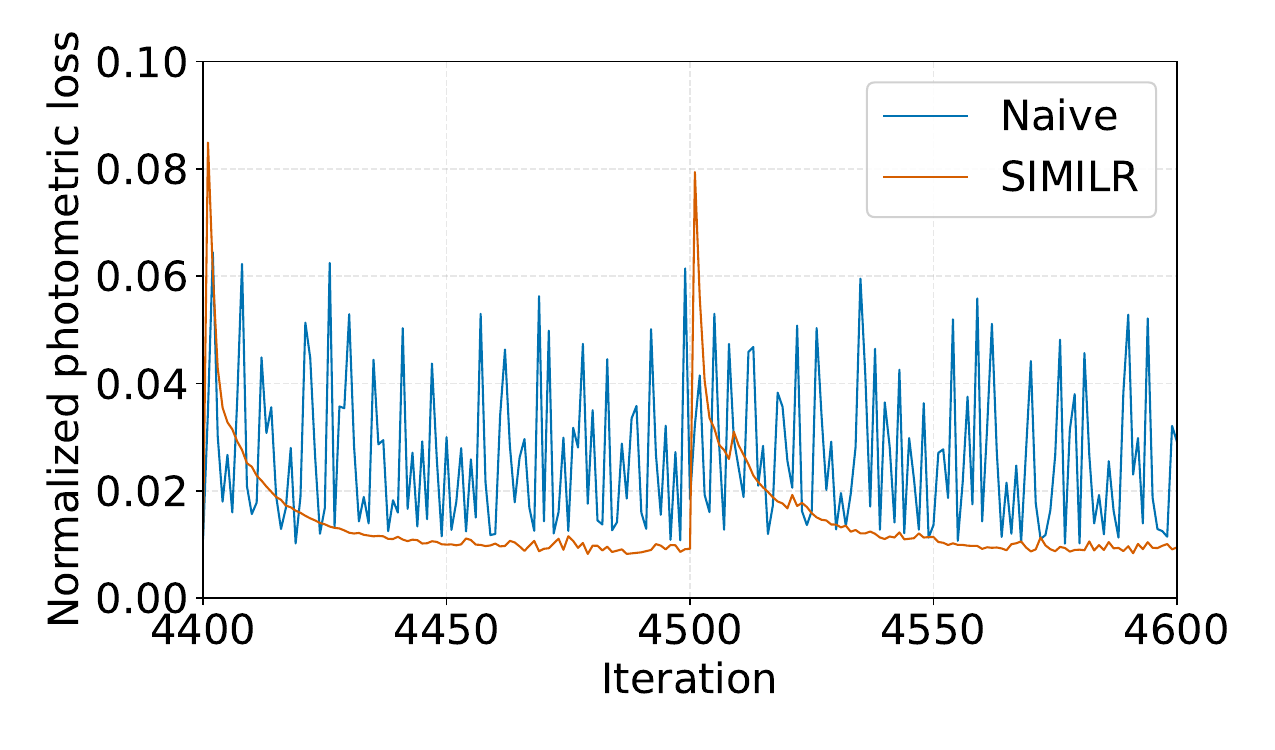}
    \caption{Zoomed view for comparison of normalized photometric loss $\mathcal{L}_{\mathrm{p}}$ between the naive baseline and \method on scene \textit{airplane}. }
    \label{fig:smooth_convergence}
\end{figure}

\paragraph{Photometric Loss Behavior.}
An important difference between \method and conventional GS, is that \method uses the same image with all available information in each iteration, whereas conventional GS uses a different image containing only a subset of available information in each iteration. \cref{fig:smooth_convergence} compares the photometric loss between \method and the naive baseline implementation described in \cref{sec:baselines} as a representative of conventional GS. The losses are normalized for visualization, since the aforementioned difference leads to losses of different orders of magnitude. As the baseline cannot handle occlusions, we choose the scene \textit{airplane} of the Dup dataset for this comparison, which shows no occlusions. Except for spikes, which occur at densification intervals, the photometric loss of \method has significantly less variance between iterations compared with the naive solution. The repetitive spikes are explained with densification operations, which are shared with conventional GS. As \method renders the object multiple times at once, the spikes are more visible as their magnitude scales with the number of instances in the scene. Fine-tuning of GS densification heuristics could potentially further improve the results.

\section{Qualitative Results}
\label{sec:qualitative_results}
\begin{figure*}
    \centering
    \raisebox{8ex}{\rotatebox{90}{Airplane\vphantom{g}} }\hfill
    \begin{subfigure}{0.16\textwidth}
        \centering
        \includegraphics[width=\textwidth]{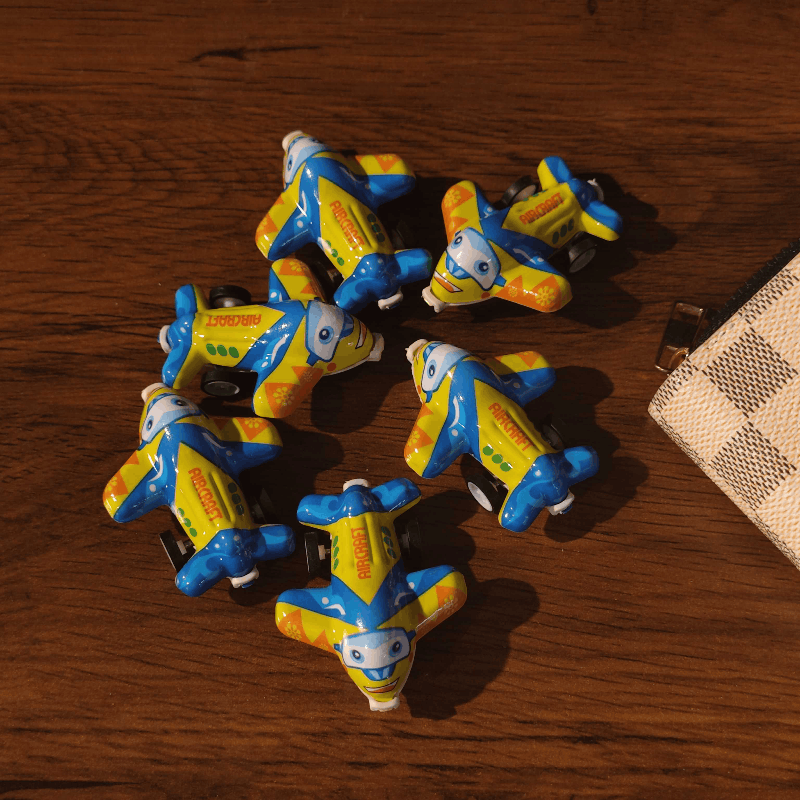}
    \end{subfigure}\hfill
    \begin{subfigure}{0.16\textwidth}
        \centering
        \includegraphics[width=\textwidth]{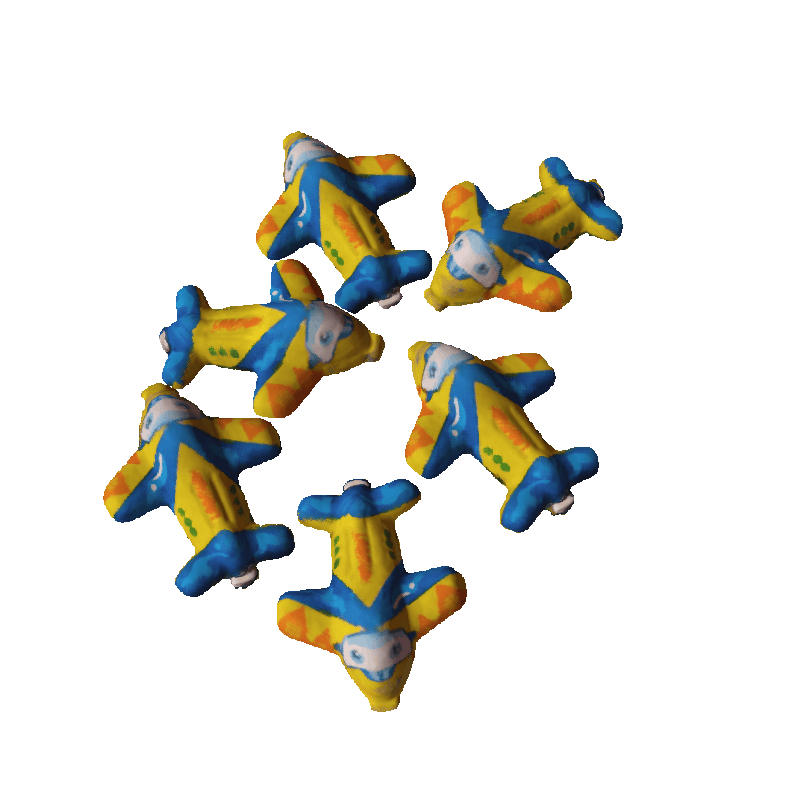}
    \end{subfigure}\hfill
        \begin{subfigure}{0.16\textwidth}
        \centering
        \includegraphics[width=\textwidth]{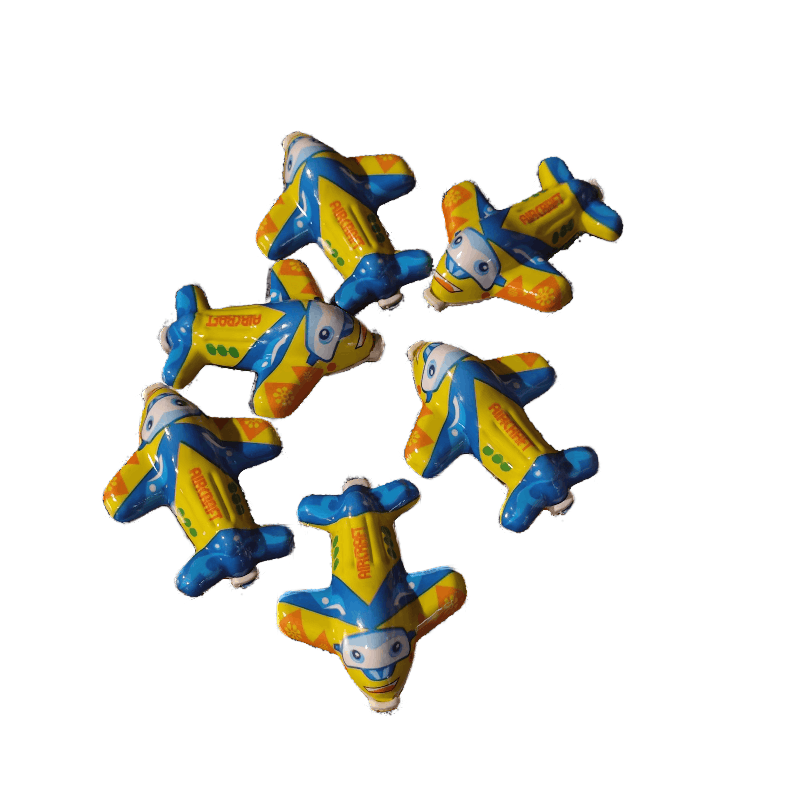}
    \end{subfigure}\hfill
    \begin{subfigure}{0.16\textwidth}
        \centering
        \includegraphics[width=\textwidth]{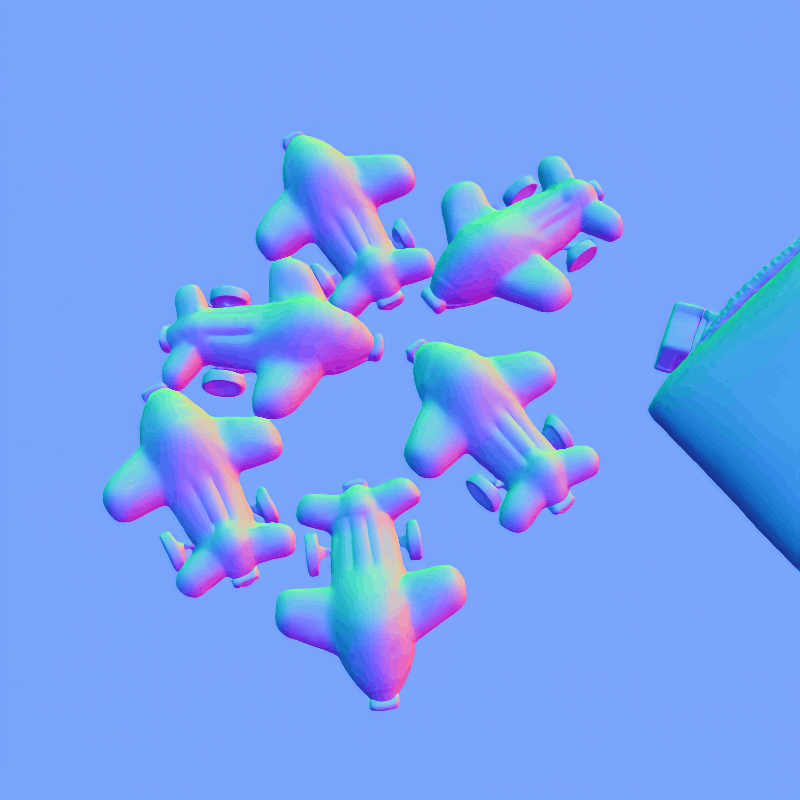}
    \end{subfigure}\hfill
        \begin{subfigure}{0.16\textwidth}
        \centering
        \includegraphics[width=\textwidth]{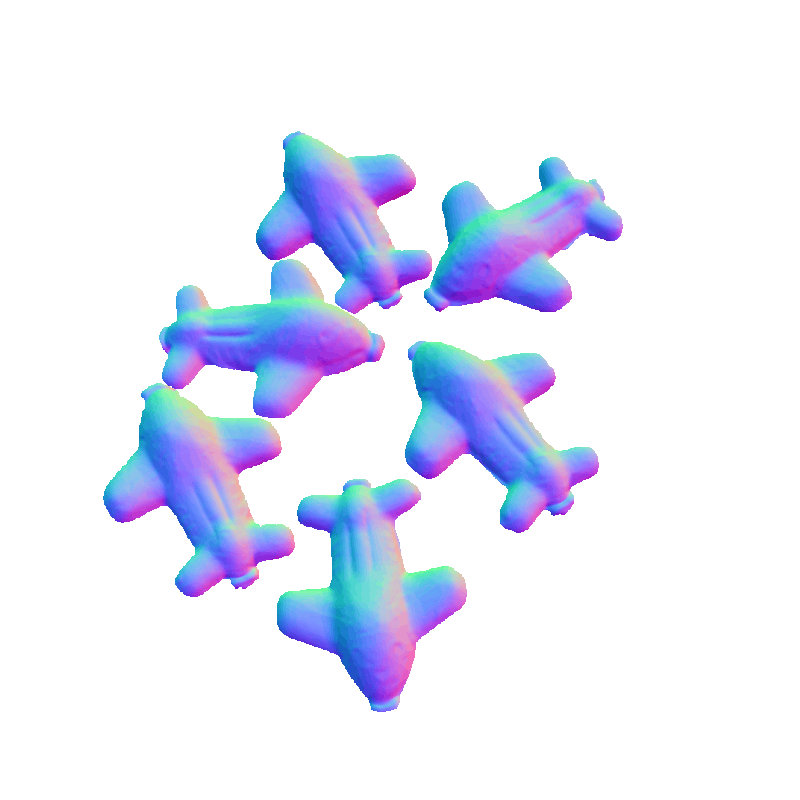}
    \end{subfigure}\hfill
            \begin{subfigure}{0.16\textwidth}
        \centering
        \includegraphics[width=\textwidth]{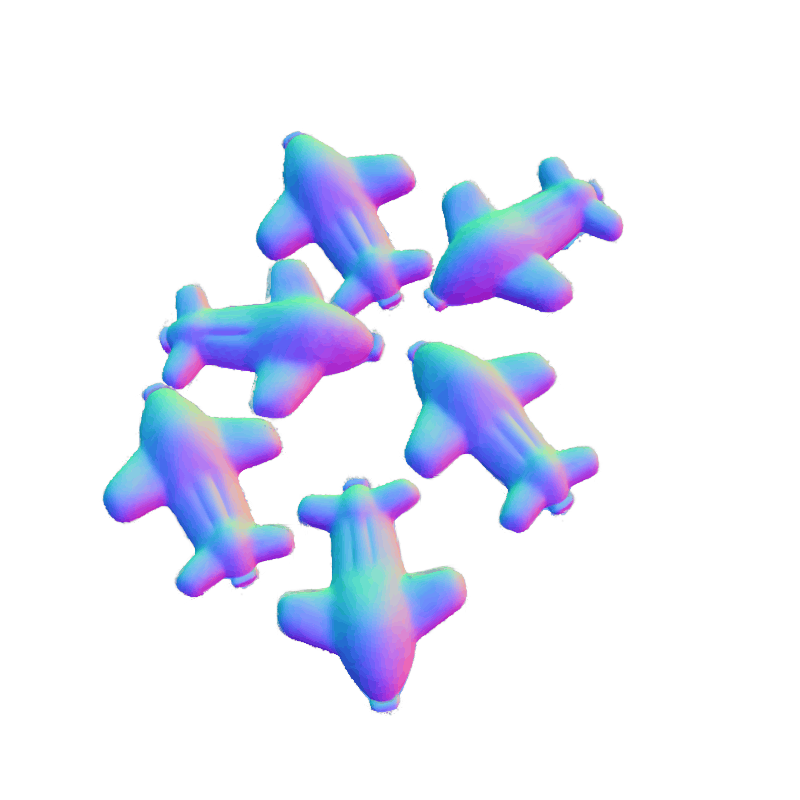}
    \end{subfigure}\\
        \raisebox{8ex}{\rotatebox{90}{Cake\vphantom{g}} }\hfill
    \begin{subfigure}{0.16\textwidth}
        \centering
        \includegraphics[width=\textwidth]{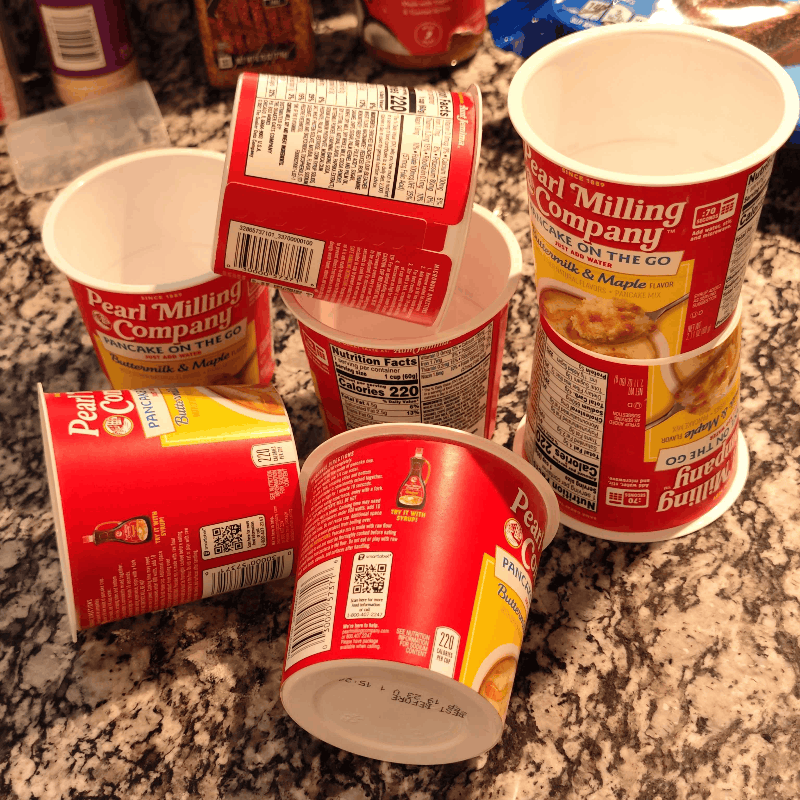}

    \end{subfigure}\hfill
    \begin{subfigure}{0.16\textwidth}
        \centering
        \includegraphics[width=\textwidth]{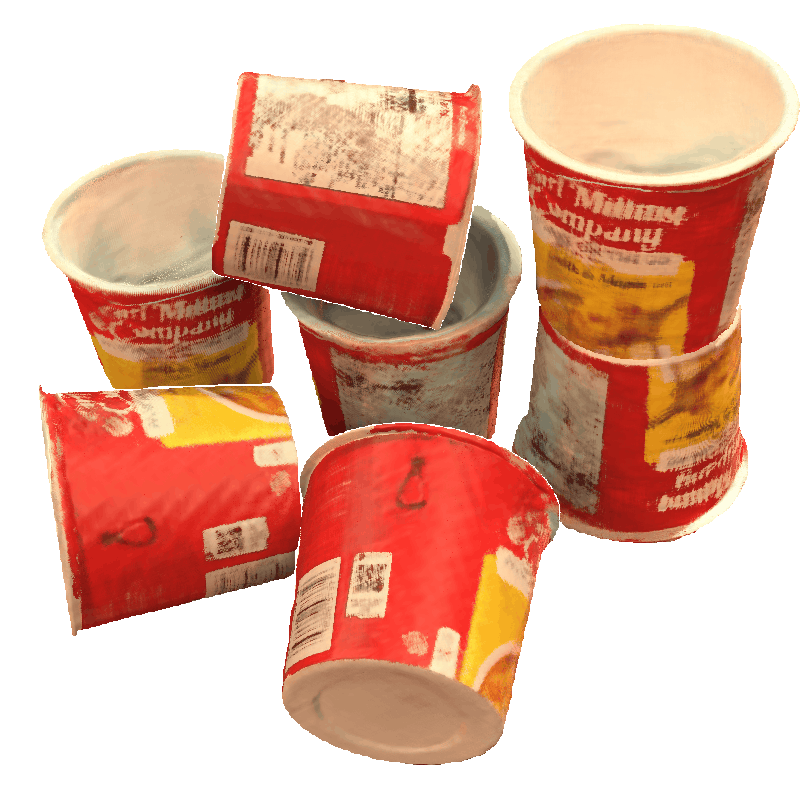}

    \end{subfigure}\hfill
        \begin{subfigure}{0.16\textwidth}
        \centering
        \includegraphics[width=\textwidth]{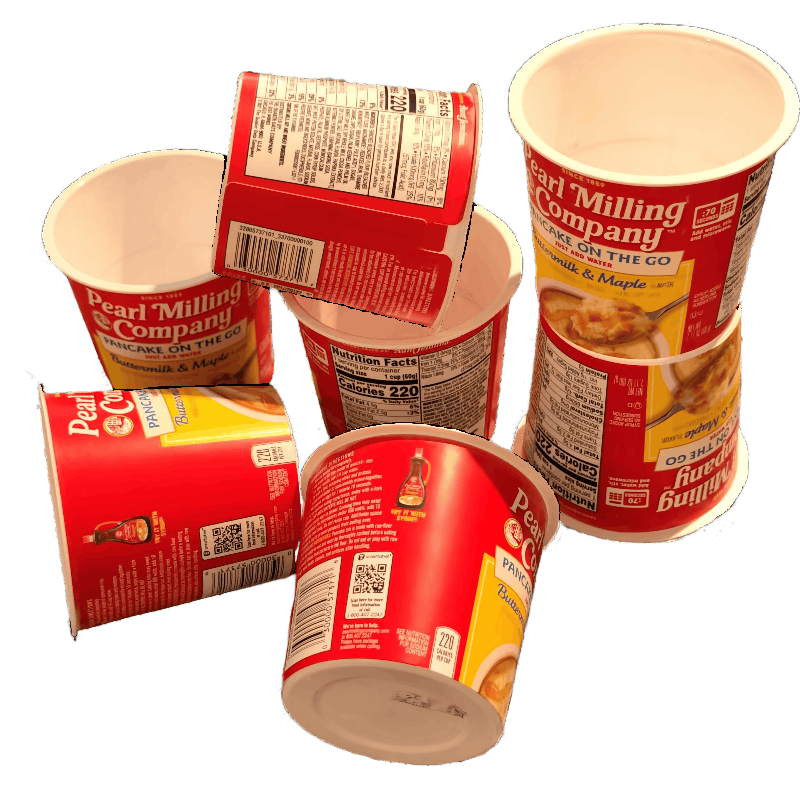}

    \end{subfigure}\hfill
    \begin{subfigure}{0.16\textwidth}
        \centering
        \includegraphics[width=\textwidth]{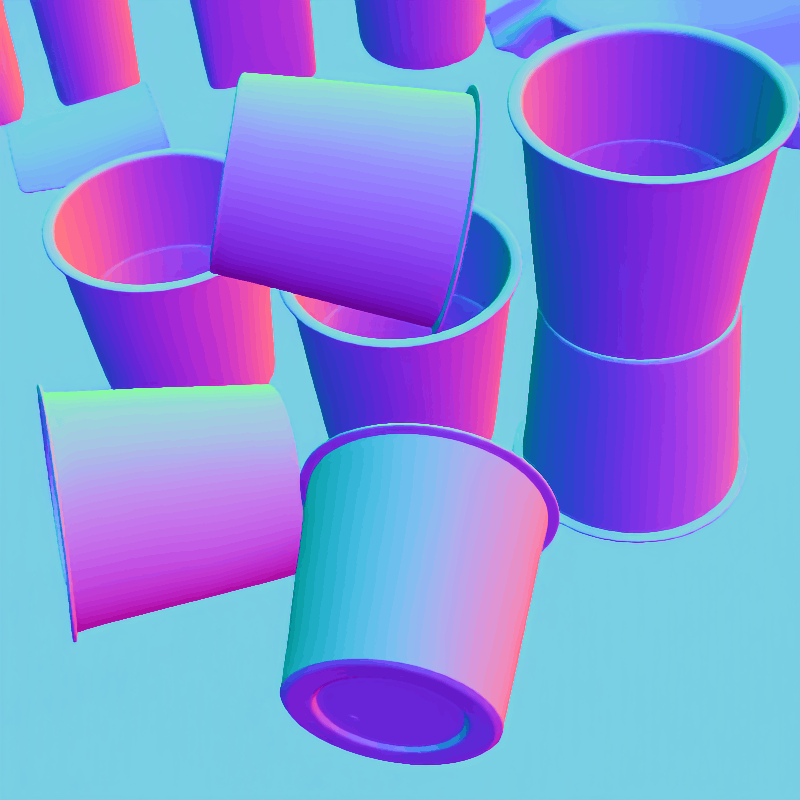}

    \end{subfigure}\hfill
        \begin{subfigure}{0.16\textwidth}
        \centering
        \includegraphics[width=\textwidth]{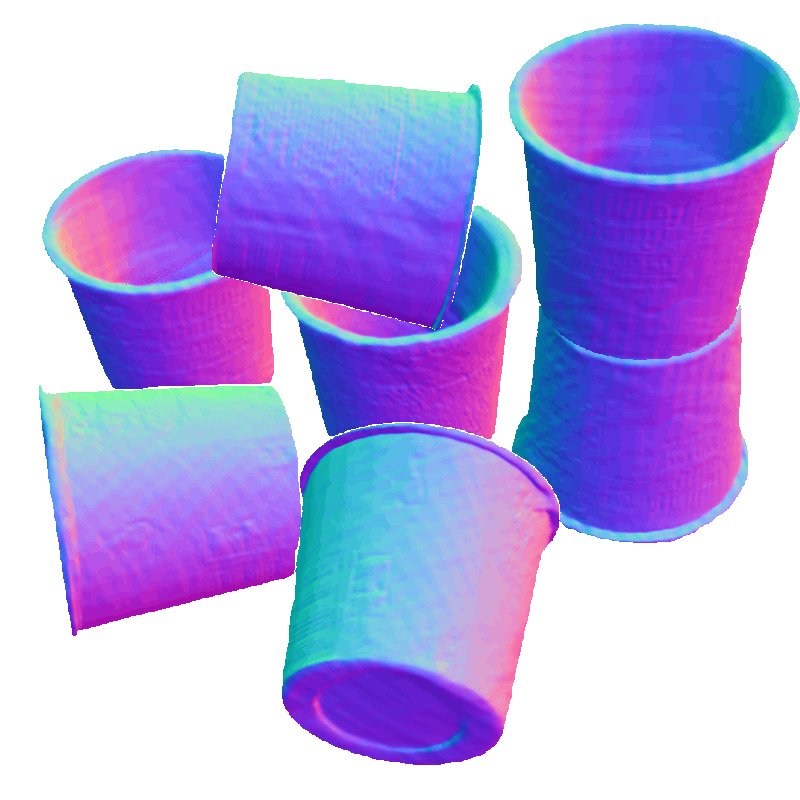}

    \end{subfigure}\hfill
            \begin{subfigure}{0.16\textwidth}
        \centering
        \includegraphics[width=\textwidth]{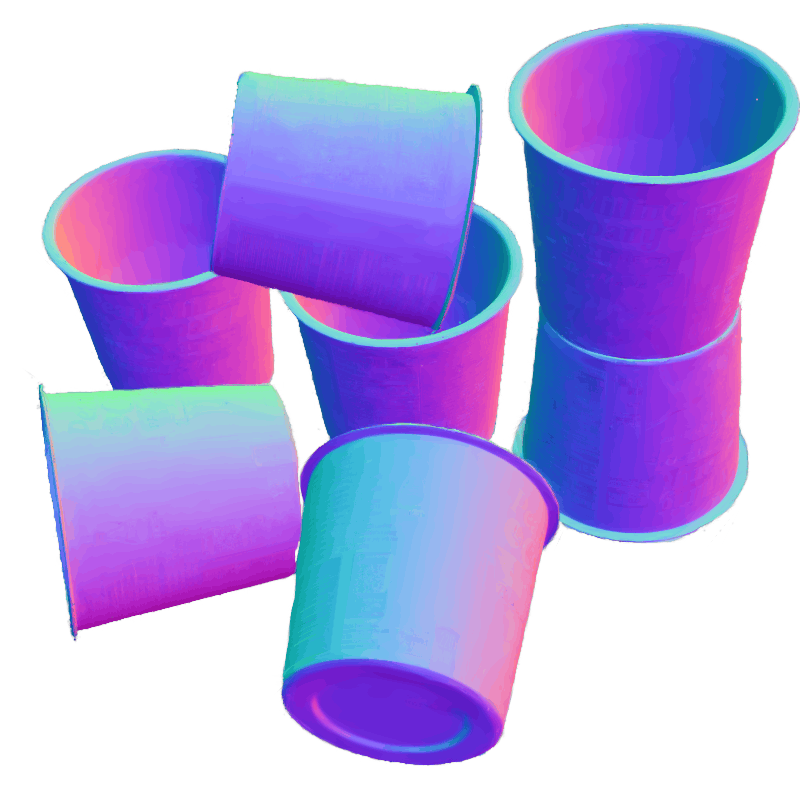}

    \end{subfigure}\\
    \raisebox{8ex}{\rotatebox{90}{Cheese\vphantom{g}} }\hfill
        \begin{subfigure}{0.16\textwidth}
        \centering
        \includegraphics[width=\textwidth]{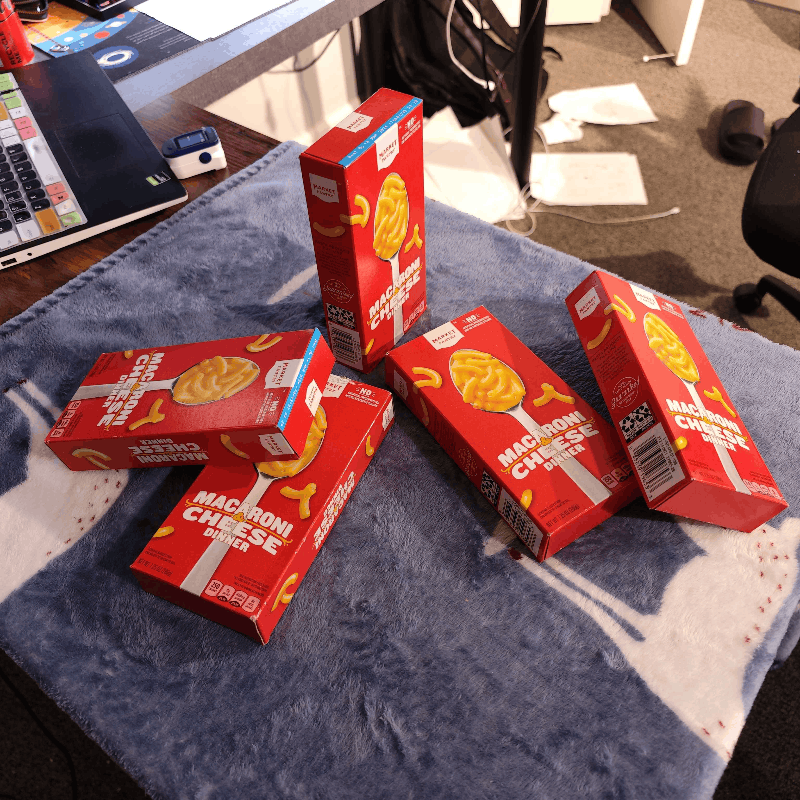}
    \end{subfigure}\hfill
    \begin{subfigure}{0.16\textwidth}
        \centering
        \includegraphics[width=\textwidth]{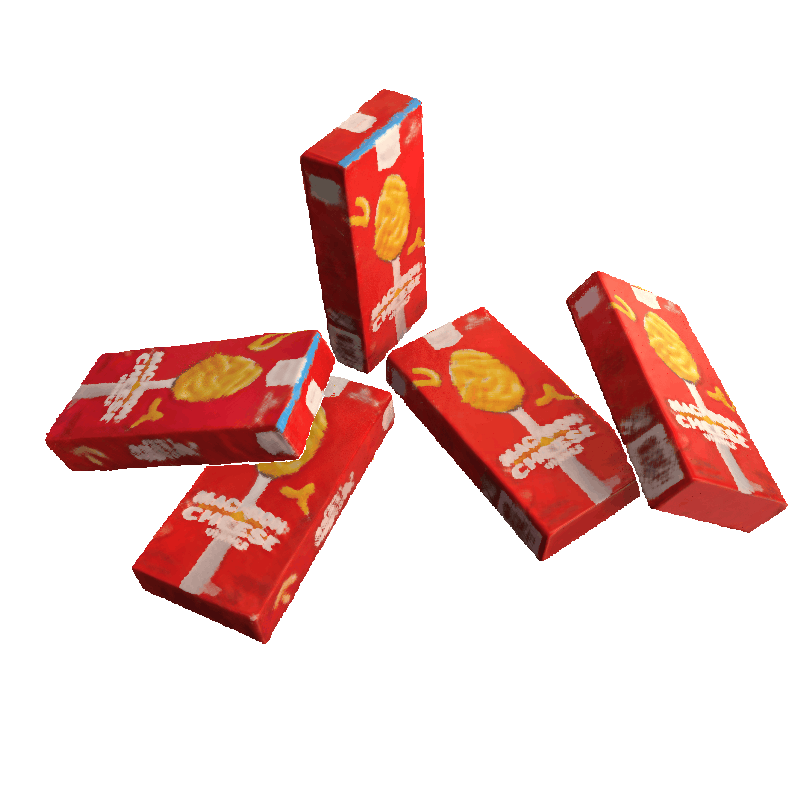}
    \end{subfigure}\hfill
        \begin{subfigure}{0.16\textwidth}
        \centering
        \includegraphics[width=\textwidth]{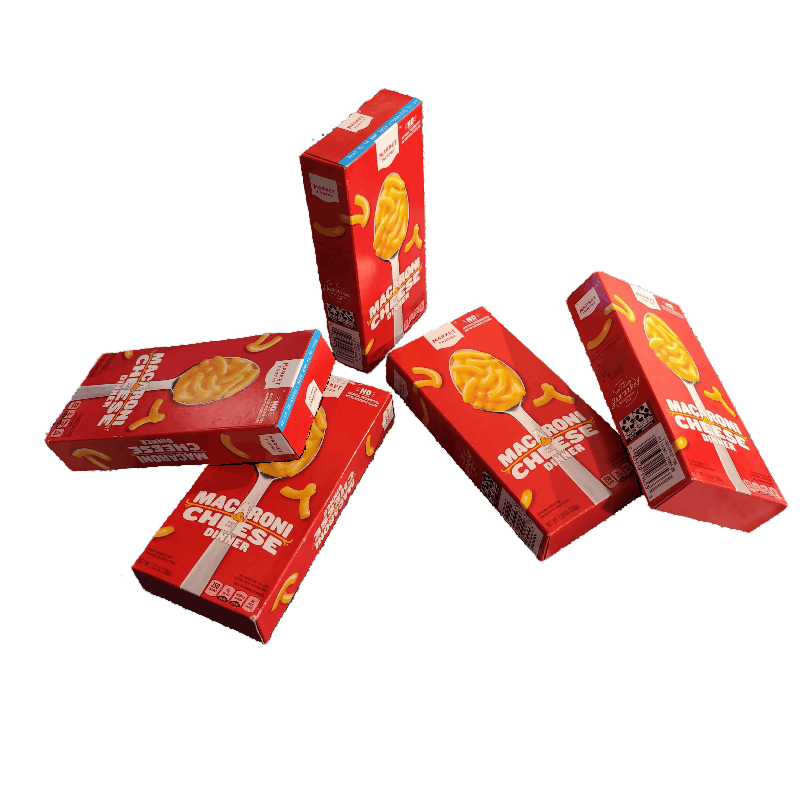}
    \end{subfigure}\hfill
    \begin{subfigure}{0.16\textwidth}
        \centering
        \includegraphics[width=\textwidth]{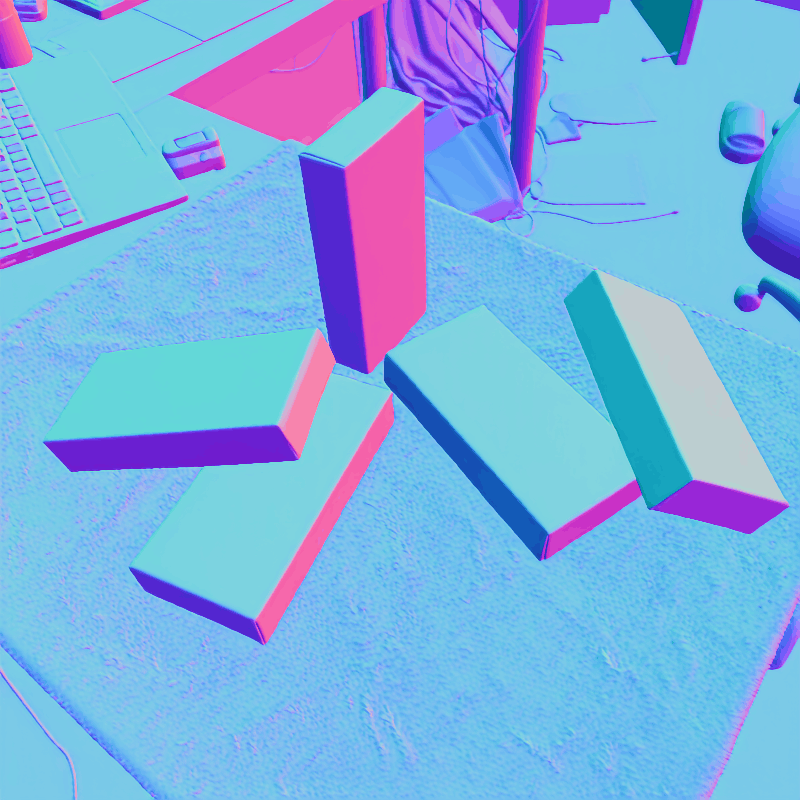}
    \end{subfigure}\hfill
        \begin{subfigure}{0.16\textwidth}
        \centering
        \includegraphics[width=\textwidth]{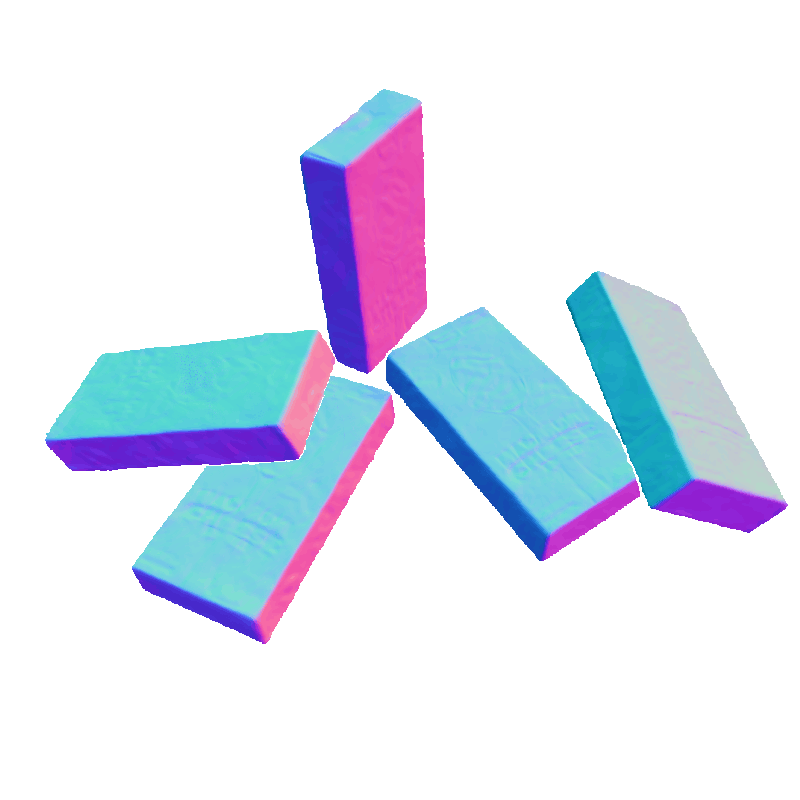}
    \end{subfigure}\hfill
            \begin{subfigure}{0.16\textwidth}
        \centering
        \includegraphics[width=\textwidth]{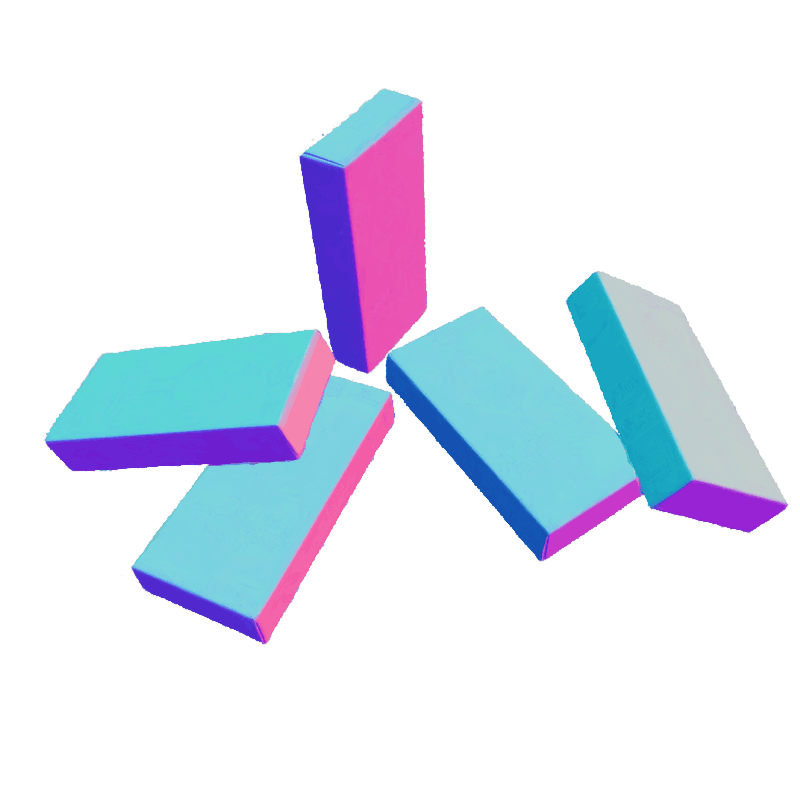}
    \end{subfigure}\\
    \raisebox{8ex}{\rotatebox{90}{Cola\vphantom{g}} }\hfill
    \begin{subfigure}{0.16\textwidth}
        \centering
        \includegraphics[width=\textwidth]{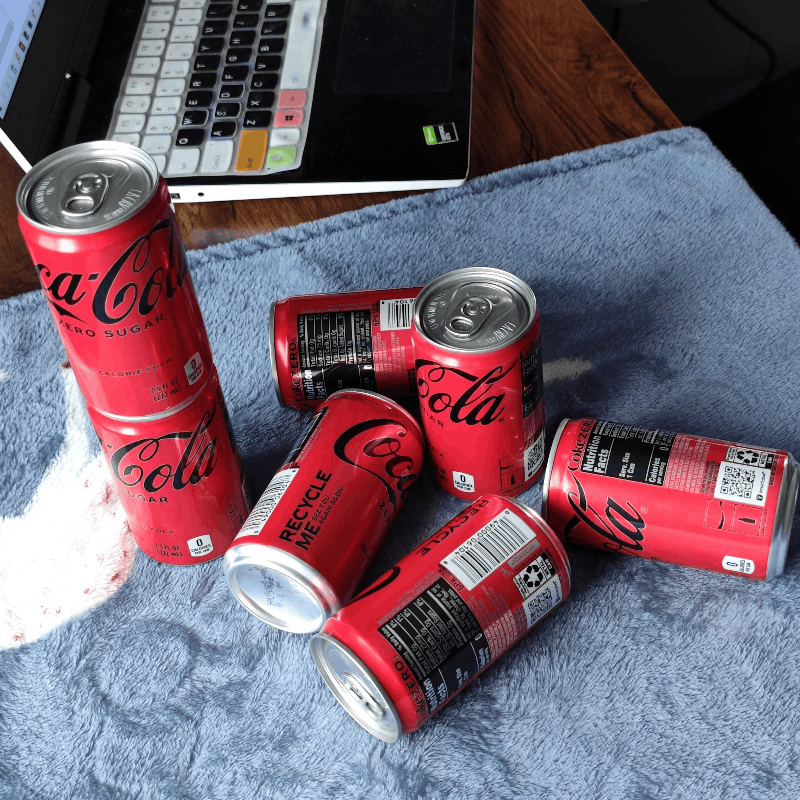}
    \end{subfigure}\hfill
    \begin{subfigure}{0.16\textwidth}
        \centering
        \includegraphics[width=\textwidth]{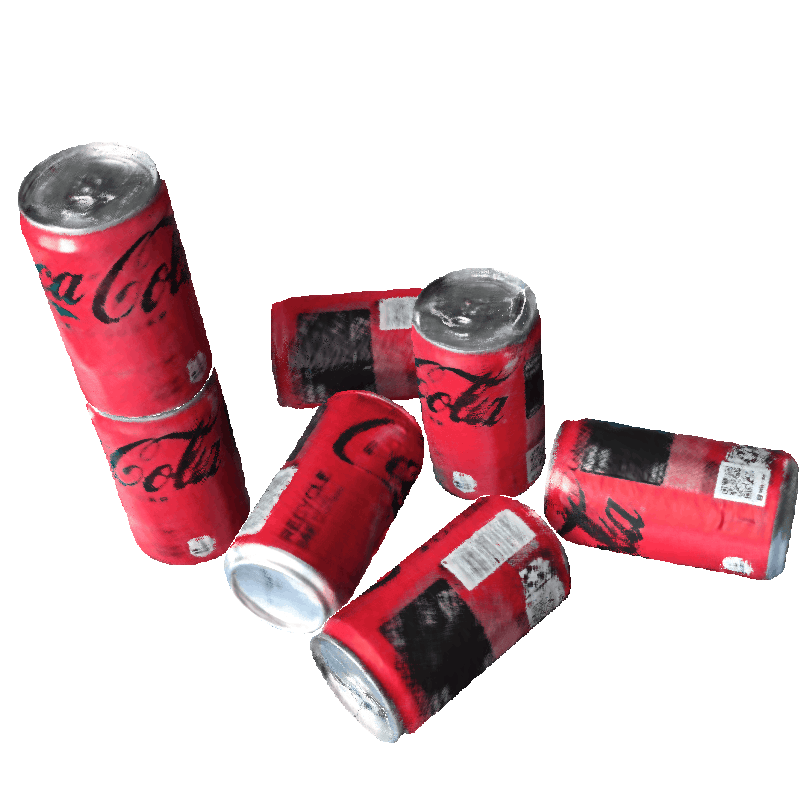}
    \end{subfigure}\hfill
        \begin{subfigure}{0.16\textwidth}
        \centering
        \includegraphics[width=\textwidth]{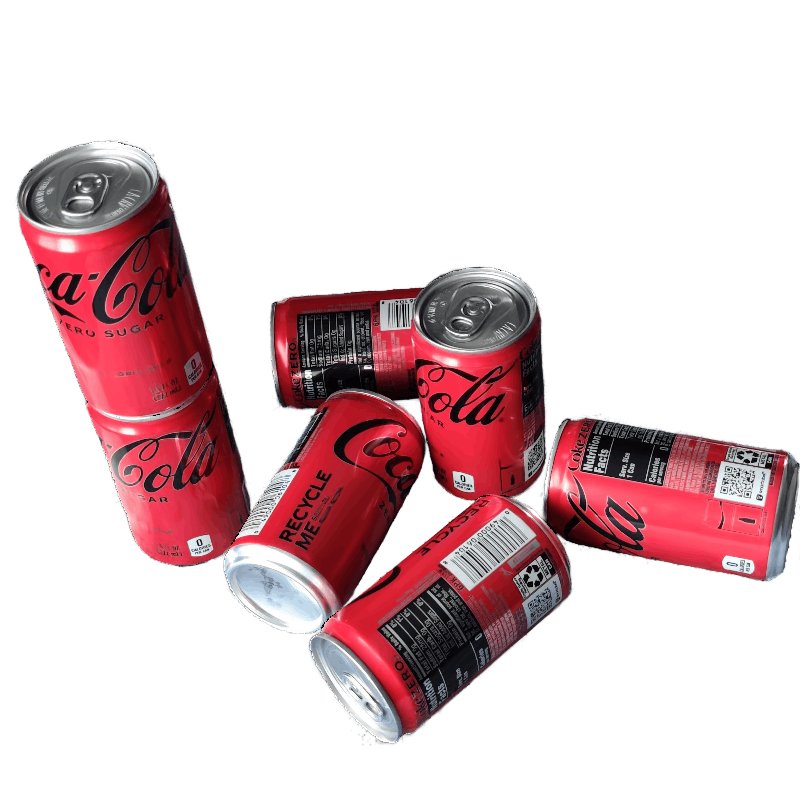}
    \end{subfigure}\hfill
    \begin{subfigure}{0.16\textwidth}
        \centering
        \includegraphics[width=\textwidth]{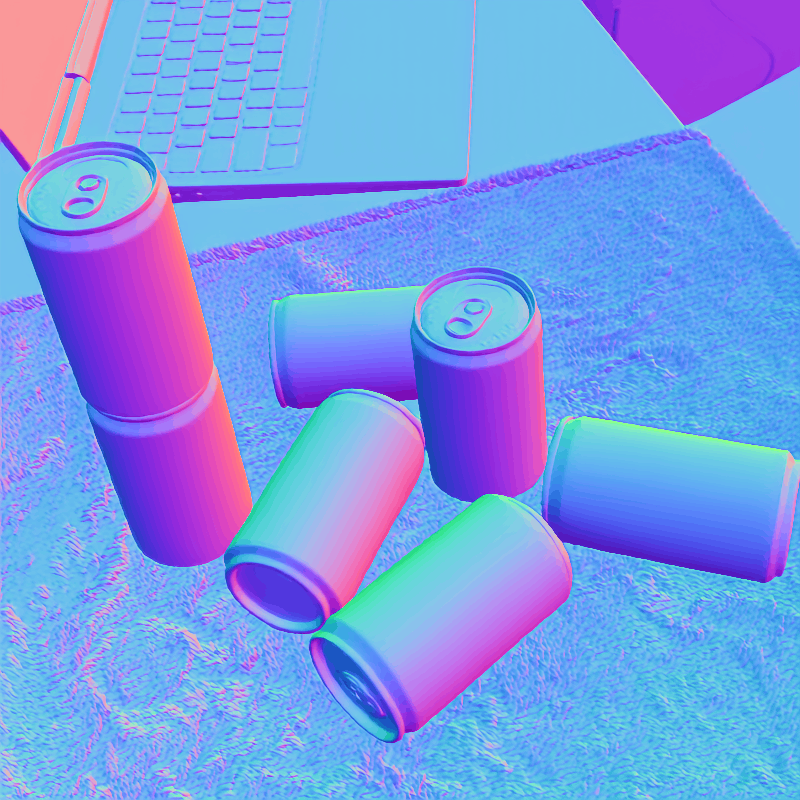}
    \end{subfigure}\hfill
        \begin{subfigure}{0.16\textwidth}
        \centering
        \includegraphics[width=\textwidth]{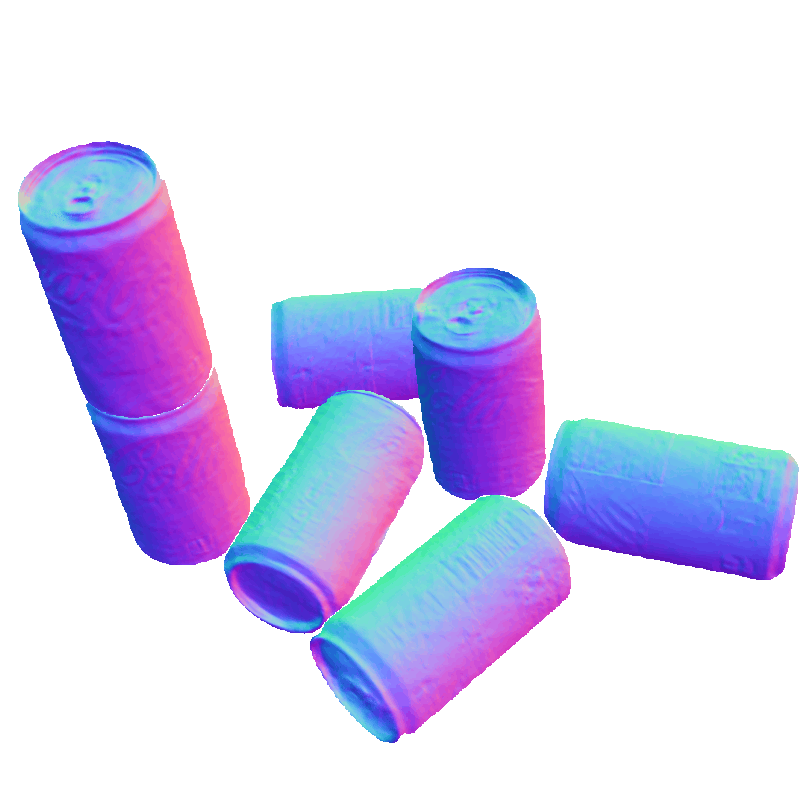}
    \end{subfigure}\hfill
            \begin{subfigure}{0.16\textwidth}
        \centering
        \includegraphics[width=\textwidth]{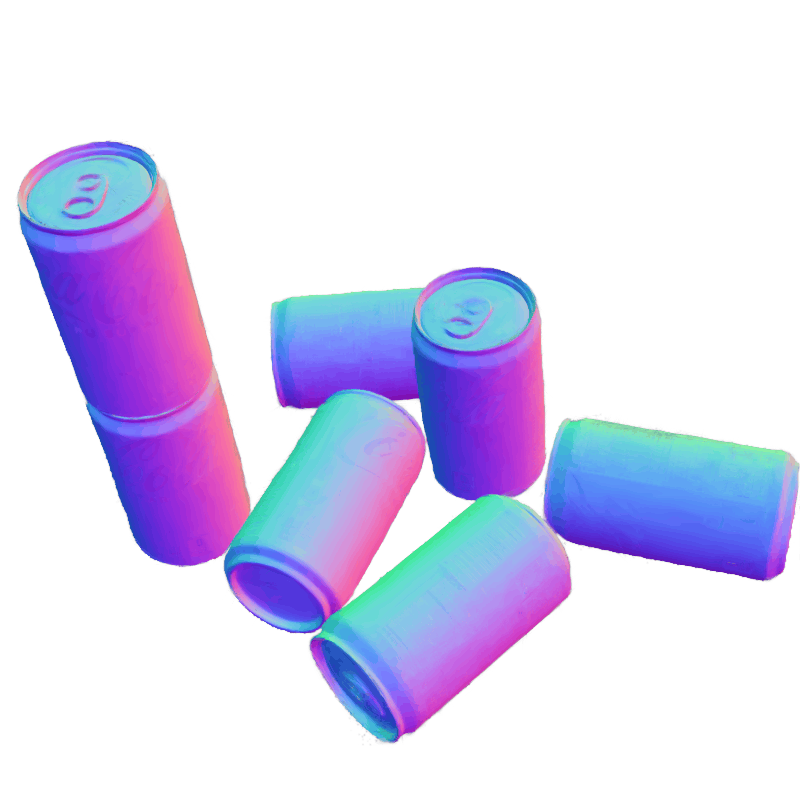}
    \end{subfigure}\\
    \raisebox{8ex}{\rotatebox{90}{Potato\vphantom{g}} }\hfill
        \begin{subfigure}{0.16\textwidth}
        \centering
        \includegraphics[width=\textwidth]{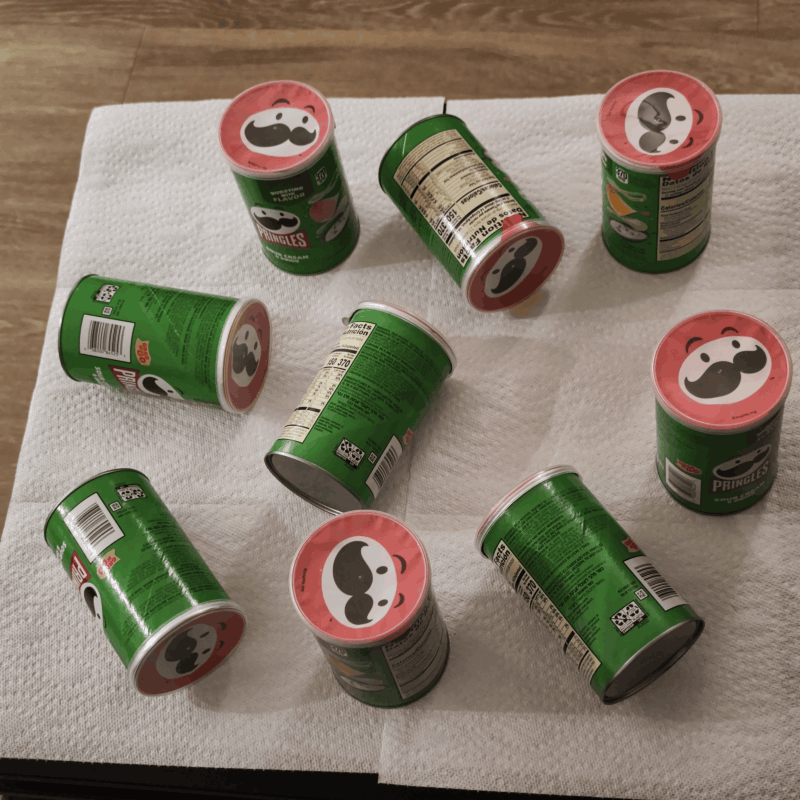}
    \end{subfigure}\hfill
    \begin{subfigure}{0.16\textwidth}
        \centering
        \includegraphics[width=\textwidth]{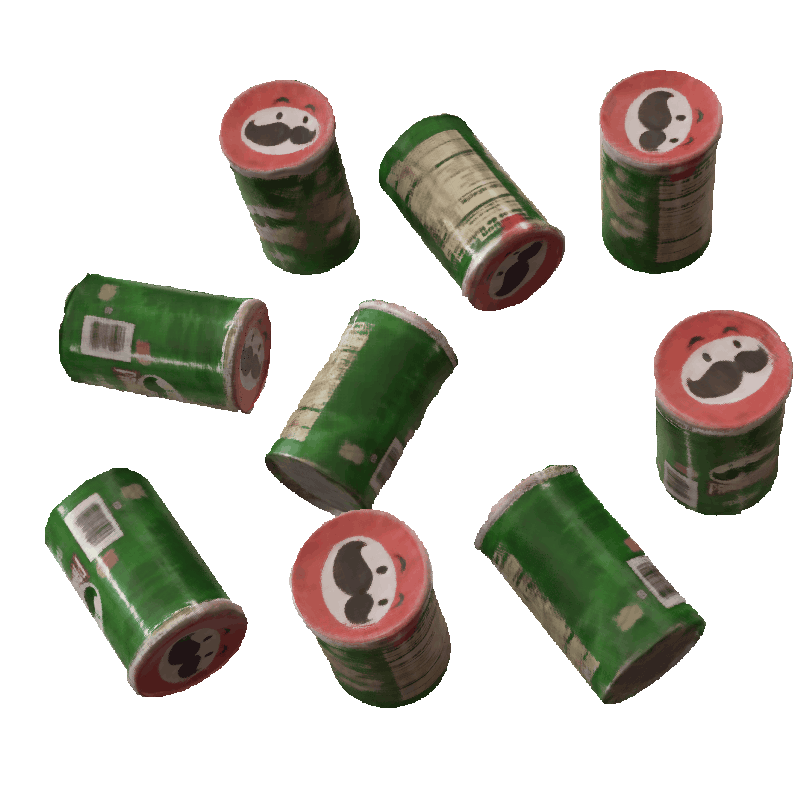}
    \end{subfigure}\hfill
        \begin{subfigure}{0.16\textwidth}
        \centering
        \includegraphics[width=\textwidth]{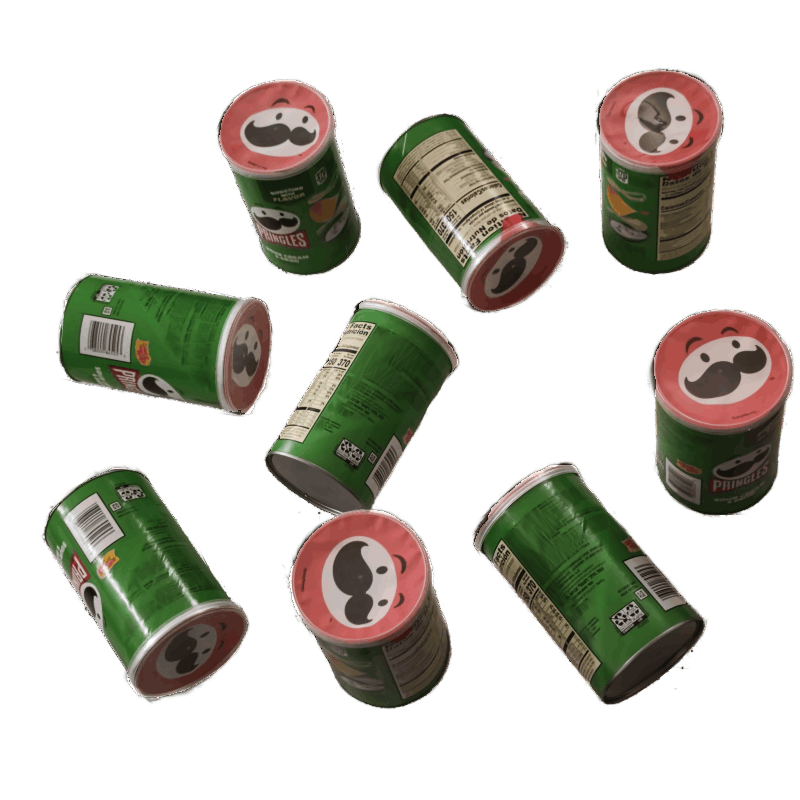}
    \end{subfigure}\hfill
    \begin{subfigure}{0.16\textwidth}
        \centering
        \includegraphics[width=\textwidth]{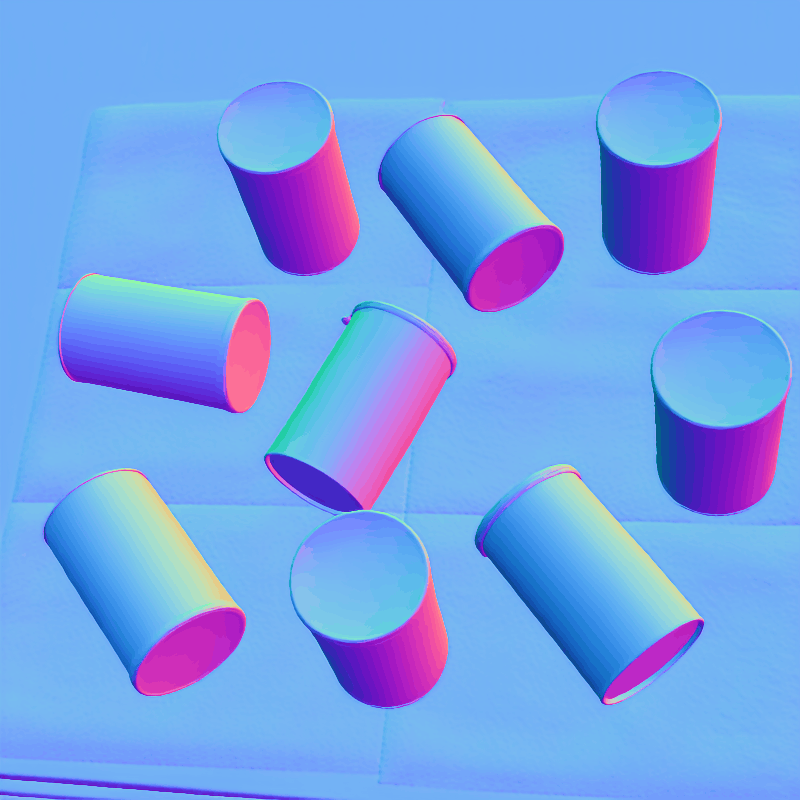}
    \end{subfigure}\hfill
        \begin{subfigure}{0.16\textwidth}
        \centering
        \includegraphics[width=\textwidth]{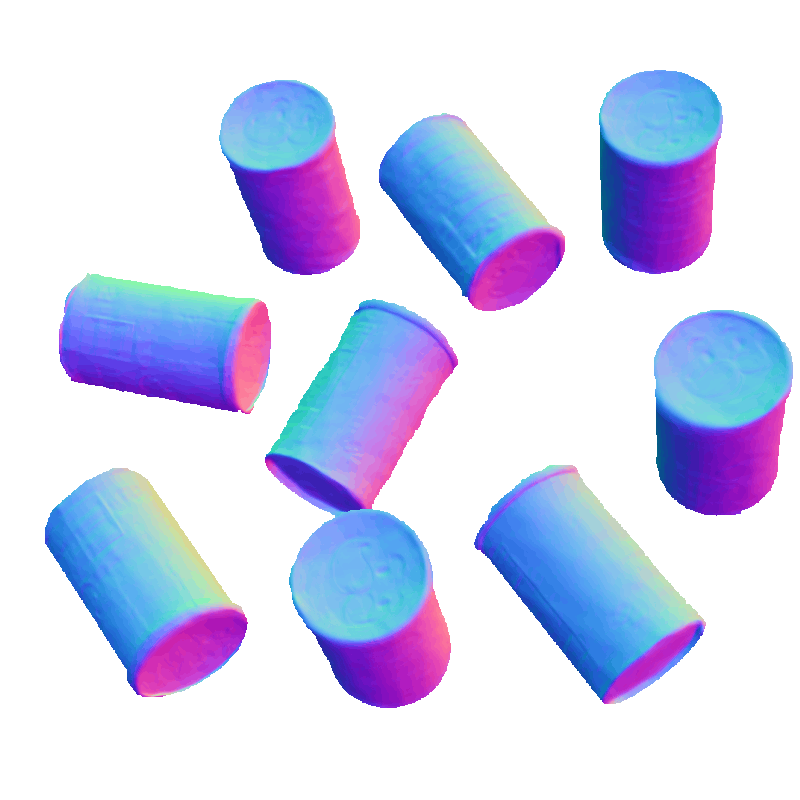}
    \end{subfigure}\hfill
            \begin{subfigure}{0.16\textwidth}
        \centering
        \includegraphics[width=\textwidth]{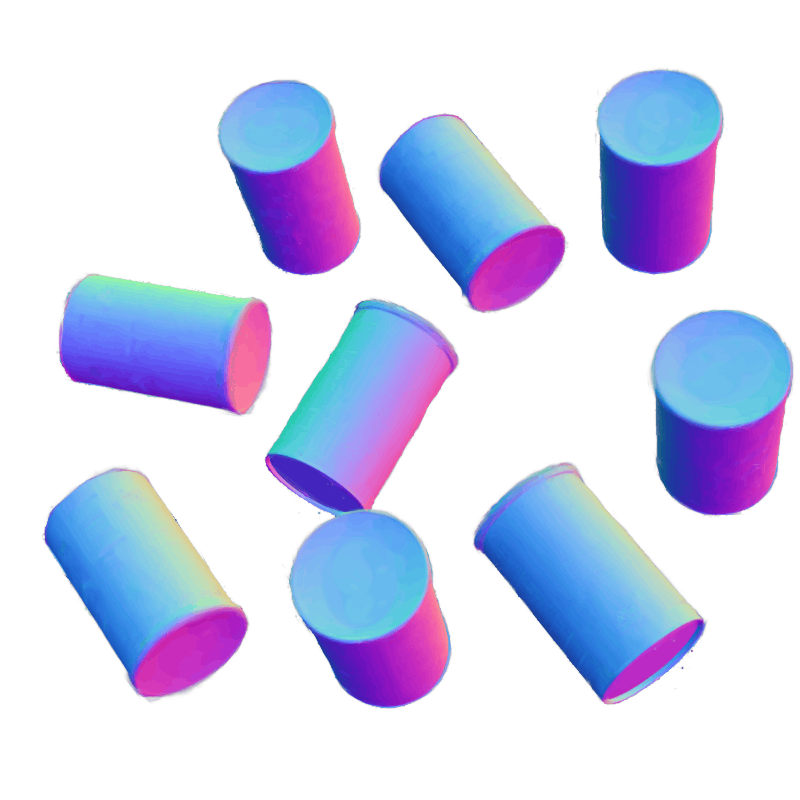}
    \end{subfigure}\\

   \raisebox{8ex}{\rotatebox{90}{Yogurt} }\hfill
    \begin{subfigure}{0.16\textwidth}
        \centering
        \includegraphics[width=\textwidth]{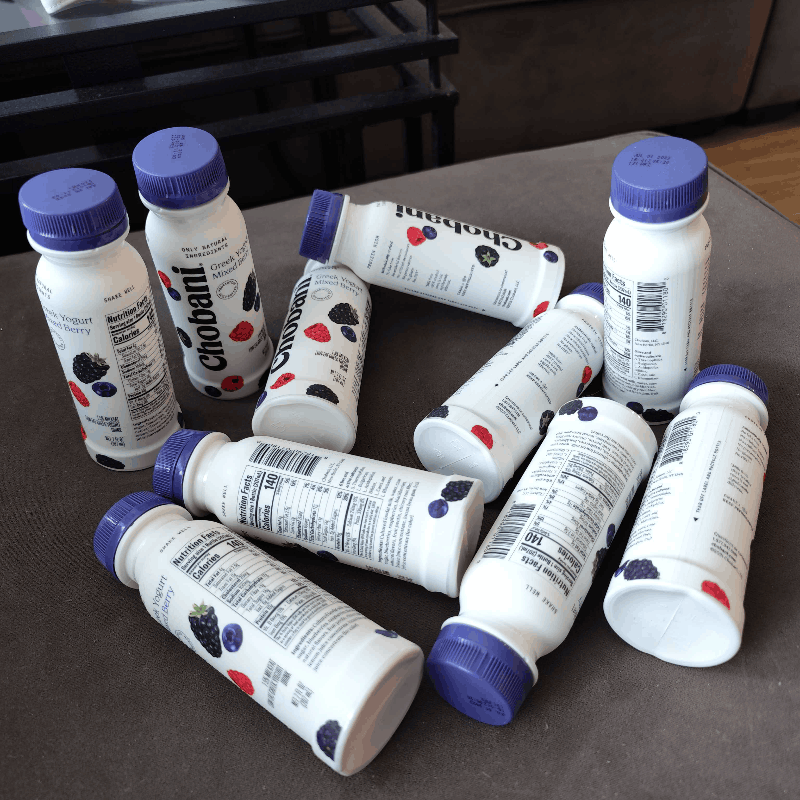}
       \caption*{Input image}
    \end{subfigure}\hfill
    \begin{subfigure}{0.16\textwidth}
        \centering
        \includegraphics[width=\textwidth]{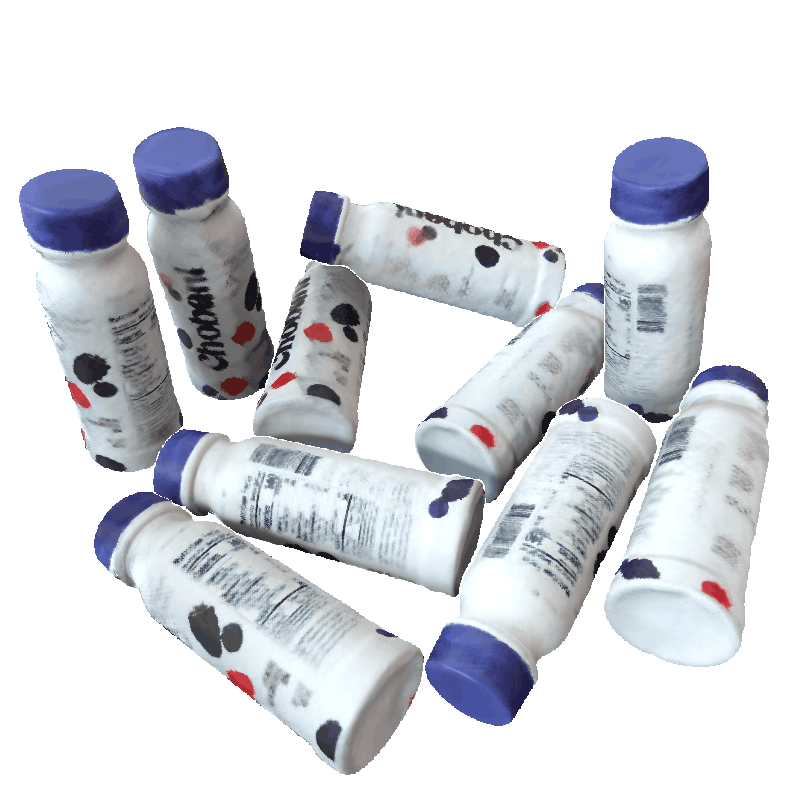}
        \caption*{RGB \they\unskip$^*$}
    \end{subfigure}\hfill
        \begin{subfigure}{0.16\textwidth}
        \centering
        \includegraphics[width=\textwidth]{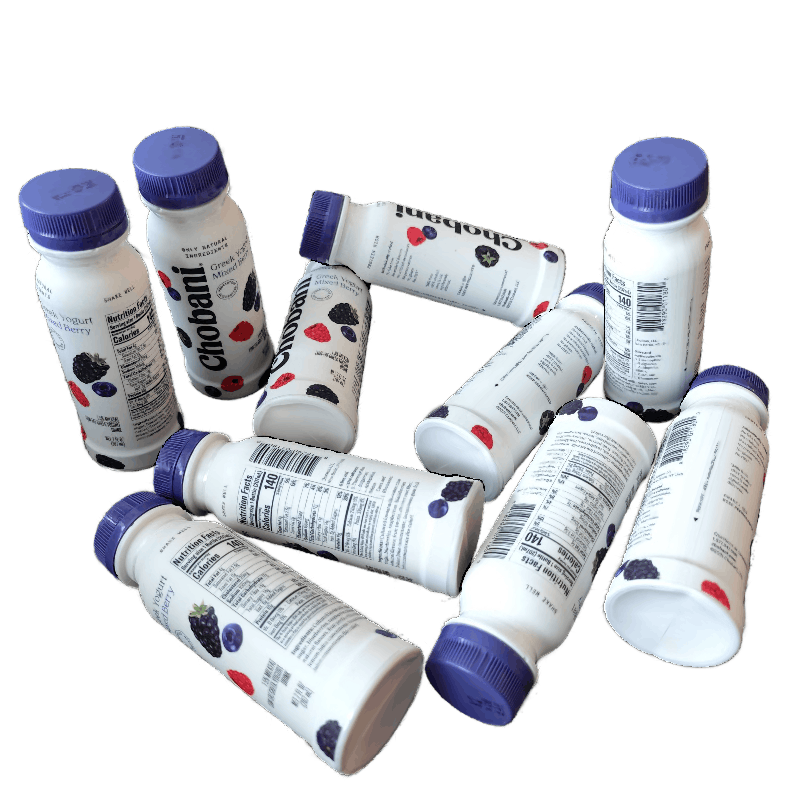}
        \caption*{RGB \method}
    \end{subfigure}\hfill
    \begin{subfigure}{0.16\textwidth}
        \centering
        \includegraphics[width=\textwidth]{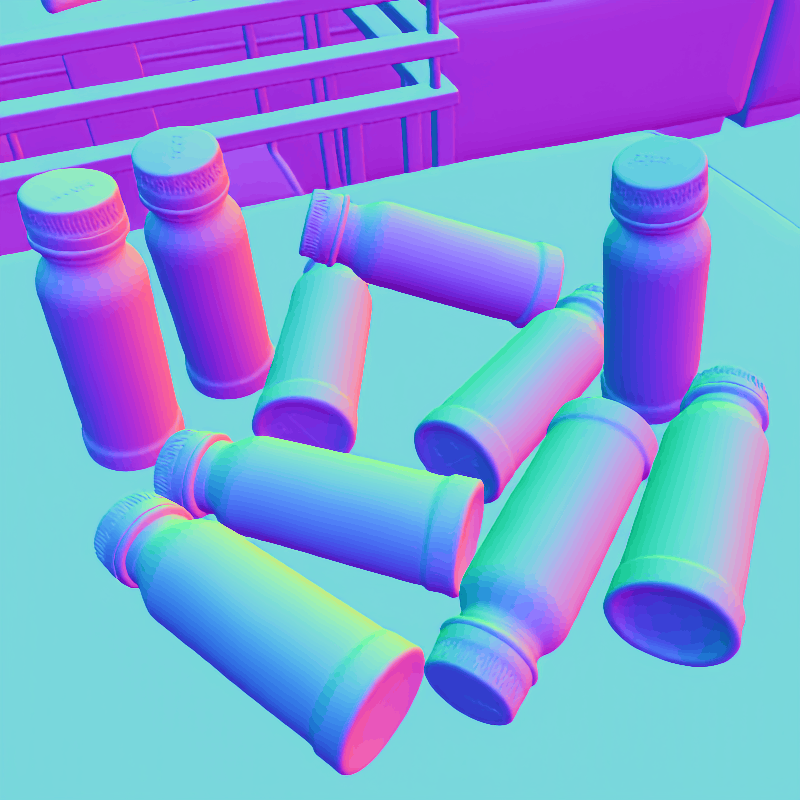}
        \caption*{Marigold V2 \cite{pavlovic2026marigold} }
    \end{subfigure}\hfill
        \begin{subfigure}{0.16\textwidth}
        \centering
        \includegraphics[width=\textwidth]{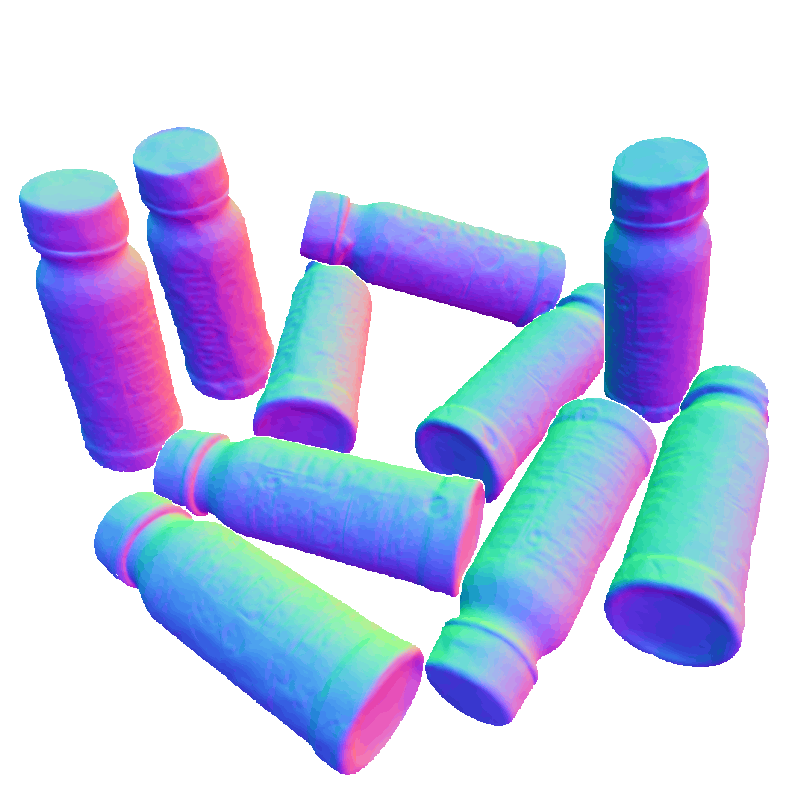}
        \caption*{\they\unskip$^*$ normals}
    \end{subfigure}\hfill
            \begin{subfigure}{0.16\textwidth}
        \centering
        \includegraphics[width=\textwidth]{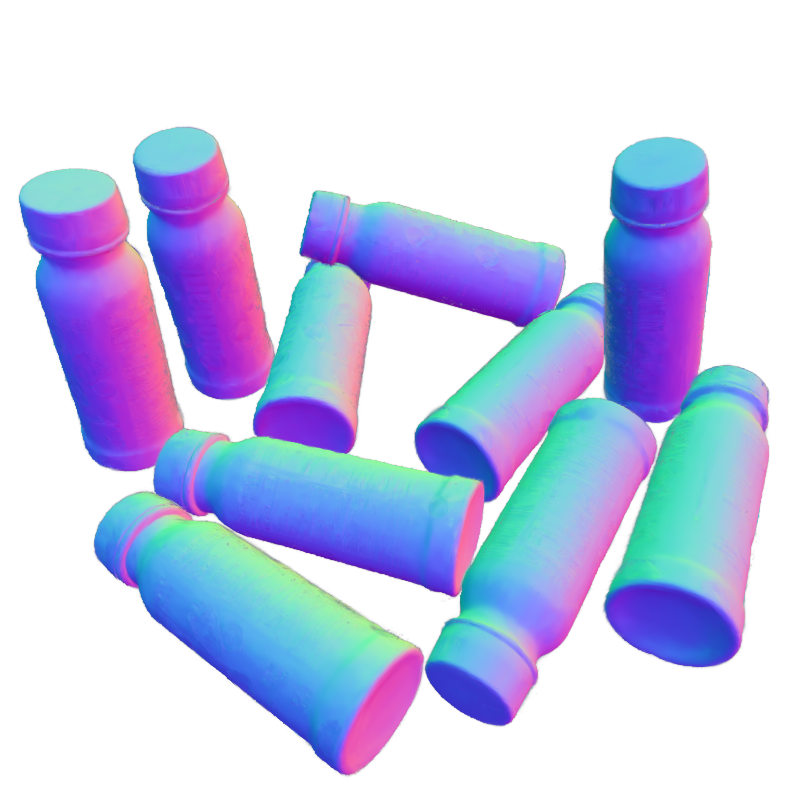}
        \caption*{\method normals}
    \end{subfigure}\\

    \caption{Qualitative results of Marigold V2 \cite{pavlovic2026marigold}, \they\unskip$^*$, and \method on real-world multi-instance images of Dup-real. Zoom in for more detail.}
    \label{fig:qualitative_comparison_real}
\end{figure*}

\paragraph{Results on Dup.} \cref{fig:qualitative_comparison_real} compares \method with \they\unskip$^*$ on scenes of the Dup-real dataset. \method renders clean normal images, that adhere better to the predicted normals of Marigold V2 than \they\unskip$^*$, while enforcing consistency between instances. At the same time, the rendered RGB images of \method are hardly distinguishable from the input image. 

\begin{figure*}
    \centering
    \raisebox{8ex}{\rotatebox{90}{Box\vphantom{g}} }\hfill
    \begin{subfigure}{0.16\textwidth}
        \centering
        \includegraphics[width=\textwidth]{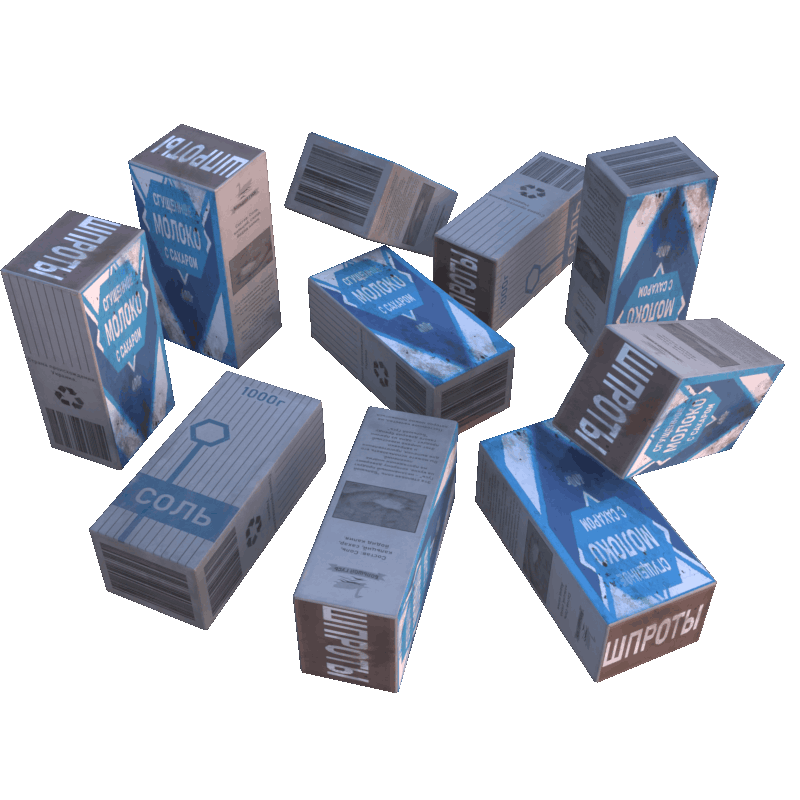}
    \end{subfigure}\hfill
    \begin{subfigure}{0.16\textwidth}
        \centering
        \includegraphics[width=\textwidth]{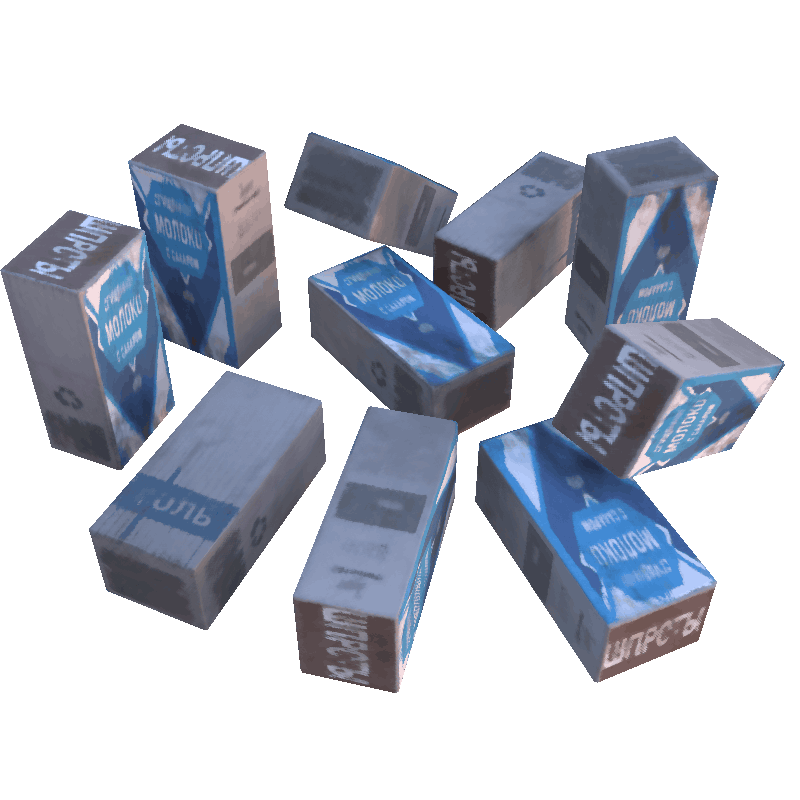}
    \end{subfigure}\hfill
    \begin{subfigure}{0.16\textwidth}
        \centering
        \includegraphics[width=\textwidth]{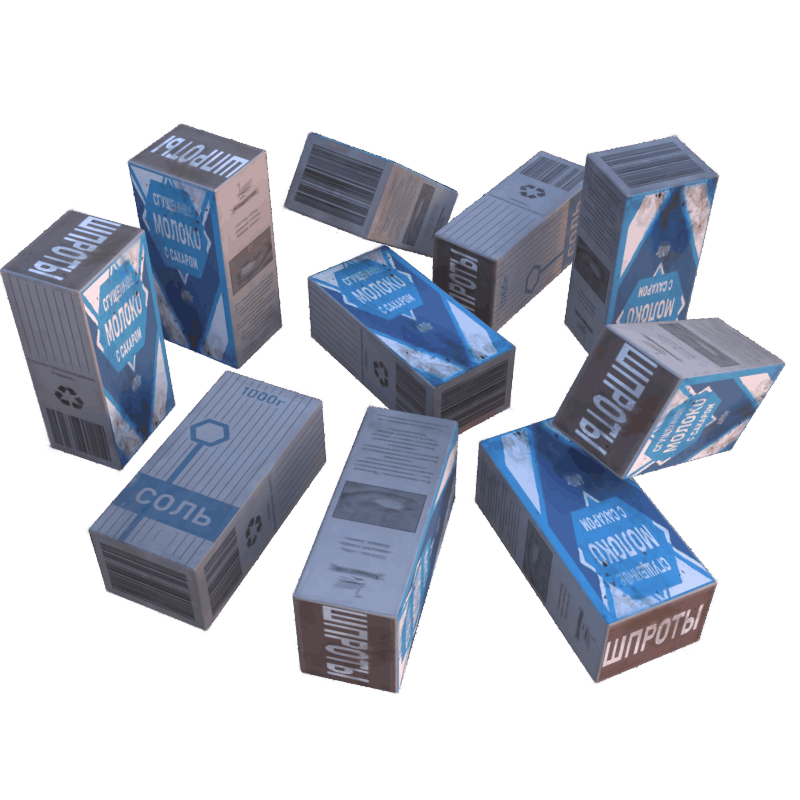}
    \end{subfigure}\hfill
    \begin{subfigure}{0.16\textwidth}
        \centering
        \includegraphics[width=\textwidth]{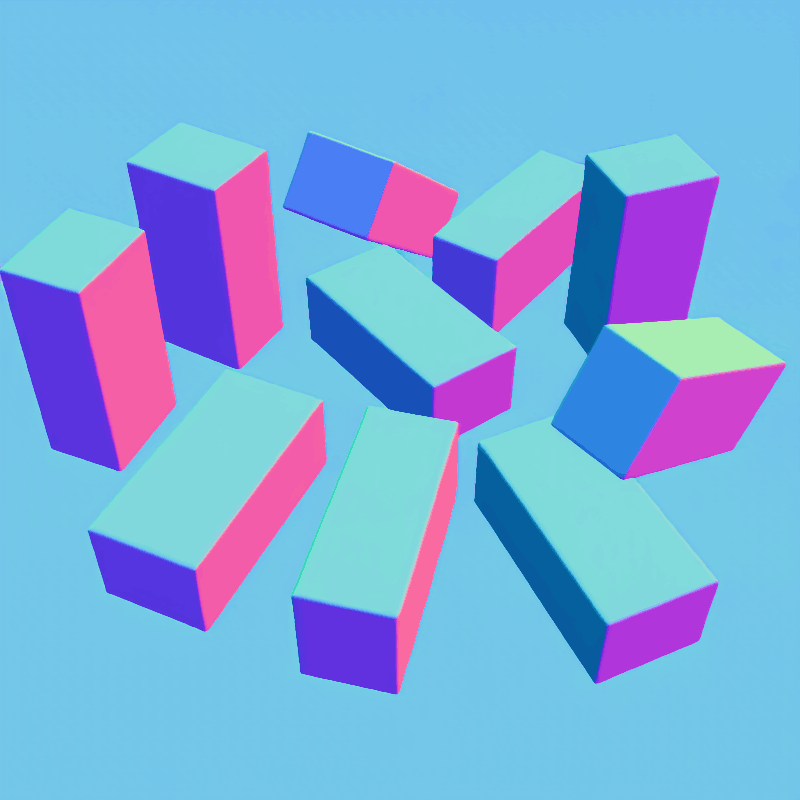}
    \end{subfigure}\hfill
    \begin{subfigure}{0.16\textwidth}
        \centering
        \includegraphics[width=\textwidth]{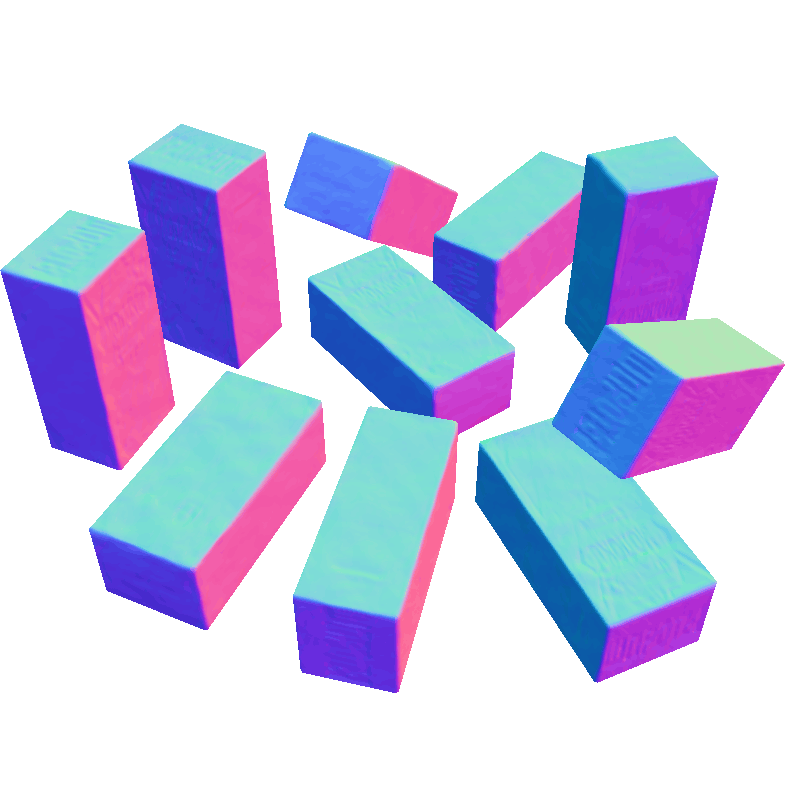}
    \end{subfigure}\hfill
    \begin{subfigure}{0.16\textwidth}
        \centering
        \includegraphics[width=\textwidth]{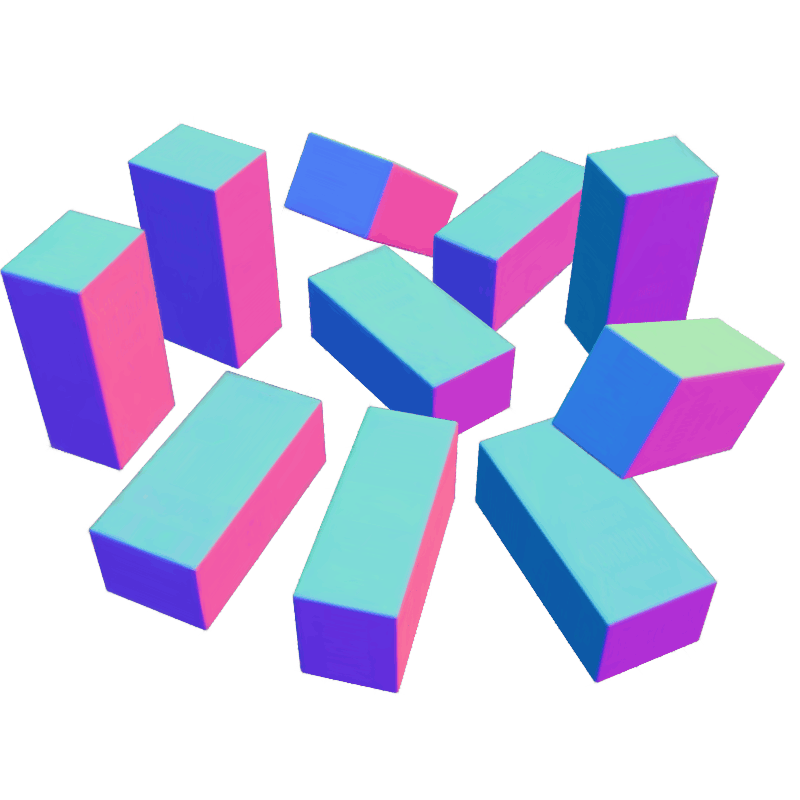}
    \end{subfigure}\\
 \raisebox{8ex}{\rotatebox{90}{Cash\vphantom{g}} }\hfill 
    \begin{subfigure}{0.16\textwidth}
        \centering
        \includegraphics[width=\textwidth]{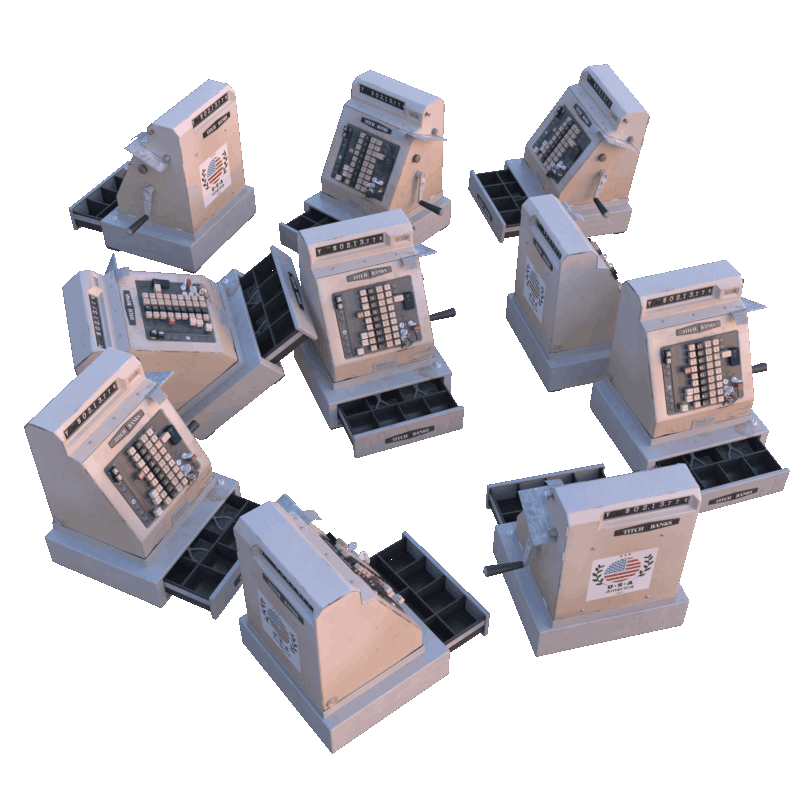}
    \end{subfigure}\hfill
    \begin{subfigure}{0.16\textwidth}
        \centering
        \includegraphics[width=\textwidth]{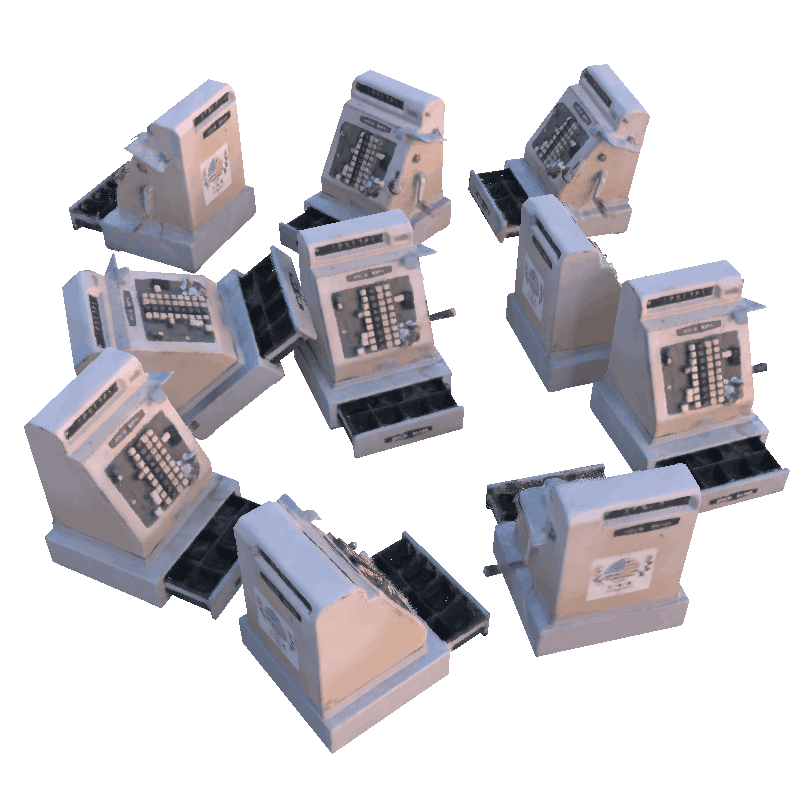}
    \end{subfigure}\hfill
    \begin{subfigure}{0.16\textwidth}
        \centering
        \includegraphics[width=\textwidth]{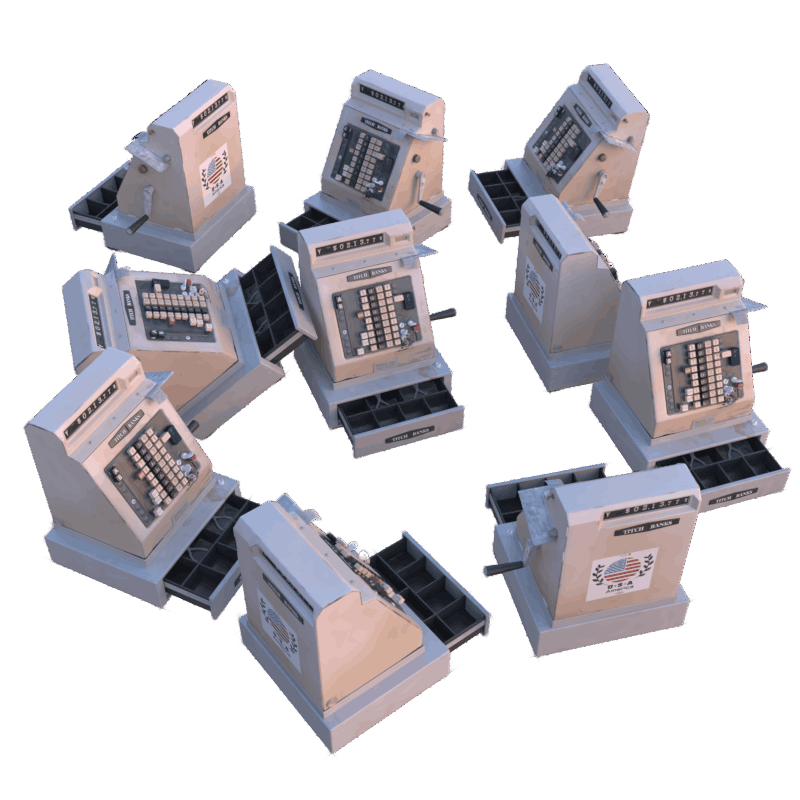}
    \end{subfigure}\hfill
    \begin{subfigure}{0.16\textwidth}
        \centering
        \includegraphics[width=\textwidth]{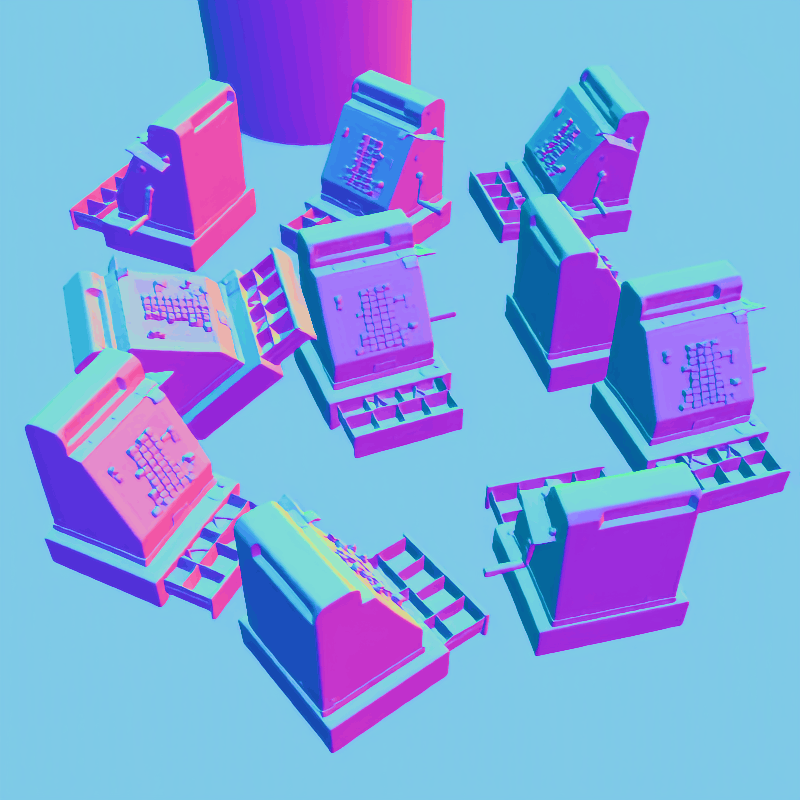}
    \end{subfigure}\hfill
    \begin{subfigure}{0.16\textwidth}
        \centering
        \includegraphics[width=\textwidth]{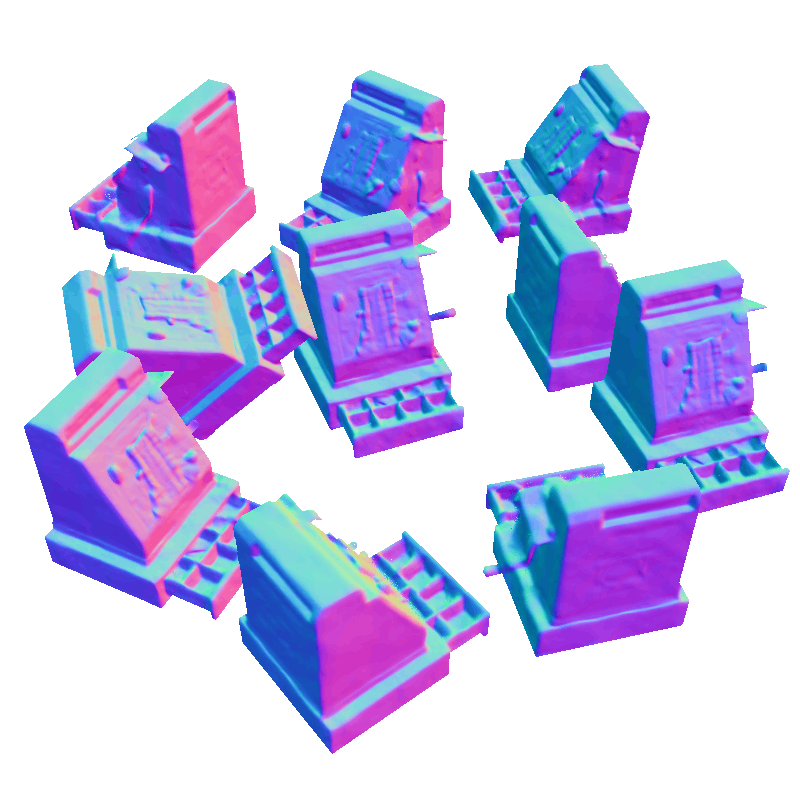}
    \end{subfigure}\hfill
    \begin{subfigure}{0.16\textwidth}
        \centering
        \includegraphics[width=\textwidth]{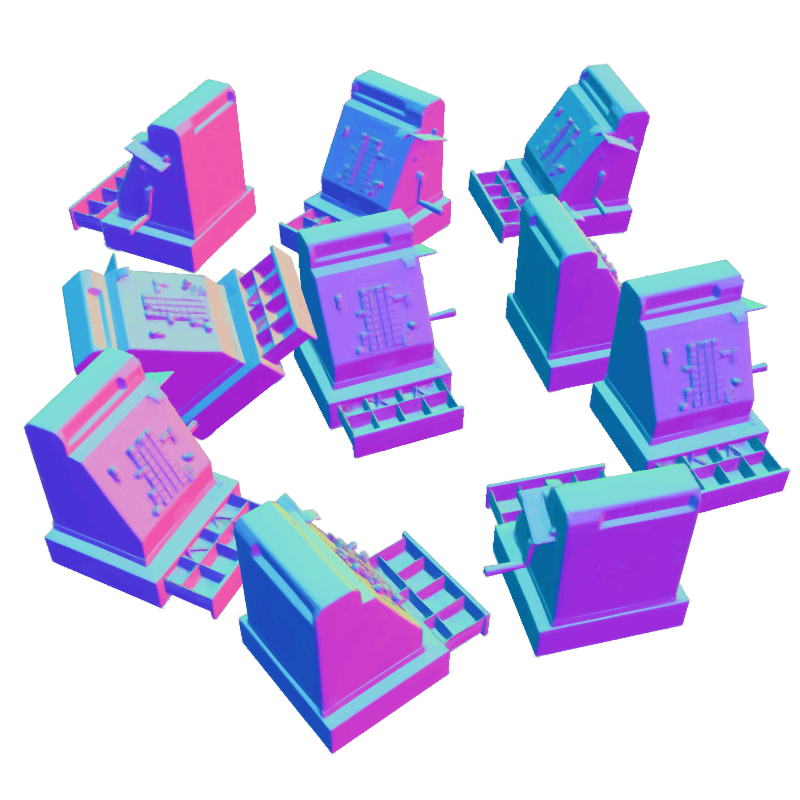}
    \end{subfigure}\\
      \raisebox{8ex}{\rotatebox{90}{Cleaner\vphantom{g}} }\hfill
    \begin{subfigure}{0.16\textwidth}
        \centering
        \includegraphics[width=\textwidth]{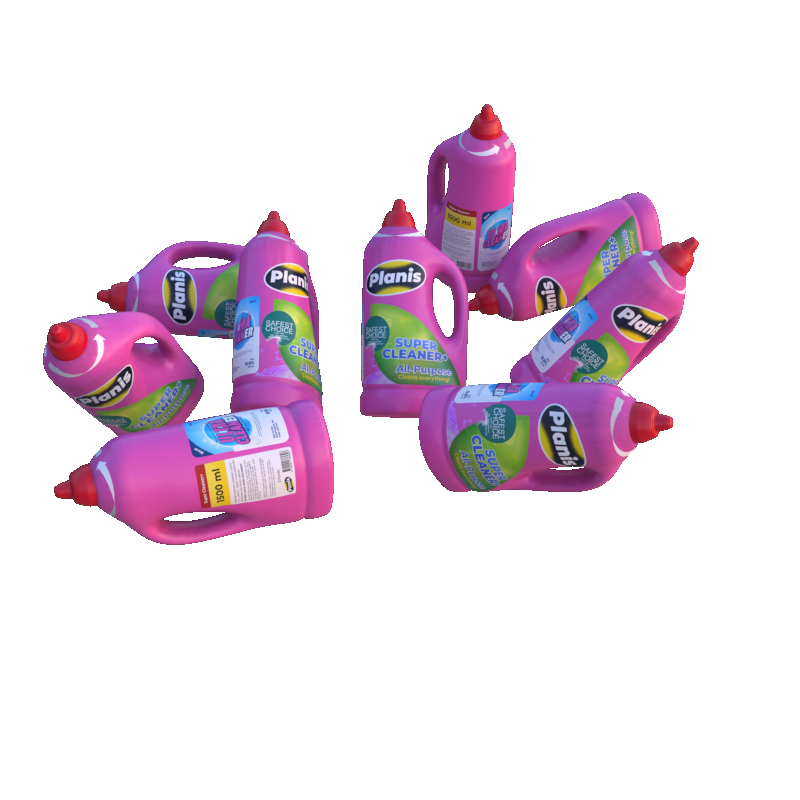}
    \end{subfigure}\hfill
    \begin{subfigure}{0.16\textwidth}
        \centering
        \includegraphics[width=\textwidth]{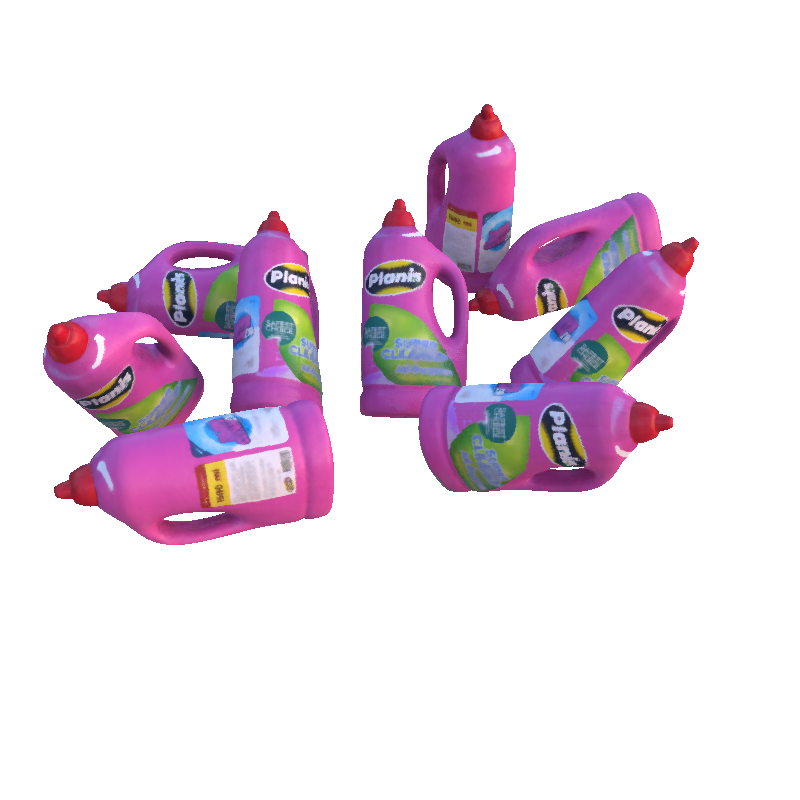}
    \end{subfigure}\hfill
    \begin{subfigure}{0.16\textwidth}
        \centering
        \includegraphics[width=\textwidth]{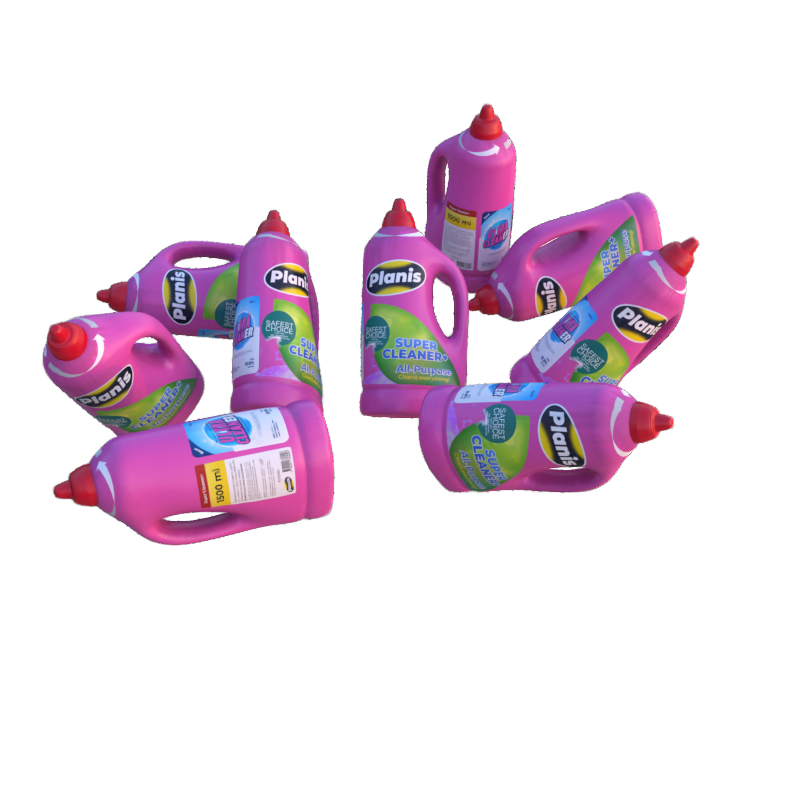}
    \end{subfigure}\hfill
    \begin{subfigure}{0.16\textwidth}
        \centering
        \includegraphics[width=\textwidth]{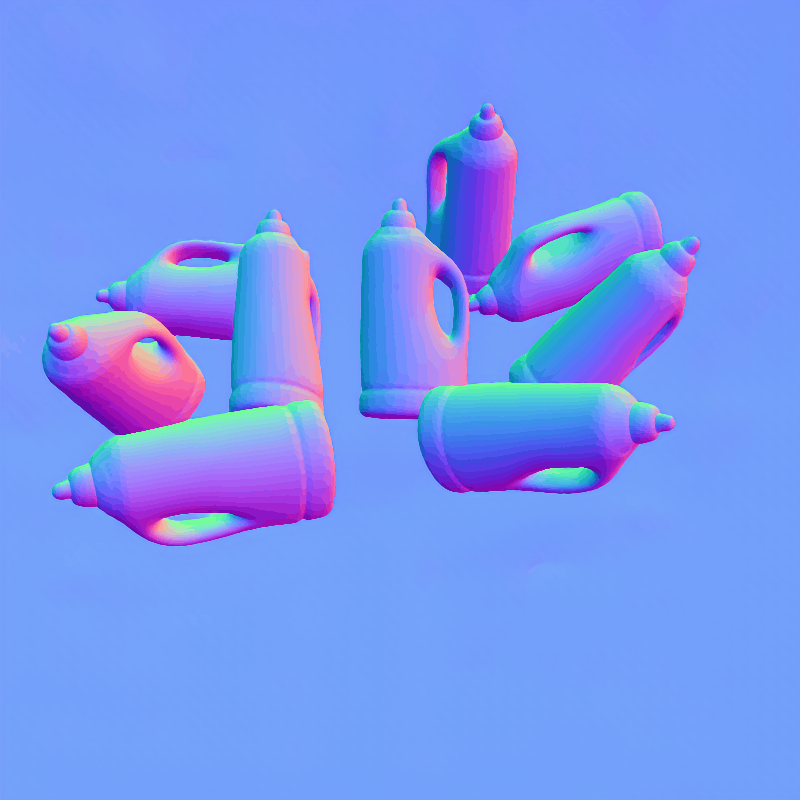}
    \end{subfigure}\hfill
    \begin{subfigure}{0.16\textwidth}
        \centering
        \includegraphics[width=\textwidth]{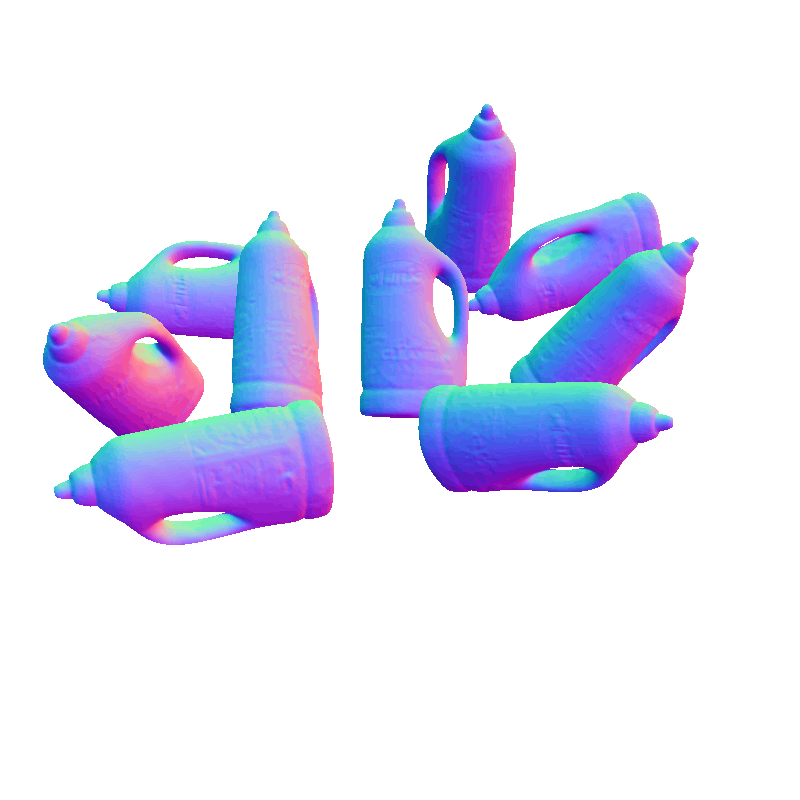}
    \end{subfigure}\hfill
    \begin{subfigure}{0.16\textwidth}
        \centering
        \includegraphics[width=\textwidth]{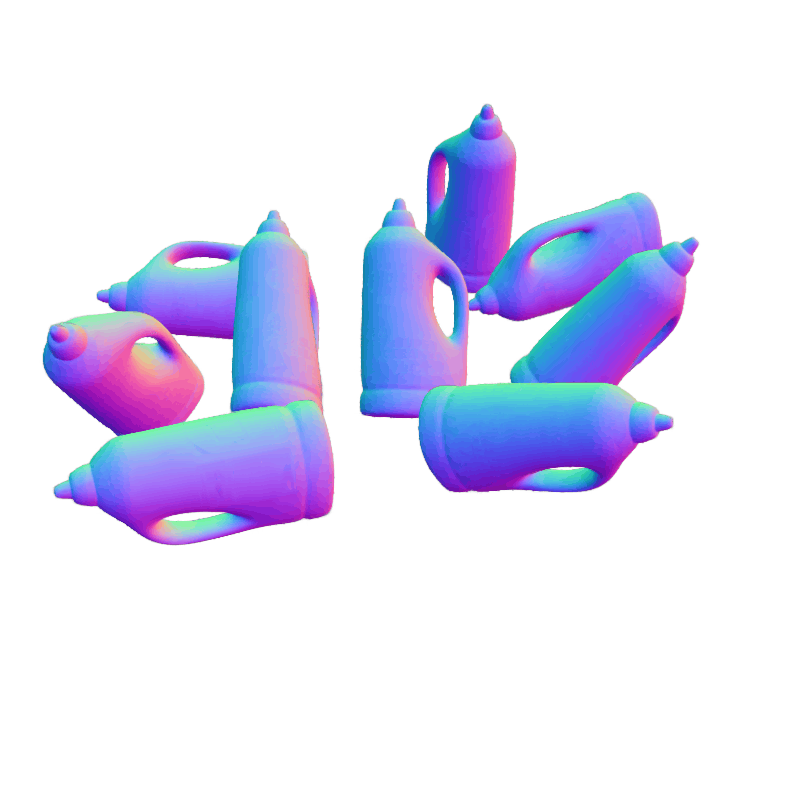}
    \end{subfigure}\\
 \raisebox{8ex}{\rotatebox{90}{Clock\vphantom{g}} }\hfill
 \begin{subfigure}{0.16\textwidth}
        \centering
        \includegraphics[width=\textwidth]{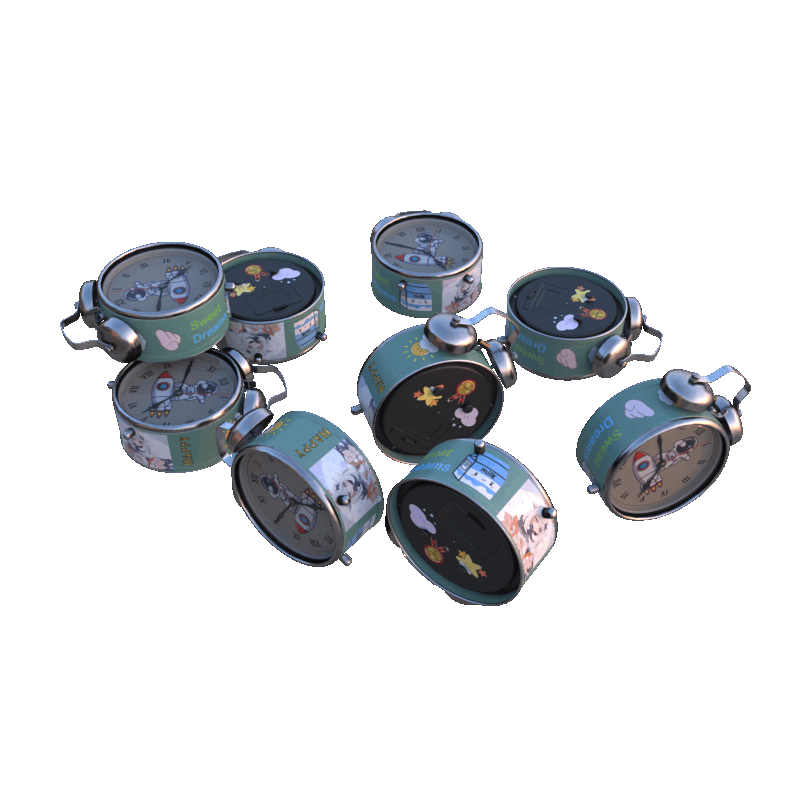}
    \end{subfigure}\hfill
    \begin{subfigure}{0.16\textwidth}
        \centering
        \includegraphics[width=\textwidth]{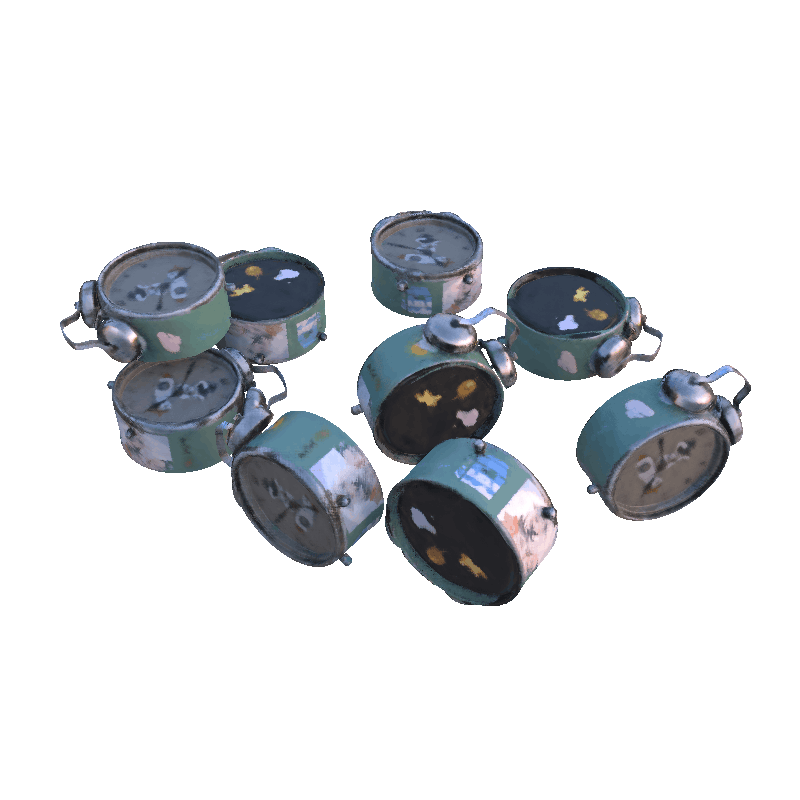}
    \end{subfigure}\hfill
    \begin{subfigure}{0.16\textwidth}
        \centering
        \includegraphics[width=\textwidth]{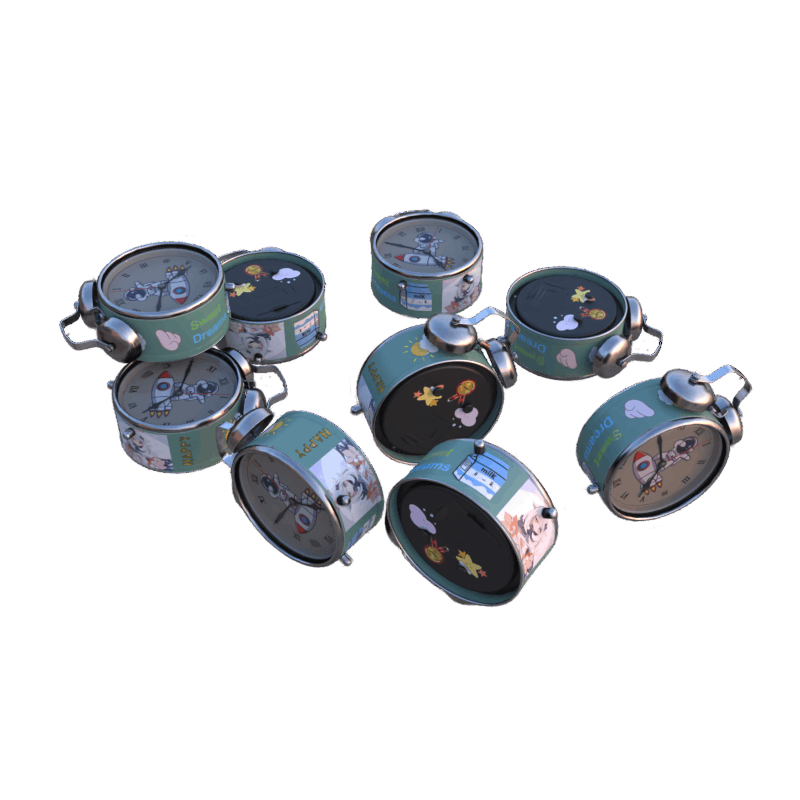}
    \end{subfigure}\hfill
    \begin{subfigure}{0.16\textwidth}
        \centering
        \includegraphics[width=\textwidth]{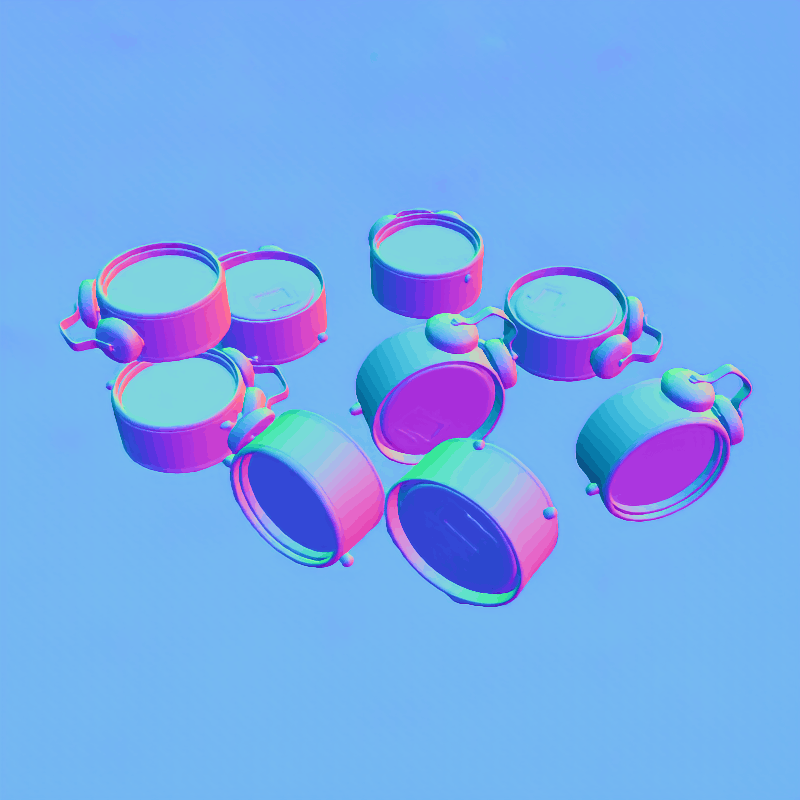}
    \end{subfigure}\hfill
    \begin{subfigure}{0.16\textwidth}
        \centering
        \includegraphics[width=\textwidth]{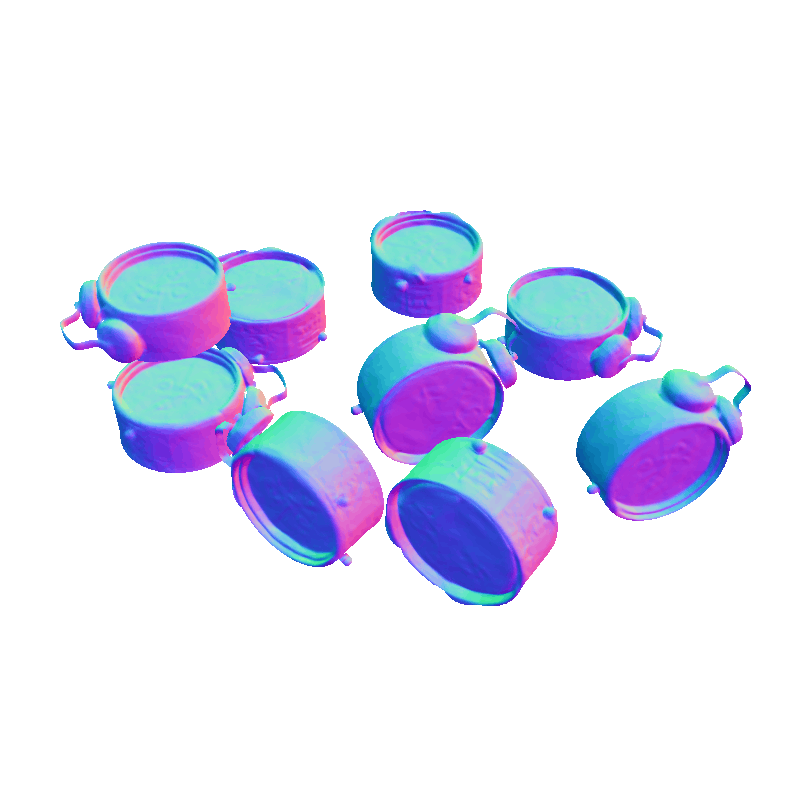}
    \end{subfigure}\hfill
    \begin{subfigure}{0.16\textwidth}
        \centering
        \includegraphics[width=\textwidth]{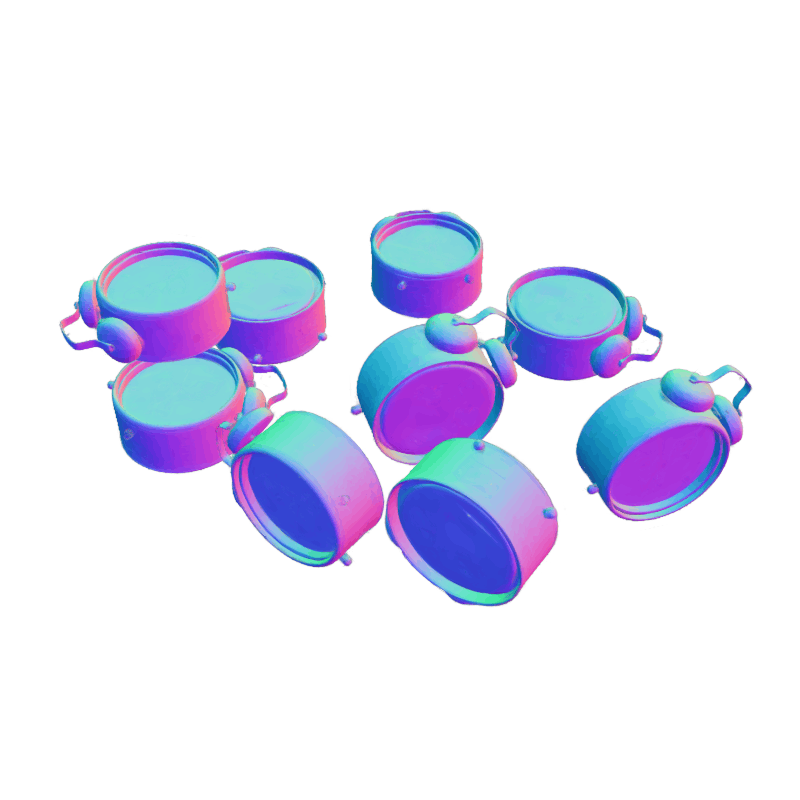}
    \end{subfigure}\\
 \raisebox{8ex}{\rotatebox{90}{Coffee\vphantom{g}} }\hfill
 \begin{subfigure}{0.16\textwidth}
        \centering
        \includegraphics[width=\textwidth]{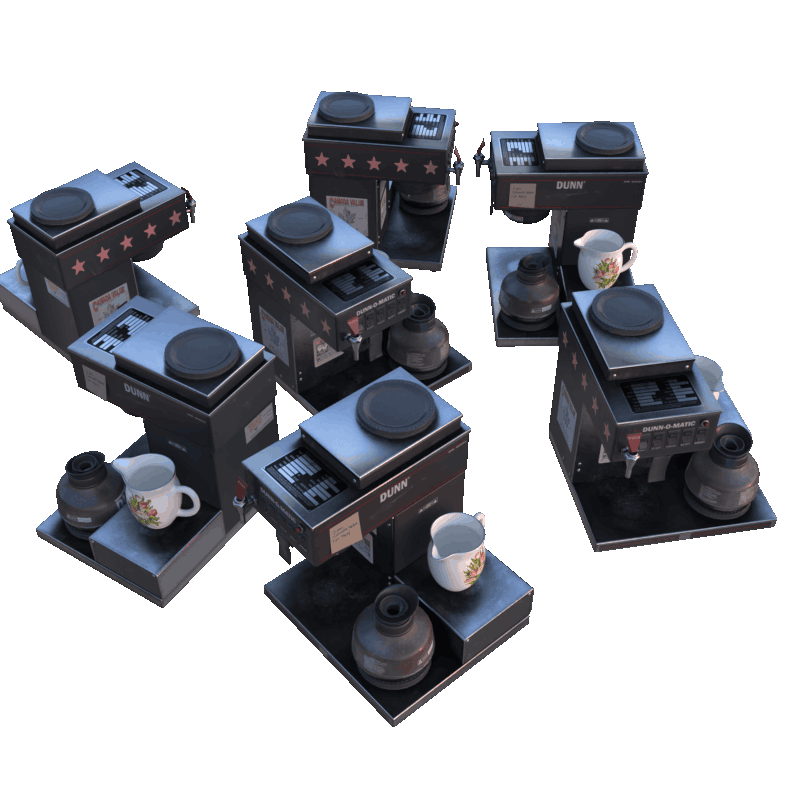}
       \caption*{Input image}
    \end{subfigure}\hfill
    \begin{subfigure}{0.16\textwidth}
        \centering
        \includegraphics[width=\textwidth]{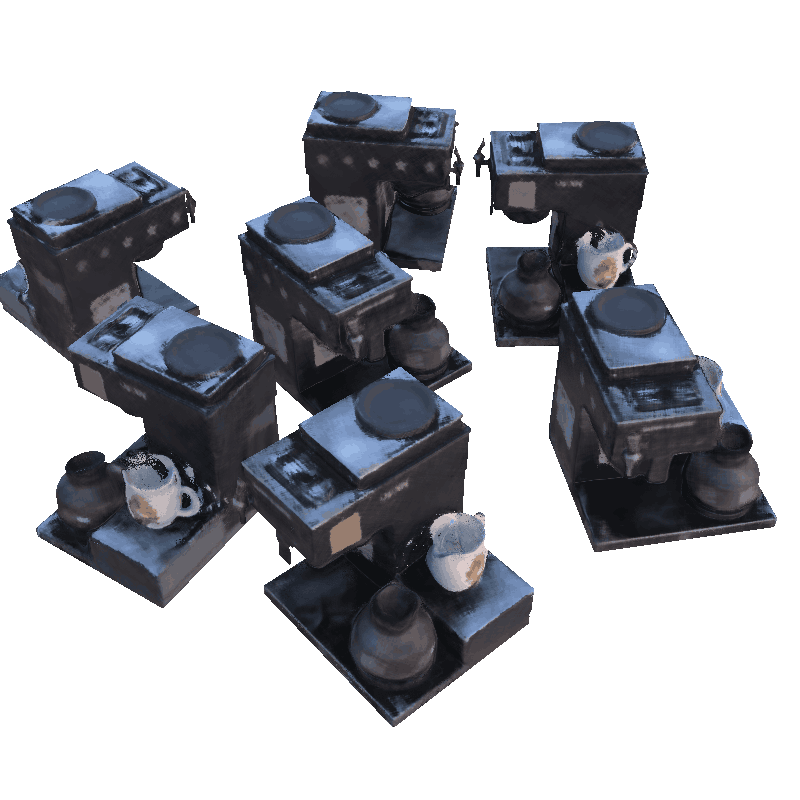}
        \caption*{RGB \they\unskip$^*$}
    \end{subfigure}\hfill
    \begin{subfigure}{0.16\textwidth}
        \centering
        \includegraphics[width=\textwidth]{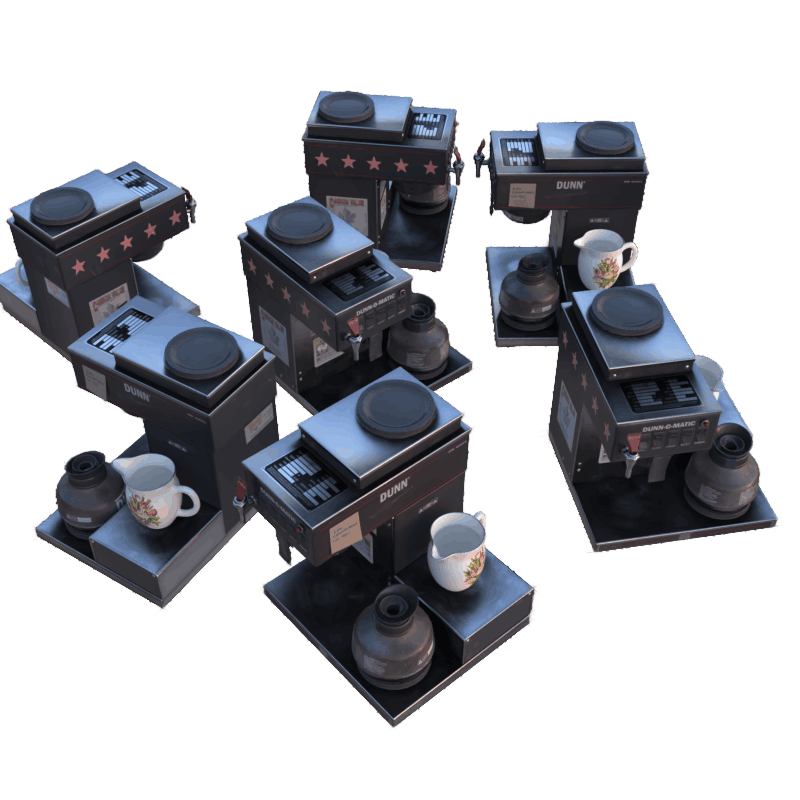}
        \caption*{RGB \method}
    \end{subfigure}\hfill
    \begin{subfigure}{0.16\textwidth}
        \centering
        \includegraphics[width=\textwidth]{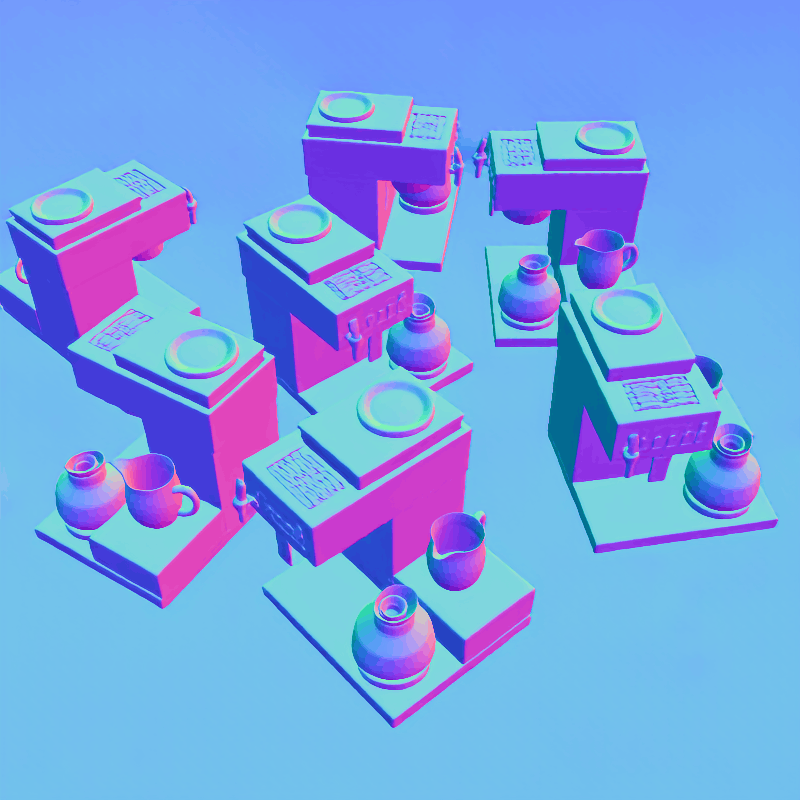}
        \caption*{Marigold V2 \cite{pavlovic2026marigold} }
    \end{subfigure}\hfill
    \begin{subfigure}{0.16\textwidth}
        \centering
        \includegraphics[width=\textwidth]{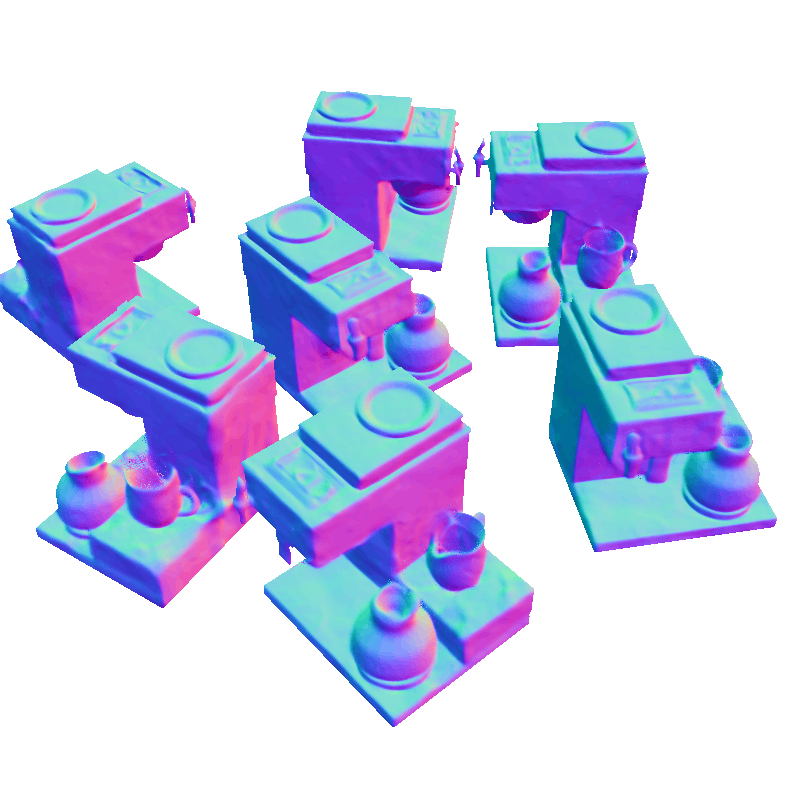}
        \caption*{\they\unskip$^*$ normals}
    \end{subfigure}\hfill
    \begin{subfigure}{0.16\textwidth}
        \centering
        \includegraphics[width=\textwidth]{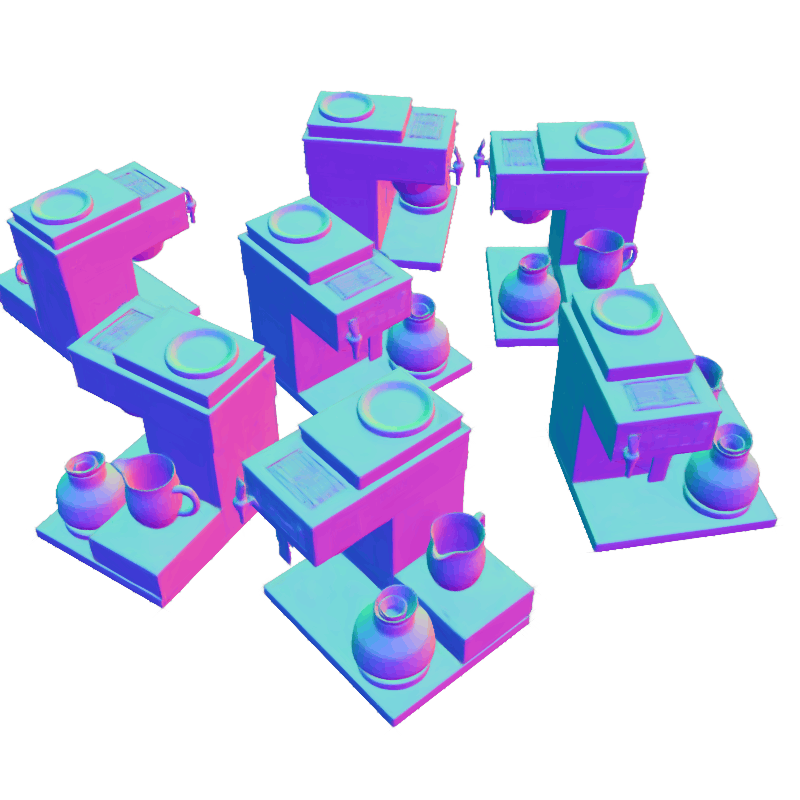}
        \caption*{\method normals}
    \end{subfigure}
    \caption{Qualitative results of Marigold V2 \cite{pavlovic2026marigold}, \they\unskip$^*$, and \method on multi-instance images of Dup-synthetic. Zoom in for more detail.}
    \label{fig:qualitative_comparison_syn1}
\end{figure*}

\begin{figure*}
    \centering
 \raisebox{8ex}{\rotatebox{90}{Fire\vphantom{g}} }\hfill
 \begin{subfigure}{0.16\textwidth}
        \centering
        \includegraphics[width=\textwidth]{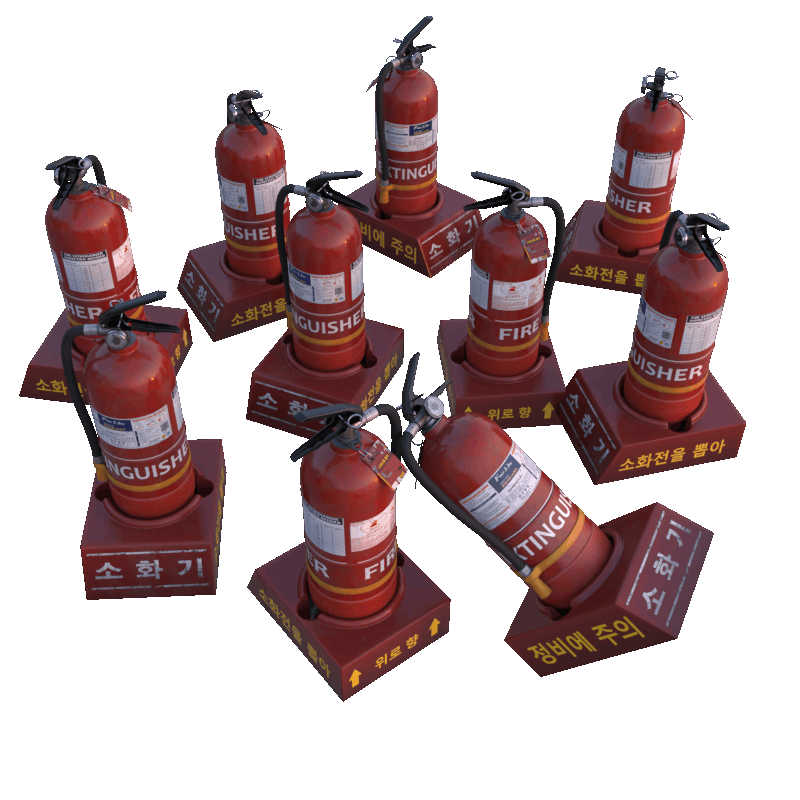}
    \end{subfigure}\hfill
    \begin{subfigure}{0.16\textwidth}
        \centering
        \includegraphics[width=\textwidth]{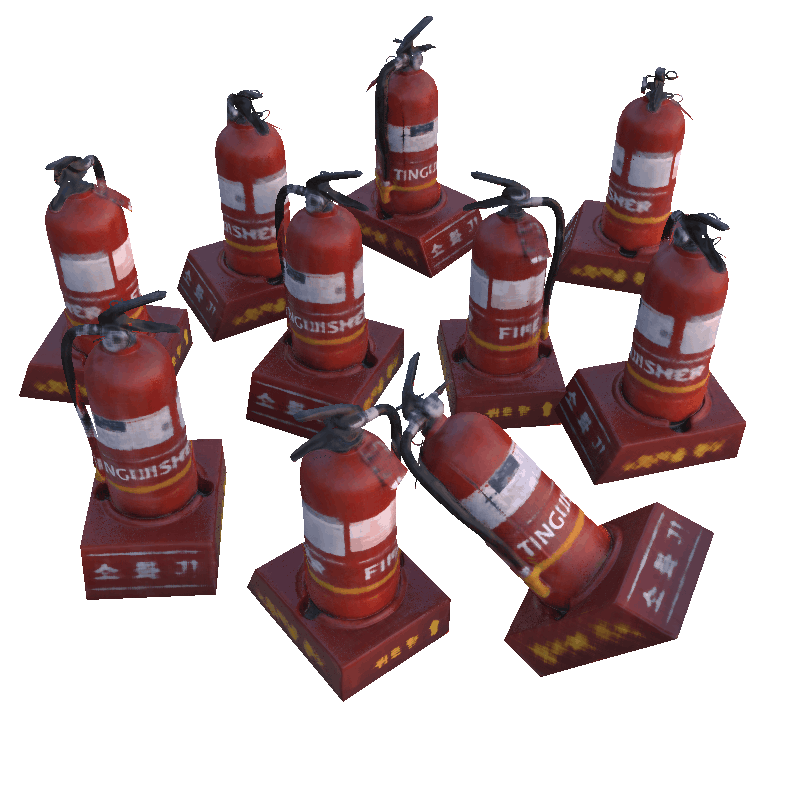}
    \end{subfigure}\hfill
    \begin{subfigure}{0.16\textwidth}
        \centering
        \includegraphics[width=\textwidth]{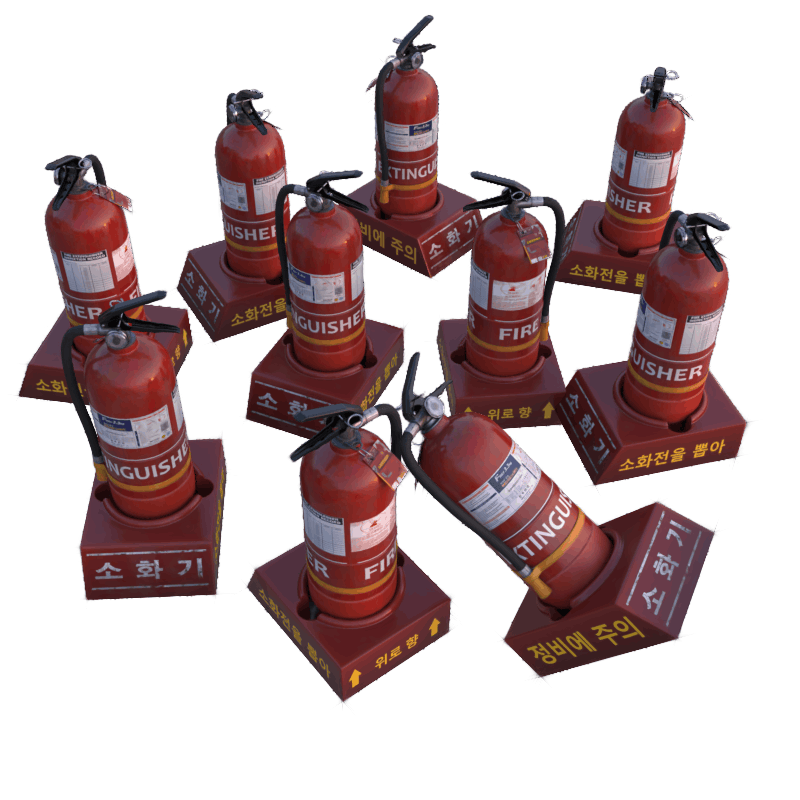}
    \end{subfigure}\hfill
        \begin{subfigure}{0.16\textwidth}
        \centering
        \includegraphics[width=\textwidth]{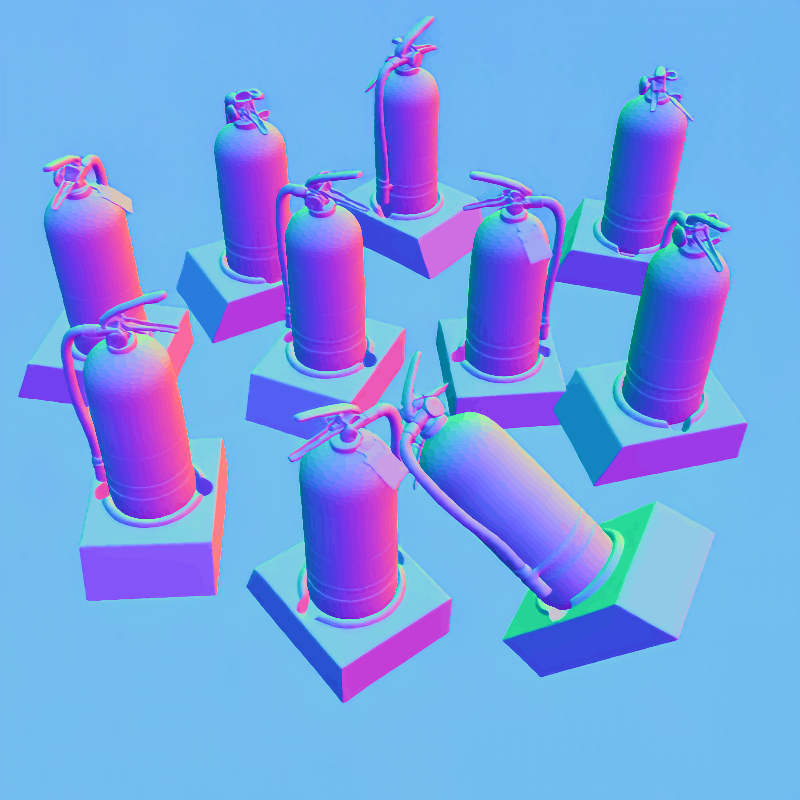}
    \end{subfigure}\hfill
    \begin{subfigure}{0.16\textwidth}
        \centering
        \includegraphics[width=\textwidth]{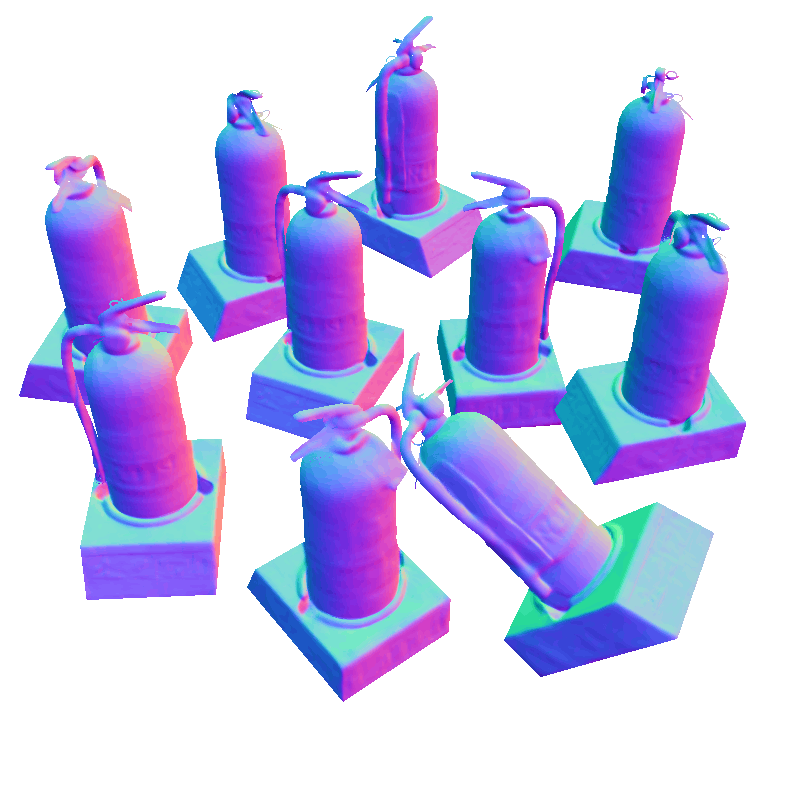}
    \end{subfigure}\hfill
    \begin{subfigure}{0.16\textwidth}
        \centering
        \includegraphics[width=\textwidth]{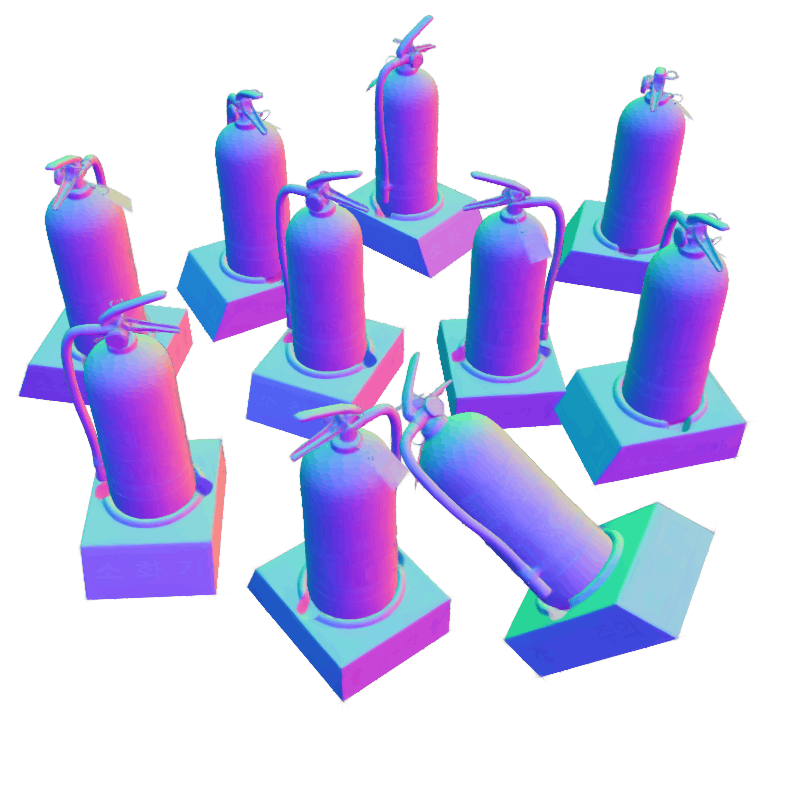}
    \end{subfigure}\\
 \raisebox{8ex}{\rotatebox{90}{Guitar\vphantom{g}} }\hfill
    \begin{subfigure}{0.16\textwidth}
        \centering
        \includegraphics[width=\textwidth]{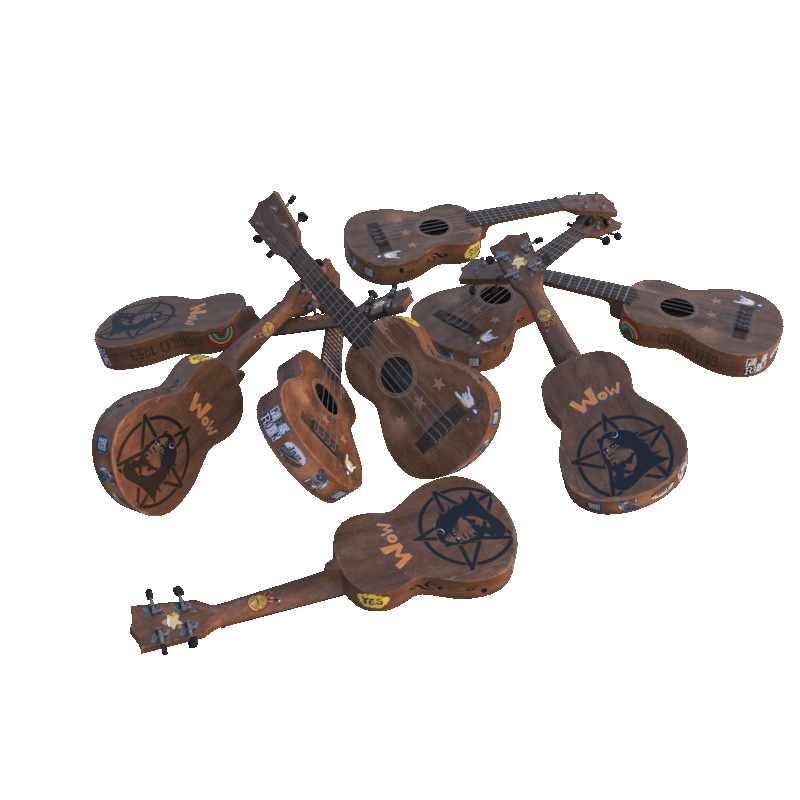}
    \end{subfigure}\hfill
    \begin{subfigure}{0.16\textwidth}
        \centering
        \includegraphics[width=\textwidth]{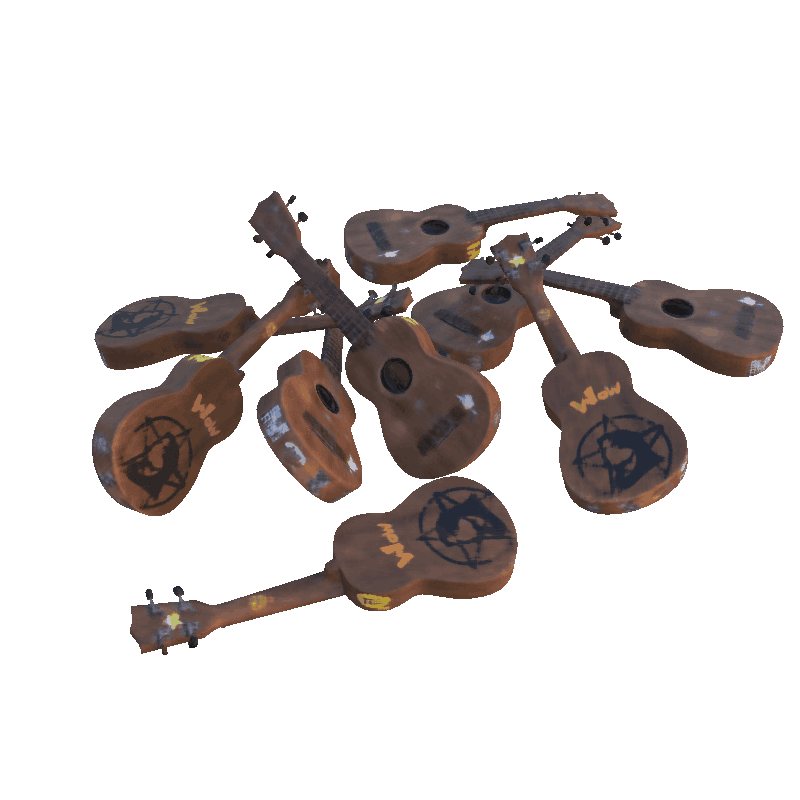}
    \end{subfigure}\hfill
    \begin{subfigure}{0.16\textwidth}
        \centering
        \includegraphics[width=\textwidth]{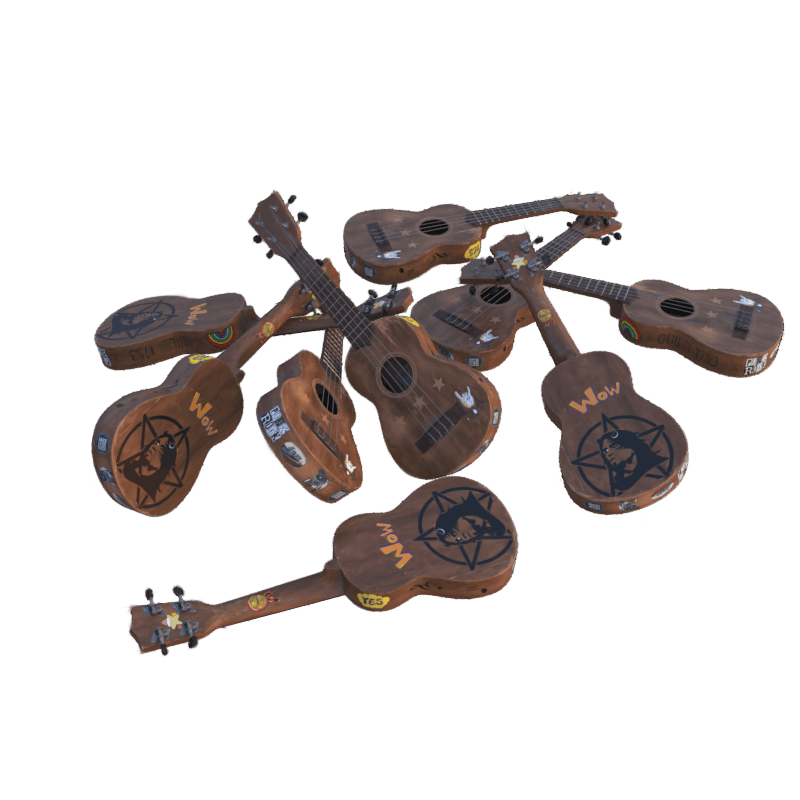}
    \end{subfigure}\hfill
    \begin{subfigure}{0.16\textwidth}
        \centering
        \includegraphics[width=\textwidth]{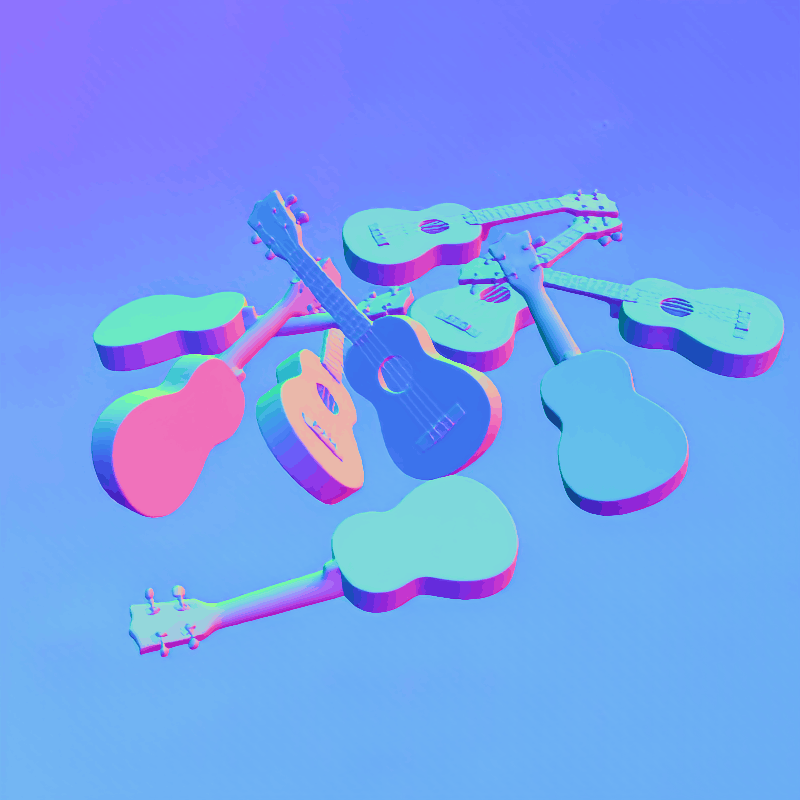}
    \end{subfigure}\hfill
    \begin{subfigure}{0.16\textwidth}
        \centering
        \includegraphics[width=\textwidth]{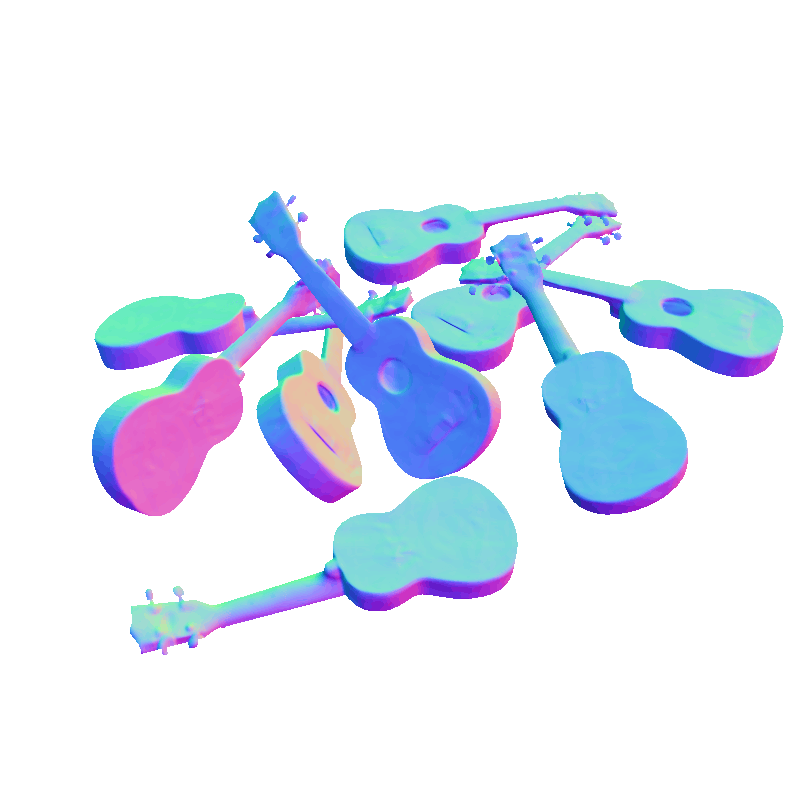}
    \end{subfigure}\hfill
    \begin{subfigure}{0.16\textwidth}
        \centering
        \includegraphics[width=\textwidth]{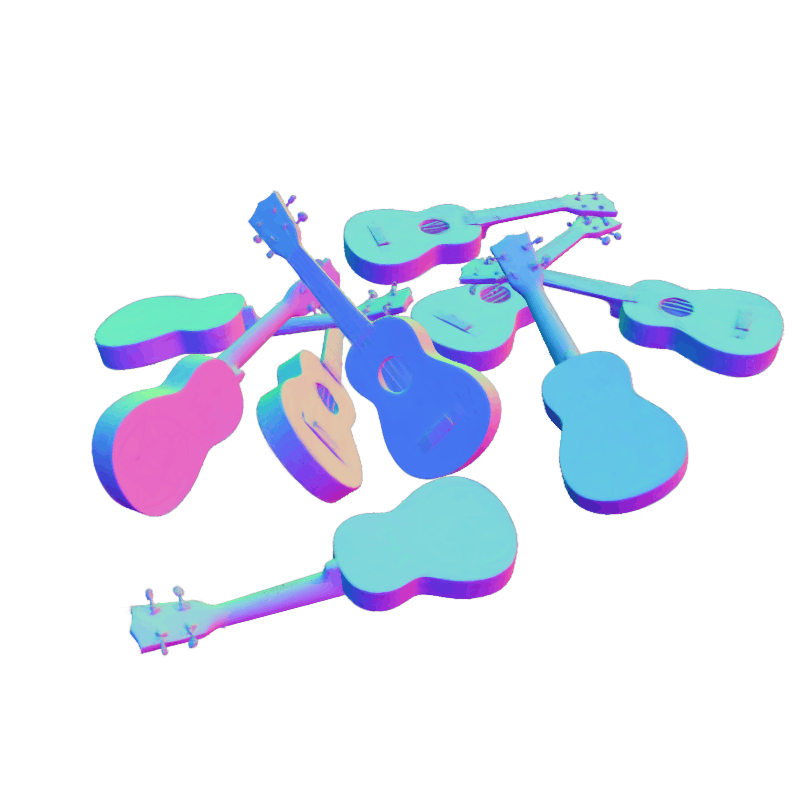}
    \end{subfigure}\\
        \raisebox{8ex}{\rotatebox{90}{Sign\vphantom{g}} }\hfill
    \begin{subfigure}{0.16\textwidth}
        \centering
        \includegraphics[width=\textwidth]{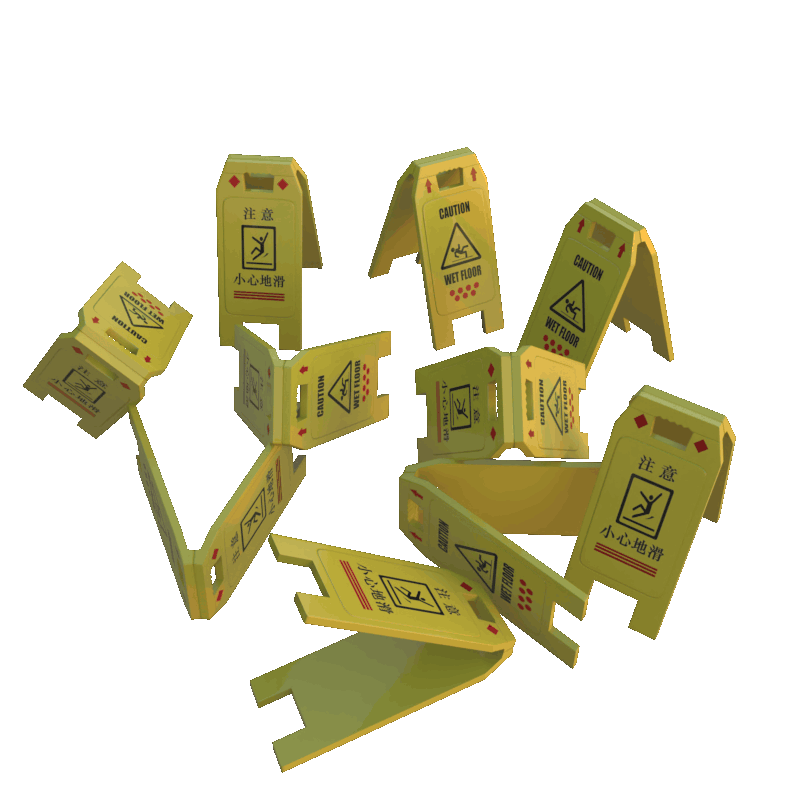}
    \end{subfigure}\hfill
    \begin{subfigure}{0.16\textwidth}
        \centering
        \includegraphics[width=\textwidth]{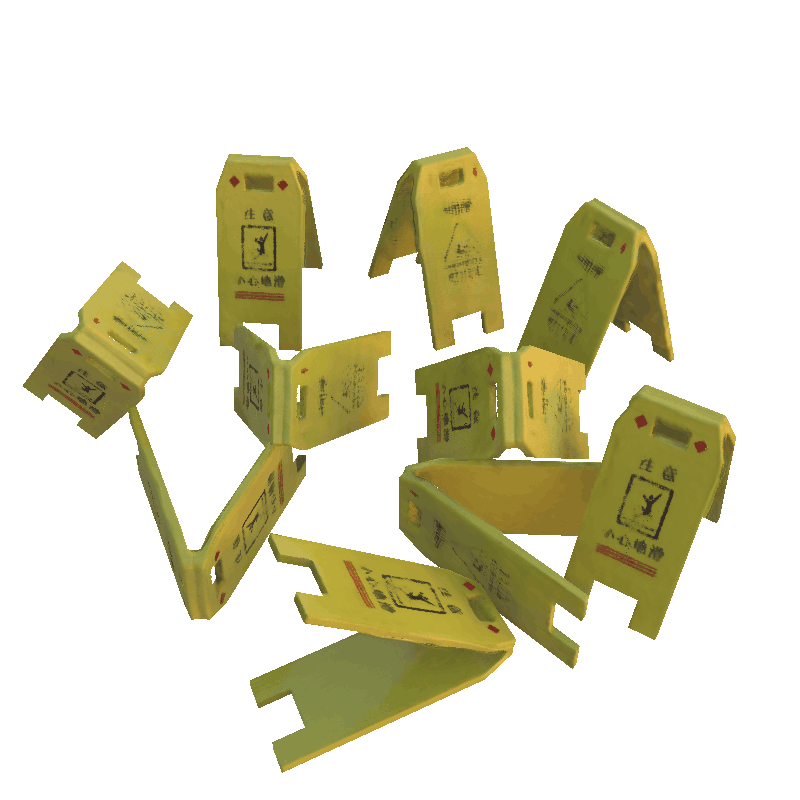}
    \end{subfigure}\hfill
    \begin{subfigure}{0.16\textwidth}
        \centering
        \includegraphics[width=\textwidth]{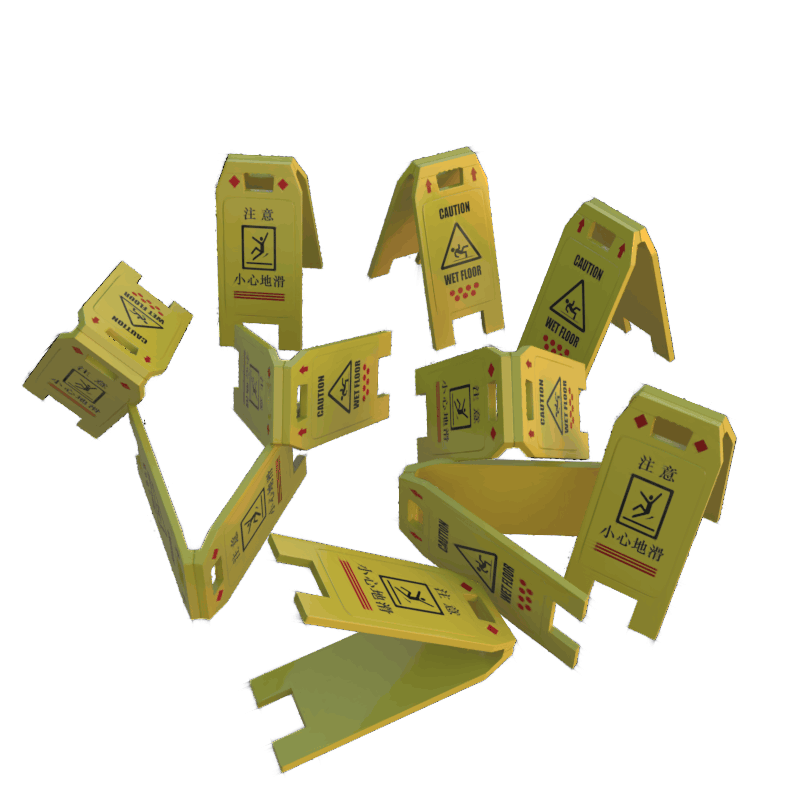}
    \end{subfigure}\hfill
    \begin{subfigure}{0.16\textwidth}
        \centering
        \includegraphics[width=\textwidth]{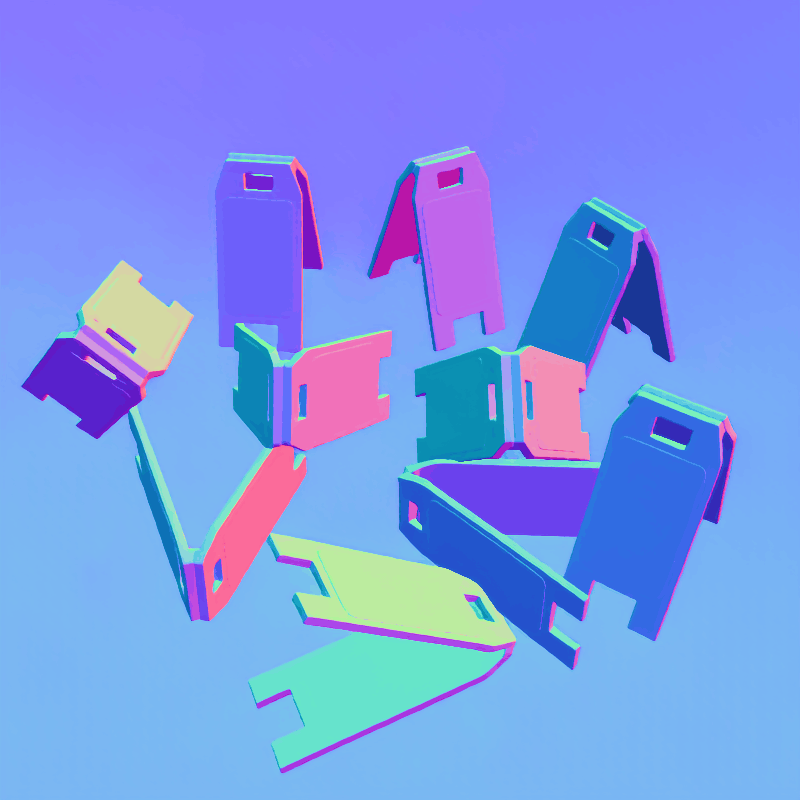}
    \end{subfigure}\hfill
    \begin{subfigure}{0.16\textwidth}
        \centering
        \includegraphics[width=\textwidth]{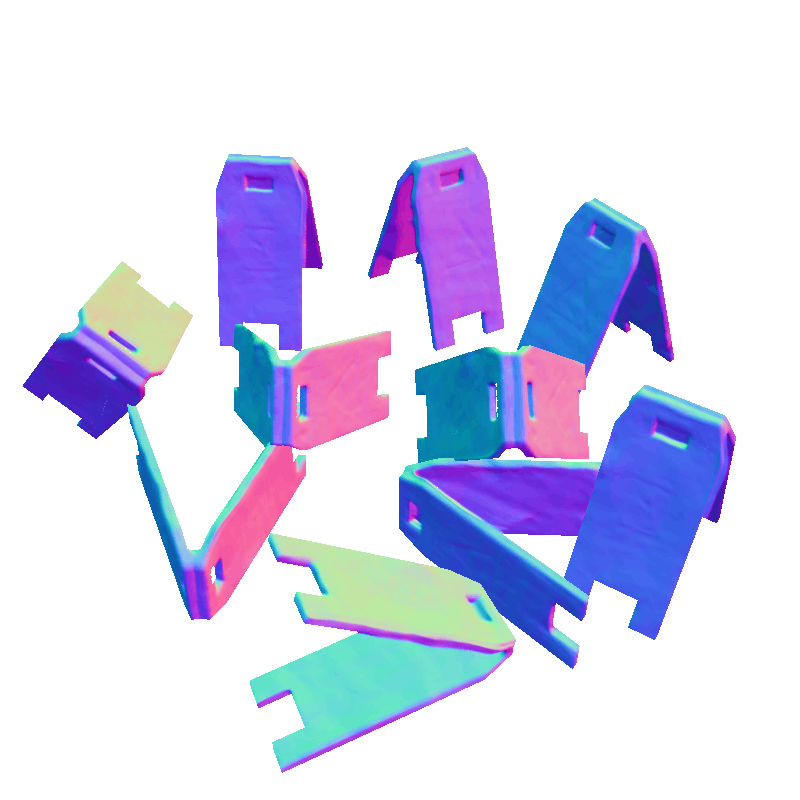}
    \end{subfigure}\hfill
    \begin{subfigure}{0.16\textwidth}
        \centering
        \includegraphics[width=\textwidth]{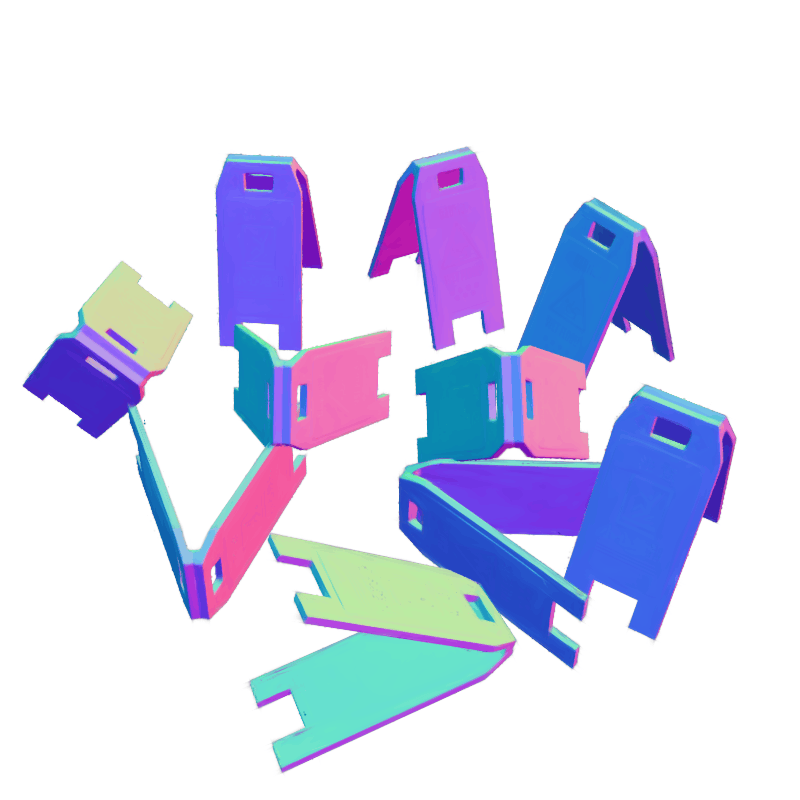}
    \end{subfigure}\\
    \raisebox{8ex}{\rotatebox{90}{Tin\vphantom{g}} }\hfill
    \begin{subfigure}{0.16\textwidth}
        \centering
        \includegraphics[width=\textwidth]{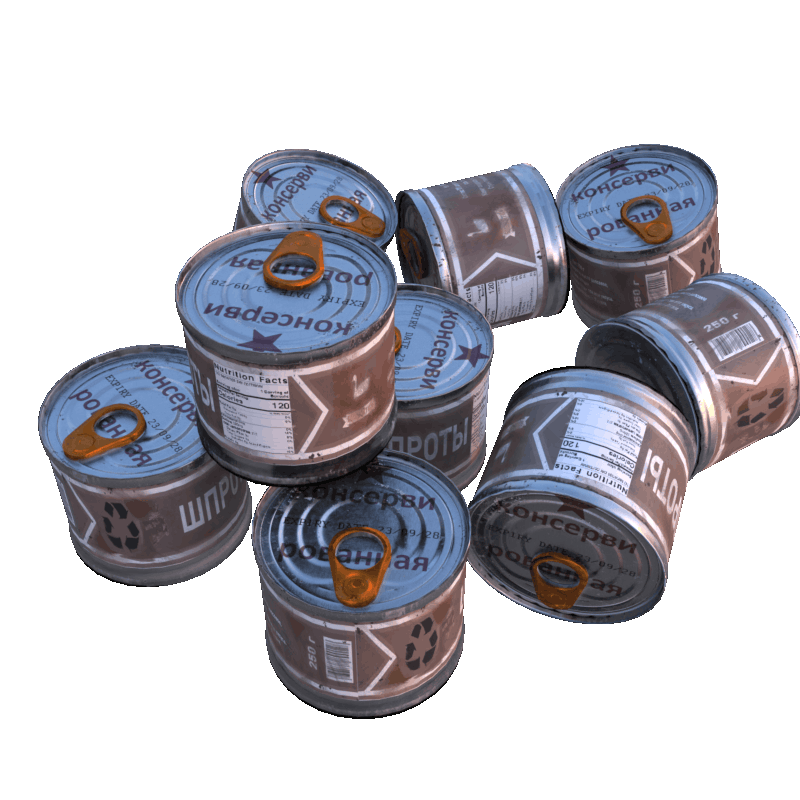}
       \caption*{Input image}
    \end{subfigure}\hfill
    \begin{subfigure}{0.16\textwidth}
        \centering
        \includegraphics[width=\textwidth]{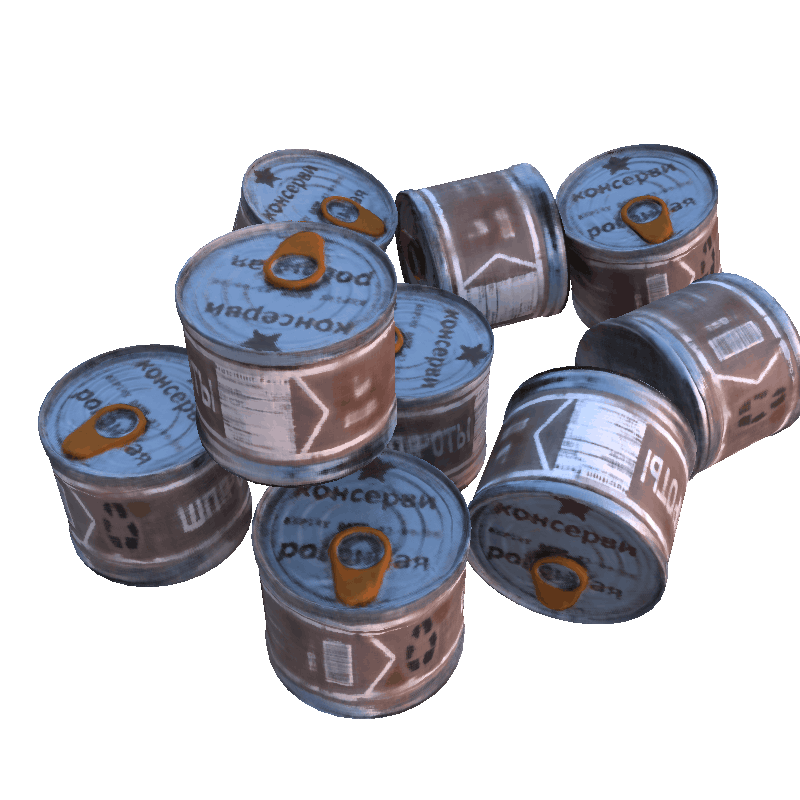}
        \caption*{RGB \they\unskip$^*$}
    \end{subfigure}\hfill
    \begin{subfigure}{0.16\textwidth}
        \centering
        \includegraphics[width=\textwidth]{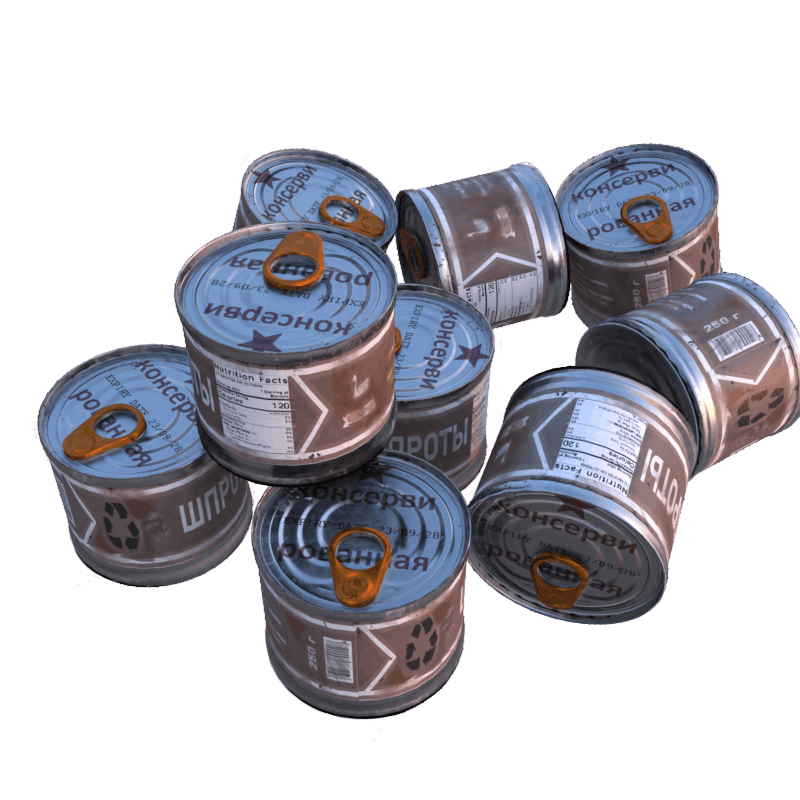}
        \caption*{RGB \method}
    \end{subfigure}\hfill
    \begin{subfigure}{0.16\textwidth}
        \centering
        \includegraphics[width=\textwidth]{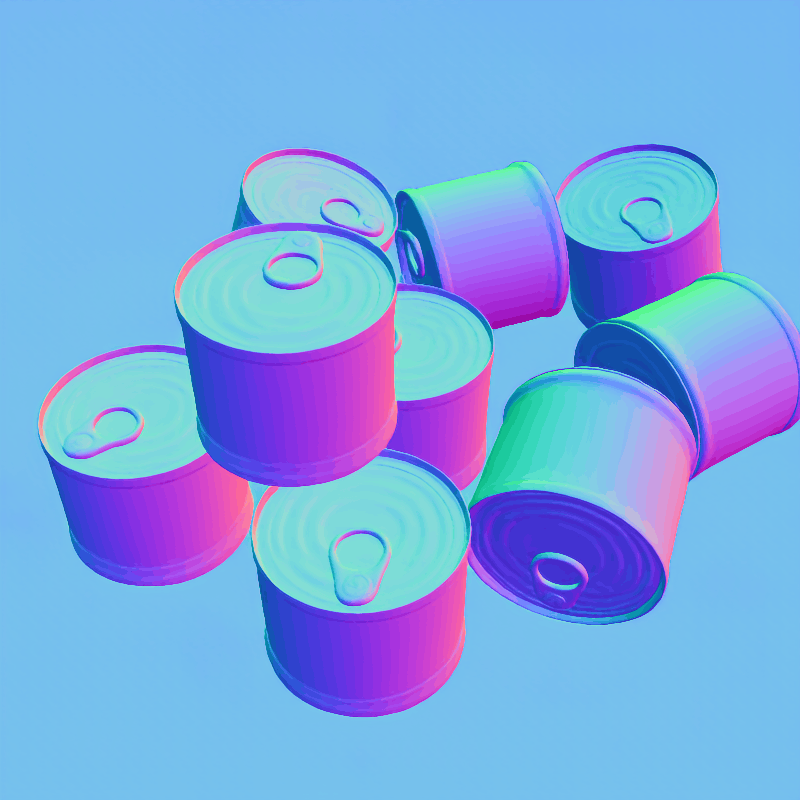}
        \caption*{Marigold V2 \cite{pavlovic2026marigold} }
    \end{subfigure}\hfill
    \begin{subfigure}{0.16\textwidth}
        \centering
        \includegraphics[width=\textwidth]{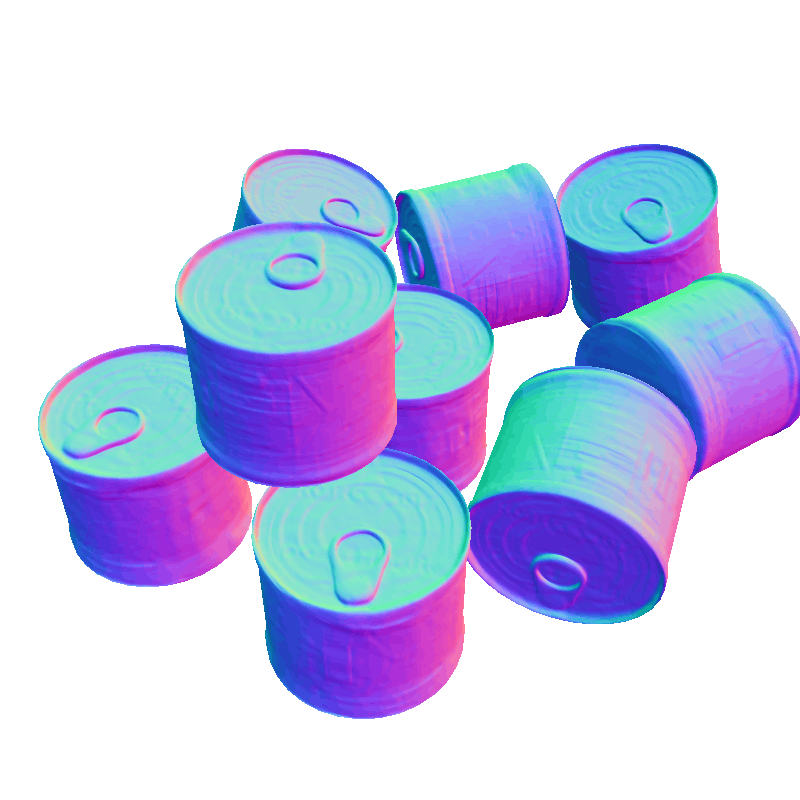}
        \caption*{\they\unskip$^*$ normals}
    \end{subfigure}\hfill
    \begin{subfigure}{0.16\textwidth}
        \centering
        \includegraphics[width=\textwidth]{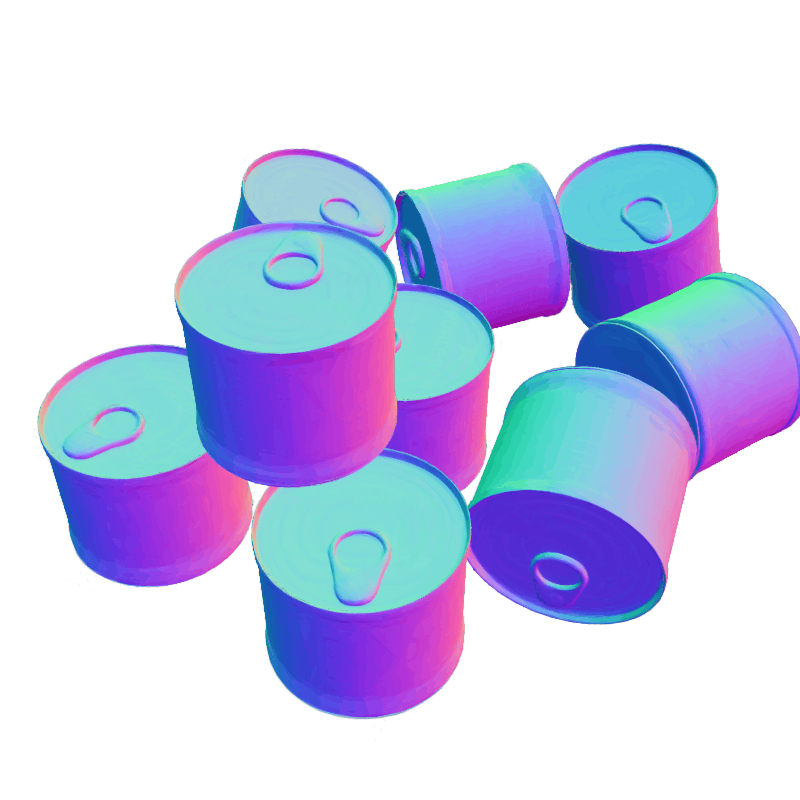}
        \caption*{\method normals}
    \end{subfigure}

    \caption{Qualitative results of Marigold V2 \cite{pavlovic2026marigold}, \they\unskip$^*$, and \method on multi-instance images of Dup-synthetic. Zoom in for more detail.}
    \label{fig:qualitative_comparison_syn2}
\end{figure*}

\cref{fig:qualitative_comparison_syn1,fig:qualitative_comparison_syn2} compare \method against \they\unskip$^*$ on Dup-synthetic, which correspond to the quantitative results in \cref{sec:experiments}. The qualitative results back up the quantitative findings, showing that \method generates more accurate RGB and normal renders compared with \they\unskip$^*$. We observe especially in the scenes \textit{cash} and \textit{clock}, that Marigold V2 predicted normals do not enforce consistency between instances, which is alleviated by \method.

\begin{figure*}
\centering
    \raisebox{8ex}{\rotatebox{90}{Fire\vphantom{g}} }\hfill
    \begin{subfigure}{0.16\textwidth}
        \centering
        \includegraphics[width=\textwidth]{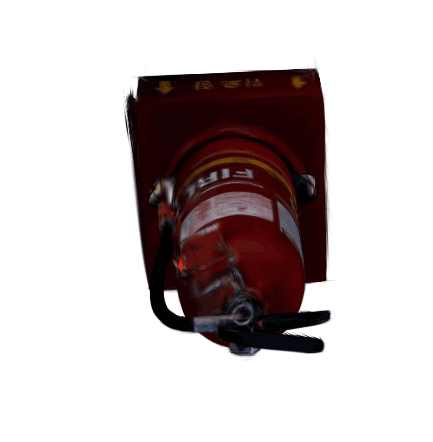}
        \caption*{(a)}
    \end{subfigure}\hfill
    \begin{subfigure}{0.16\textwidth}
        \centering
        \includegraphics[width=\textwidth]{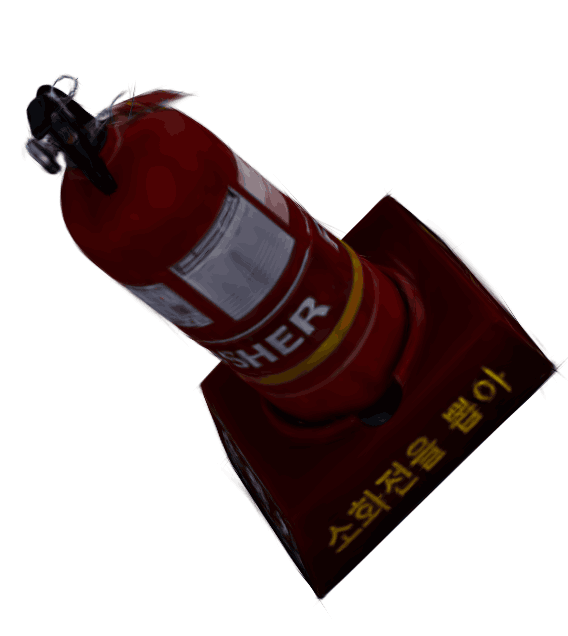}
        \caption*{(b)}
    \end{subfigure}\hfill
    \begin{subfigure}{0.16\textwidth}
        \centering
        \includegraphics[width=\textwidth]{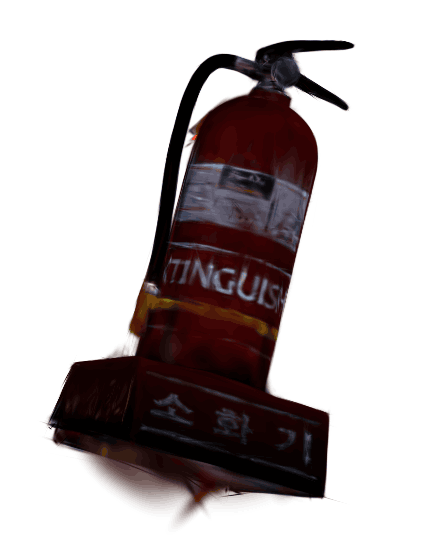}
        \caption*{(c)}
    \end{subfigure}\hfill
    \begin{subfigure}{0.16\textwidth}
        \centering
        \includegraphics[width=\textwidth]{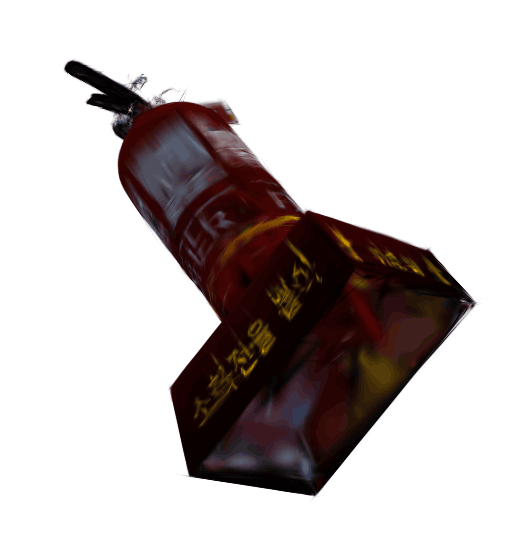}
        \caption*{(d)}
    \end{subfigure}\\
        \raisebox{8ex}{\rotatebox{90}{XYZ3\vphantom{g}} }\hfill
    \begin{subfigure}{0.16\textwidth}
        \centering
        \includegraphics[width=\textwidth]{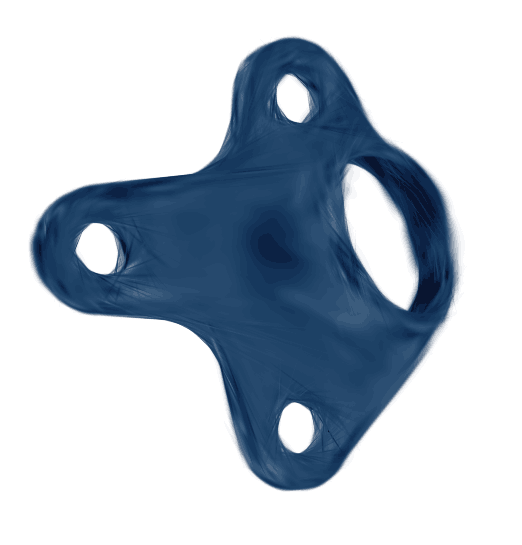}
        \caption*{(a)}
    \end{subfigure}\hfill
        \begin{subfigure}{0.16\textwidth}
        \centering
        \includegraphics[width=\textwidth]{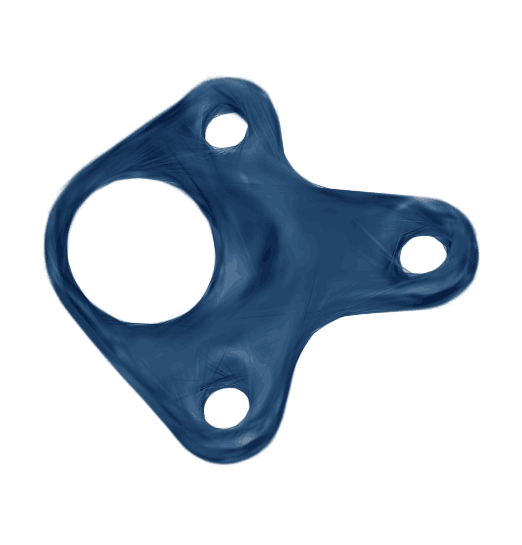}
        \caption*{(b)}
    \end{subfigure}\hfill
    \begin{subfigure}{0.16\textwidth}
        \centering
        \includegraphics[width=\textwidth]{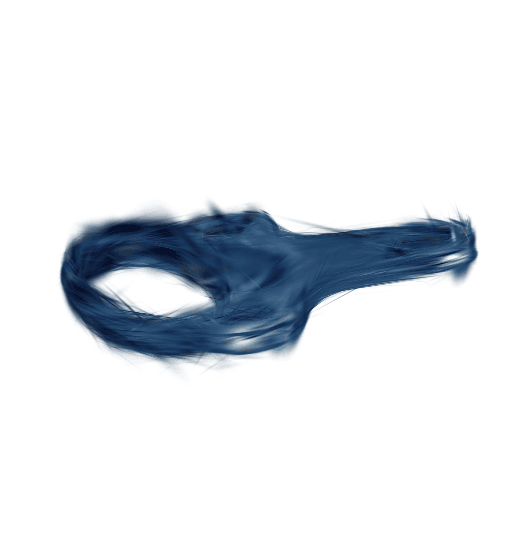}
        \caption*{(c)}
    \end{subfigure}\hfill
    \begin{subfigure}{0.16\textwidth}
        \centering
        \includegraphics[width=\textwidth]{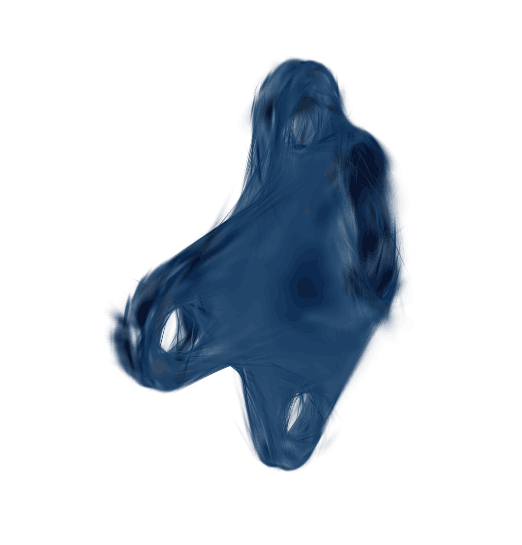}
        \caption*{(d)}
    \end{subfigure}

    \caption{Examples for novel view synthesis with \method on XYZ-IBD (top) and Dup-synthetic (bottom). We show examples for both view interpolation (a, b) and view extrapolation (c, d). }
    \label{fig:novel_views}
\end{figure*}

\paragraph{Novel View Synthesis Results.}
As a GS method, \method can trivially be exploited for novel view synthesis. We show examples in \cref{fig:novel_views}. When rendering the object in poses that are not frequently observed in the input image (pose extrapolation), \method leads to less accurate novel views, which is a known limitation of GS methods. Although only a small number of views of the object is visible in the single multi-instance image, still \method is able to generate convincing novel views.

\begin{figure*}
\centering
    \raisebox{12ex}{\rotatebox{90}{Fire\vphantom{g}} }\hfill
    \begin{subfigure}{0.35\textwidth}
        \centering
        \rotatebox{180}{\includegraphics[width=\textwidth]{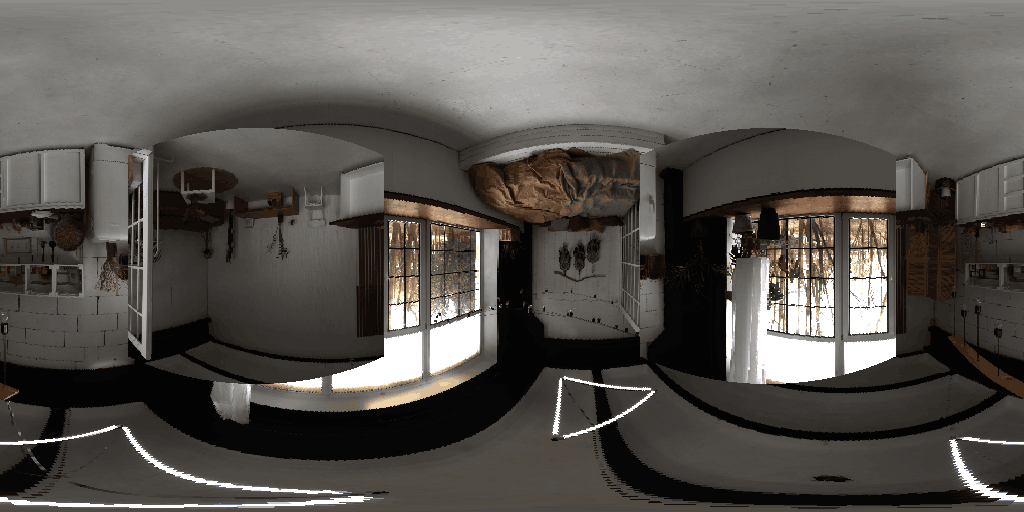}}
      \caption*{Environment map}
    \end{subfigure}\hfill
    \begin{subfigure}{0.2\textwidth}
        \centering
        \includegraphics[width=\textwidth]{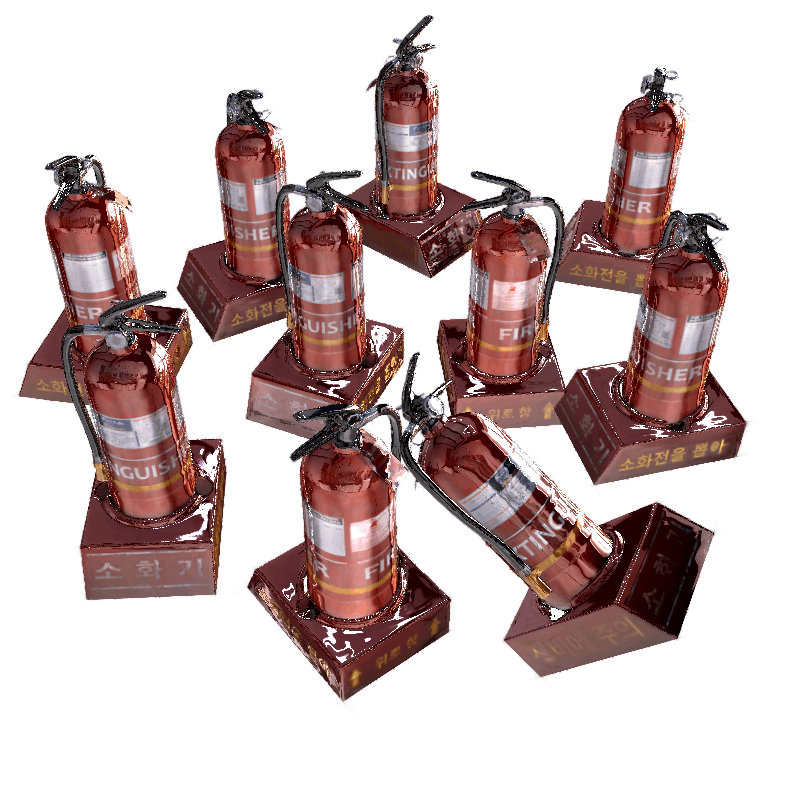}
        \caption*{Relighting}
    \end{subfigure}\hfill
    \begin{subfigure}{0.2\textwidth}
        \centering
        \includegraphics[width=\textwidth]{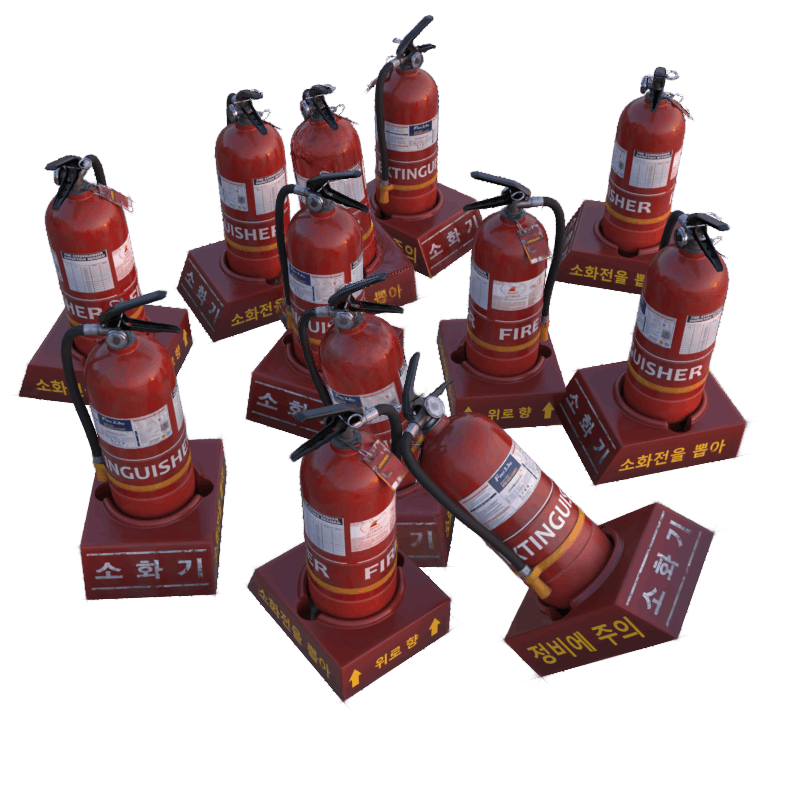}
       \caption*{Object insertion}
     \end{subfigure}
  
    \caption{Examples for relighting and object insertion with \method. The first column shows the envmap which is used for the relighting in the second column. Object insertion application is shown in the third column.}
    \label{fig:relighting}
\end{figure*}

\paragraph{Deferred Rendering Results.} The focus and objective of \method is the 3D reconstruction and object pose estimation of multi-instance images, but methodically \method achieves this by decomposing geometry and material properties from scene illumination. By our decision to model specular effects via the light probes, we do not need to explicitly model visibility during training. This comes at the cost of baking reflections and shadows into the light probes, but allows for a cheap two-stage deferred rendering (\cref{sec:preliminary}). Therefore, relighting and object insertion with a different environment map lead to slightly incorrect results. Nevertheless, we showcase object insertion and relighting results, ignoring visibility in \cref{fig:relighting}. The renders show that \method can do these tasks, even if it is not tuned to them. 

\begin{figure*}[tbp]

    \centering
\rowlabel{1}%
\begin{subfigure}[c]{0.15\textwidth}
    \centering
    \includegraphics[width=\textwidth]{graphics/15_14.png}
\end{subfigure}\hfill
\begin{subfigure}[c]{0.15\textwidth}
    \centering
    \includegraphics[height=2cm,trim={16cm 8cm 13cm 4cm},clip]{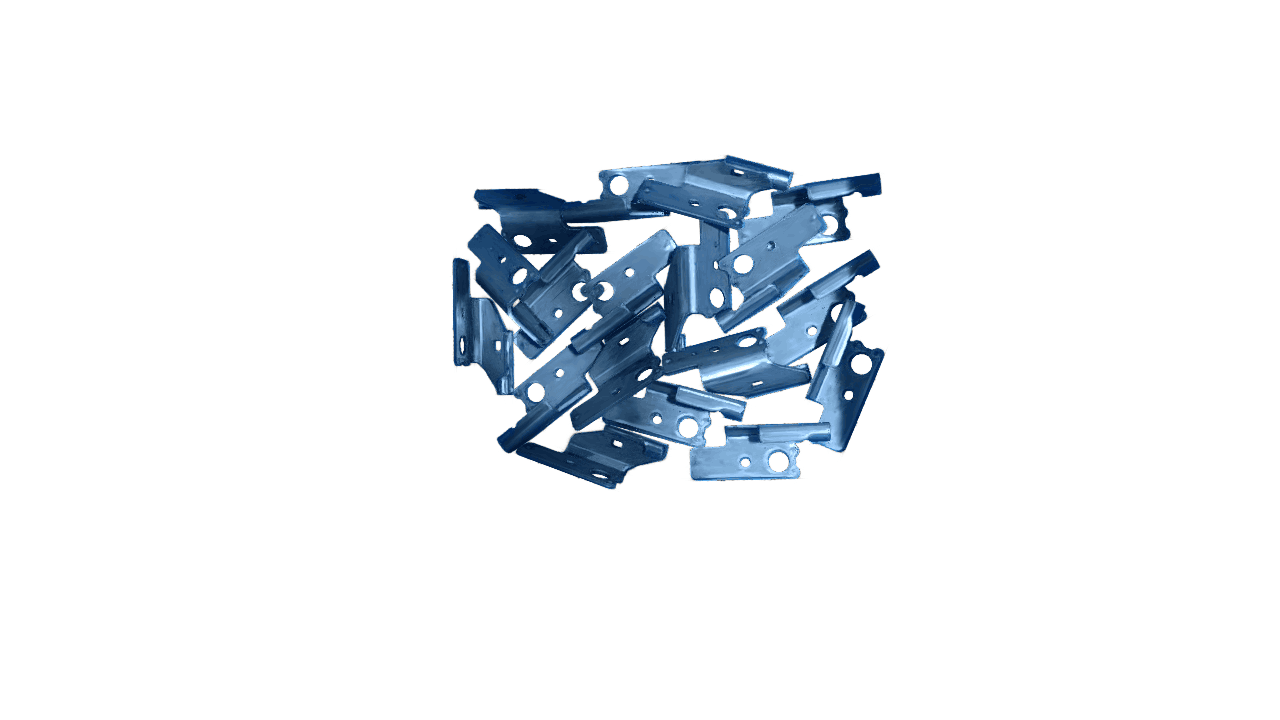}
\end{subfigure}\hfill
\begin{subfigure}[c]{0.15\textwidth}
    \centering
    \includegraphics[height=2cm,trim={16cm 8cm 13cm 4cm},clip]{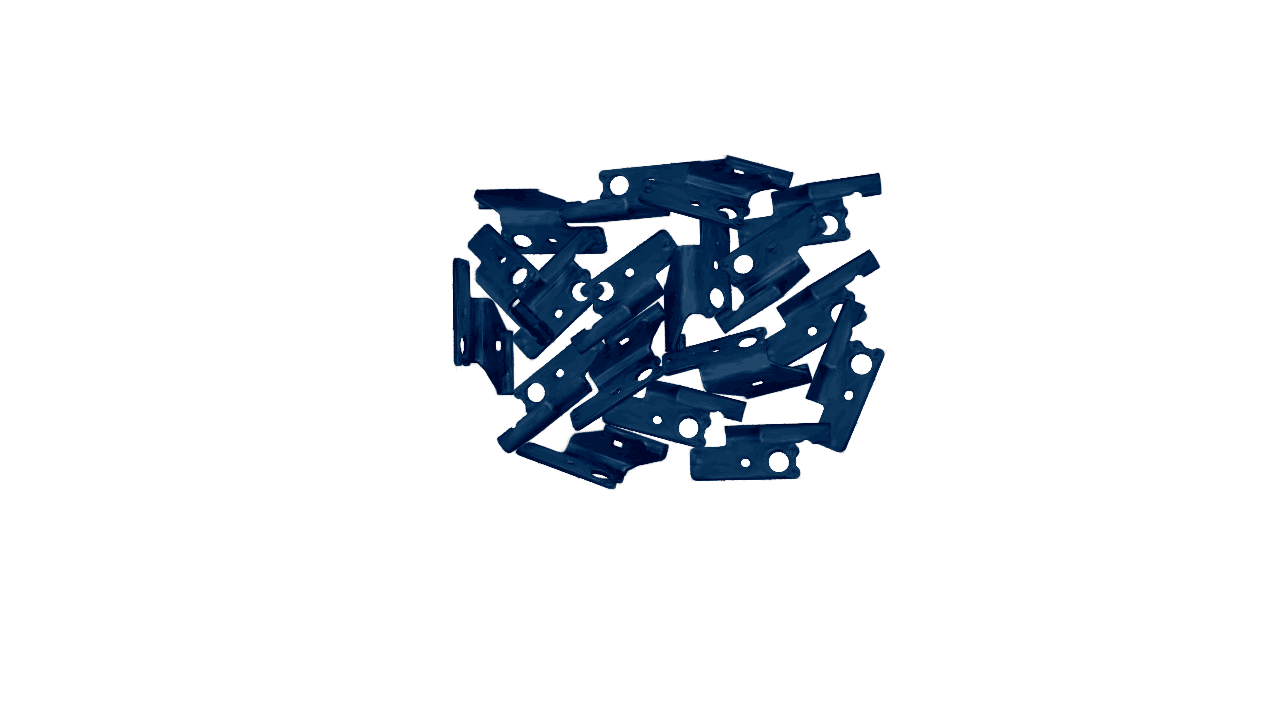}
\end{subfigure}\hfill
\begin{subfigure}[c]{0.15\textwidth}
    \centering
    \includegraphics[height=2cm,trim={16cm 8cm 13cm 4cm},clip]{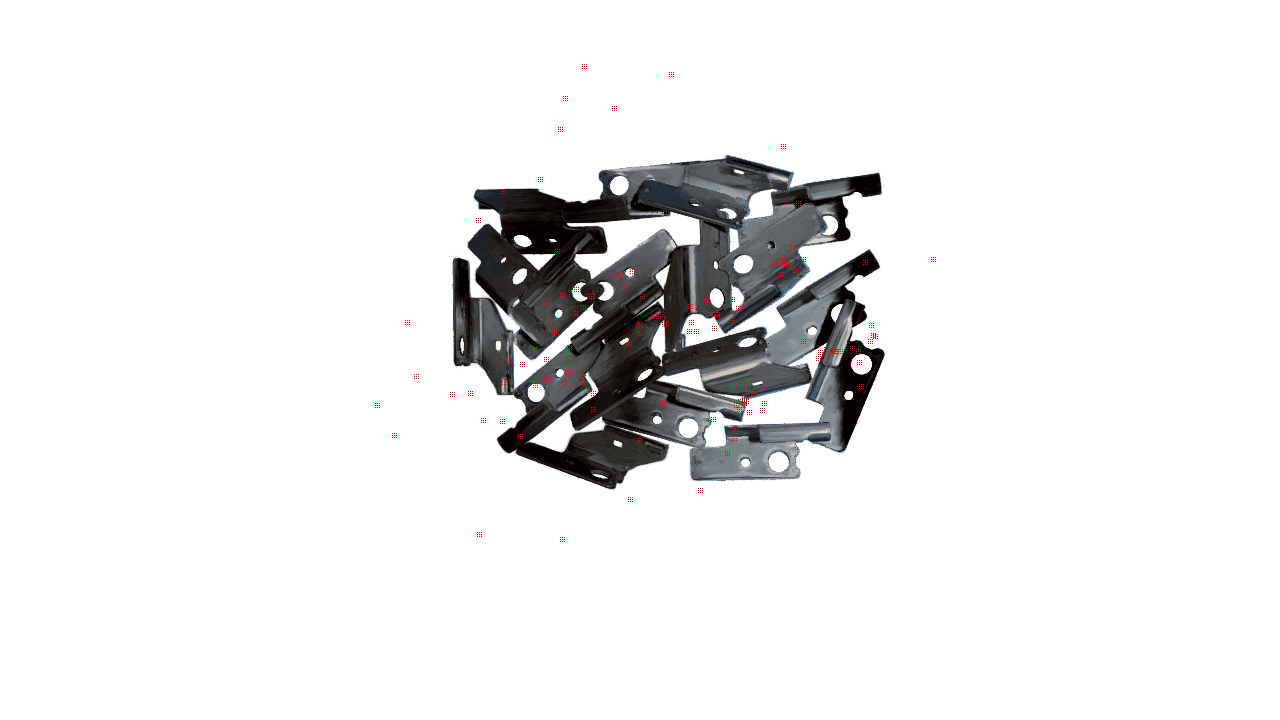}
\end{subfigure}\hfill
\begin{subfigure}[c]{0.15\textwidth}
    \centering
    \includegraphics[height=2cm,trim={16cm 8cm 13cm 4cm},clip]{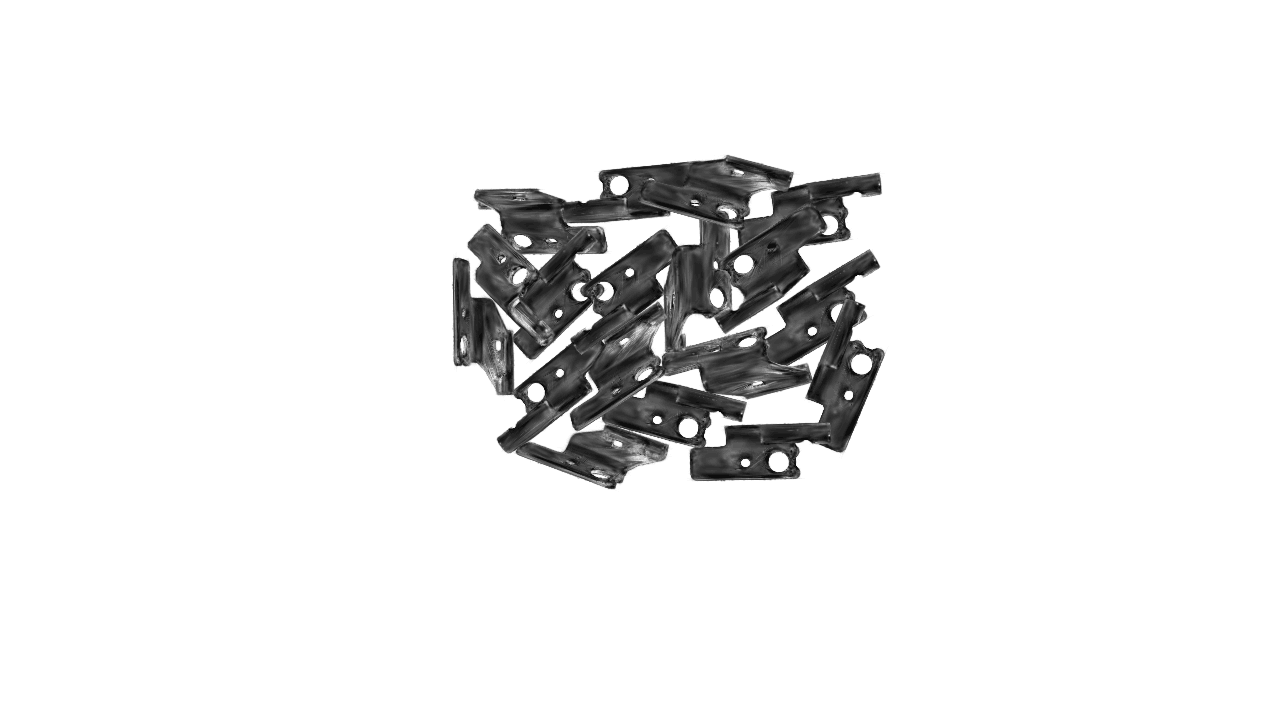}
\end{subfigure}\\
\rowlabel{2}%
\begin{subfigure}[c]{0.15\textwidth}
    \centering
    \includegraphics[width=\textwidth]{graphics/60_6.png}
\end{subfigure}\hfill
\begin{subfigure}[c]{0.15\textwidth}
    \centering
    \includegraphics[height=2cm,trim={14cm 6cm 12cm 4cm},clip]{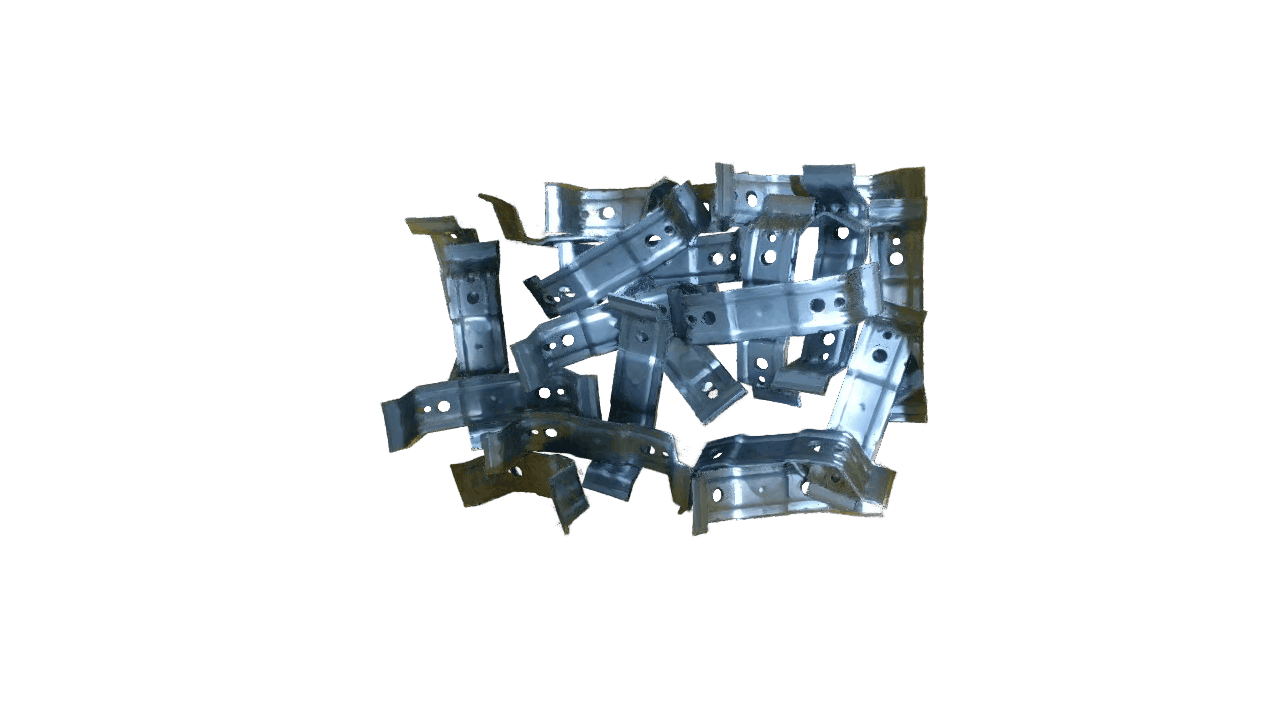}
\end{subfigure}\hfill
\begin{subfigure}[c]{0.15\textwidth}
    \centering
    \includegraphics[height=2cm,trim={14cm 6cm 12cm 4cm},clip]{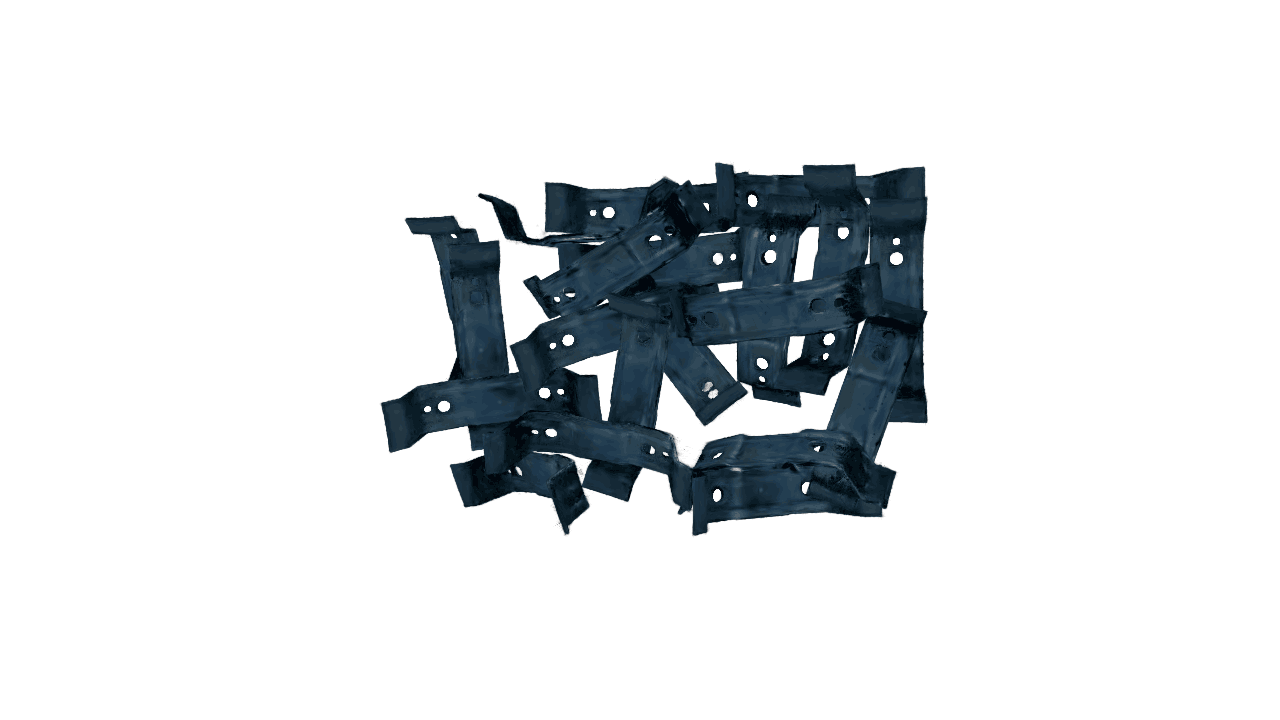}
\end{subfigure}
\hfill
\begin{subfigure}[c]{0.15\textwidth}
    \centering
    \includegraphics[height=2cm,trim={14cm 6cm 12cm 4cm},clip]{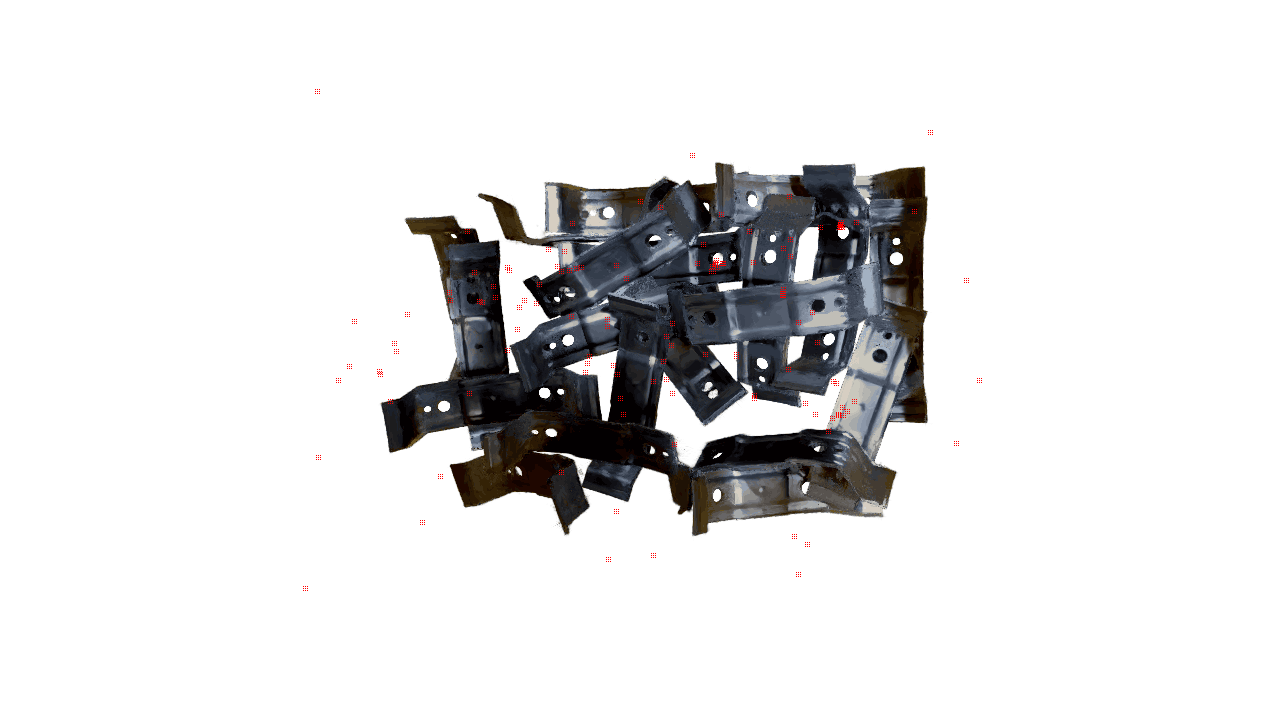}
\end{subfigure}\hfill
\begin{subfigure}[c]{0.15\textwidth}
    \centering
    \includegraphics[height=2cm,trim={14cm 6cm 12cm 4cm},clip]{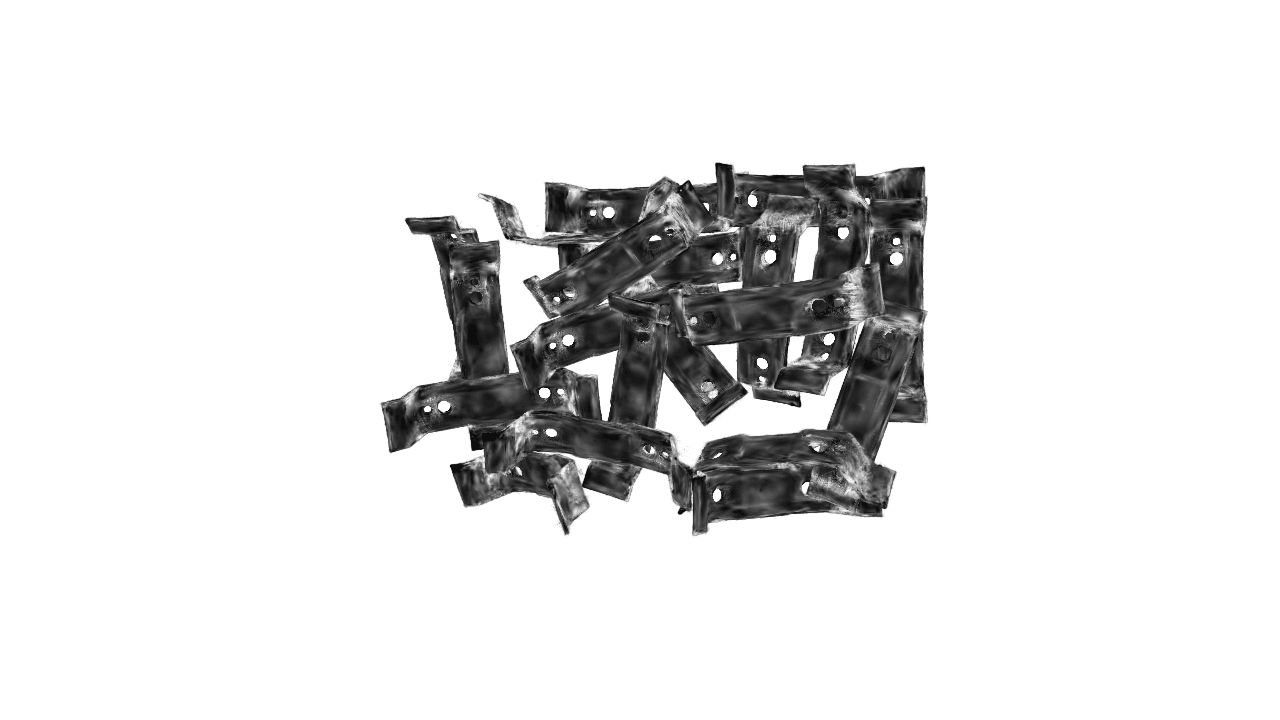}
\end{subfigure}\\
\rowlabel{3}%
\begin{subfigure}[c]{0.15\textwidth}
    \centering
    \includegraphics[width=\textwidth]{graphics/40_17.png}
\end{subfigure}\hfill
\begin{subfigure}[c]{0.15\textwidth}
    \centering
    \includegraphics[height=2cm,trim={14cm 6cm 7cm 4cm},clip]{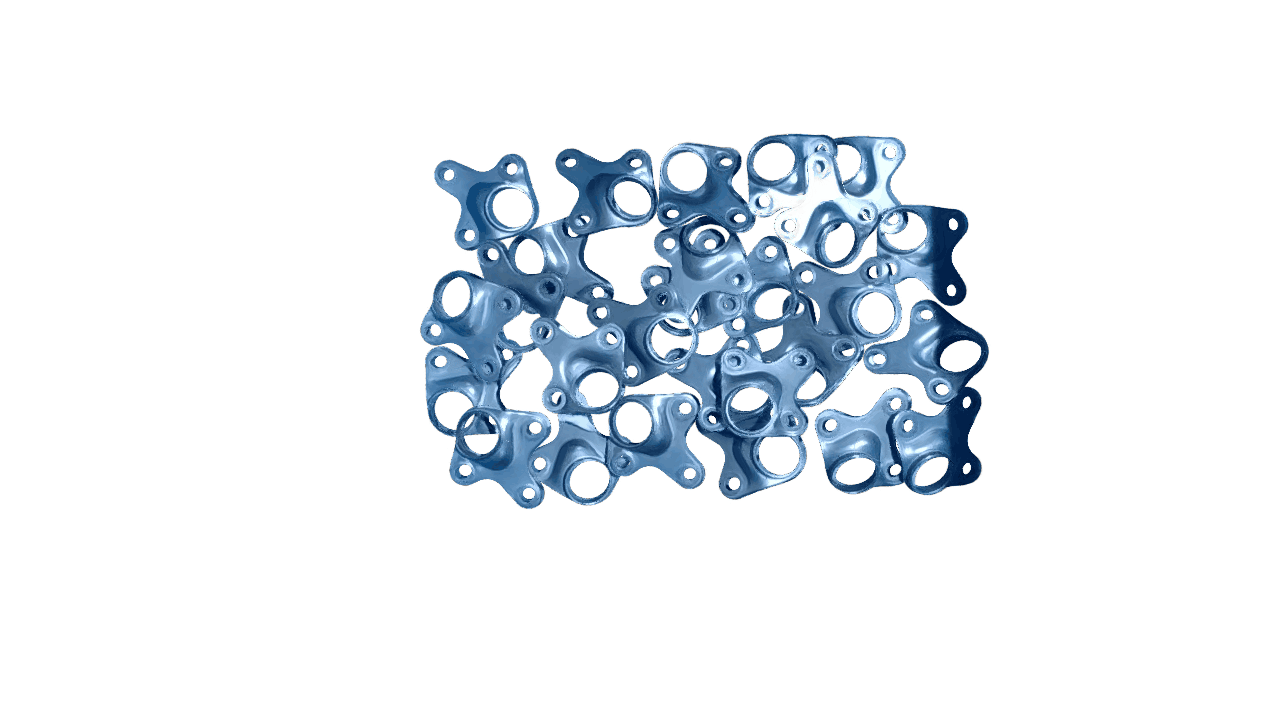}
\end{subfigure}\hfill
\begin{subfigure}[c]{0.15\textwidth}
    \centering
    \includegraphics[height=2cm,trim={14cm 6cm 7cm 4cm},clip]{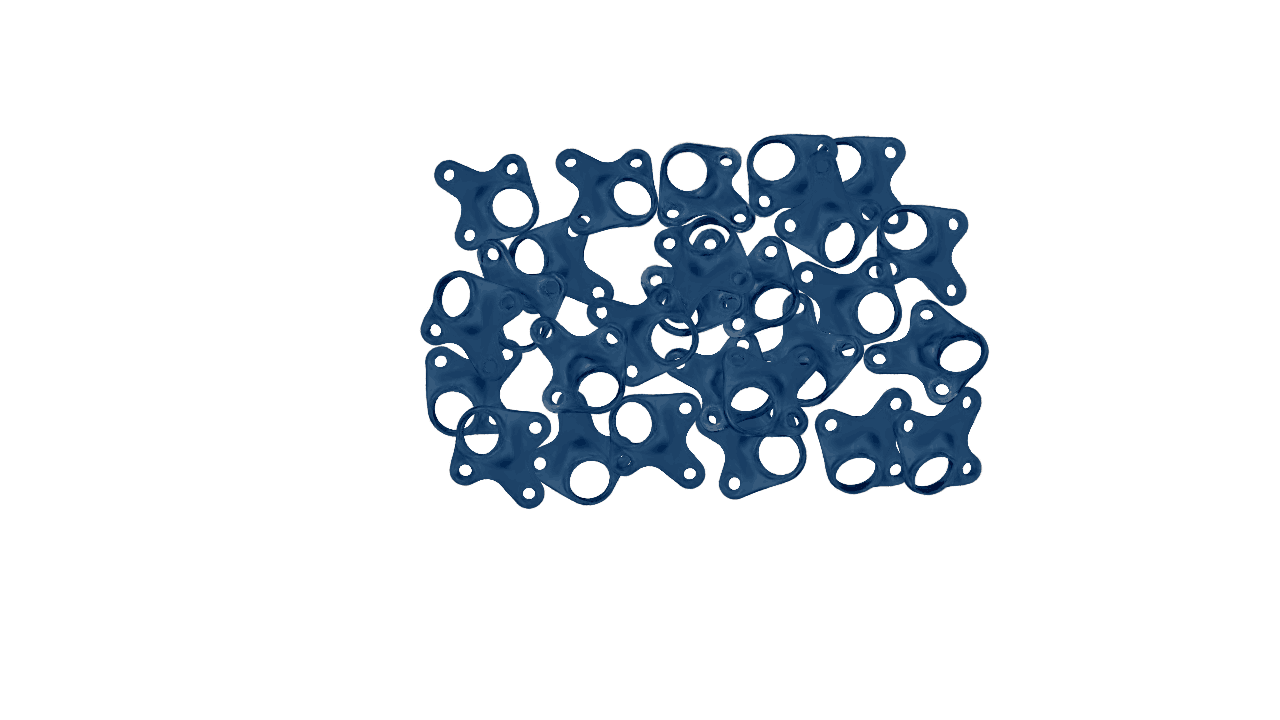}
\end{subfigure}\hfill
\begin{subfigure}[c]{0.15\textwidth}
    \centering
    \includegraphics[height=2cm,trim={14cm 6cm 7cm 4cm},clip]{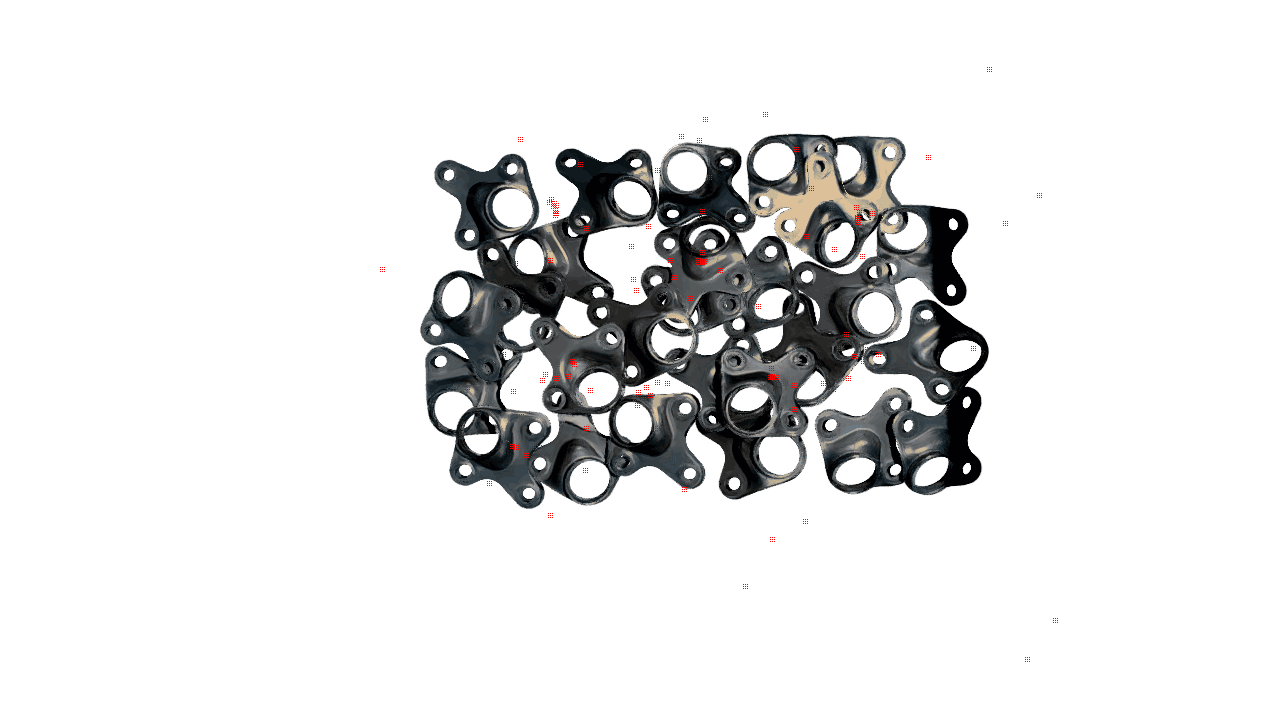}
\end{subfigure}\hfill
\begin{subfigure}[c]{0.15\textwidth}
    \centering
    \includegraphics[height=2cm,trim={14cm 6cm 7cm 4cm},clip]{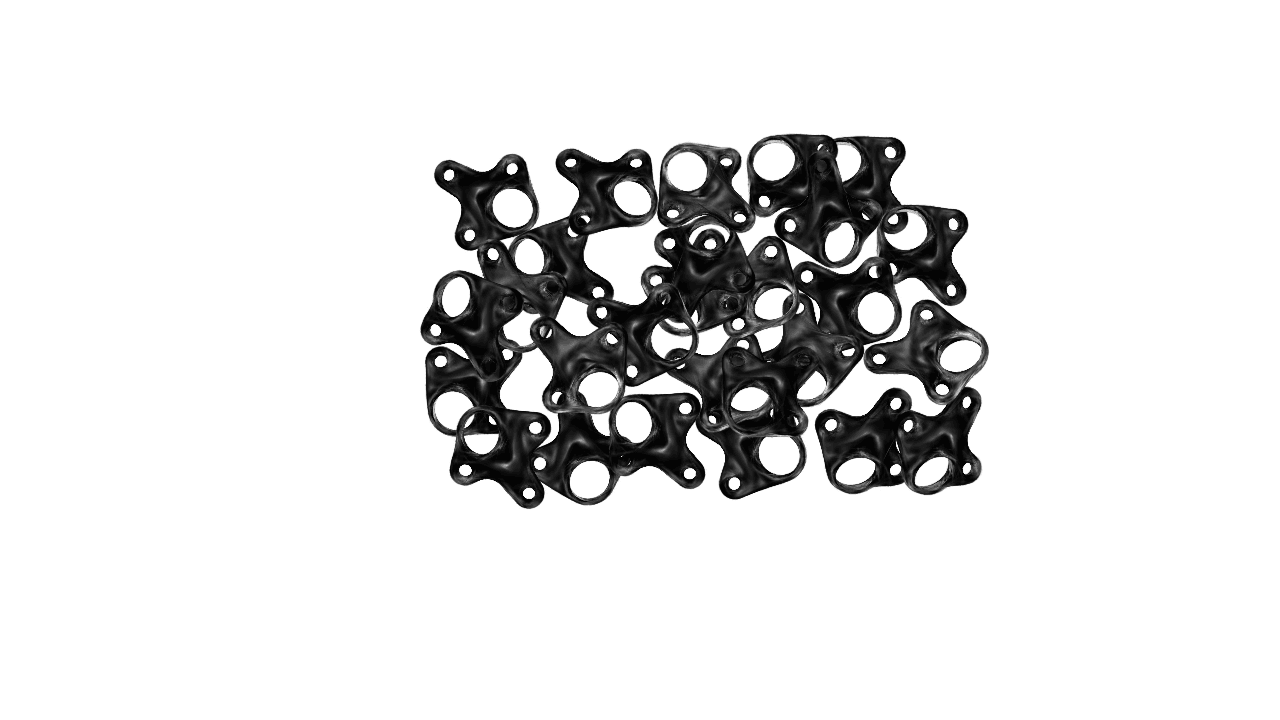}
\end{subfigure}
\\
\rowlabel{4}%
\begin{subfigure}[c]{0.15\textwidth}
    \centering
    \includegraphics[width=\textwidth]{graphics/30_34/000034_000000.png}
    \caption*{\centering Input image\\ \vphantom{p}}
\end{subfigure}\hfill
\begin{subfigure}[c]{0.15\textwidth}
    \centering
    \includegraphics[height=2cm,trim={13cm 6cm 10cm 4cm},clip]{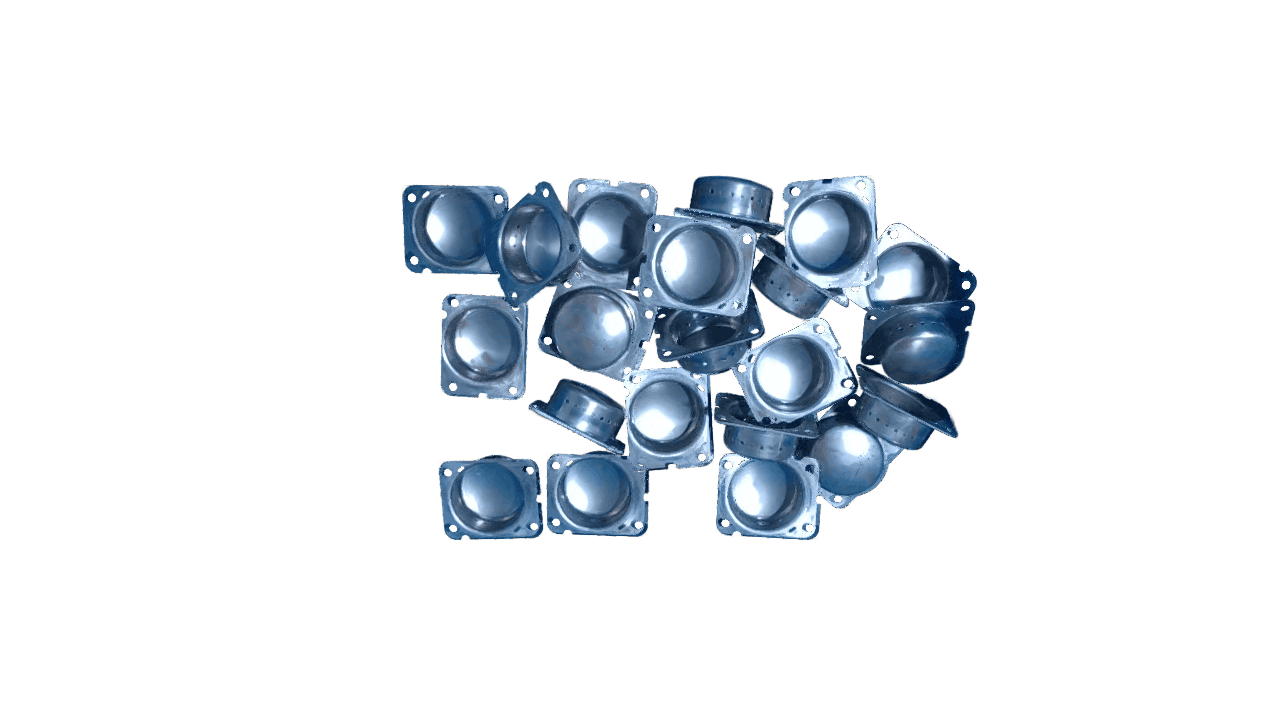}
    \caption*{\centering Rendered RGB\\ \vphantom{p}}
\end{subfigure}\hfill
\begin{subfigure}[c]{0.15\textwidth}
    \centering
    \includegraphics[height=2cm,trim={13cm 6cm 10cm 4cm},clip]{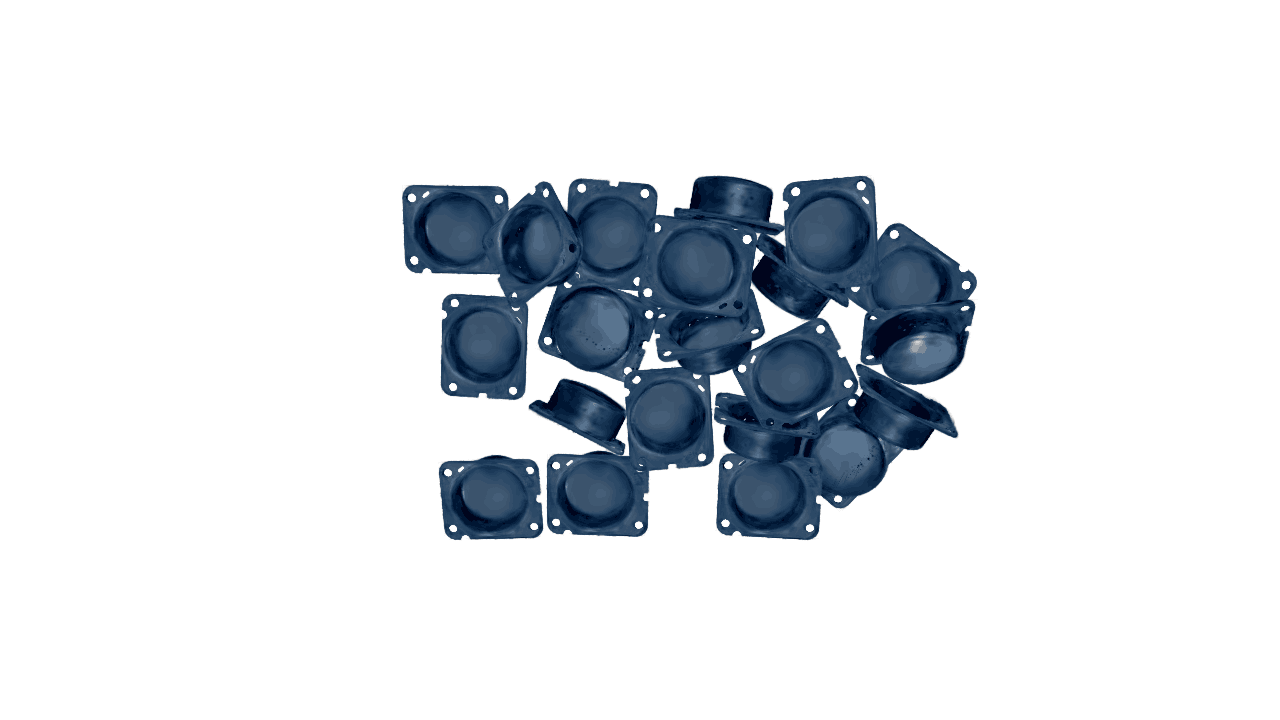}
    \caption*{\centering Rendered diffuse \\color}
\end{subfigure}\hfill
\begin{subfigure}[c]{0.15\textwidth}
    \centering
    \includegraphics[height=2cm,trim={13cm 6cm 10cm 4cm},clip]{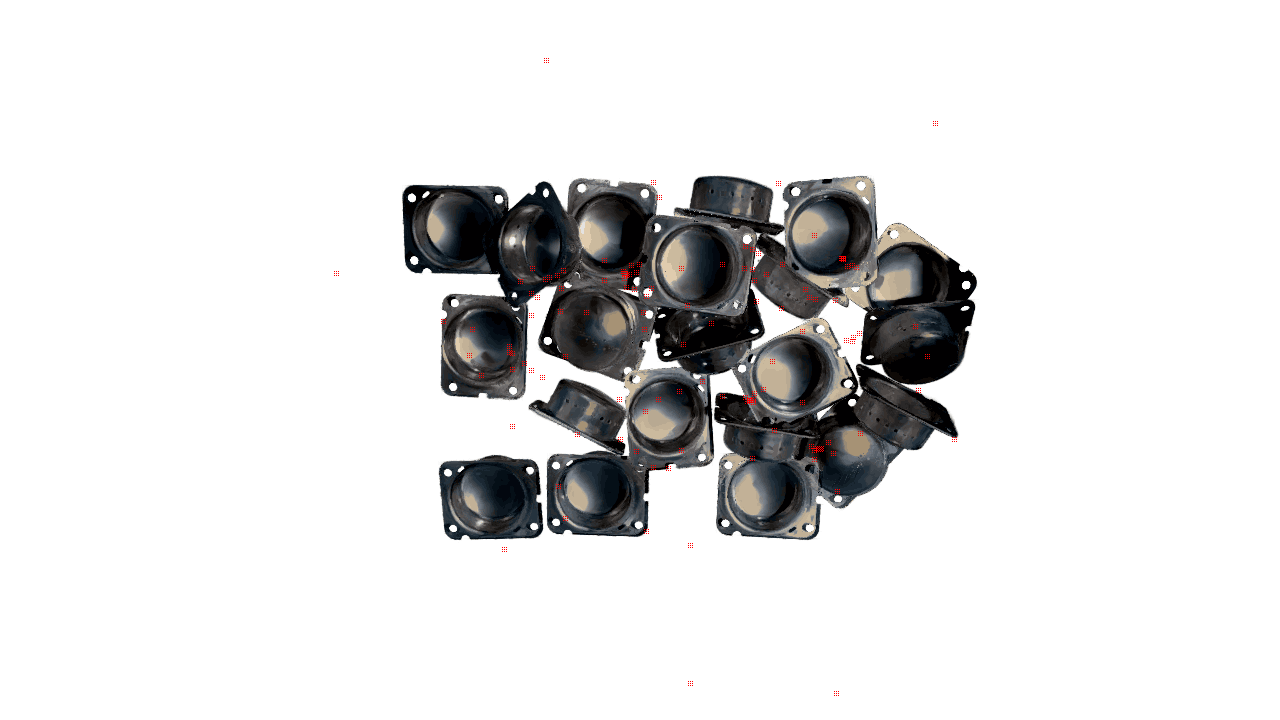}
    \caption*{\centering Rendered specular \\ with light probes}
\end{subfigure}\hfill
\begin{subfigure}[c]{0.15\textwidth}
    \centering
    \includegraphics[height=2cm,trim={13cm 6cm 10cm 4cm},clip]{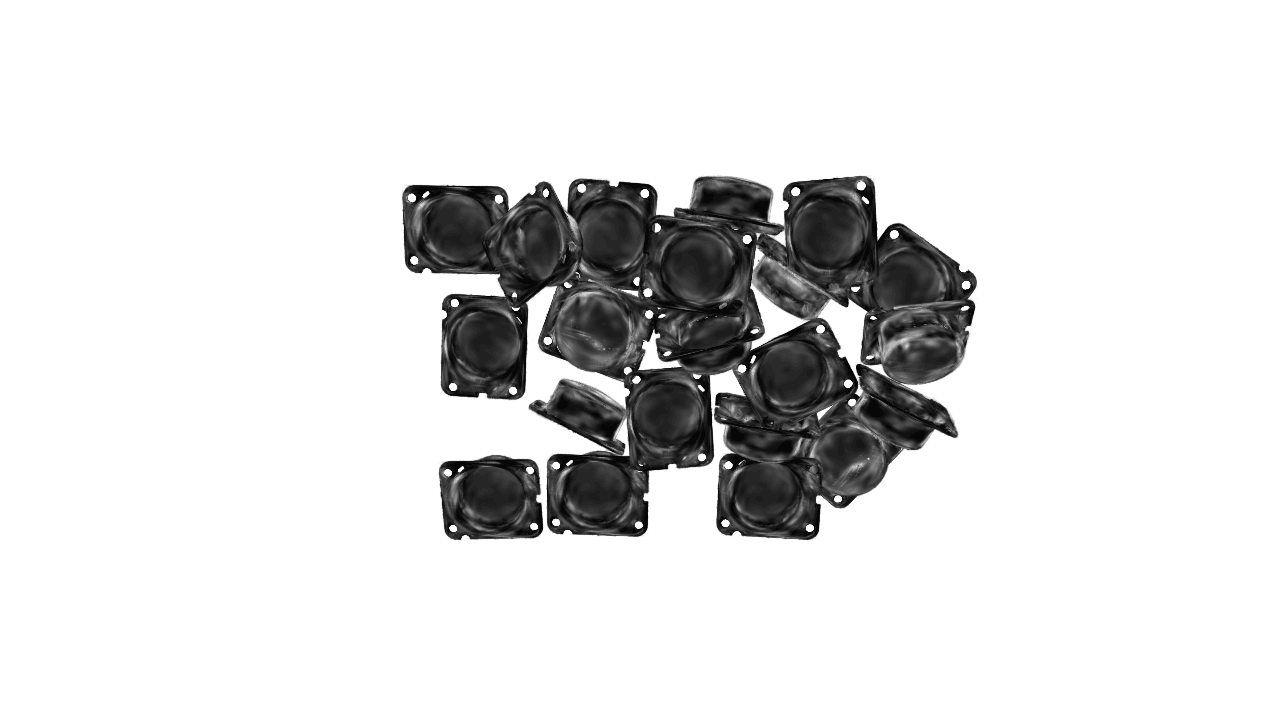}
    \caption*{\centering Rendered roughness\\ \vphantom{p}}
\end{subfigure}
\\
    \caption{Deferred rendering of \method on real-world data of XYZ-IBD 1-4. Positions of light probes are visualized as red dots. Zoom in for more detail.}
    \label{fig:xyz_diffuse_roughness}
\end{figure*}

In \cref{fig:xyz_diffuse_roughness} we show the learned deferred rendering properties of \method on the real-world XYZ-IBD dataset. The diffuse color renders show mostly no specular highlights, though exceptions exist, for example in scene XYZ4. The majority of specular highlights get explained by the light probes in combination with the roughness. Light probes are dispersed throughout the scenes, with clusters in regions with high inter- and intra-object occlusions and reflections, where they are needed.

